\documentclass[10pt,twocolumn,letterpaper]{article}

\usepackage{cvpr}              

\usepackage[dvipsnames]{xcolor}

\definecolor{cvprblue}{rgb}{0.21,0.49,0.74}
\usepackage{graphicx}
\usepackage{amsmath}
\usepackage{mathtools}
\usepackage{booktabs}
\usepackage{comment}
\usepackage{xcolor}
\usepackage{float}
\usepackage{subcaption}
\usepackage{pifont}
\newcommand{\mypara}[1]{\vspace{3pt}\noindent\textbf{#1}}
\usepackage[pagebackref,breaklinks,colorlinks,citecolor=cvprblue]{hyperref}
\usepackage{epsfig}
\usepackage{graphicx}
\usepackage{amsmath}
\usepackage{mathtools}
\usepackage{subcaption}
\usepackage{wasysym}
\usepackage{booktabs}
\usepackage{array}

\def\paperID{270} 
\def\confName{3DV\xspace}
\def\confYear{2025\xspace}

\DeclareMathOperator*{\argmin}{arg\,min}
\newcommand{\OURS}{HeadCraft}

\title{\OURS: Modeling High-Detail Shape Variations for Animated 3DMMs}

\author{
Artem Sevastopolsky$^1$  \qquad Philip-William Grassal$^2$  \qquad Simon Giebenhain$^1$ \\ ShahRukh Athar$^3$  \qquad Luisa Verdoliva$^{4,1}$  \qquad Matthias Nie{\ss}ner$^1$\\[0.3cm]
$^1$ Technical University of Munich (TUM), Germany \hspace{0.3cm}
$^2$ Copresence AG, Germany \\
$^3$ Stony Brook University, US \hspace{0.3cm}
$^4$ University of Naples Federico II, Italy
}

\begin{document}


\twocolumn[{%
\renewcommand\twocolumn[1][]{#1}%
\vspace{-0.75cm}
\maketitle
\vspace{-0.5cm}
\includegraphics[width=\textwidth,trim={0.2cm 7.3cm 0.65cm 0.1cm},clip]{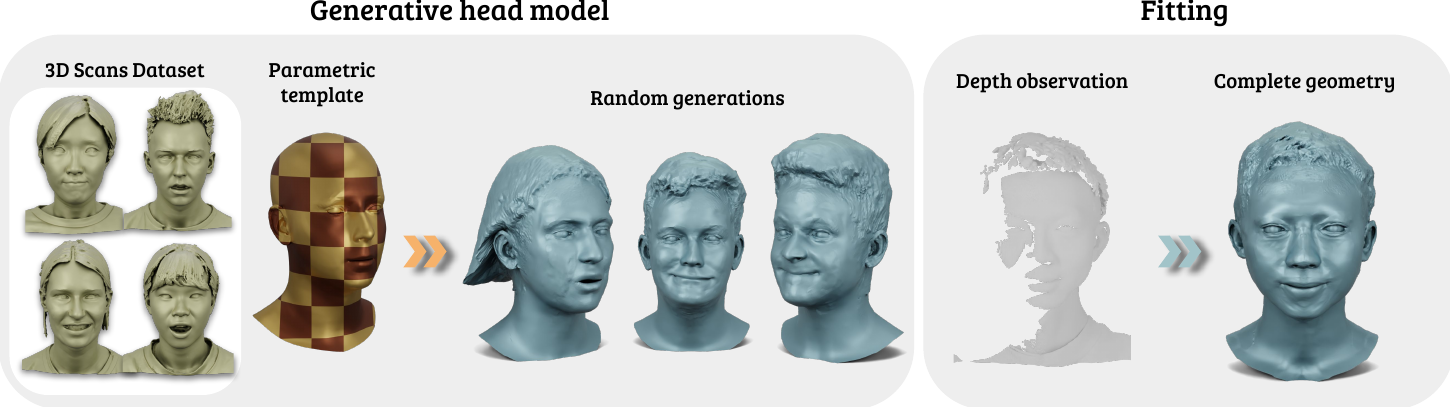}
\vspace{-0.5cm}
\captionof{figure}{
We present HeadCraft, a generative model for highly-detailed human heads, ready for animation. 
Our method is trained on 2D displacement maps collected by registering a parametric template head with free surface displacements to a large set of 3D head scans.
%
The resulting model is highly versatile and its latent code can be fit to an arbitrary depth observation. }
\label{fig:teaser}
\vspace{0.5cm}
}]
\begin{abstract}
\vspace{-1.0cm}
%
Current advances in human head modeling allow to generate plausible-looking 3D head models via neural representations, such as NeRFs and SDFs. 
%
Nevertheless, constructing complete high-fidelity head models with explicitly controlled animation remains an issue.
Furthermore, completing the head geometry based on a partial observation, e.g. coming from a depth sensor, while preserving a high level of detail is often problematic for the existing methods.
%
We introduce a generative model for detailed 3D head meshes on top of an articulated 3DMM which allows explicit animation and high-detail preservation at the same time.
%
Our method is trained in two stages.
%
First, we register a parametric head model with vertex displacements to each mesh of the recently introduced NPHM dataset of accurate 3D head scans.
The estimated displacements are baked into a hand-crafted UV layout.
%
Second, we train a StyleGAN model in order to generalize over the UV maps of displacements, which we later refer to HeadCraft.
%
%
%
%
The decomposition of the parametric model and high-quality vertex displacements allows us to animate the model and modify the regions semantically.
%
We demonstrate the results of unconditional sampling, fitting to a scan and editing.
The code and data are available at \url{https://seva100.github.io/headcraft}.
\end{abstract}

\vspace{-0.1cm}
\section{Introduction}

The ability to create lifelike 3D head models is crucial for many applications, ranging from video game character design to virtual try-on experiences and medical simulations. 3D Morphable Models (3DMMs)~\cite{3dmm} constitute an essential and robust tool for basic 3D head geometry estimation and tracking that enables its further reconstruction and animation. 
Furthermore, 3DMMs, along with their consistent UV parameterization of the surface, are commonly used in constructing virtual avatars as an approach for approximate surface representation or regularization ~\cite{mvp,nha,gnarf,insta}.

Constructing (neural) 3DMMs capable of representing the diverse distribution over human heads while disentangling identity and expressions and capturing a high degree of detail is challenging.
At the same time, such neural representation would ideally be compatible with well-established methods for animation and tracking, which are usually mesh-based and are required to fulfill certain real-time constraints.

Despite the recent progress of implicit representations, such as neural radiance fields (NeRFs)~\cite{nerf} and signed distance functions (SDFs)~\cite{deepsdf}, the most prominent 3DMMs~\cite{3dmm,flame,bfm} are based on a template mesh (i.e. feature explicit geometry), and principal component analysis (PCA), to represent identity and expression variations. Mesh-based 3DMMs can usually be robustly fitted to videos, easily animated and integrate well with established graphics pipelines.
However, these approaches are fundamentally limited by both the mesh resolution and representational capacity of its underlying (multi-)linear statistical model. 
In our work, we improve on both of these aspects and leverage recent advances in generative image modeling by using StyleGAN2~\cite{stylegan2} to predict highly detailed geometry in the UV space, which is independent of the mesh resolution.

A different line of work attempts to build 3DMMs based on neural SDFs instead of meshes (e.g. ~\cite{i3dmm, nphm}), thereby enabling reconstruction at arbitrary resolutions.
However, the incorporation of such implicit representations is limited due to a lack of compatibility with standard graphics systems and animation tools. Furthermore, SDF-based approaches often require costly evaluations, e.g. via marching cubes~\cite{lorensen1998marching}, to extract an implicit surface, which can hinder the real-time application of these methods.

Inspired by the combination of these ideas, in this research, we introduce a generative model that allows for animation and tracking and preserves a high level of detail.
At the heart of our approach lies the idea of combining an explicit parametric head model (FLAME~\cite{flame}) with surface displacements complementing the low geometry detail of the head model.
FLAME is an example of a 3D Morphable Model~\cite{3dmm, egger20203d} with a fixed set of vertices and fixed topology, constructed as a linear statistical model over the heads with point-to-point correspondence and further controlled by shape and expression latent codes.
We register a highly subdivided FLAME mesh template with free vertex displacements to all 3D head scans in the NPHM~\cite{nphm} dataset to obtain the necessary training data. 
To facilitate as high level of detail in the displacements as possible, they are fitted in two steps.
First, the optimization problem is solved for vector displacements with strong regularization that penalizes very hard for self-intersections of the deformed mesh regions.
Afterwards, a separate optimization step refines the displacements only along the normals of the deformed vertices while keeping the regularization weight low.
These displacements are baked into a predefined UV layout.
Finally, we train a StyleGAN2~\cite{stylegan2} model to generalize over this set of baked 2D displacement maps.
This novel architecture allows us to operate at a resolution higher than the conventional FLAME template, enabling the generation of highly detailed and animatable 3D head models.

To validate the efficacy and practical utility of our approach, we evaluate it in several settings. 
The diversity and fidelity of the generated 3D head meshes are quantitatively and visually compared to other methods w.r.t. the real head scans from the FaceVerse dataset \cite{faceverse}, both in UV space and rendered image space.
We also explore the applicability of our approach in fitting the latent representation of the generative model to complete or incomplete point cloud data and demonstrate its animation and manipulation capabilities.


\begin{figure*}[t!]
\vspace{-0.6cm}
\begin{center}
\includegraphics[width=\textwidth,trim={0.4cm 6cm 0.4cm 0cm},clip]{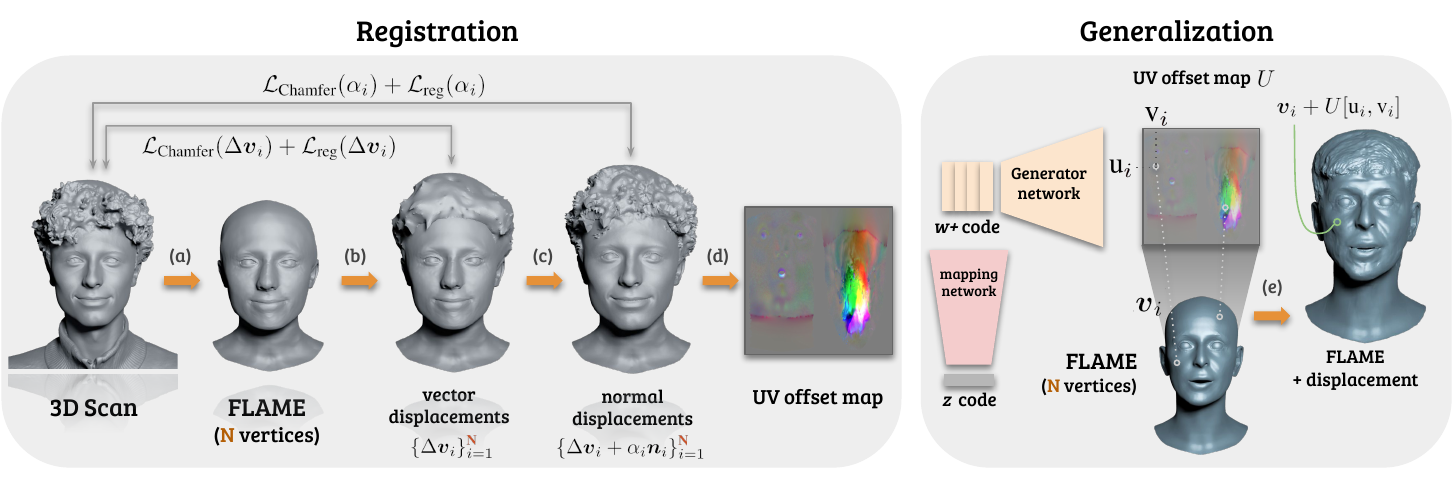}
\vspace{-0.7cm}
\captionof{figure}{
An overview of the method. 
In the registration stage, we \textbf{(a)} fit the FLAME template by the face landmarks to the scan from the NPHM dataset and highly subdivide it, \textbf{(b)} optimize for the vertex displacements in $\mathbb{R}^3$ to fit the rough geometry with strong regularizations, \textbf{(c)} optimize for the scalar refinements of the displacements along the normal directions, and \textbf{(d)} bake the displacements into a UV offset map. 
To generalize over the UV offset maps, we train a StyleGAN2~\cite{stylegan2} model.
After training, the offsets can be applied to an arbitrary FLAME template by subdividing it and \textbf{(e)} querying the generated UV offset map with the (u, v) locations of the FLAME vertices.
}
\label{fig:overview}
\end{center}
\vspace{-0.5cm}
\end{figure*}

To summarize, our contributions are as follows:
\begin{itemize}
    \item We introduce a two-stage registration procedure to craft high-detail displacement maps on top of 3DMMs from 3D scanning data. This enables the application of 2D generative models to tackle the 3D generative task.
    \item We propose a generative model operating in the displacement maps domain to enhance the low-frequency geometry of FLAME with details and extend its shape space to a range of variations.
    \item We demonstrate the versatility of our method through unconditional sampling, interpolation, semantic geometry transfer, and 3D completion based on partial depth observations.
\end{itemize}

\section{Related Work}
%
%
Many recent solutions to computer vision problems involving human bodies and heads are built on statistical body models, forming the foundation for building personalized avatars \cite{nha, insta, detailedhumanavatars, nerface, peoplefromclothing}, motion tracking \cite{face2face, deca}, scan registration \cite{nphm}, controlling image synthesis \cite{stylerig}, and more.
Their line of research is divided into two major branches.

\mypara{Mesh-based Models.}
Pioneering work in the field \cite{3dmm} proposed 3D morphable models (3DMMs) for the human faces' identity, expression, and appearance representation.
Their model is built around a 3D template mesh and linear parametric blendshapes derived from PCAs over 3D scan data.
With new datasets and registration procedures, their work has been extended from faces to heads \cite{flame, ict, otherhead}, hands \cite{mano}, full bodies \cite{smpl}, or combinations of these \cite{smpl-x, ghum}.
The template mesh has a fixed topology which provides consistent UV unwrapping and enables fitting to know surface correspondences, e.g. semantic regions and landmarks.
Yet, it limits the representative power w.r.t. the overall shape and level of detail beyond what the template provides.
Downstream approaches compensate this by optimizing displacements \cite{rome, nha, detailedhumanavatars, dynamicsurface, peoplefromclothing, facescape, ict} or additional implicit geometry \cite{delta, authenticavatars} on top of the mesh.
Displacements are applied either per-vertex individually \cite{rome, nha, detailedhumanavatars, peoplefromclothing} or as a displacement map over the whole surface using the consistent UV unwrapping of the template \cite{facescape, ict, deca}.
Some approaches infer the displacement maps from images or texture reprojections for specific individuals \cite{facescape, deca}.
By making use of high-quality scans, the authors of \cite{ict} demonstrate that GANs \cite{gan, stylegan, stylegan2} for image generation can be leveraged to learn a generative model over displacement maps in the face region.
%
Our method takes this idea further by learning a generative model for displacements over the whole head. It showcases that even the surface geometry of large and complicated hairstyles can be represented with high fidelity.
The resulting outputs of our full-head generative model provide quality approaching the scan level while exploiting the UV surface correspondences of the underlying 3DMM and enabling further animation through the rigging of the template.
%

\mypara{Implicit Models.}
The recent success of implicit SDFs \cite{deepsdf} and neural radiance fields \cite{nerf} in 3D modeling has also motivated applying them for statistical body models.
Most implicit models learn shape and appearance in a canonical reference space \cite{nphm, i3dmm, imGHUM, headnerf, h3dnet} or as displacements on top of an existing model \cite{PhoMoH}.
For better generalization and detail preservation, some approaches use a composition of implicit SDFs to model the canonical space \cite{imGHUM, nphm}.
Articulation and animation are modeled either directly in canonical coordinates \cite{headnerf, PhoMoH}, through implicit deformation fields \cite{nphm, i3dmm}, explicit joints \cite{imGHUM} or blendshape deformations borrowed from explicit 3DMMs \cite{IMAvatar, anifacegan}.
While the aforementioned methods rely on multi-view data and aligned 3D scans, a separate line of research demonstrates that statistical shape and appearance priors can also be learned from unstructured image collections  \cite{pigan, eg3d, lolnerf, stylesdf, ag3d}.

Implicit approaches do not rely on topology and shape templates.
This allows them to fit more detail and complex shapes such as hair \cite{nphm, PhoMoH, i3dmm} and even glasses \cite{eg3d, stylesdf}.
Yet, it prevents consistent surface correspondences between different samples, which need to be explicitly learned \cite{PhoMoH, imGHUM}.
As our approach uses a mesh-based template, it does not suffer from these limitations and has an explicit model for animation.
Still, we are able to show that we can provide a comparable level of detail as implicit methods and also model hair surface geometry which has not been achieved with a generative, explicit shape model before. 

\newcommand{\diverseVisualsPicSize}{0.12\textwidth}
\begin{table*}[h!]
    \centering
    \setlength{\tabcolsep}{0pt}
    \begin{tabular}{p{.4cm}ccccccc}
        \raisebox{2.5\normalbaselineskip}[0pt][0pt]{\rotatebox[origin=c]{90}{\small\textbf{Ours}}} & 
        \includegraphics[width=\diverseVisualsPicSize,trim={0cm 2cm 1cm 0cm},clip]{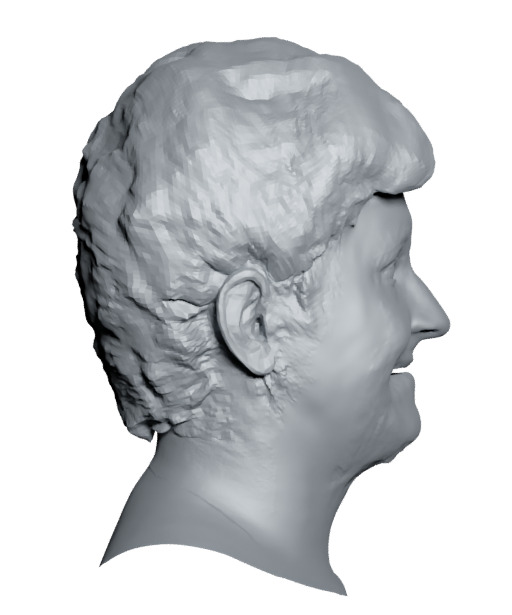} &
        \includegraphics[width=\diverseVisualsPicSize,trim={0cm 2cm 1cm 0cm},clip]{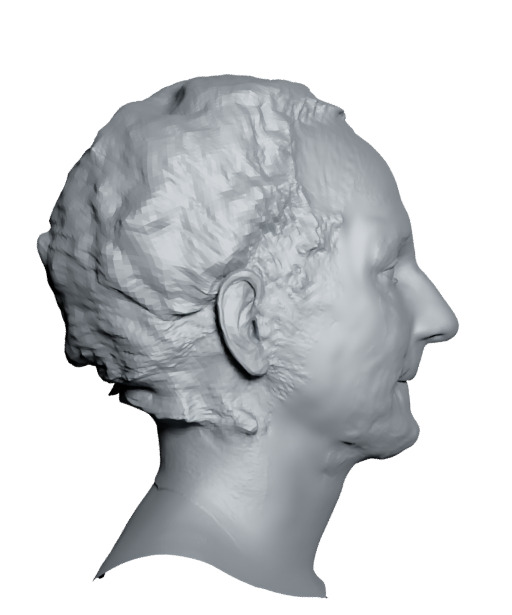} & 
        \includegraphics[width=\diverseVisualsPicSize,trim={0cm 2cm 1cm 0cm},clip]{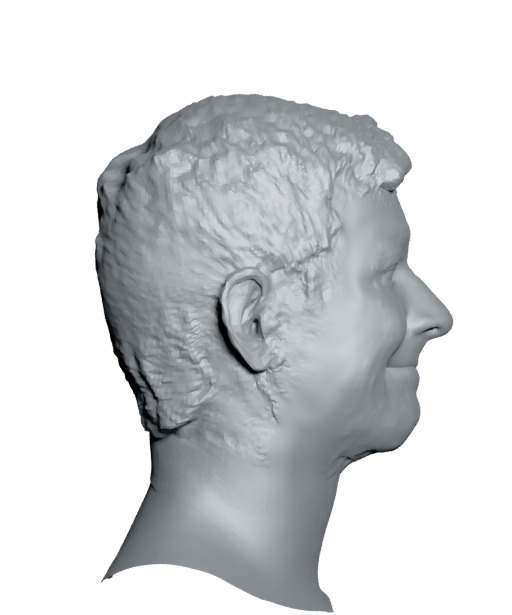} & 
        \includegraphics[width=\diverseVisualsPicSize,trim={0cm 2cm 1cm 0cm},clip]{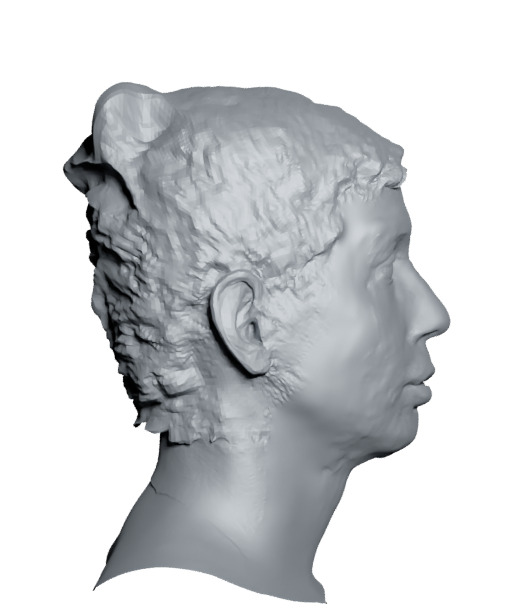} & 
        \includegraphics[width=\diverseVisualsPicSize,trim={0cm 2cm 1cm 0cm},clip]{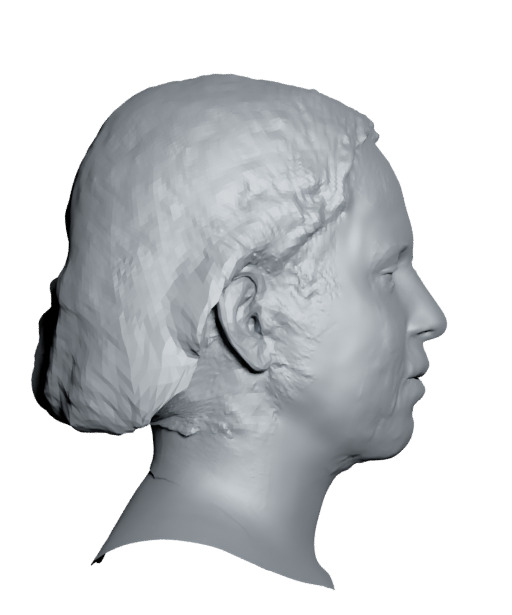} & 
        \includegraphics[width=\diverseVisualsPicSize,trim={0cm 2cm 1cm 0cm},clip]{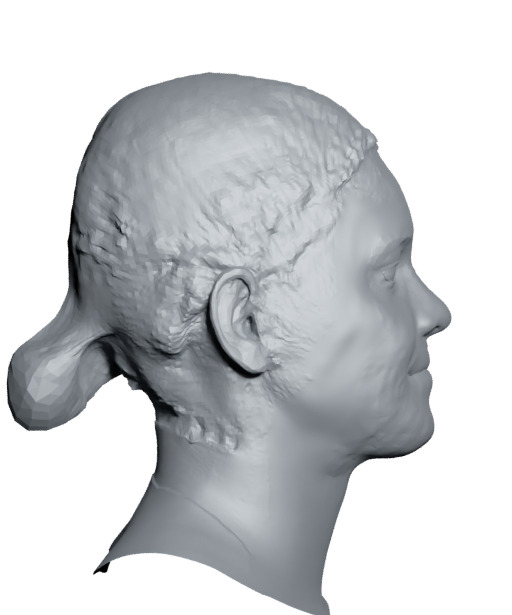} & 
        \includegraphics[width=\diverseVisualsPicSize,trim={0cm 2cm 1cm 0cm},clip]{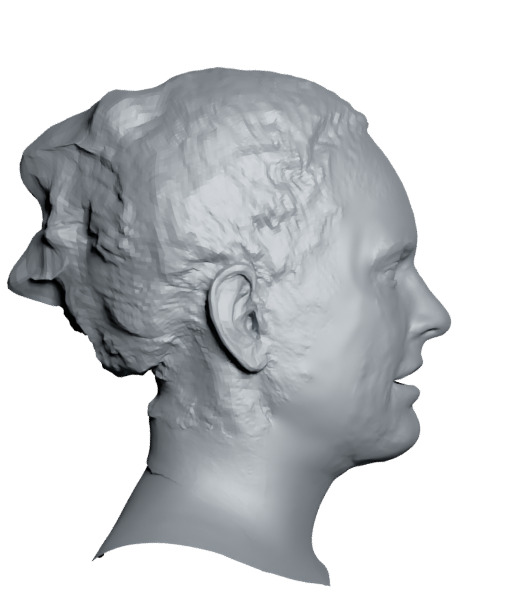} \\
        \raisebox{2.28\normalbaselineskip}[0pt][0pt]{\rotatebox[origin=c]{90}{\small NPHM~\cite{nphm}}} & 
        \includegraphics[width=\diverseVisualsPicSize,trim={0cm 11cm 0cm 9cm},clip]{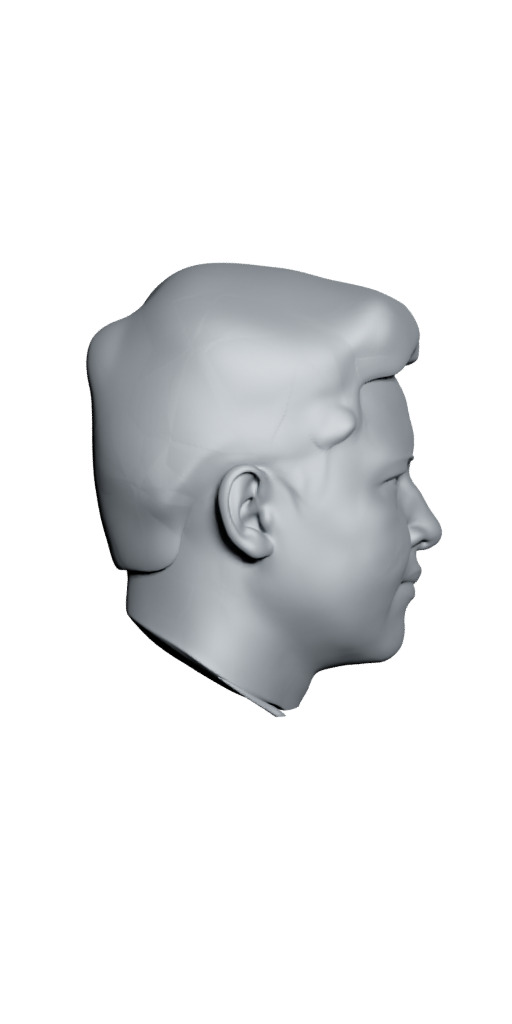} & 
        \includegraphics[width=\diverseVisualsPicSize,trim={0cm 11cm 0cm 9cm},clip]{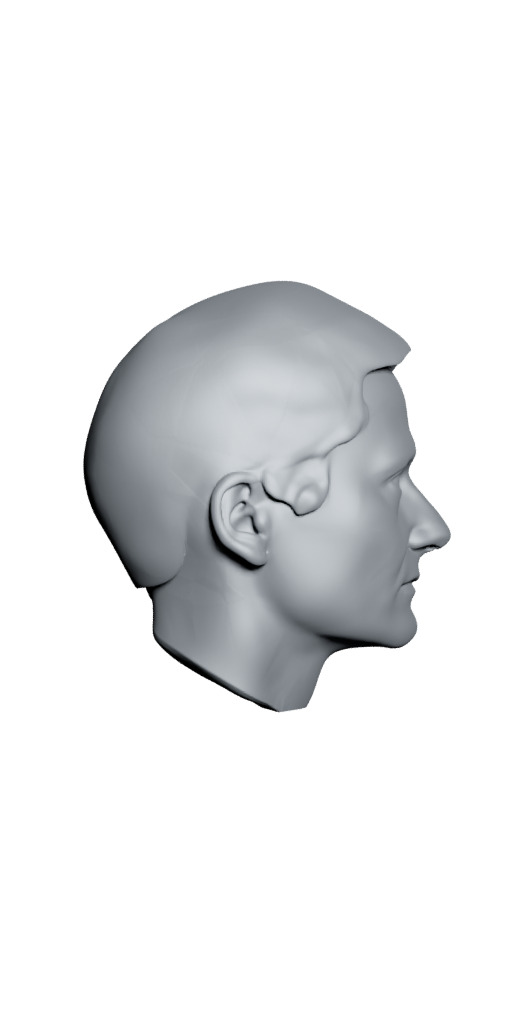} & 
        \includegraphics[width=\diverseVisualsPicSize,trim={0cm 11cm 0cm 9cm},clip]{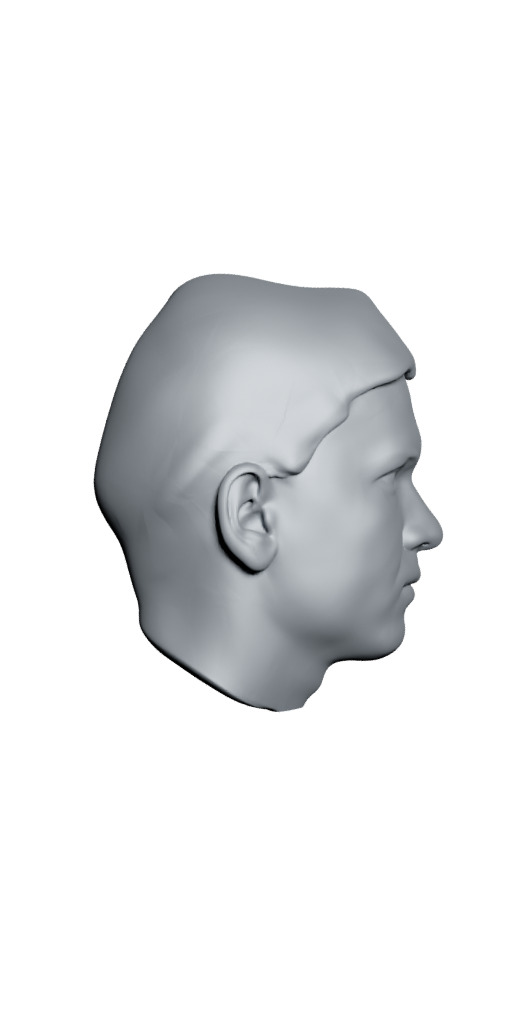} & 
        \includegraphics[width=\diverseVisualsPicSize,trim={0cm 11cm 0cm 9cm},clip]{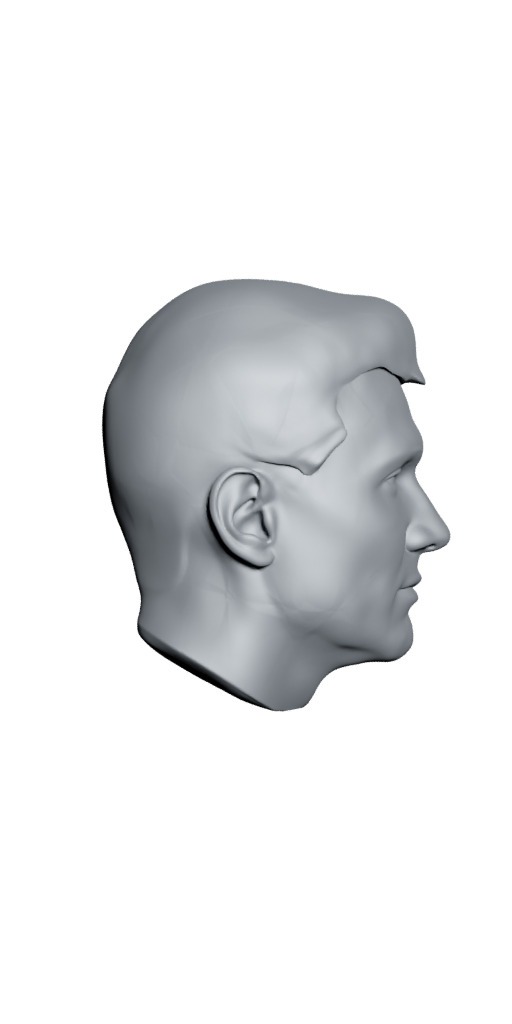} & 
        \includegraphics[width=\diverseVisualsPicSize,trim={0cm 11cm 0cm 9cm},clip]{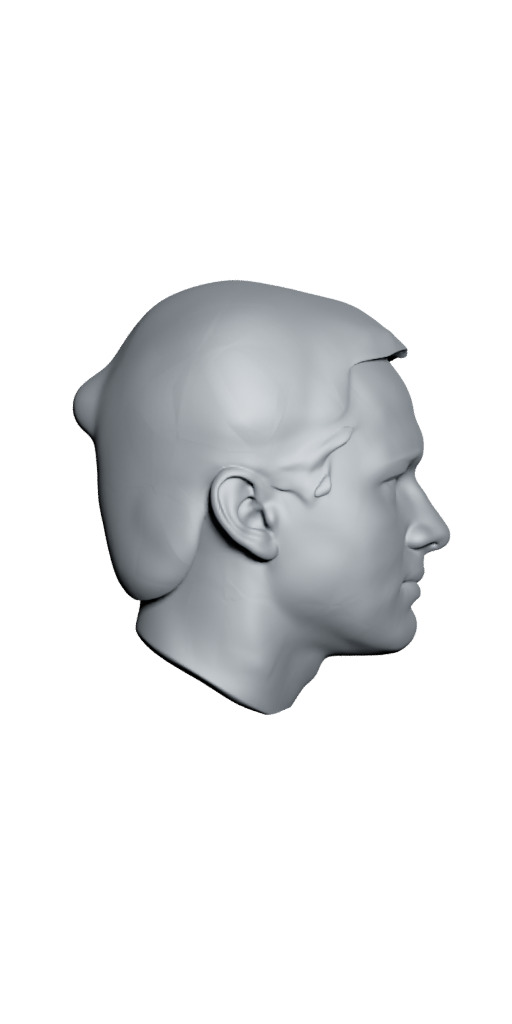} & 
        \includegraphics[width=\diverseVisualsPicSize,trim={0cm 11cm 0cm 9cm},clip]{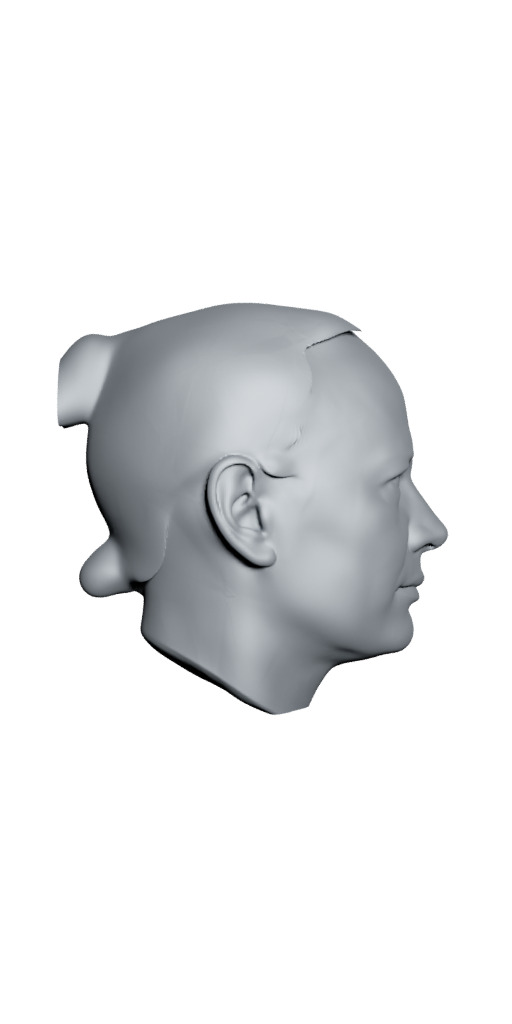} & 
        \includegraphics[width=\diverseVisualsPicSize,trim={0cm 11cm 0cm 9cm},clip]{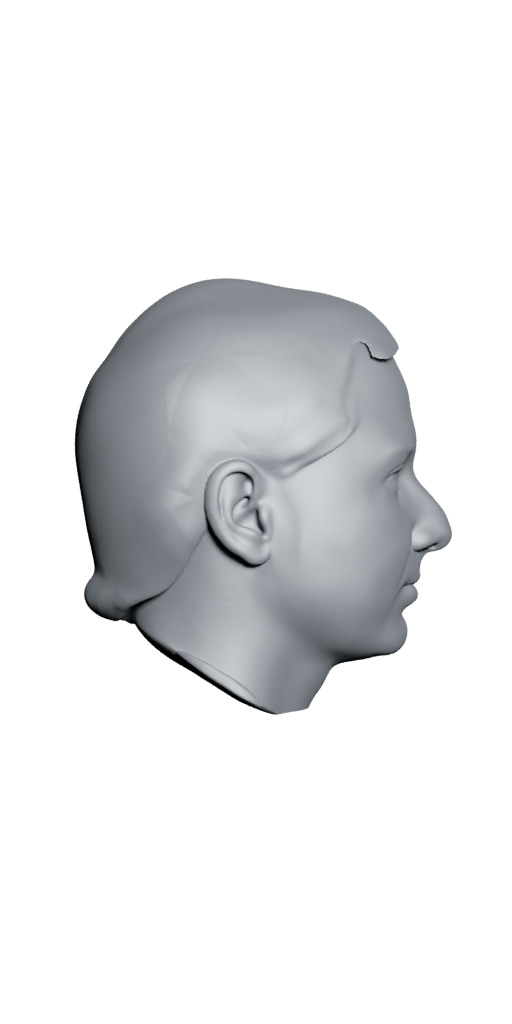}  \\
        %
        %
        \raisebox{2.4\normalbaselineskip}[0pt][0pt]{\rotatebox[origin=c]{90}{{\small ROME}$_{\;\text{\textcolor{gray}{(linear)}}}$~\cite{rome}}} & 
        \includegraphics[width=\diverseVisualsPicSize,trim={0cm 2cm 0cm 2.5cm},clip]{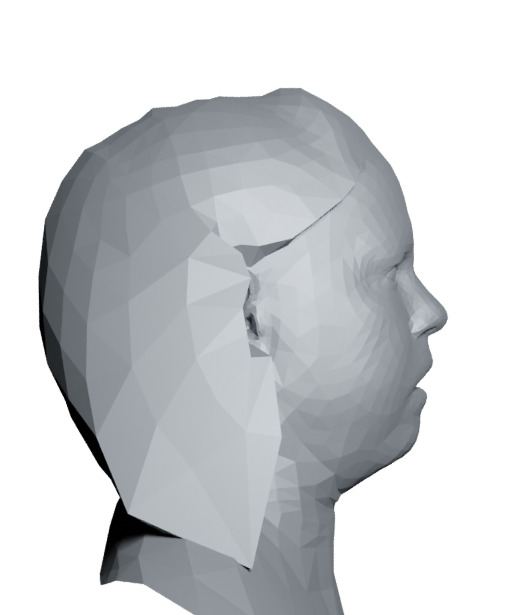} & 
        \includegraphics[width=\diverseVisualsPicSize,trim={0cm 2cm 0cm 2.5cm},clip]{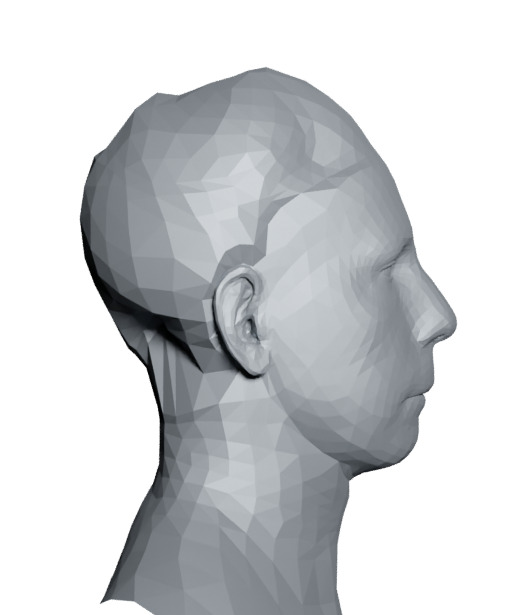} & 
        \includegraphics[width=\diverseVisualsPicSize,trim={0cm 2cm 0cm 2.5cm},clip]{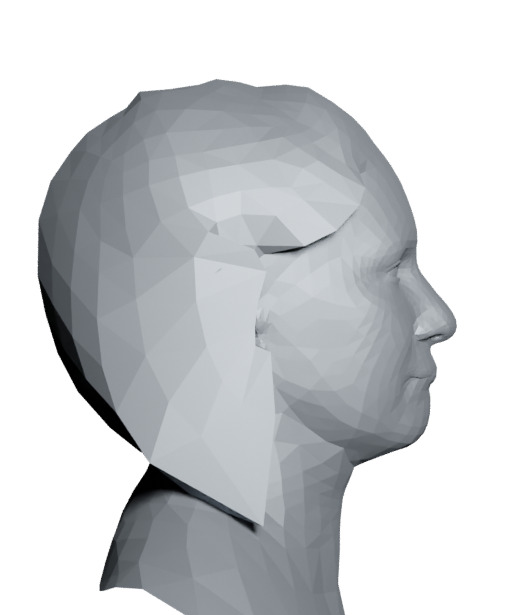} & 
        \includegraphics[width=\diverseVisualsPicSize,trim={0cm 2cm 0cm 2.5cm},clip]{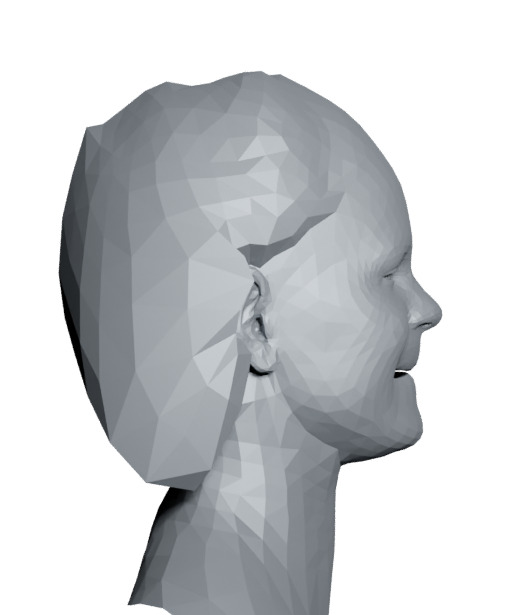} & 
        \includegraphics[width=\diverseVisualsPicSize,trim={0cm 2cm 0cm 2.5cm},clip]{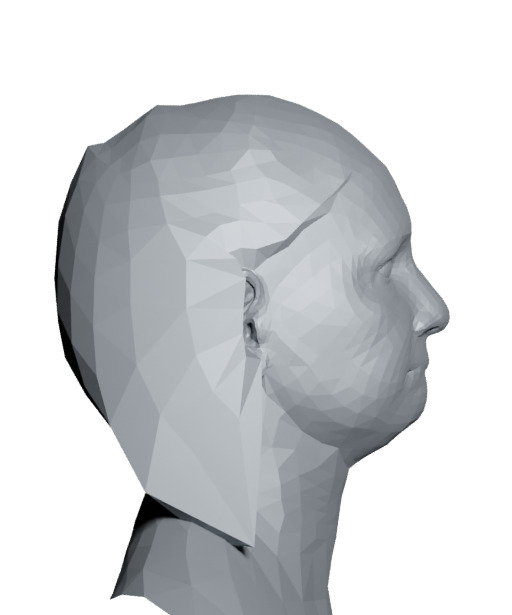} & 
        \includegraphics[width=\diverseVisualsPicSize,trim={0cm 2cm 0cm 2.5cm},clip]{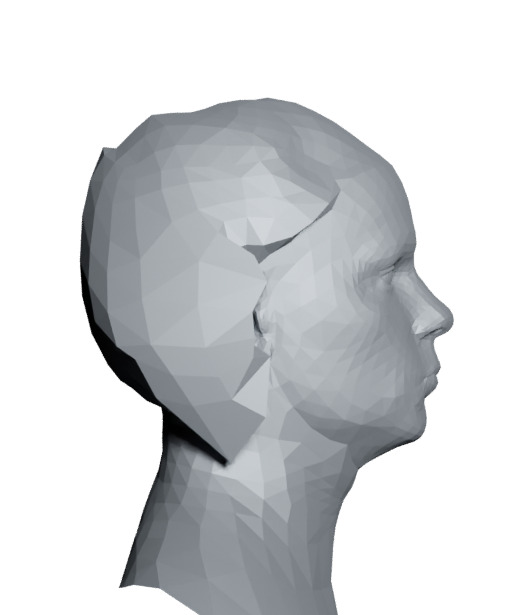} & 
        \includegraphics[width=\diverseVisualsPicSize,trim={0cm 2cm 0cm 2.5cm},clip]{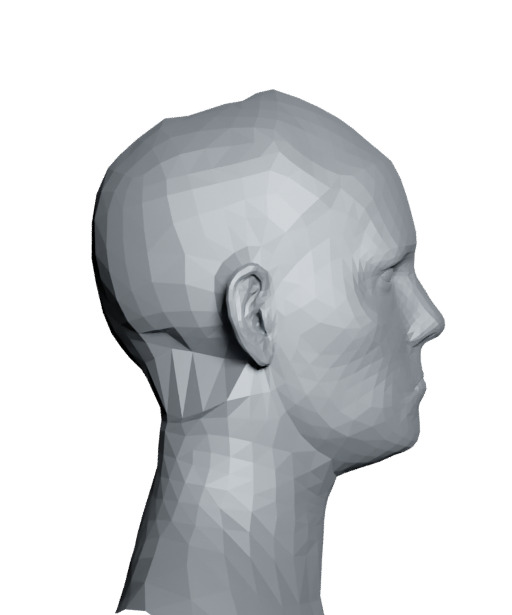}  \\
        \raisebox{2.5\normalbaselineskip}[0pt][0pt]{\rotatebox[origin=c]{90}{\small PCA baseline}} &  
        \includegraphics[width=\diverseVisualsPicSize,trim={0cm 2cm 1cm 1cm},clip]{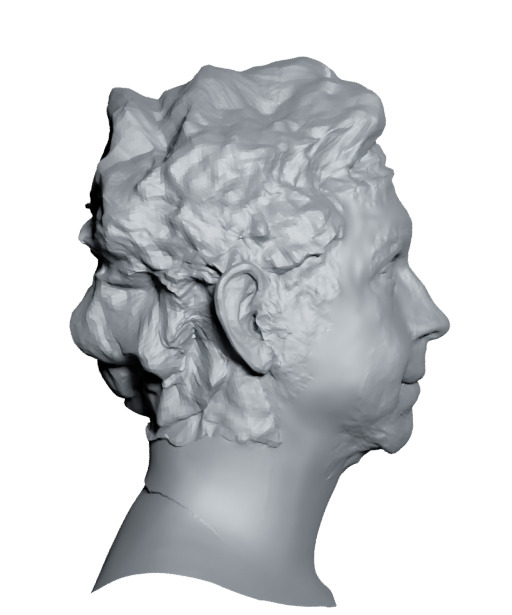} & 
        \includegraphics[width=\diverseVisualsPicSize,trim={0cm 2cm 1cm 1cm},clip]{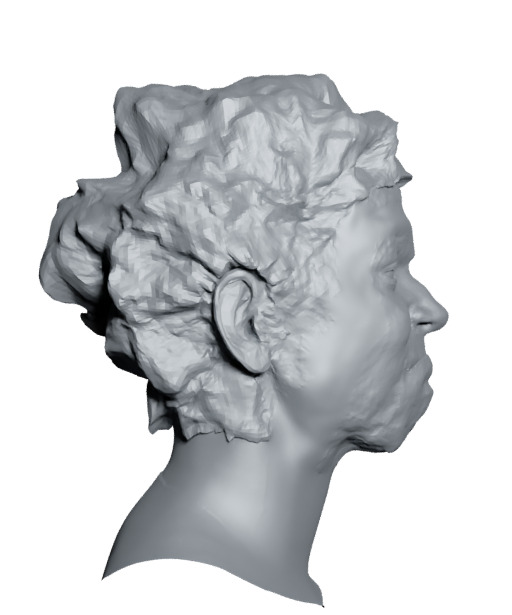} & 
        \includegraphics[width=\diverseVisualsPicSize,trim={0cm 2cm 1cm 1cm},clip]{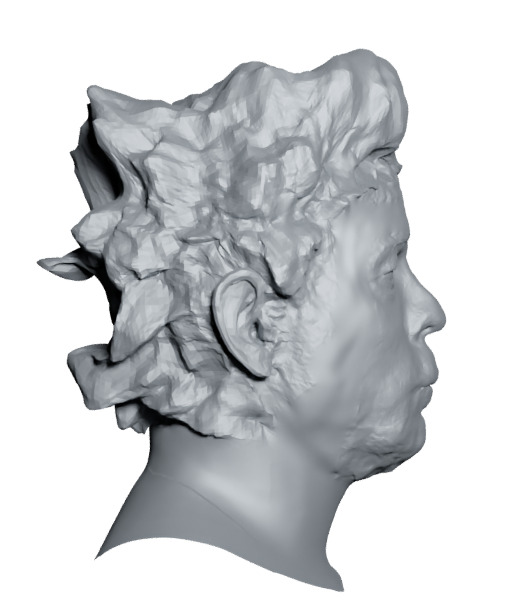} & 
        \includegraphics[width=\diverseVisualsPicSize,trim={0cm 2cm 1cm 1cm},clip]{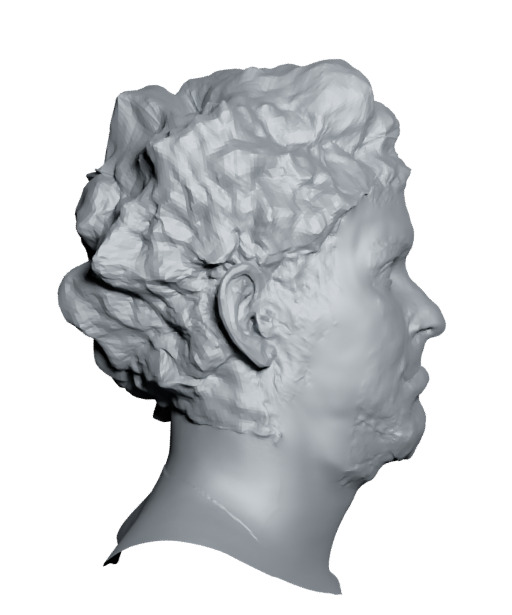} & 
        \includegraphics[width=\diverseVisualsPicSize,trim={0cm 2cm 1cm 1cm},clip]{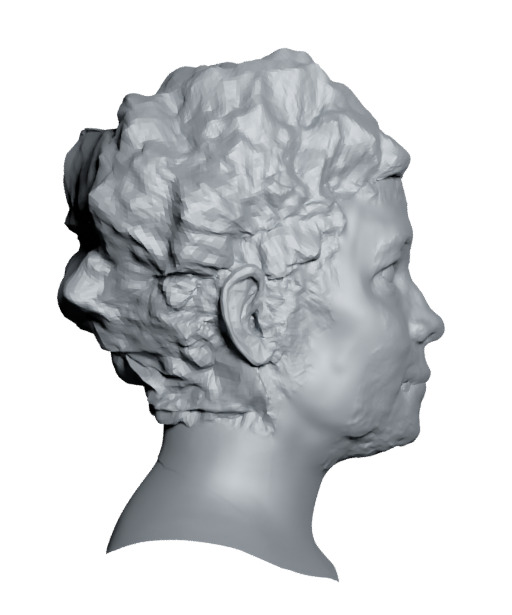} & 
        \includegraphics[width=\diverseVisualsPicSize,trim={0cm 2cm 1cm 1cm},clip]{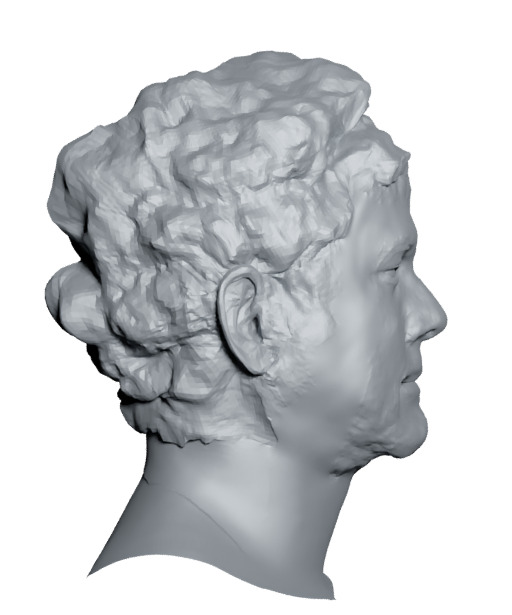} & 
        \includegraphics[width=\diverseVisualsPicSize,trim={0cm 2cm 1cm 1cm},clip]{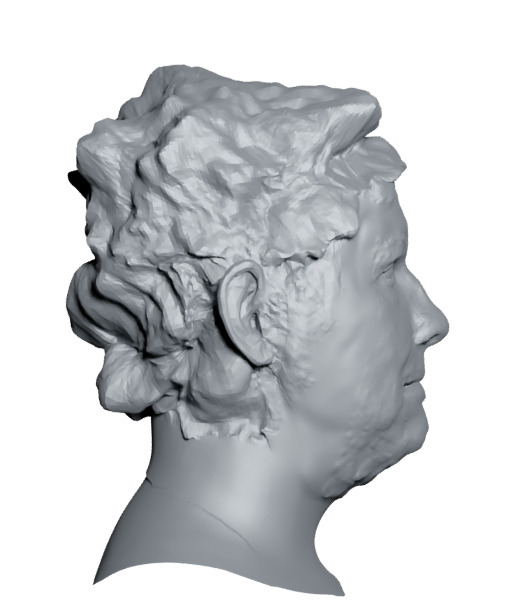} 
    \end{tabular}
    \vspace{-0.2cm}
    \captionof{figure}{
    Visual comparison of fidelity and diversity of the meshes generated by various methods.
    For \textit{Ours}, random FLAMEs are sampled from Gaussian distribution with statistics calculated over the NPHM dataset; same for the \textit{PCA baseline} pre-fitted to our UV registrations. 
    Meshes from \textit{NPHM} are obtained by sampling the latent codes and running marching cubes over the generated SDF representations. 
    We demonstrate higher variability of produced head geometry and better details than the other methods. 
    %
    }
    \vspace{-0.3cm}
    \label{fig:diverse_visuals}
\end{table*}

\section{HeadCraft}

We cast the task of learning high-fidelity distribution over 3D heads as a 2D generative problem by leveraging the UV-space of an existing 3DMM~\cite{flame}. This allows us to rely on the well-explored body of research on CNN-based 2D generative models~\cite{stylegan2, stylegan2-ada}.

To this end, we register a dataset of high-end 3D head scans~\cite{nphm} in the FLAME head model \cite{flame} topology and bake the details into a displacement map in UV space, as explained further in Sec.~\ref{subsec:flame_reg}.
Subsequently, in Sec.~\ref{subsec:gen_model}, we train a 2D generative model to learn the distribution over the displacement maps, which can be converted into detailed 3D displacements compatible with the underlying FLAME model.
An overview of HeadCraft is presented in Fig.~\ref{fig:overview}.

\subsection{Displacements registration procedure}
\label{subsec:flame_reg}

The purpose of this step is to create a dataset of 2D displacement maps representing the geometric information beyond the representational capacity of classical 3DMMs.
To this end, we compute displacements from a FLAME~\cite{flame} mesh to the high quality 3D head scans from the NPHM dataset~\cite{nphm}.
Let us consider a scanned mesh $\mathcal{P} = (V^\textrm{gt}, \mathcal{F}^\textrm{gt})$ with vertices $V^\textrm{gt} \in \mathbb{R}^{|V^\textrm{gt}| \times 3},$ and faces $\mathcal{F^\textrm{gt}} \in \{1, \dots, |V^\textrm{gt}|\}^{|\mathcal{F}^\textrm{gt}| \times 3}$.

In order to find the appropriate FLAME parameters for the scan, we follow the rigid alignment optimization procedure outlined in the NPHM work~\cite{nphm}. 
This procedure requires face landmarks to be known, which can be annotated manually or, as provided with the dataset in our case, calculated via 2D face landmark detectors on the projections of the colored scans and lifted to 3D. 
This way, we obtain a FLAME template, corresponding to the given scan, and subdivide it via Butterfly algorithm~\cite{butterfly_algorithm}.
We will refer to the template after subdivision as to $F = (V, \mathcal{F}, \mathcal{U}_\mathcal{F})$, where $V \in \mathrm{R}^{|V| \times 3}$ are the vertices coordinates,  $\mathcal{F} \in \mathrm{R}^{|\mathcal{F}| \times 3}$ are the corresponding faces, and $\mathcal{U}_\mathcal{F} \in \mathrm{R}^{|\mathcal{F}| \times 3 \times 2}$ are the texture coordinates of each vertex in a triangle. 
Note that using triangle coordinates instead of vertex coordinates is important due to the presence of a seam in the FLAME model, thus making UVs for the seam vertices ambiguous.

As FLAME basis does not represent hair or face details, we define these in a form of vertex displacements and learn them in two stages. 
During the first stage, we optimize the loss function $\mathcal{L}_\textrm{Stage 1} (D)$ for additive vector displacements $D_\textrm{Stage 1} \in \mathbb{R}^{|V| \times 3}$ of the vertices:
\begin{equation}
    \begin{aligned}
        \mathcal{L}(D, V, \mathcal{F}, V^\textrm{gt} | \boldsymbol{\lambda}) &= 
            \lambda^\textrm{Chamfer} \, \mathcal{L}_\textrm{Chamfer}(V+D, V^\textrm{gt}) \\
        &+   \lambda^\textrm{edge} \,\mathcal{L}_\textrm{edge}(V+D, \mathcal{F}) \\
        &+   \lambda^\textrm{lapl} \, \mathcal{L}_\textrm{lapl}(V+D, \mathcal{F})
    \end{aligned}
\end{equation}

\begin{equation}
    \mathcal{L}_\textrm{Stage 1} (D) = \mathcal{L}(D, V, \mathcal{F}, V^\textrm{gt}\, |\, \boldsymbol{\lambda}_\textrm{Stage 1}) 
\end{equation}

Hyperparameters $\boldsymbol{\lambda}_\textrm{Stage 1} = (\lambda_\textrm{Stage 1}^\textrm{Chamfer}, \lambda_\textrm{Stage 1}^\textrm{edge}, \lambda_\textrm{Stage 1}^\textrm{lapl})$ define the Chamfer matching term weight, the weight of edge length regularization and standard Laplacian regularization. 
In this stage, the weight of regularizations is high in order to prevent self-intersections that can occur when regressing vector displacements.
Also, we only optimize the vector displacements for the hair region.
%

In the second stage, we optimize the loss function $\mathcal{L}_\textrm{Stage 2} (\boldsymbol{\alpha})$ for displacements $D_\textrm{Stage 2} \in \mathbb{R}^{|V| \times 3}$ that are only allowed to move over the normals of the previously displaced vertices: 

\begin{equation}
    D_\textrm{Stage 2} = D_\textrm{Stage 1} + N \astrosun\, \boldsymbol{\alpha},
\end{equation}

\noindent where $N \in \mathbb{R}^{|V| \times 3}$ corresponds to the normals, calculated by numerical difference for vertices deformed after the Stage 1, and $\astrosun$ defines the element-wise product of rows of $N$ and elements of $\alpha$ (each normal $\boldsymbol{n}_i$ is multiplied by the respective amplitude $\alpha_i$). 
$\mathcal{L}_\textrm{Stage 2} (\boldsymbol{\alpha})$ is expressed through the same basic loss expression:
\begin{equation}
    \mathcal{L}_\textrm{Stage 2} (\boldsymbol{\alpha}) = \mathcal{L}(D_\textrm{Stage 1} + N \astrosun\, \boldsymbol{\alpha}, V, \mathcal{F}, V^\textrm{gt}\, |\, \boldsymbol{\lambda}_\textrm{Stage 2}),
\end{equation}

\noindent while hyperparameters $\boldsymbol{\lambda}_\textrm{Stage 2}$ are selected with relatively lower regularization weights. 
This allows for fitting high-frequency details while maintaining the same rough shape of the regressed shape.
At this stage, we allow both hair and face regions to deform, while subtle parts such as ears and eyeballs are fixed from moving.

Finally, we bake the displacements $ D_\textrm{Stage 2} $ into a UV map $U \in \mathbb{R}^{\textrm{H} \times \textrm{W} \times 3}$ by rendering it onto the UV space with known texture coordinates $\,\mathcal{U}_\mathcal{F}$ and triangles $\mathcal{F}$. 

The registration procedure is repeated for the dataset consisting of multiple 3D scans, resulting in a set of UV displacement maps $(U_1, \dots, U_S)$.

\subsection{Generative model}
\label{subsec:gen_model}

The described registration procedure allows us to relax the problem of 3D head geometry generation into a problem of generation of 2D UV displacement maps, which allows us to apply a 2D generative model. 
We have selected StyleGAN2 for that purpose due to its capability of generalizing over relatively small datasets of images~\cite{stylegan2-ada,diffaugment,amazing_stylegan_repo} while maintaining close-to-SoTA image generation capabilities~\cite{stylegan2}.
The model consists of a mapping network and a generator network, which we will refer to as $f(z)$ together, where $z \in \mathcal{Z} \subset \mathbb{R}^\textrm{D}$ is a latent code sampled from a standard normal distribution during training (with truncation trick~\cite{truncation_trick} at the inference time). 
The generator produces a UV displacement map $U = f(z)$, which we can apply to an arbitrary (anyhow densely subdivided) FLAME template $F = (V, \mathcal{F}, \mathcal{U}_\mathcal{F})$ by querying the map $U$ with its texture coordinates $\mathcal{U}_\mathcal{F}$ to obtain the respective vertex displacements.
The final mesh $M = (V + D(U), \mathcal{F}, \mathcal{U}_\mathcal{F})$ is different from the subdivided FLAME only in terms of the vertex locations.
We later demonstrate visually that the generated displacements could be applied to an arbitrary template.    

\mypara{UV layout.}
Importantly, we modify the UV embedding of the FLAME template mesh into a 2D plane (see the Supplementary for the illustration). 
Doing so results in a more favorable layout for the 2D generative model with a limited receptive field. As we demonstrate further, the large distance in the UV space of the head's back left and right parts leads to inconsistent generations.
The new layout allows us to stack the left- and right-hand sides into a single-channel displacement map with better seam alignment.

\mypara{Post-processing.} 
Since the UV map $U$ is generated in the UV layout that contains a seam, we expect StyleGAN to resolve it in general, i.e. produce similar displacements in the face and scalp region near the same seam vertex.
Still, there is no dedicated supervision during StyleGAN training that ensures that it always happens and that the border is preserved pixel-perfect.
Because of that, we apply Laplacian smoothing~\cite{vollmer1999improved} to the mesh $M$ in the $K$-vertex vicinity of the seam.
Additional smoothness of the face region is achieved by applying Laplacian smoothing to the facial skin, neck, scalp, eyeballs, and inner mouth region with different strength to the subdivided FLAME template in advance, before adding the displacements.
Technical details are provided in the Supplementary Material.
\begin{table}[b!]
    \vspace{-0.4cm}
    \centering
    \resizebox{.8\linewidth}{!}{
        \begin{tabular}{lccccccc}
            \toprule
                         & FID ↓          & KID ↓           & MMD ↓         & JSD ↓          & COV ↑             \\ \midrule
            Ours         & \textbf{68.00} & \textbf{0.065}  & \textbf{6.51} & 21.33 & \textbf{53.85\%} \\
            PCA          &  126.31  &  0.165       &  10.47  &  20.16         & 23.08\%          \\
            NPHM
                        & 139.82         & 0.170             &  7.80  & \textbf{19.06}         & 46.15\%          \\
            ROME
                            &  169.65  &  0.204   &  10.02  &  23.19   &  32.69\%  \\ 
            FLAME
                        & 198.85         & 0.262             &  12.95  & 23.89          & 5.77\%           \\ \bottomrule
        \end{tabular}
    }
    \caption{The comparison of quality and diversity of random samples generated by each of the methods. 
    FID and KID measure the similarity of the generated mesh renderings vs. the renderings of the ground truth meshes in FaceVerse dataset,
    while 3D metrics MMD, JSD, COV assess the similarity of generated and real point clouds distributions.
    MMD is multiplied by $10^3$ and JSD by $10^2$.
    }
    %
    \label{table:uncond}
\end{table}

\section{Experiments}

We evaluate HeadCraft's generative capabilities by examining its unconditional generation performance in Sec.~\ref{subsec:results} and ablate over several important aspects. 
Additionally, in Sec.~\ref{subsec:applications}, we demonstrate the way HeadCraft can be beneficial in several important downstream applications, such as 3D head completion from a partial point cloud, its seamless integration with the FLAME expression space, and semantic geometry transfer. Before presenting these results, we describe implementation details, as well as our chosen metrics and baselines in Sec.~\ref{subsec:implementation_details}, \ref{subsec:metrics}, and \ref{subsec:baselines}, respectively.

\subsection{Implementation details}
\label{subsec:implementation_details}

\noindent \textbf{Datasets.} 
Our method is trained on the 3D real scans of human heads from NPHM~\cite{nphm} dataset, namely, 6975 high-resolution scans of 327 diverse identities captured by two 3D scanners each.
To quantitatively evaluate the generative performance of all compared methods, we use FaceVerse~\cite{faceverse} dataset as a source of the ground truth scans not intersecting with our training data by identities.

\noindent \textbf{Registration procedure.} 
We use Adam optimizer with learning rate of $3 \cdot 10^{-2}$ for the first stage and $3 \cdot 10^{-4}$ for the second stage. 
The hyperparameters $\boldsymbol{\lambda}_\textrm{Stage 1} = (\lambda_\textrm{Stage 1}^\textrm{Chamfer},\, $ $\lambda_\textrm{Stage 1}^\textrm{edge}, \,\lambda_\textrm{Stage 1}^\textrm{lapl})$ equal to $(2 \cdot 10^3, 2 \cdot 10^5, 10^4)$. For the second stage, $\boldsymbol{\lambda}_\textrm{Stage 2} = (2 \cdot 10^4, 2 \cdot 10^4, 10^4)$.
In the Chamfer loss, we additionally apply correspondences pruning by distance of 1.0, which defines that all the correspondences between source and target with the distance more than 1.0 in the NPHM coordinate system are automatically discarded.
This has been introduced for more consistent gradual learning of displacements, such that at each optimization step, only the nearest points affect the deformation learning.

\noindent \textbf{Generative modeling.} 
As the 2D generative model, we choose StyleGAN2-ADA~\cite{stylegan2-ada} with all augmentations turned off (since they wouldn't yield valid UV maps in our case) and 8 mapping network layers. To stablize the GAN training, we utilize a high gradient penalty of 4.0 for the discriminator. The learning rates are $2 \cdot 10^{-3}$ for the generator and $1 \cdot 10^{-3}$ for the discriminator. We train it for 95K steps with the batch size of 8 and $256 \times 256$ image resolution.

\subsection{Metrics}
\label{subsec:metrics}
To evaluate the quality of the generated samples in section~\ref{subsec:results}, we rely on both 2D metrics computed on renderings, as well as, on metrics directly compute in 3D. In the following we describe all employed metrics in details.

Firstly, to evaluate the visual plausibility of the generated geometry, we render 2195 ground truth meshes from FaceVerse and the same number of meshes generated by each method with highly metallic material from eight distinct viewpoints, uniformly sampled along the circular trajectory in the horizontal plane. 
The FID~\cite{fid} and KID~\cite{kid} perceptual metrics are calculated for all generated and ground truth renderings from a given viewpoint and then macro-averaged over eight viewpoints.
Secondly, we compare the distributions of point clouds sampled from the generated and ground truth meshes.
To do that, we sample 10K points from each of the 2195 generated and the same number of ground truth meshes and calculate several 3D similarity metrics.
Jensen-Shannon Divergence (JSD) is evaluated by comparing the distributions of generated and ground truth points, splat into a voxel grid (in our case, of $512^3$ voxels).
Minimum Matching Distance (MMD) is a measure of 3D object realism that, for each ground truth sample, involves evaluating the distance to the most similar sample in the generated set.
Similarly, Coverage (COV) indicates the percentage of the ground truth samples, for which the nearest neighbor among all ground truth and generated samples falls into the generated set.
%
%
More detailed description of the 3D metrics can be found in~\cite{yang2019pointflow,erkocc2023hyperdiffusion}.

\subsection{Baselines}
\label{subsec:baselines}
We compare against the recent state-of-the-art head model\linebreak NPHM~\cite{nphm}, which uses neural SDFs to represent the head geometry. 
We additionally compare against\linebreak ROME~\cite{rome}, an alternative approach that models complete head geometry including hair using a FLAME template mesh. ROME is trained in an unsupervised fashion on the large-scale video dataset VoxCeleb2~\cite{voxceleb2}.

Furthermore, we compare against a PCA-based baseline, whereas a linear PCA basis is fitted to our UV displacement maps, and provide the numbers for random FLAME samples without added displacements as a reference. 

While NPHM, PCA baseline, and Ours have been fitted to exactly the same training dataset, for ROME, the checkpoint from the public repository has been used.

\subsection{Results}
\label{subsec:results}

\noindent \textbf{Unconditional sampling.} 
In Fig.~\ref{fig:diverse_visuals}, we compare the difference in details and diversity of the unconditional samples produced by our method to the ones produced by NPHM~\cite{nphm} and ROME~\cite{rome} methods. 
For Ours, PCA baseline and ROME, a FLAME with random shape, expression and jaw parameters are sampled from normal distribution for every head mesh, in accordance with the statistics precalculated over the NeRSemble dataset~\cite{nersemble}.
For ROME, we sample the FLAME displacements from a linear model provided by the authors of ROME as the sampling strategy proposed by the ROME authors. 
Visually, we observe both higher diversity and better representation of details than for all baselines.
The details of the facial region are generally the sharpest for ROME, PCA baseline, and Ours, due to the use of the FLAME template.
In Table~\ref{table:uncond}, we also quantify the level of detail and variety of the generated meshes w.r.t.~the full head scans from the FaceVerse dataset~\cite{faceverse} that has not been used for training.
In addition, we demonstrate how much the generated samples deviate from the NPHM training set in Fig.~\ref{fig:memorization}.
\textit{Nearest} is found by comparing the generated displacement map to the maps of registered ground truth displacements for all training scans by L2 distance over the scalp.

The results in Table~\ref{table:uncond} indicate that the renderings from our method appear more realistic than of the other methods, with either PCA baseline or NPHM performing similar according to different subsets of the metrics.
Close promixity to NPHM by MMD, COV, JSD could be explained by training on exactly the same dataset.

\newcommand{\memorizationTablePicSize}{0.15\linewidth}

\begin{table}[]
    \centering
    \setlength{\tabcolsep}{0pt}
    \begin{tabular}{cccccccc}
    {\small Generated} & {\small Nearest} & \hspace{0.3cm} & {\small Generated} & {\small Nearest} & \hspace{0.3cm} & {\small Generated} & {\small Nearest}\\
    \includegraphics[width=\memorizationTablePicSize,trim={4cm 1cm 0.5cm 1cm},clip]{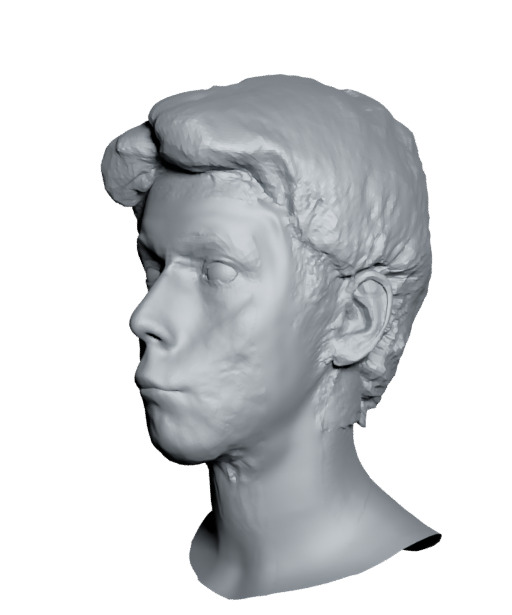} 
    & \includegraphics[width=\memorizationTablePicSize,trim={4cm 1cm 0.5cm 1cm},clip]{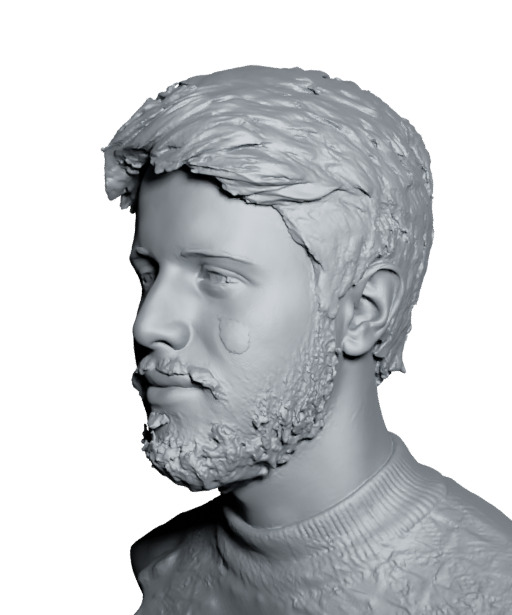}
    & \hspace{0.3cm}
    & \includegraphics[width=\memorizationTablePicSize,trim={4cm 1cm 0.5cm 1cm},clip]{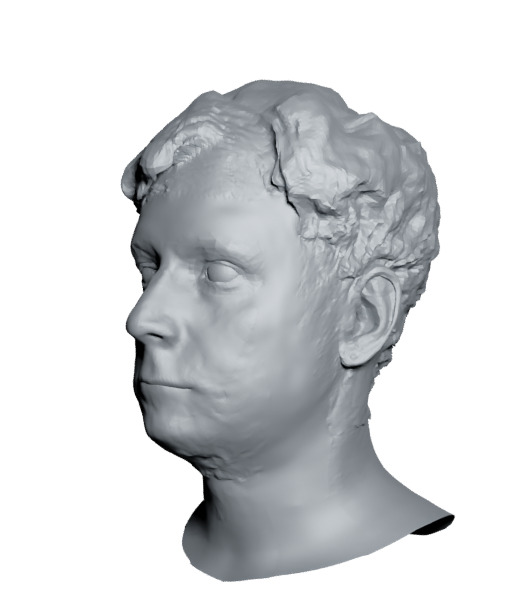}  
    & \includegraphics[width=\memorizationTablePicSize,trim={4cm 1cm 0.5cm 1cm},clip]{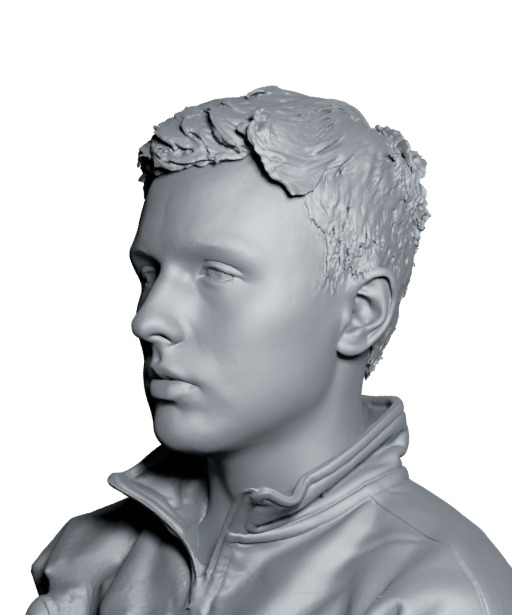} 
    & \hspace{0.3cm}
    & \includegraphics[width=\memorizationTablePicSize,trim={4cm 1cm 0.5cm 1cm},clip]{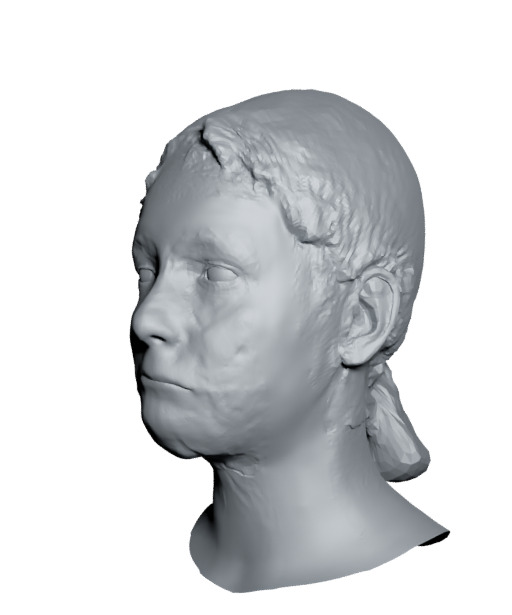}  
    & \includegraphics[width=\memorizationTablePicSize,trim={4cm 1cm 0.5cm 1cm},clip]{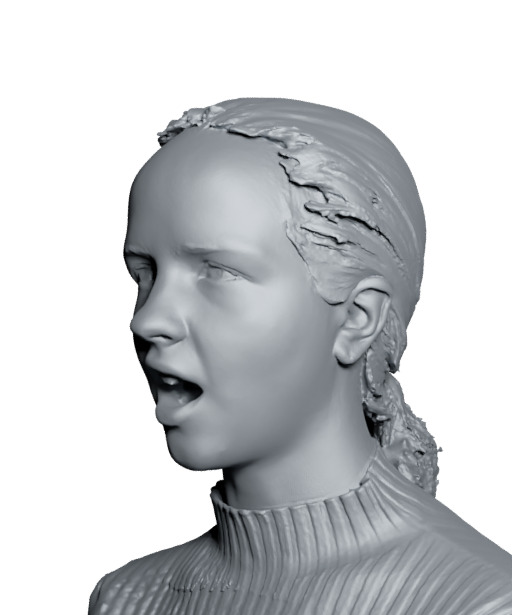} 
    \end{tabular}
    \vspace{-0.3cm}
    \captionof{figure}{
    Randomly generated samples from HeadCraft and the corresponding nearest neighbors in the NPHM dataset among the scans used for training.
    $L_2$ distance over the scalp part of the displacement maps was used.
    Displacements were added to a random FLAME template for all samples.
    }
    \label{fig:memorization}
    \vspace{-0.4cm}
\end{table}

\mypara{Ablating over the choice of the generative model architecture.}
We compare StyleGAN to other state-of-the-art generative model architectures, namely of VAE~\cite{kingma2019introduction} and VQ-VAE~\cite{van2017neural} family, with ResNet-18 encoder and decoder. 
For VQ-VAE, the sampling from the latent space is implemented via training PixelCNN autoregressive model~\cite{van2016pixel}.
The results are presented in Table~\ref{table:arch_ablation} and can be visually assessed in the Supplementary.

\begin{table}[b!]
    \centering
    \vspace{-0.1cm}
    \resizebox{.8\linewidth}{!}{
        \begin{tabular}{lccccccc}
        \toprule
              & FID ↓          & KID ↓          & MMD ↓         & JSD ↓          & COV ↑             \\ \midrule
        Ours         & \textbf{68.00} & \textbf{0.065}  & \textbf{6.51} & 21.33 & \textbf{53.85\%} \\
        SG $\rightarrow$ VAE
                    &  112.03     &   0.130      &  6.78    & 21.67  &   47.12\%               \\
        SG $\rightarrow$ VQ-VAE
                    & 124.17 & 0.151  &  7.15   &    21.93  &  43.27\%     \\
        PCA          &  126.31  &  0.165     &  10.47  &  \textbf{20.16}         & 23.08\%          \\ \bottomrule 
        \end{tabular}
    }
    \caption{Ablation over the generative model design. 
    VAE and VQ-VAE follow the ResNet-18 encoder and decoder architecture, while \textit{Ours} is based on StyleGAN2.
    We also include PCA baseline scores here as a reference.
    }
    \label{table:arch_ablation}
    \vspace{0.1cm}
\end{table}
\newcommand{\registrationProcessTablePicSize}{0.23\linewidth}

\begin{table}[h!]
    \centering
    \setlength{\tabcolsep}{0pt}
    \begin{tabular}{cccc}
         \hspace{0.15cm} (a) Stage 1 \hspace{0.15cm} &  \hspace{0.15cm} (b) Stage 2 \hspace{0.15cm}  & \hspace{0.15cm} (c) Stage 1, \hspace{0.15cm}   & \hspace{0.4cm}  (d) Ours \hspace{0.2cm}  \\
        only & only & $\boldsymbol{\lambda}=\boldsymbol{\lambda}_\textrm{Stage 2}$ &  \\
        \includegraphics[width=\registrationProcessTablePicSize,trim={2cm 9cm 0cm 9cm},clip]{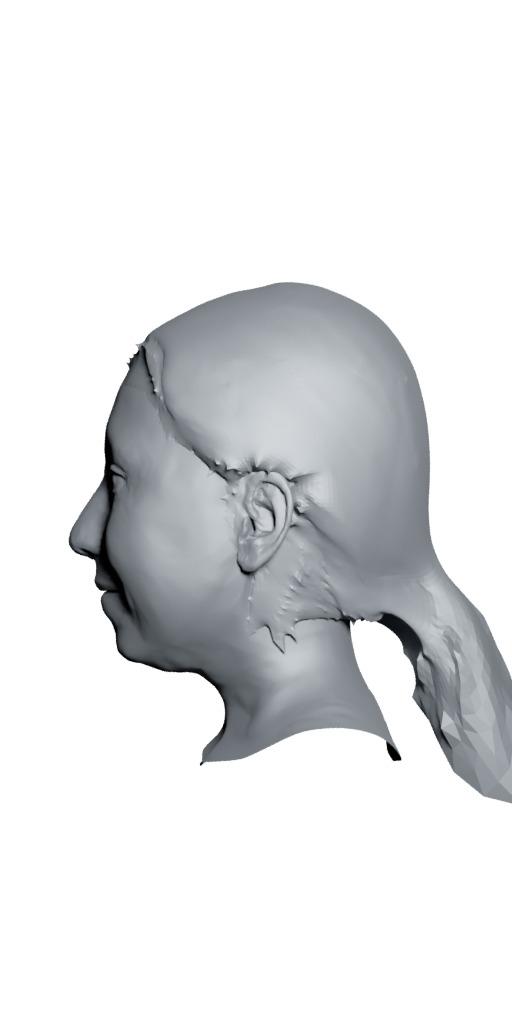} 
        & \includegraphics[width=\registrationProcessTablePicSize,trim={2cm 9cm 0cm 9cm},clip]{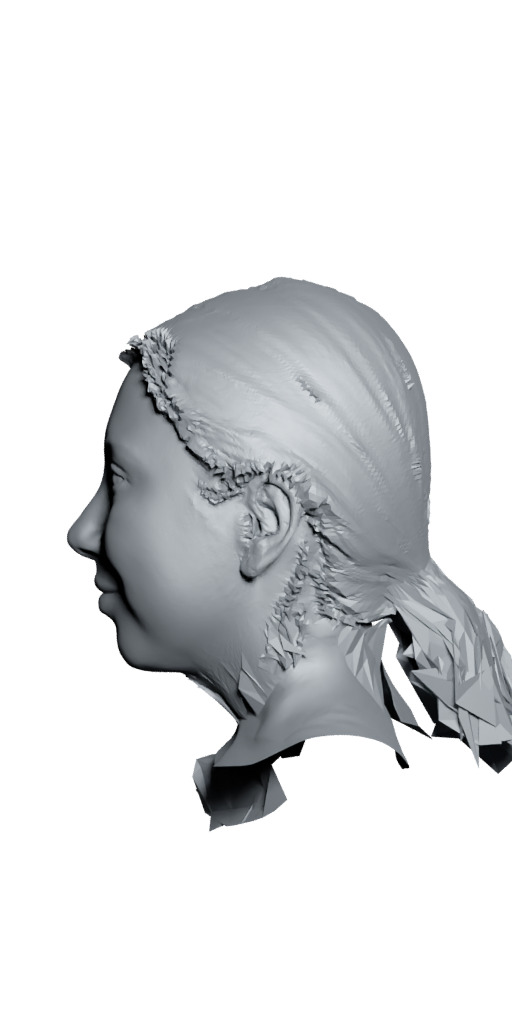}
        & \includegraphics[width=\registrationProcessTablePicSize,trim={2cm 9cm 0cm 9cm},clip]{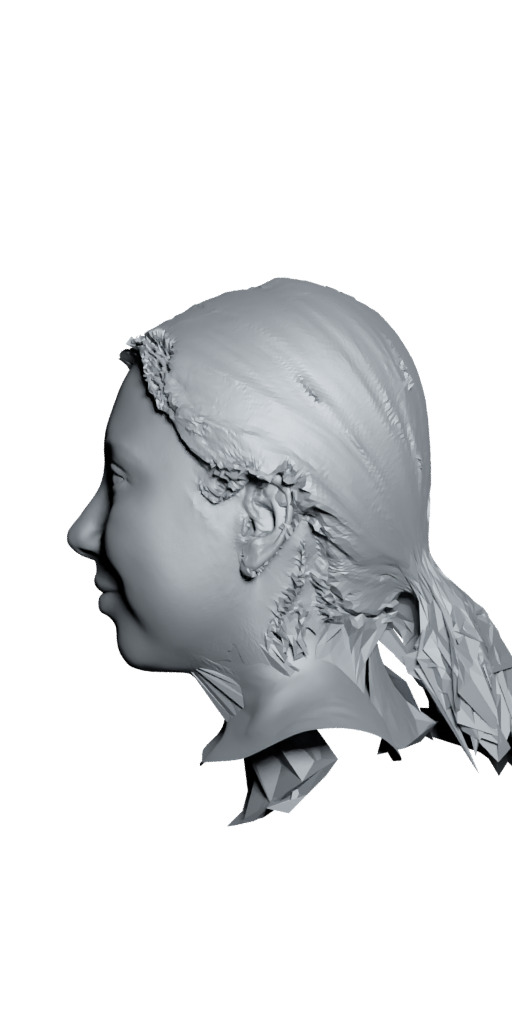}
        & \includegraphics[width=\registrationProcessTablePicSize,trim={2cm 9cm 0cm 9cm},clip]{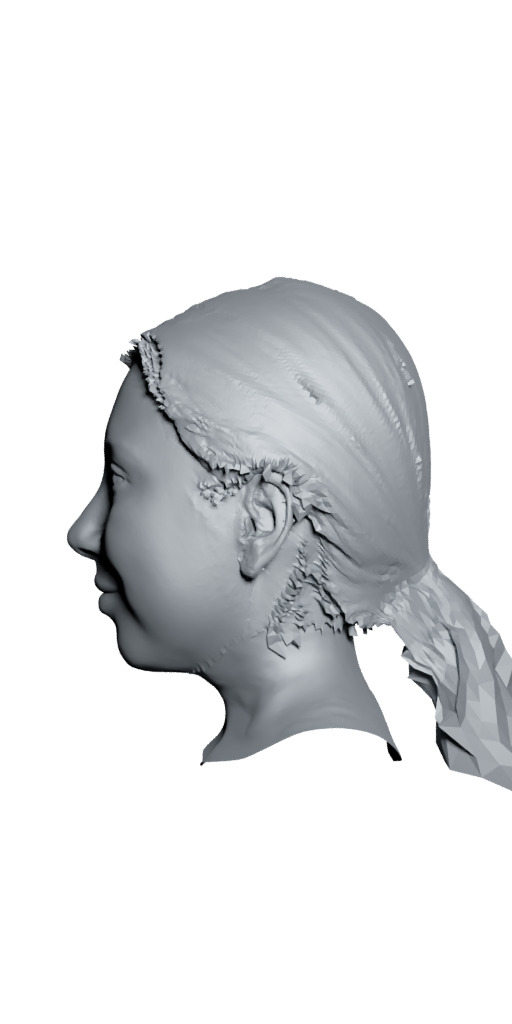}
        \\
        \fbox{\includegraphics[width=0.16\linewidth,trim={10cm 10cm 2cm 21cm},clip]{fig/4_experiments/registration_process/vector_only/step_4000.jpg}}
        & \fbox{\includegraphics[width=0.16\linewidth,trim={10cm 10cm 2cm 21cm},clip]{fig/4_experiments/registration_process/normal_only/step_4000.jpg}}
        & \fbox{\includegraphics[width=0.16\linewidth,trim={10cm 10cm 2cm 21cm},clip]{fig/4_experiments/registration_process/vector_lambda2/step_4000.jpg}}
        & \fbox{\includegraphics[width=0.16\linewidth,trim={10cm 10cm 2cm 21cm},clip]{fig/4_experiments/registration_process/ours/step_4000.jpg}}  
        \\
    \end{tabular}
    \vspace{-0.1cm}
    \captionof{figure}{
    Ablation over the one-stage vs. two-stage registration.
    Regressing only vector displacements (a) yields too smooth geometry, and learning them only along the normals (b) introduces spikes -- just like running the first stage with smaller $\boldsymbol{\lambda}$ (c).
    }
    \label{fig:registration_progress}
    \vspace{-0.2cm}
\end{table} 

\mypara{Behavior of the registration procedure.} 
In Fig.~\ref{fig:registration_progress}, we demonstrate the advantage of the two-stage registration procedure, described in Subsec.~\ref{subsec:flame_reg}, over omitting one of the stages.
As can be seen, keeping only the vector displacement optimization results in too rough shape, and relaxing the regularization constraints yields significant artifacts such as self-intersections and spikes.
Running the normal displacement stage without any preliminary vector displacement stage performs similarly to our two-stage procedure but produces artifacts for long hair that does not trivially project onto the surface.
In turn, it can produce the mappings between template vertices and scan vertices, inconsistent across various samples for the long hair parts.

\begin{figure}[b!]
    \vspace{-0.5cm}
    \includegraphics[width=\linewidth,trim={0cm 6cm 0cm 0cm},clip]{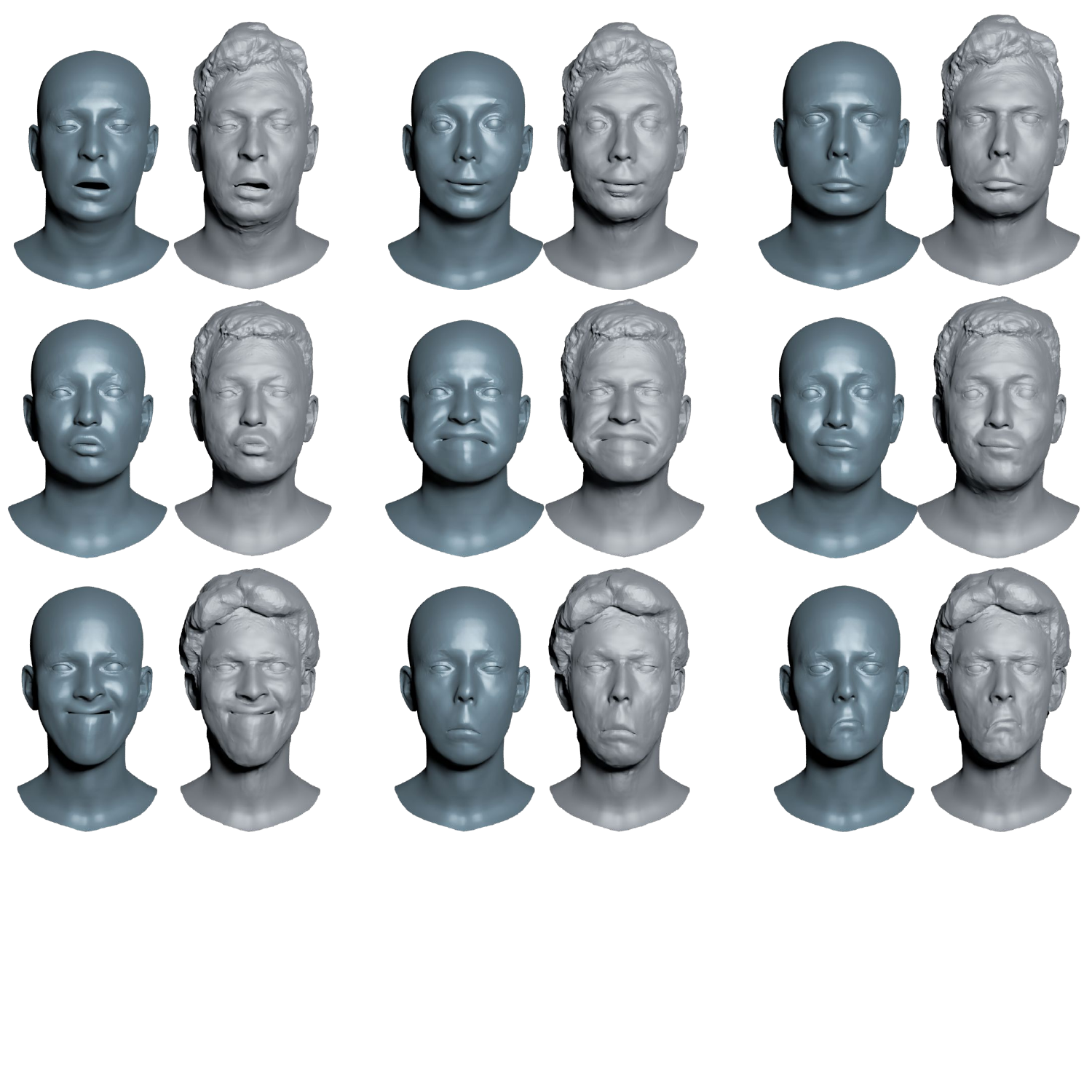}
    \vspace{-0.5cm}
    \caption{Demonstration of the animation capabilities of the model.
    Each of the sequences is created by adding the randomly generated displacements from HeadCraft to the FLAME template with varying extreme expression parameters.
    }
    \label{fig:animation}
    \vspace{-0.3cm}
\end{figure}

\subsection{Applications}
\label{subsec:applications}

\noindent \textbf{Fitting the latent code to a depth map.}
Our model can act as a prior for completing the partial observations, e.g. when they come from a depth sensor.
To evaluate the performance of the model in that scenario, we demonstrate the completion capabilities of the model over a number of scans from NPHM corresponding to the subjects unseen during training.
For each of these scans, we project their depth onto random viewpoints in the frontal hemisphere and project it back to 3D to construct partial point clouds.
To obtain a partial UV map to be completed, we run our registration procedure with a few modifications to fit a part of the scan.
Namely, we only fit the points within the convex hull of the partial point cloud, apply stronger edge length regularization weight, and constrain the points at the border of the allowed region from moving.
The final mask of observed UV texels is refined by only selecting those points that turn out to be close to the partial point cloud.
Finally, a latent code of HeadCraft explaining the partial UV map is found via StyleGAN inversion techniques.
More technical details of the partial registration and inversion are provided in the Supplementary.
The fitting quality can be evaluated by the visual comparison in Fig.~\ref{fig:fitting_result}. 

\begin{table*}[h!]
    \centering
    \renewcommand{\arraystretch}{0}
    \resizebox{.9\linewidth}{!}{
        \setlength{\tabcolsep}{0pt}
        \begin{tabular}{cccccp{0.9cm}ccccc}
            Depth & \multicolumn{2}{c}{Completion} & \multicolumn{2}{c}{Ground truth} &  & Depth & \multicolumn{2}{c}{Completion} & \multicolumn{2}{c}{Ground truth} \\
            \includegraphics[width=0.08\textwidth,trim={10cm 3cm 9.5cm 2cm},clip]{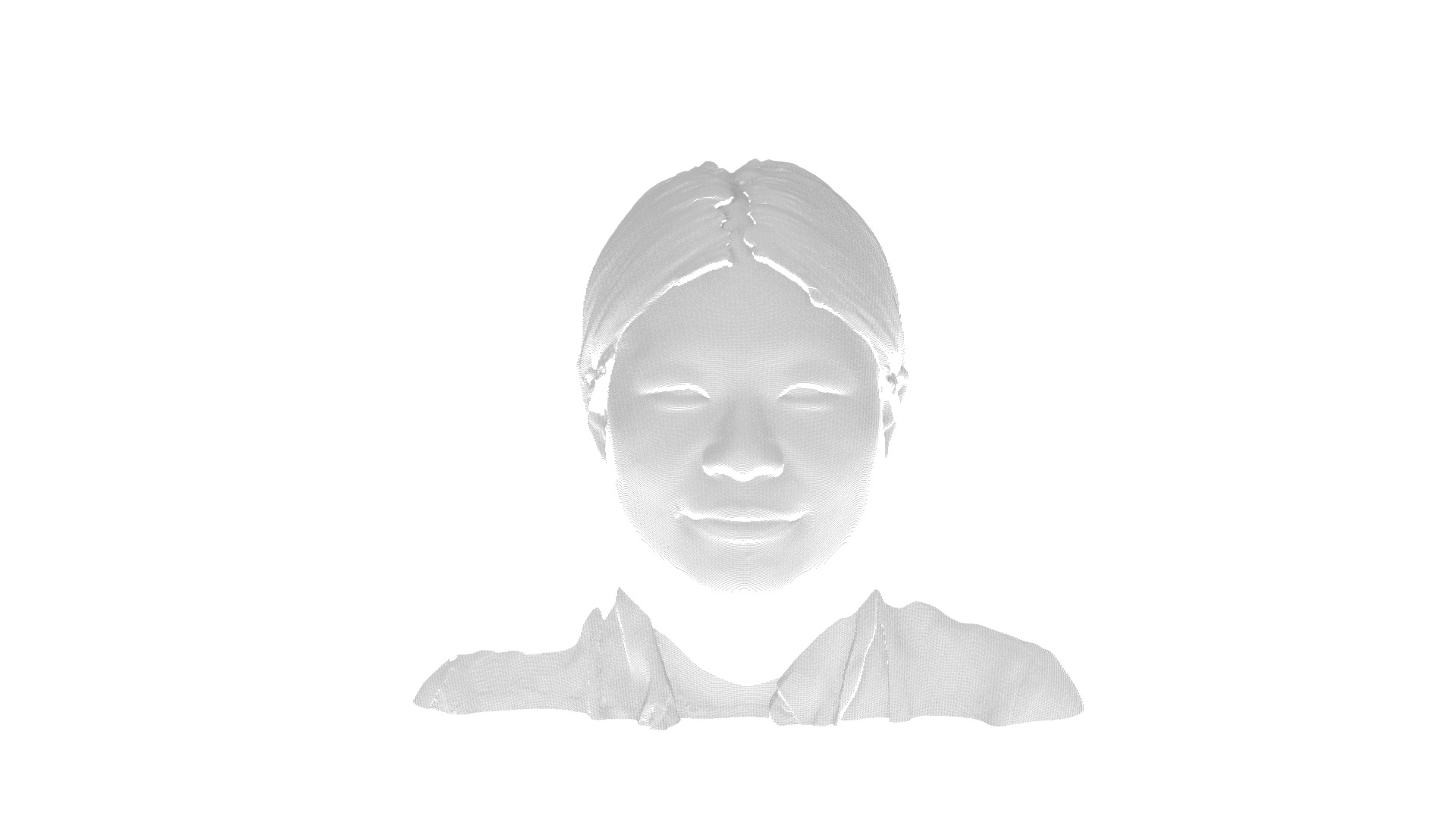}
            & \includegraphics[width=0.09\textwidth,trim={3cm 4cm 3cm 1cm},clip]{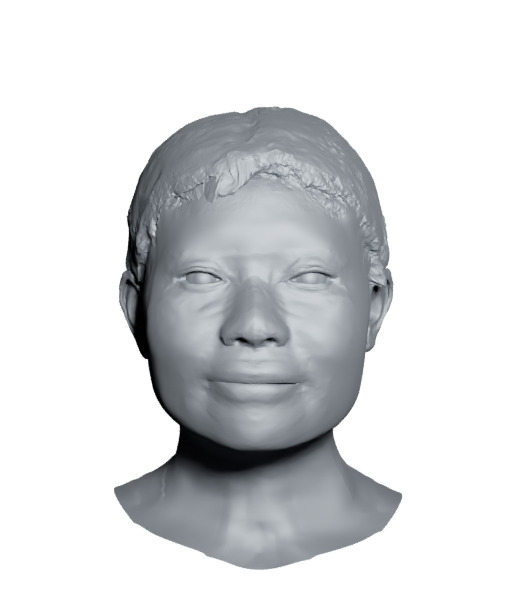}   
            & \includegraphics[width=0.11\textwidth,trim={0cm 2cm 1cm 1cm},clip]{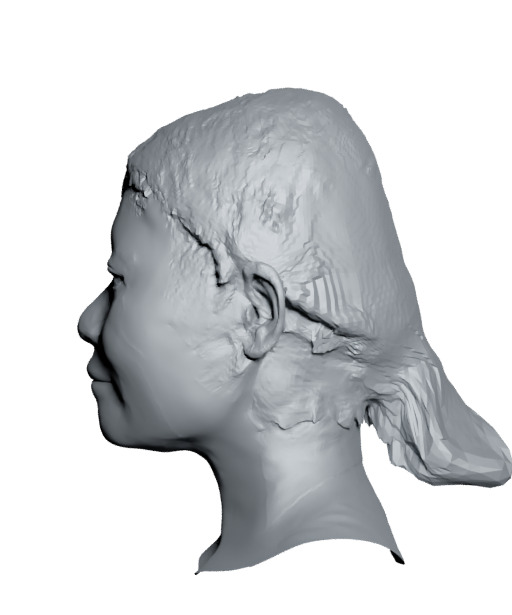}  
            & \includegraphics[width=0.09\textwidth,trim={3cm 4cm 3cm 1cm},clip]{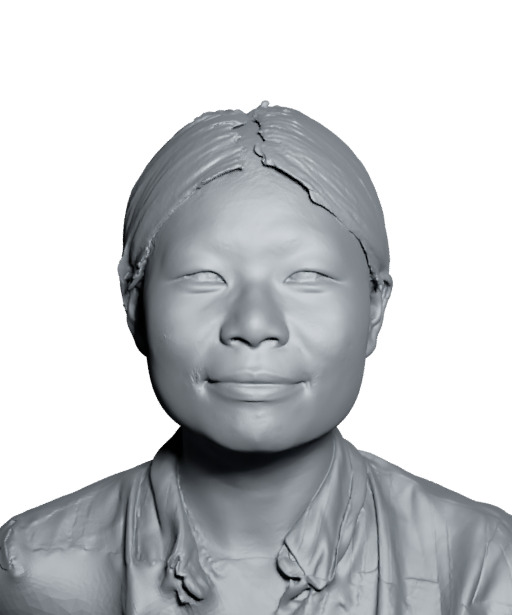}
            & \includegraphics[width=0.11\textwidth,trim={0cm 2cm 1cm 1cm},clip]{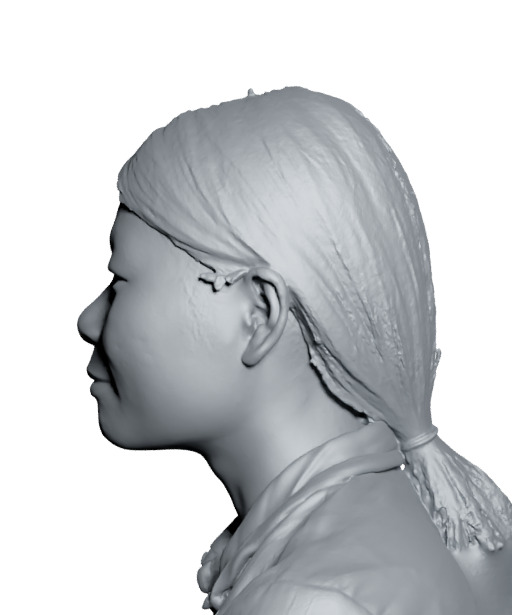}   
            & 
            & \includegraphics[width=0.08\textwidth,trim={10cm 3cm 9.5cm 2cm},clip]{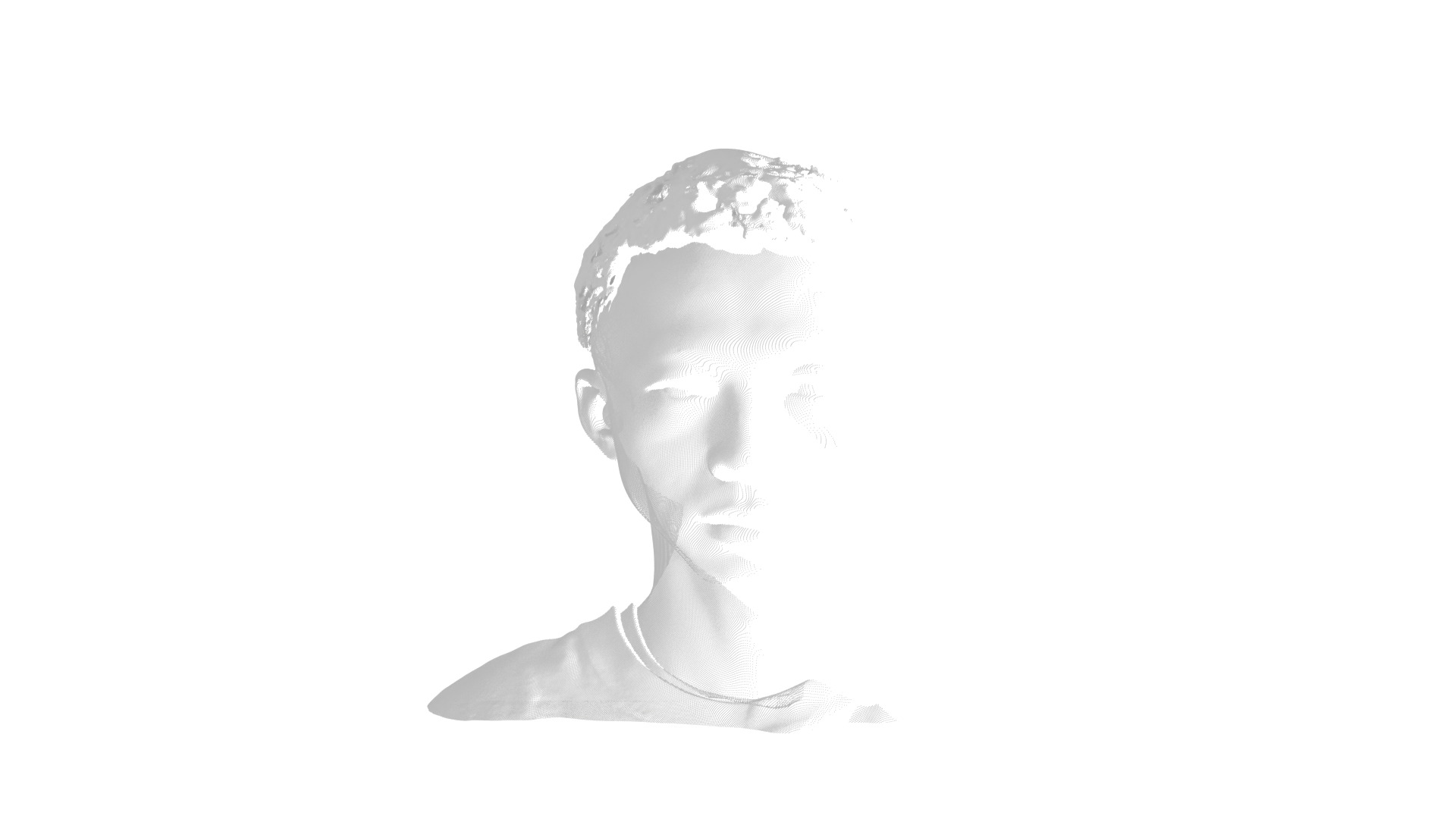}
            & \includegraphics[width=0.09\textwidth,trim={3cm 4cm 3cm 1cm},clip]{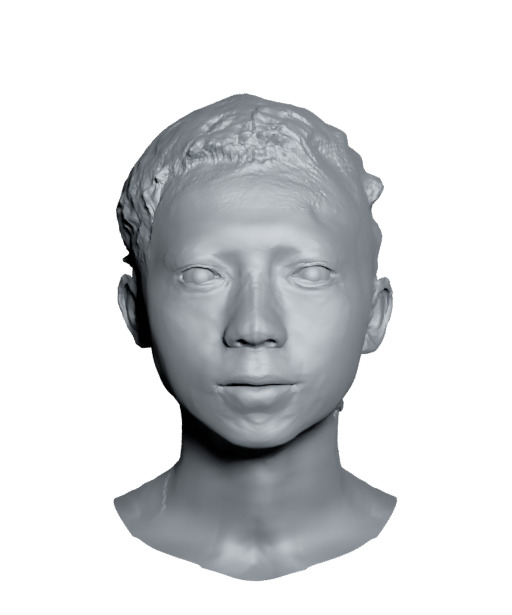}   
            & \includegraphics[width=0.11\textwidth,trim={0cm 2cm 1cm 1cm},clip]{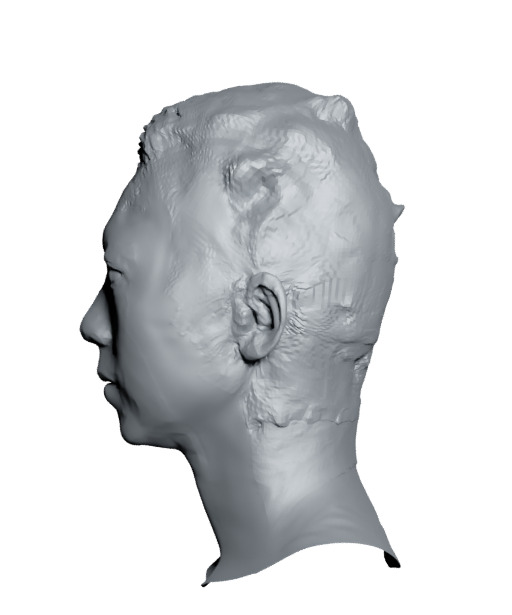}  
            & \includegraphics[width=0.09\textwidth,trim={3cm 4cm 3cm 1cm},clip]{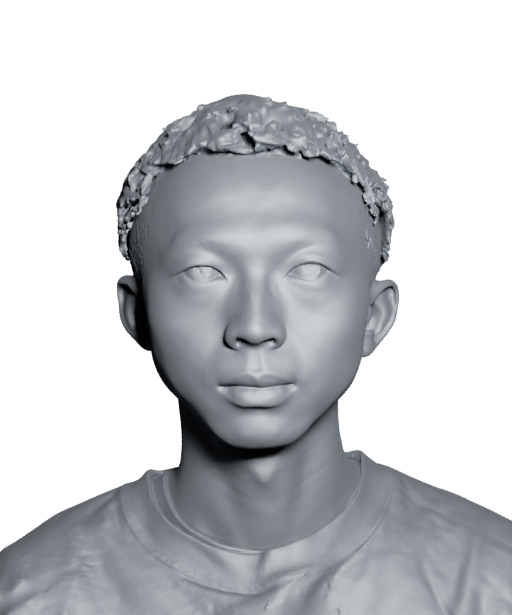}
            & \includegraphics[width=0.11\textwidth,trim={0cm 2cm 1cm 1cm},clip]{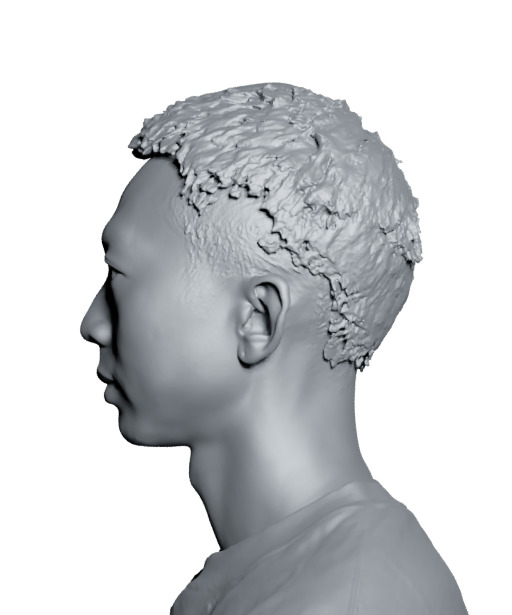}   
            \\
            \includegraphics[width=0.08\textwidth,trim={10cm 3cm 9.5cm 2cm},clip]{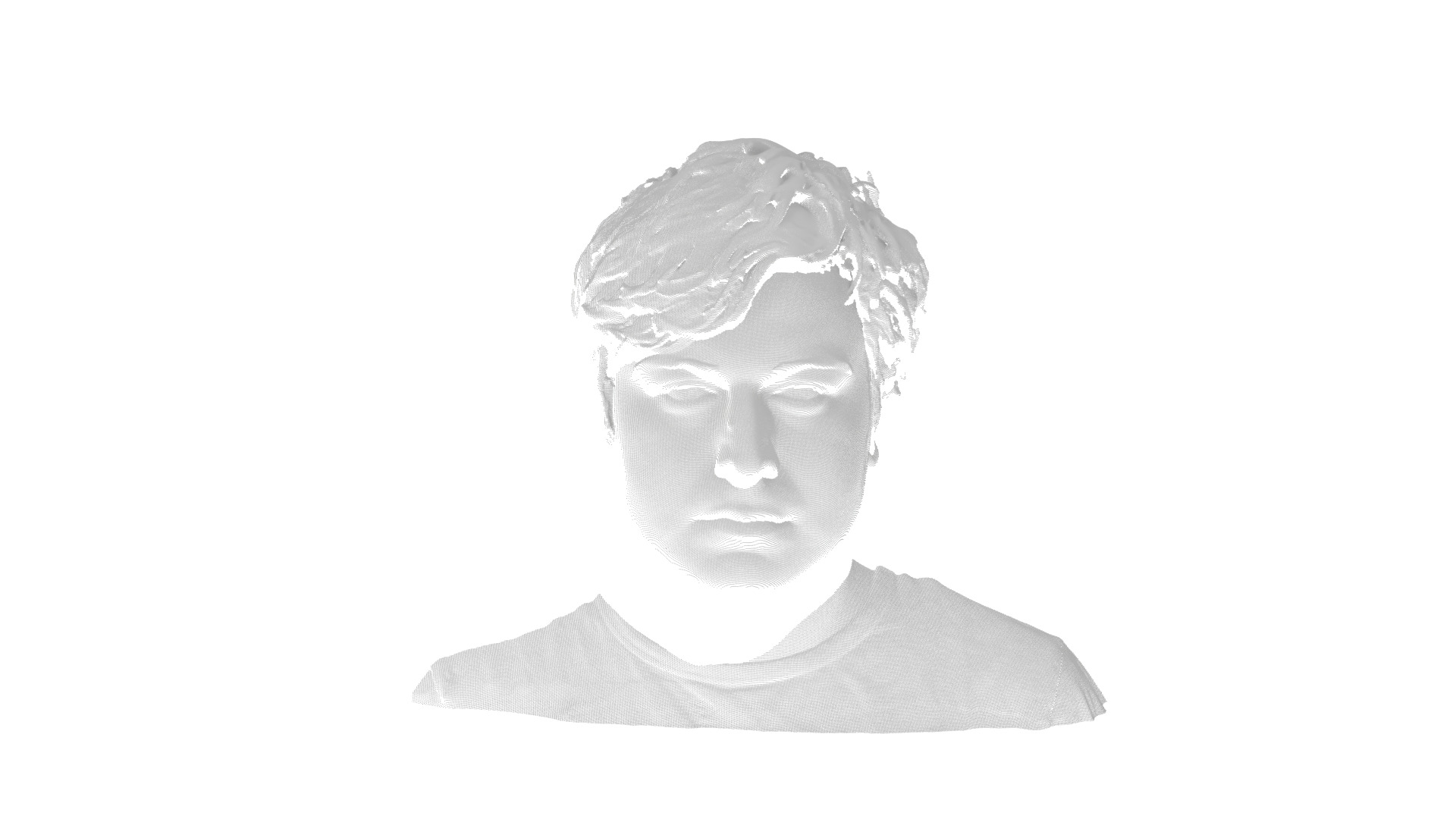}
            & \includegraphics[width=0.09\textwidth,trim={3cm 4cm 3cm 1cm},clip]{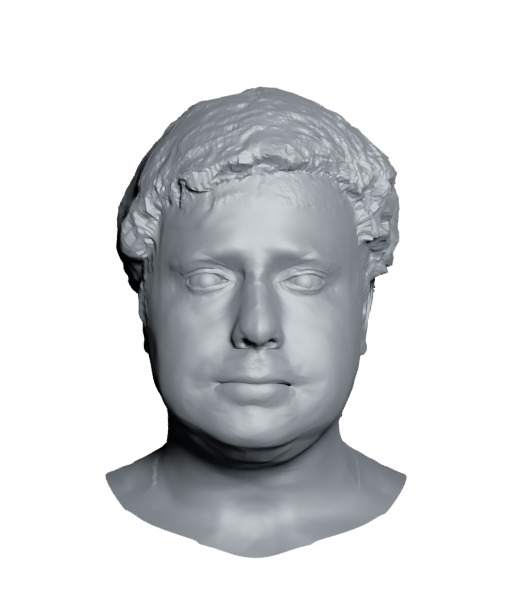}   
            & \includegraphics[width=0.11\textwidth,trim={0cm 2cm 1cm 1cm},clip]{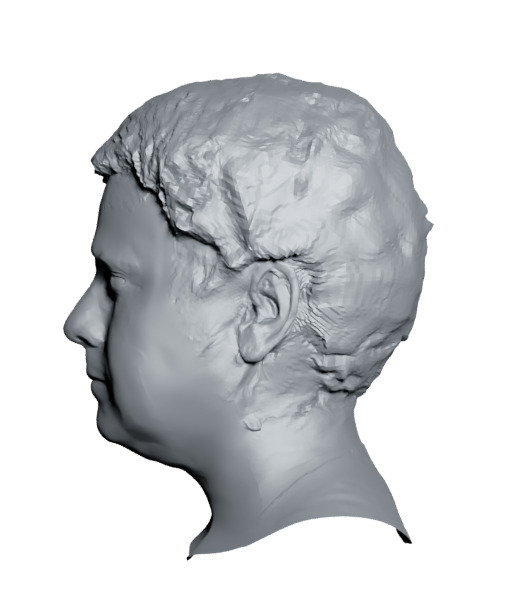}  
            & \includegraphics[width=0.09\textwidth,trim={3cm 4cm 3cm 1cm},clip]{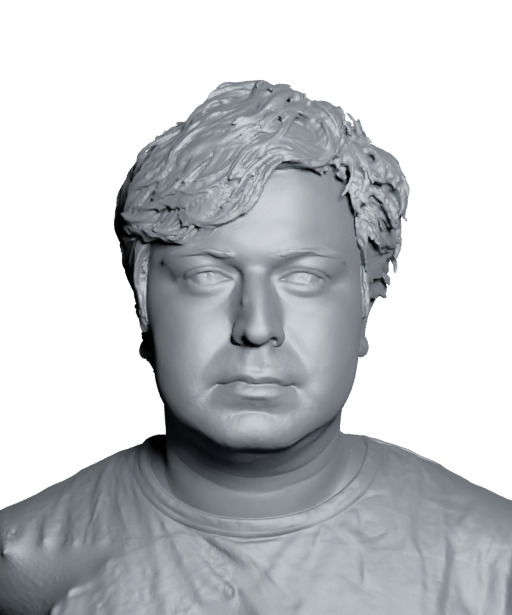}
            & \includegraphics[width=0.11\textwidth,trim={0cm 2cm 1cm 1cm},clip]{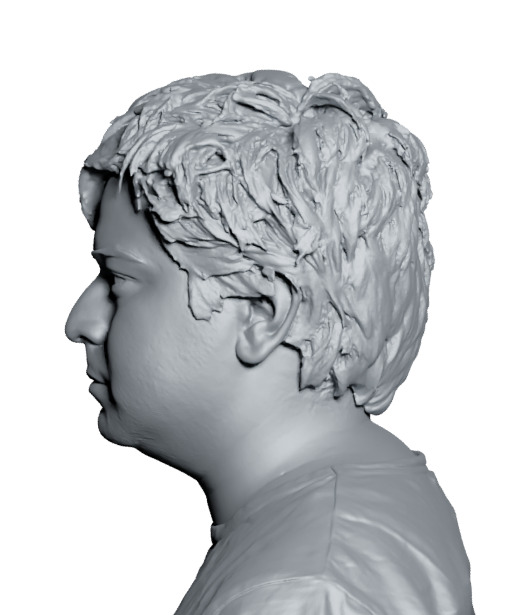}   
            & 
            & \includegraphics[width=0.08\textwidth,trim={10cm 3cm 9.5cm 2cm},clip]{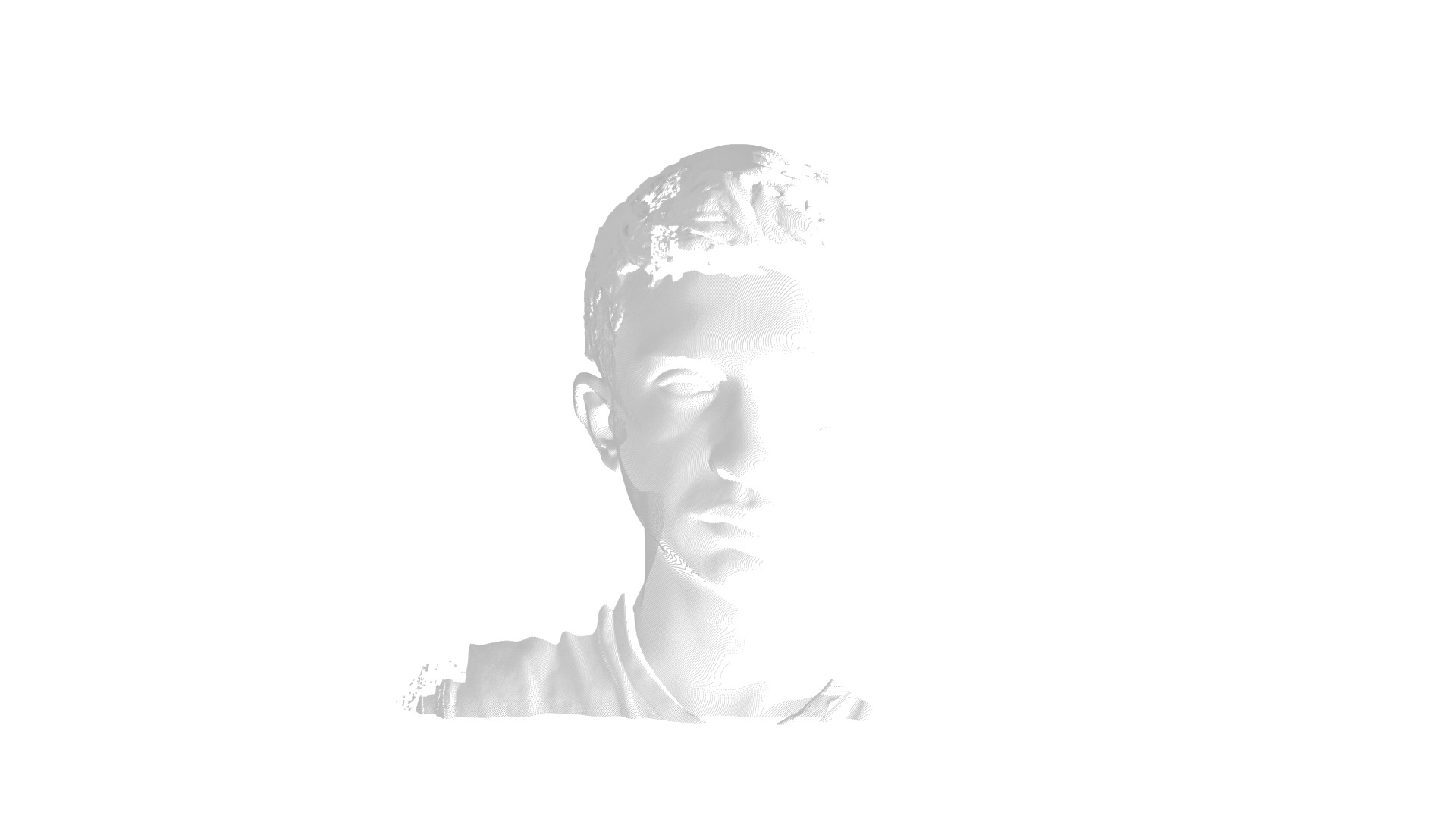}
            & \includegraphics[width=0.08\textwidth,trim={3cm 4cm 3cm 1cm},clip]{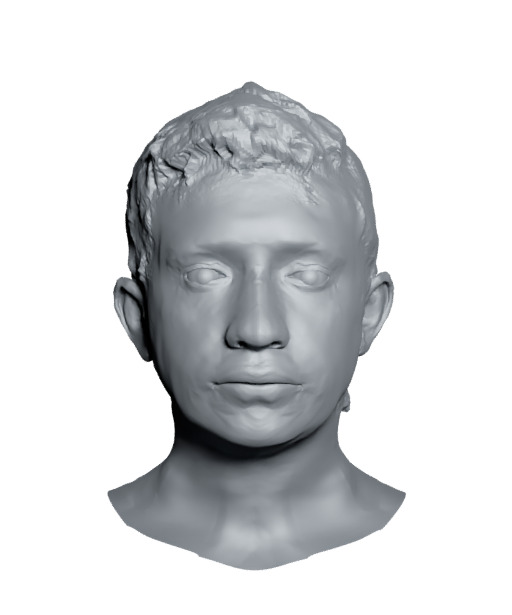}   
            & \includegraphics[width=0.1\textwidth,trim={0cm 2cm 1cm 1cm},clip]{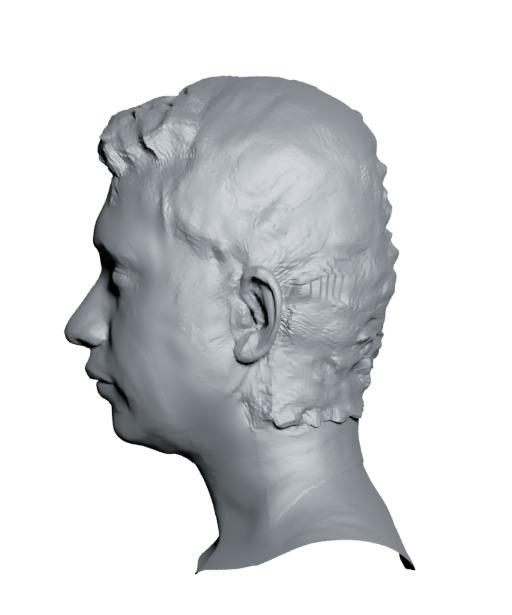}  
            & \includegraphics[width=0.08\textwidth,trim={3cm 4cm 3cm 1cm},clip]{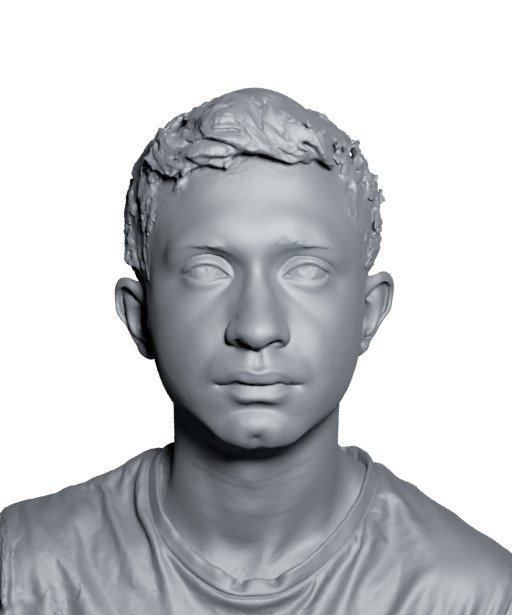}
            & \includegraphics[width=0.1\textwidth,trim={0cm 2cm 1cm 1cm},clip]{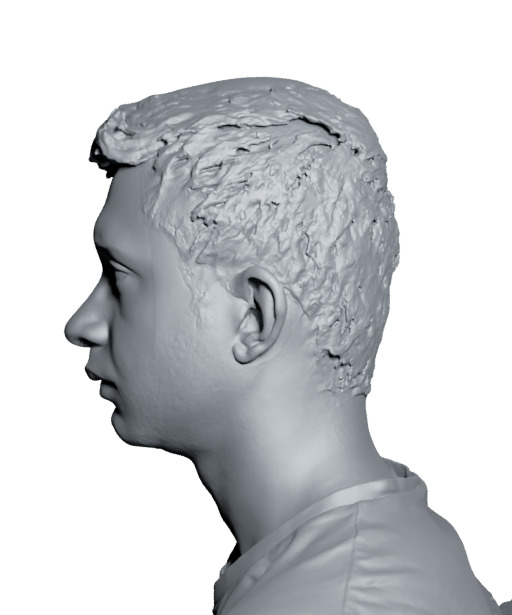}   
            \\[0.2cm] \midrule
            %
            Depth & Ours & NPHM & Gr.~truth &  &  &  & Depth & Ours & NPHM & Gr.~truth  \\
            \includegraphics[width=0.08\textwidth,trim={10cm 3cm 9.5cm 2cm},clip]{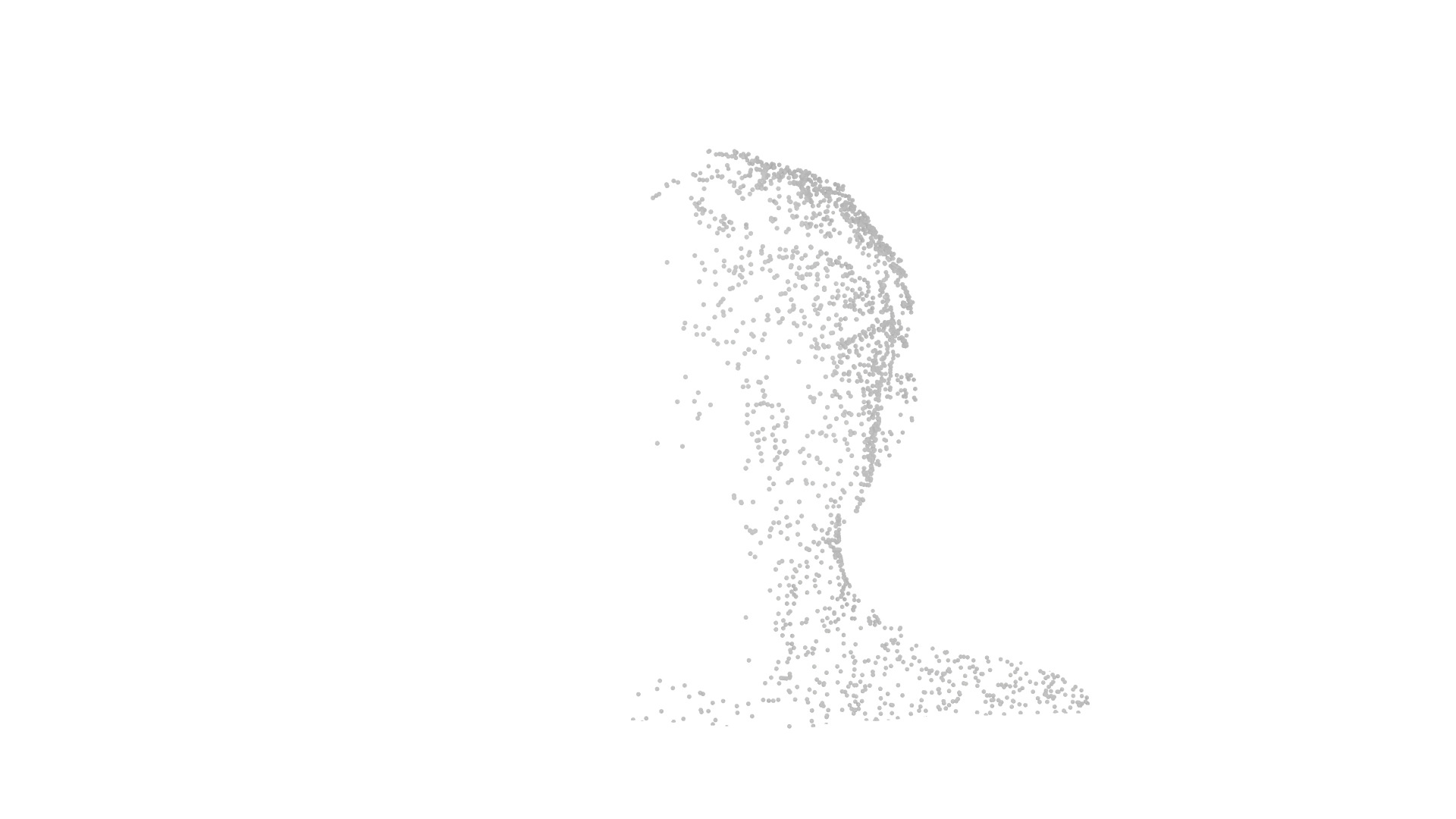}
            & \includegraphics[width=0.11\textwidth,trim={0cm 2cm 1cm 1cm},clip]{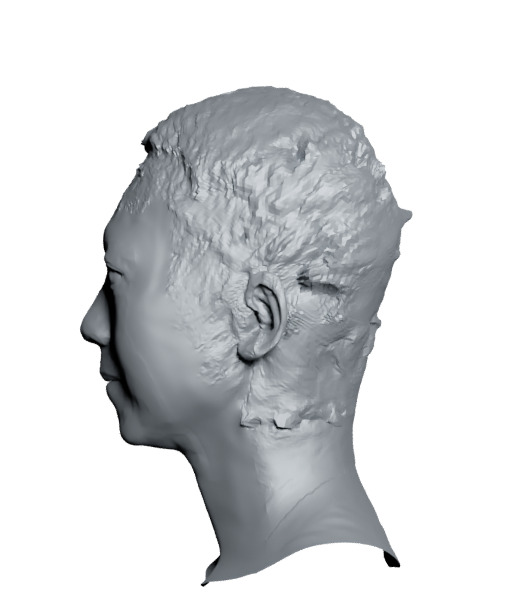}  
            & \includegraphics[width=0.11\textwidth,trim={0cm 2cm 1cm 1cm},clip]{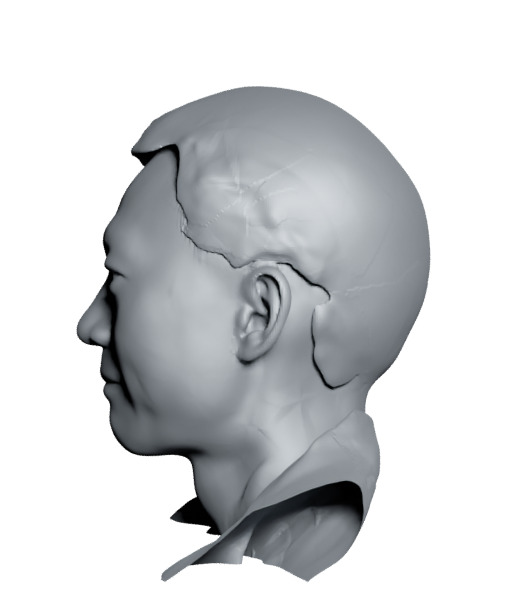} 
            & \includegraphics[width=0.11\textwidth,trim={0cm 2cm 1cm 1cm},clip]{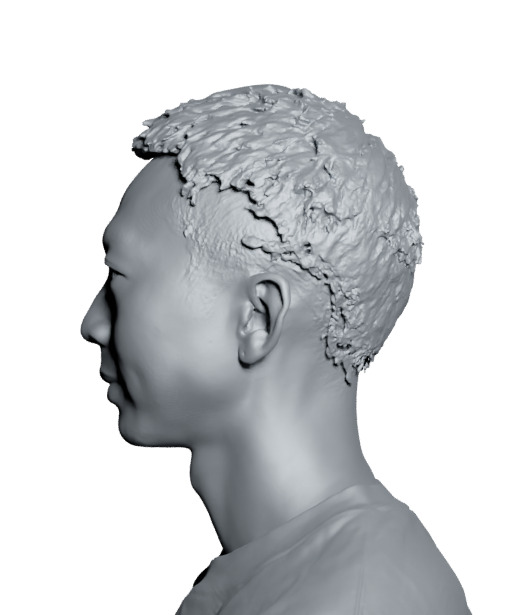}
            &
            &
            & 
            & \includegraphics[width=0.08\textwidth,trim={10cm 3cm 9.5cm 2cm},clip]{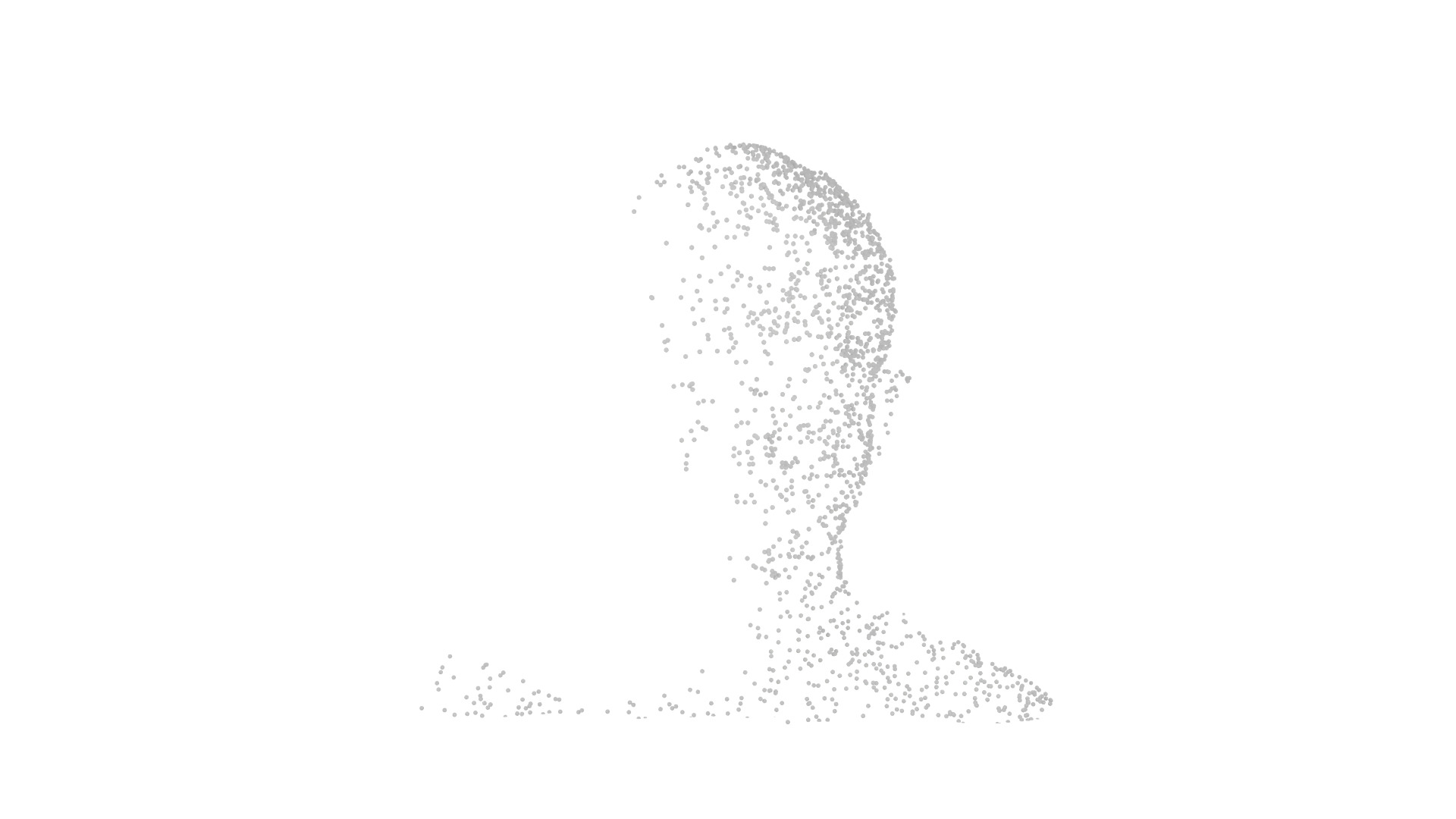}
            & \includegraphics[width=0.11\textwidth,trim={0cm 2cm 1cm 1cm},clip]{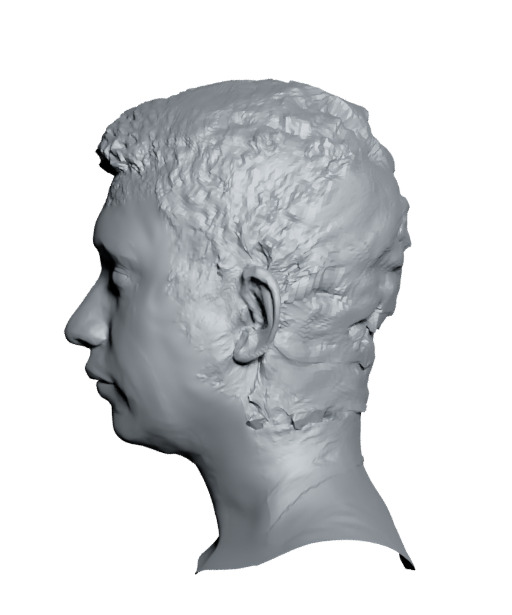}  
            & \includegraphics[width=0.11\textwidth,trim={0cm 2cm 1cm 1cm},clip]{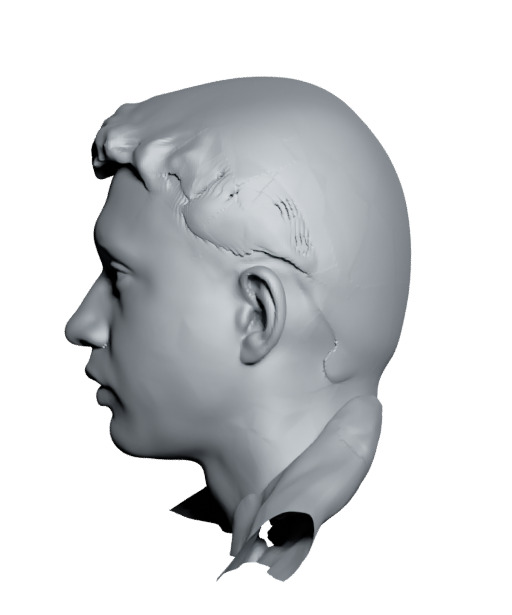} 
            & \includegraphics[width=0.11\textwidth,trim={0cm 2cm 1cm 1cm},clip]{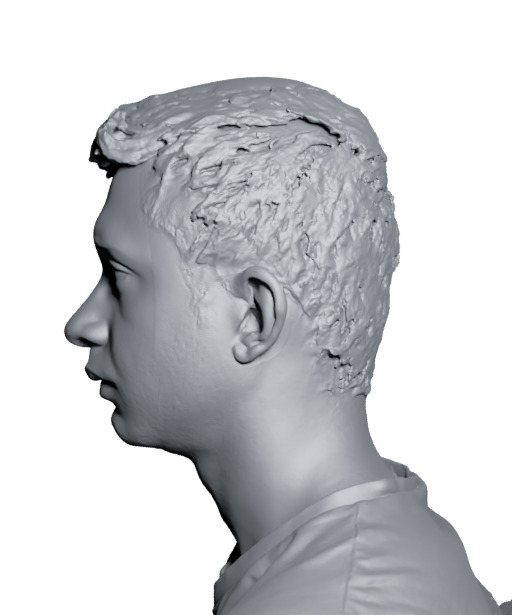} \\
            \includegraphics[width=0.08\textwidth,trim={10cm 3cm 9.5cm 2cm},clip]{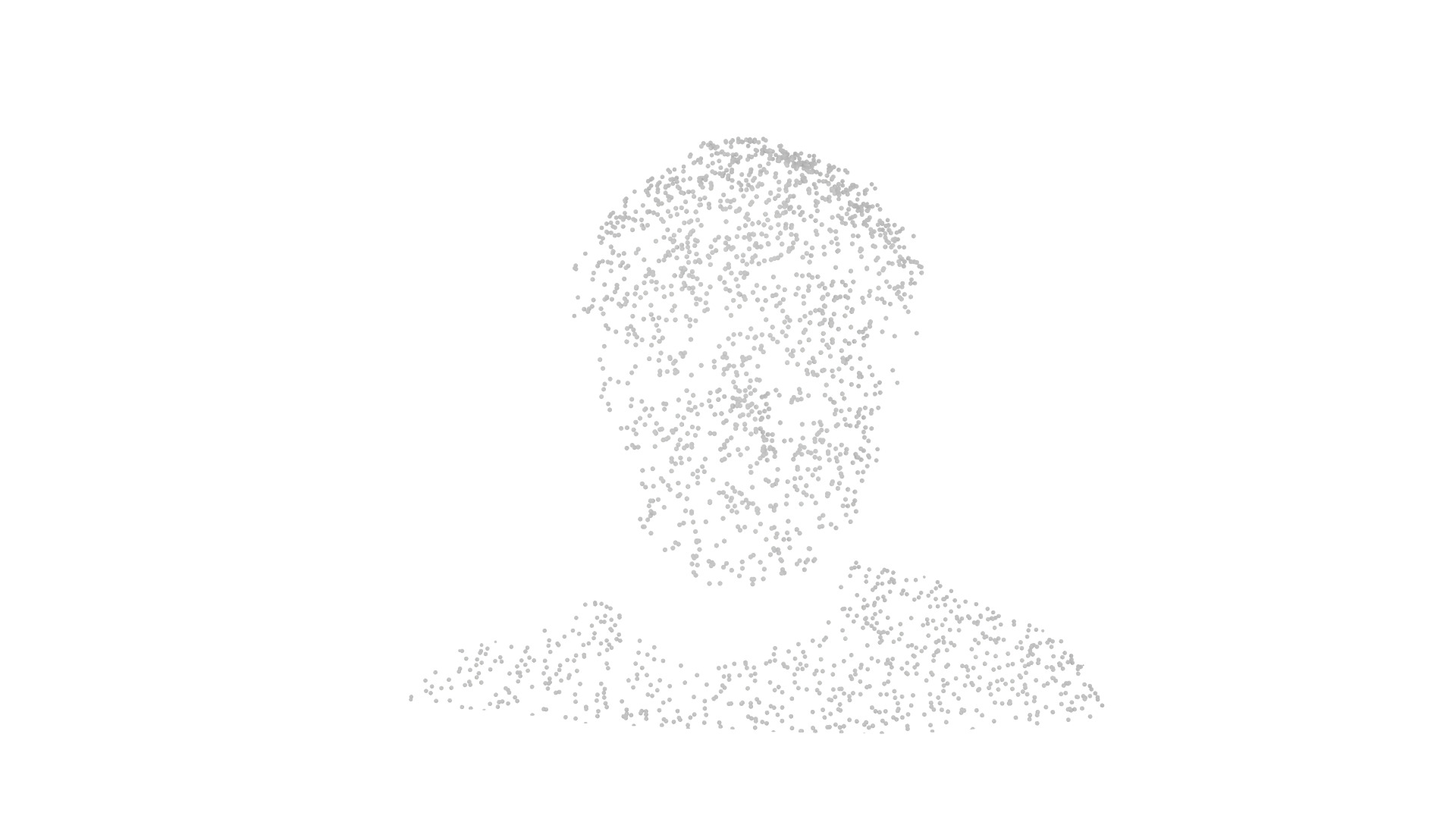}
            & \includegraphics[width=0.11\textwidth,trim={0cm 2cm 1cm 1cm},clip]{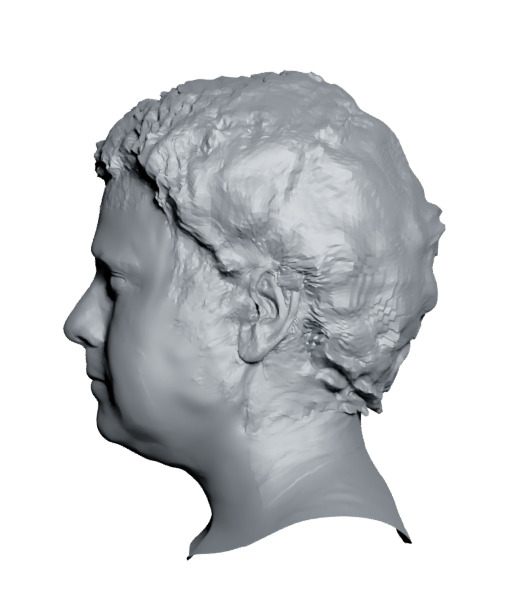}  
            & \includegraphics[width=0.11\textwidth,trim={0cm 2cm 1cm 1cm},clip]{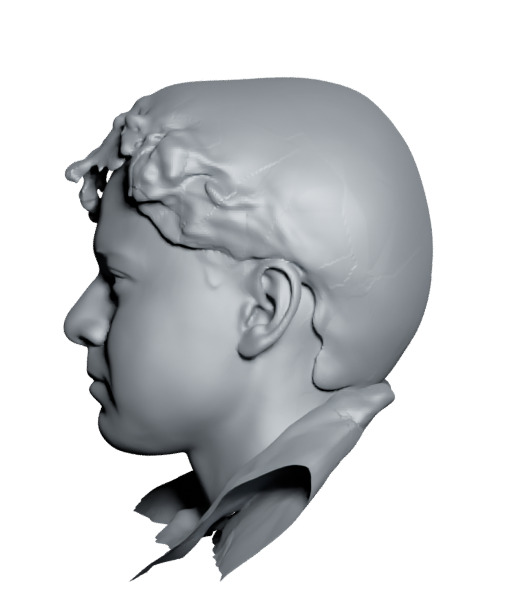} 
            & \includegraphics[width=0.11\textwidth,trim={0cm 2cm 1cm 1cm},clip]{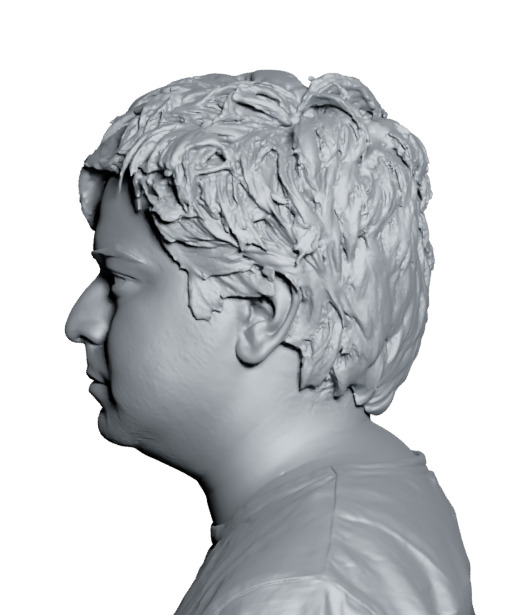}
            &
            &
            & 
            & \includegraphics[width=0.08\textwidth,trim={10cm 3cm 9.5cm 2cm},clip]{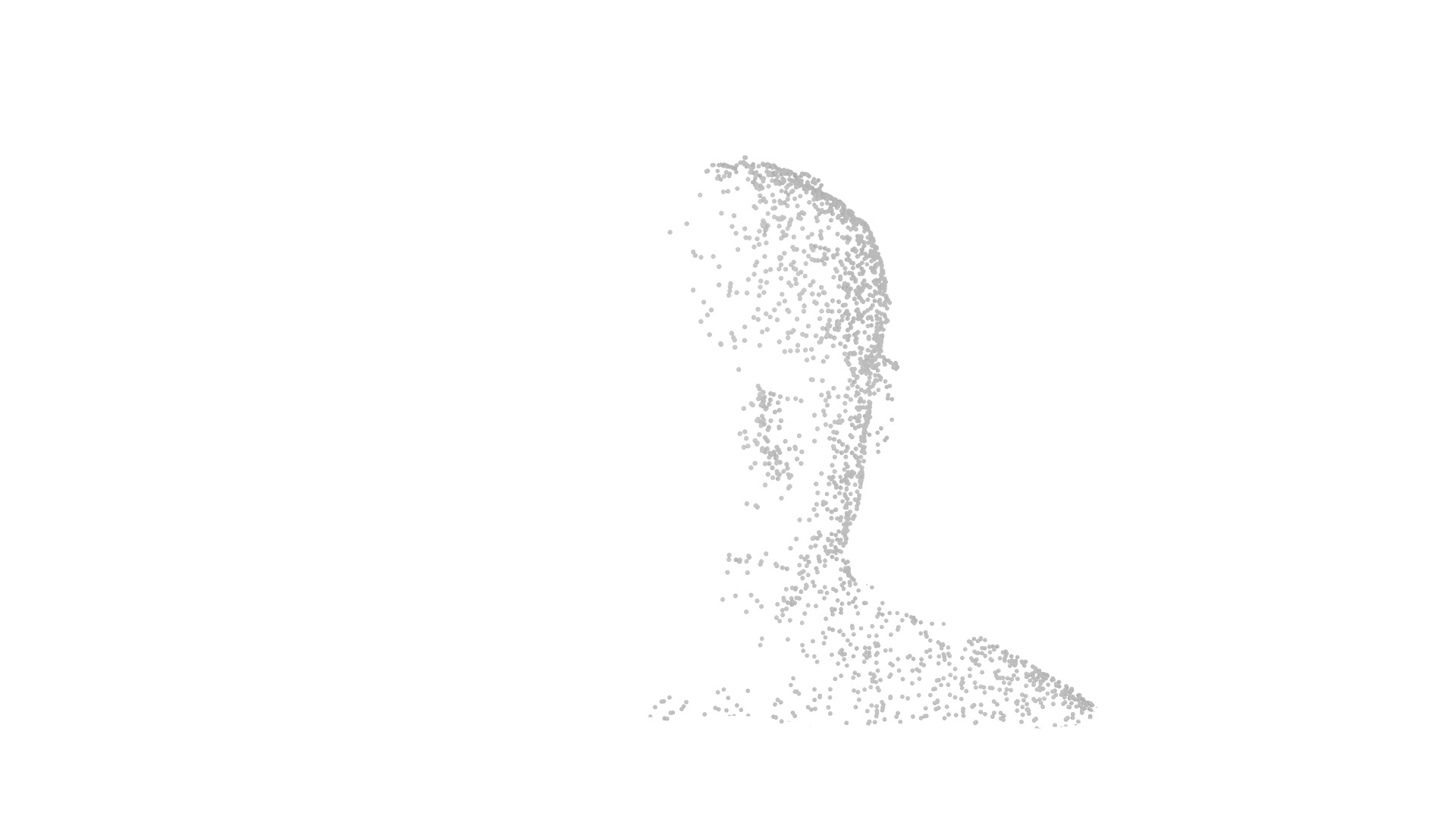}
            & \includegraphics[width=0.11\textwidth,trim={0cm 2cm 1cm 1cm},clip]{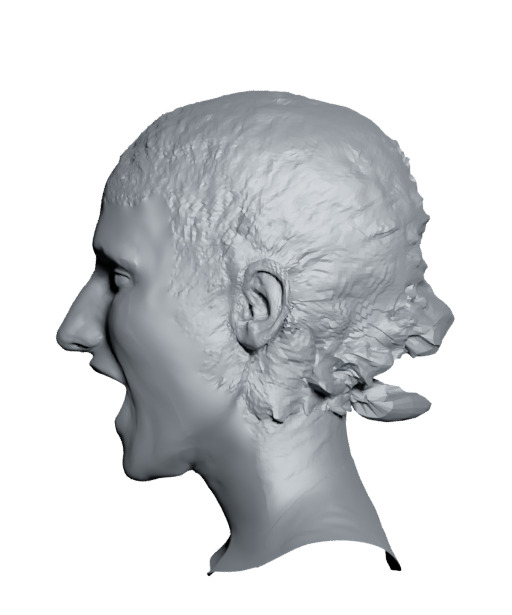}  
            & \includegraphics[width=0.11\textwidth,trim={0cm 2cm 1cm 1cm},clip]{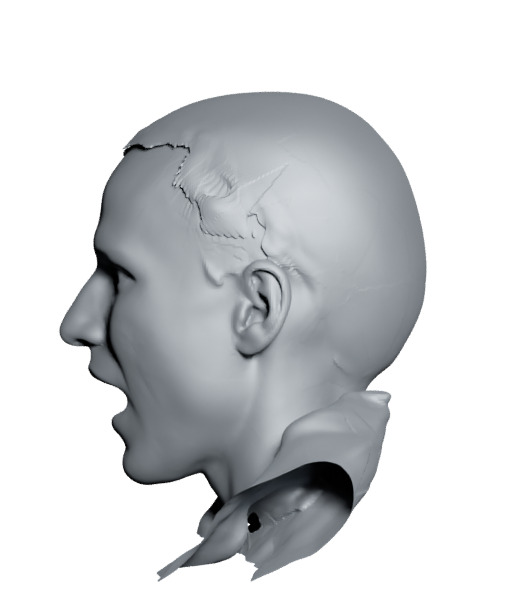} 
            & \includegraphics[width=0.11\textwidth,trim={0cm 2cm 1cm 1cm},clip]{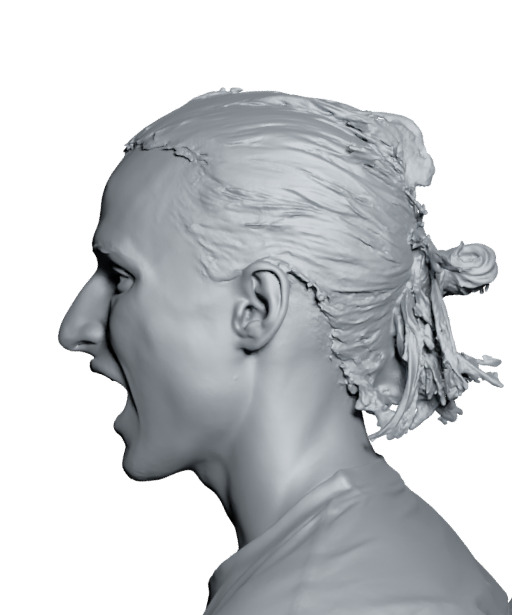}
        \end{tabular}
    }
    \vspace{0cm}
    \captionof{figure}{
    Demonstration of geometry completion aided by the HeadCraft model.
    Here, we extract depth maps from scans from the NPHM dataset, unseen during training, and try to complete them by finding the appropriate latent representation of StyleGAN.
    As a necessary intermediate step, we first apply our registration procedure to the partial point cloud to locate the points in the UV space of the template.
    The optimal latent is found by minimizing the discrepancy of the complete UV map and registered partial UV map in the observed regions.
    HeadCraft is also capable of estimating plausible details for a very sparse point cloud (1\% of \# points) -- see the last row.
    }
    \label{fig:fitting_result}
    \vspace{-0.2cm}
\end{table*}    
\begin{figure*}[h!]
    \centering
    \includegraphics[width=0.85\linewidth,trim={0.5cm 5.5cm 0.5cm 0cm},clip]{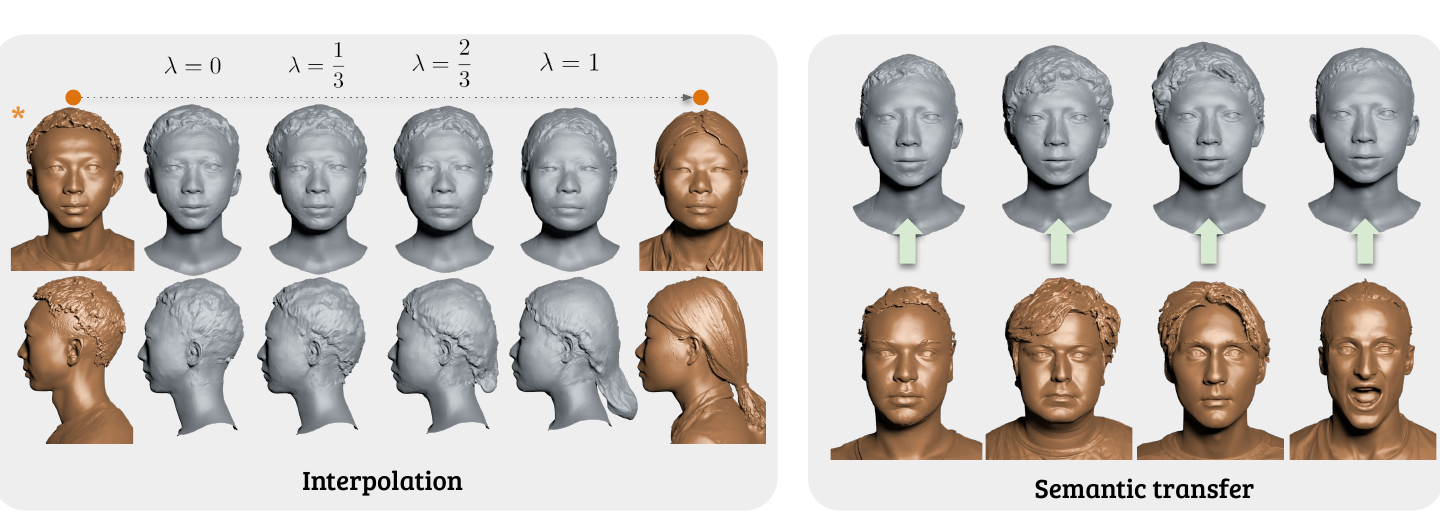}
    \vspace{-0.2cm}
    \caption{
    Semantic editing. 
    Interpolating the latent representations of HeadCraft \textit{(left)} allows us to smoothly change the person's appearance from one to another.
    To do that, we fit the latent codes for two real scans in the NPHM dataset (brown, top rows), unseen during training, and blend them together with a $\lambda$ weight.
    Likewise, we can transfer surface variations from one person to another \textit{(right)}.
    The source person is marked with {\textcolor{orange}{$\star$}} on the left and the driving person is in the bottom row on the right.
    }
    \label{fig:interpolation}
    \vspace{-0.2cm}
\end{figure*}

The capabilities of fitting the model to the full scan, e.g. created from Structure-from-Motion (SfM), are demonstrated as a part of the
semantic editing experiments in Fig.~\ref{fig:interpolation} (top rows, the result of the latent fitting, $\lambda=\{0, 1\}$).

\mypara{Animation.} 
The decomposition of the parametric model and the displacements allows us to animate the complete head model.
In our experiments, we take real multi-view video sequences with talking people from the NeRSemble dataset~\cite{nersemble} and obtain shape, expression, jaw, and head pose parameters for each time frame of the speech by running a FLAME tracker for each sequence.
For each of the sample shapes, estimated from the sequences, we reenact the corresponding FLAME with estimated expression parameters, subdivide the template and query a randomly pre-sampled UV displacement map from HeadCraft.
Since the template is also deforming over time, we rotate the displacements according to the changing surface normals of the template.
The reenactment results on NeRSemble are available in the Supplementary Video.
In Fig.~\ref{fig:animation}, for higher clarity, we demonstrate rigging with randomly sampled FLAME shapes and a small number of extreme expressions generated artificially by randomly setting a subset of the first ten expression components of FLAME to $\pm 2$.


\noindent \textbf{Interpolation between the displacements.}
In Fig.~\ref{fig:interpolation}, we show how interpolating the 
latent code of our generative model influences the change of the geometry. 
Further interpolations are presented in the Supplementary Video.

\mypara{Semantic transfer from one scan to another.} 
Access to the shared UV space allows us to modify the geometry semantically.
In Fig.~\ref{fig:interpolation}, the transfer of the scalp region from one ground truth NPHM scan, unseen during training, to the other is shown.
The transfer is performed via fitting the latent representation of HeadCraft to the driver scan (the source of displacements) and feeding it to the model.
The extracted displacements are later applied to the source scan.
\section{Discussion}

In this work, a generative model for 3D human heads is presented.
We demonstrate the efficacy of the hybrid approach involving an underlying animatable parametric model and a neural vertex displacement modeller.
Most importantly, our method allows to model high-quality shape variations while maintaining the realistic animation capability, and the inversion framework allows us to find a suitable latent representation to either represent a full head scan or a part of it that could come from e.g. the depth sensor.
A direction of the possible future work could be focused on incorporating an appearance model for color and material-based relighting and a physical model of hair movement, based on, for instance, hair strands, to support more realistic animation.
The code and the dataset of displacement registrations will be released to the public.

\noindent \textbf{Acknowledgments.}
We gratefully acknowledge the support of this research by a TUM-IAS Hans Fischer Senior
Fellowship, the ERC Starting Grant Scan2CAD (804724) and the Horizon Europe vera.ai project (101070093).
We also thank Yawar Siddiqui, Alexey Artemov, Justus Thies for helpful advice, Tobias Kirschstein for his assistance with the NeRSemble,
Taras Khakhulin for his help with the ROME baseline,
Angela Dai for the video voiceover.
{
    \small
    \bibliographystyle{ieeenat_fullname}
    \bibliography{bibliography}
}

\clearpage
\begin{appendix}
\section{Method: Technical Details}

\subsection{Displacement registration procedure}
\label{subsec:supmat_registration}


Here we explain the procedure in more detail.
The vertices and displacements are modeled in the NPHM coordinate system, aligned with the scans, and the scaling of $30\times$ applied.
The implementation of Butterfly subdivision of the FLAME template from MeshLab~\cite{meshlab} was used. 
The parameters of the subdivision are constant across the scans and equal to 3 subdivision iterations with a threshold of 42.5. 
The subdivision produces around 100K vertices and 200K triangles for the original template consisting of 5023 vertices and 9976 triangles and smooths the surface.
The description of the optimization problem features individual loss terms.
The expanded expression for the terms are as follows.
The Chamfer term $ \mathcal{L}_\textrm{Chamfer}(P_1, P_2) $ quantifies the distance between the point clouds $P_1 \in \mathbb{R}^{|P_1| \times 3}$ and $P_2 \in \mathbb{R}^{|P_2| \times 3}$ is supposed to be differentiable w.r.t. the points of $P_1$.
In our work, we apply the version pruned by the distance of the correspondences, i.e. when the Euclidean distance between point and its matched version exceeds the predefined threshold $d$, this correspondence is not accounted in the loss term.

{\small
\begin{align*}
    \mathcal{L}_\textrm{Chamfer} & (P_1, P_2) = \\
    & \frac 1{\sum_{p \in P_1} [d(p, nn(p, P_2)) \le d]} \cdot \\
    & \cdot \sum\limits_{p \in P_1} \left( d(p, nn(p, P_2)) \cdot [d(p, nn(p, P_2)) \le d] \right) \\
    & + \frac 1{\sum_{p \in P_2} [d(p, nn(p, P_1)) \le d]} \cdot  \\ 
    & \cdot \sum\limits_{p \in P_2} \left( d(p, nn(p, P_1))  \cdot [d(p, nn(p, P_1)) \le d] \right),
\end{align*}
}

\noindent where $d(\cdot, \cdot)$ stands for the Euclidean distance between two points in space and $nn(p, P) = \argmin\limits_{p' \in P} d(p, p')$ is the nearest neighbor of $p$ in a point cloud $P$.

Edge length regularization is defined as follows.

\begin{align}
    \mathcal{L}_\textrm{edge} (V, \mathcal{F}) = 
     \frac 1{|E|} \sum_{(e_a, e_b) \in E} d^2(V_{e_a}, V_{e_b}),
\end{align}

\noindent where $E = E(\mathcal{F})$ is a set of graph edges derived from the faces $\mathcal{F}$.
To construct it, we consider each face bringing three new edges and later leave only the unique edges in $E$.

Laplacian term is defined as the Euclidean distance between the vertex and its neighbors, which can be efficiently calculated via computing sparse Laplacian $L = L(V, \mathcal{F})$ of the graph:

\begin{align}
    \mathcal{L}_\textrm{lapl} (V, \mathcal{F}) = 
     \sqrt{\sum_{v \in V} \| Lv \|_2^2}
\end{align}

The outer norm is used instead of e.g. L1 averaging to enforce the uniform smoothness of the mesh and avoid spikes that tend to appear otherwise (see, e.g., the documented example in \href{https://github.com/facebookresearch/pytorch3d/issues/432}{PyTorch3D~\cite{ravi2020pytorch3d} repository}).

During the vector displacement stage, only the scalp region (defined by the semantic mask shipped with FLAME~\cite{flame}) is optimized. 
During the normal displacement stage, we also unfreeze the facial region but keep the neck, eyeballs, ears, and inner mouth region frozen (the latter is annotated manually in Blender~\cite{blender} package and is frozen because of its absence in the ground truth scans, as it is placed fully interior).
Each stage takes around 3 min for 1K steps on a single NVIDIA RTX 2080 Ti GPU.
We used the PyTorch3D~\cite{ravi2020pytorch3d} functions for implementation of all the loss terms.

In Fig.~\ref{fig:uv_layout}, we show how the seam was annotated for the custom UV layout.
%

\begin{table*}[h!]
    \setlength{\tabcolsep}{5pt}    
        \begin{center}
            \begin{tabular}{cc|cc}
                \includegraphics[width=.17\linewidth]{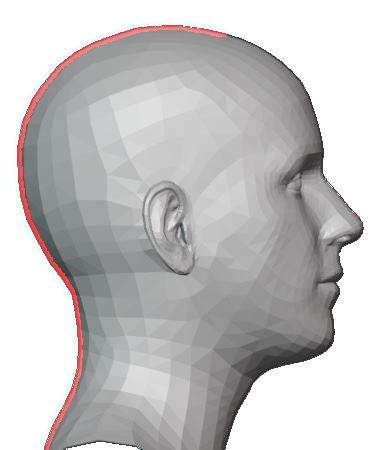} &
                \includegraphics[width=.14\linewidth]{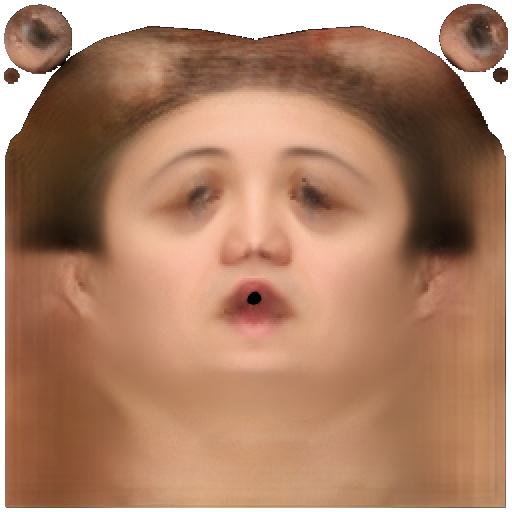} & 
                \includegraphics[width=.17\linewidth]{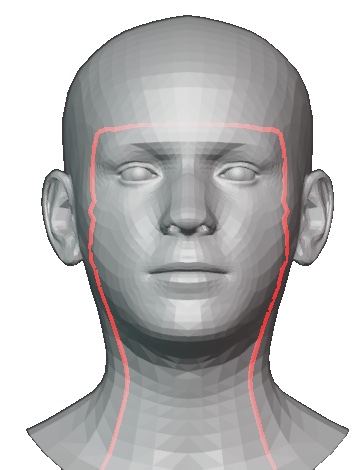} &
                \scalebox{2}[1]{\includegraphics[width=.14\linewidth]{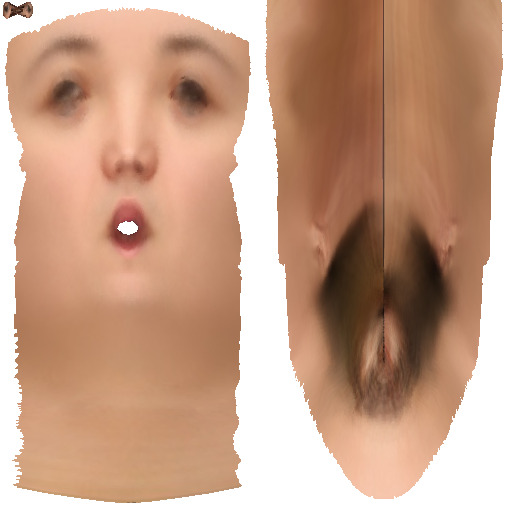}}
            \end{tabular}
            \vspace{-0.1cm}
            \captionof{figure}{
                Comparison of the FLAME layouts.
                The standard, commonly used unwrapping for FLAME \textit{(left)} features a seam corresponding to the vertical line in the back of the head and pays more attention to the facial region than to the scalp.
                In the hand-crafted custom layout that we employ \textit{(right)}, a different seam around the face border is selected, thus making the regions of face and scalp separated and of similar size to avoid unnecessary breaks and expand the region of interest. Lower neck is not modeled by our method and hence made partially unseen in UV.
            }
            \label{fig:uv_layout}
        \end{center}
\end{table*}

Blender~\cite{blender}~4.0 package was used to annotate the seam for the custom UV layout.
The seam is manually annotated to be symmetric w.r.t. the reflection symmetry of FLAME.
Firstly, the layout was constructed via ''Unwrap`` Blender tool and adjusted manually via local translation and scaling tools to better fit the available space.
After that, the UVs for vertices in the left part of the head were assigned with the mirrored UVs of the vertices in the right part to account for symmetry.
This is repeated for the face and scalp separately.
Finally, the face, scalp, and eyeballs parts were scaled and translated to fit the unit square.
Regions, for which displacements are not modeled, are either given smaller space in the layout (eyeballs) or moved outside of the unit square (unmodeled parts of the neck).

The displacement vectors entries typically belong to the [-2, +2] range, while some large shape variations (e.g. a ponytail) can introduce the offsets into a large range up to [-20, +20]. 
We clip any displacements, obtained after full registration, to the [-20, +20] range.
As the last step, the displacements are rendered in UV space, and each UV map is saved as uint16 image files linearly renormalized from [-20, +20] to [0, $2^{16} - 1$].
Saving in uint16 (double-byte intensity value) instead of the widely used uint8 
(single-byte intensity value) is important, since most of the displacements vector entries are concentrated around the small neighborhood of zero and the precision can be lost when renormalizing from [-20, +20] to [0, $2^8 - 1$] instead and discretizing.
Saving UV maps as raw files would otherwise facilitate much slower training of the generative model due to the time-consuming loading and memory usage overhead.

\begin{table}[]
    \vspace{0.3cm}
    \setlength{\tabcolsep}{0.2pt}
    \begin{tabular}{cccccccccc}
        \raisebox{1.9\normalbaselineskip}[0pt][0pt]{\rotatebox[origin=c]{90}{{\small Standard UV}}}\hspace{0.2cm}
        & \includegraphics[width=0.14\linewidth,trim={2cm 0cm 2cm 0cm},clip]{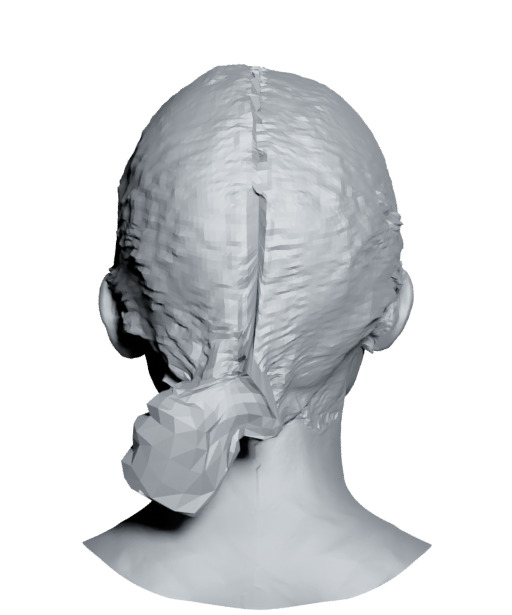} 
        & \includegraphics[width=0.14\linewidth,trim={2cm 0cm 2cm 0cm},clip]{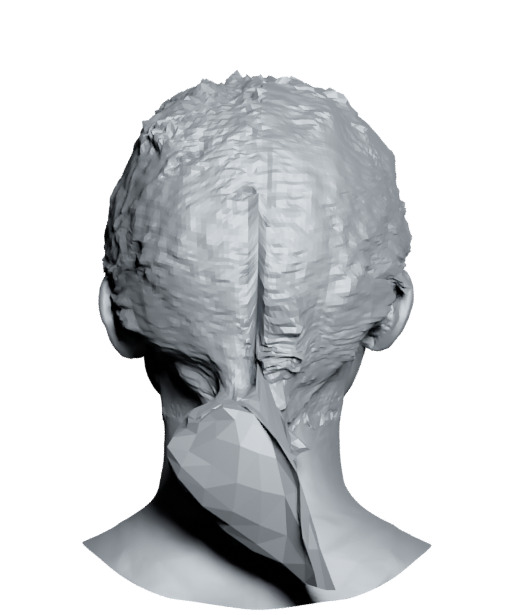} 
        & \includegraphics[width=0.14\linewidth,trim={2cm 0cm 2cm 0cm},clip]{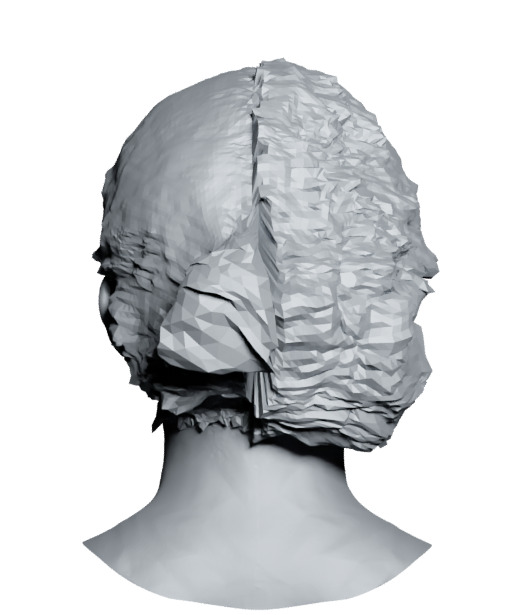}
        & \hspace{0.3cm}
        & \raisebox{1.9\normalbaselineskip}[0pt][0pt]{\rotatebox[origin=c]{90}{\textbf{{\small Custom UV}}}}\hspace{0.2cm}
        & \includegraphics[width=0.14\linewidth,trim={2cm 0cm 2cm 0cm},clip]{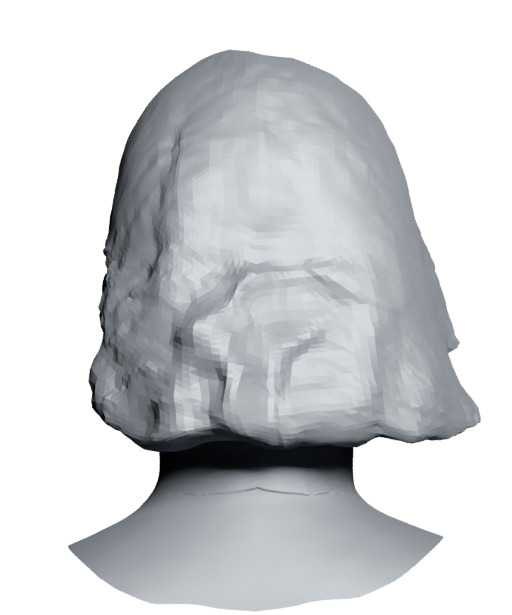} 
        & \includegraphics[width=0.14\linewidth,trim={2cm 0cm 2cm 0cm},clip]{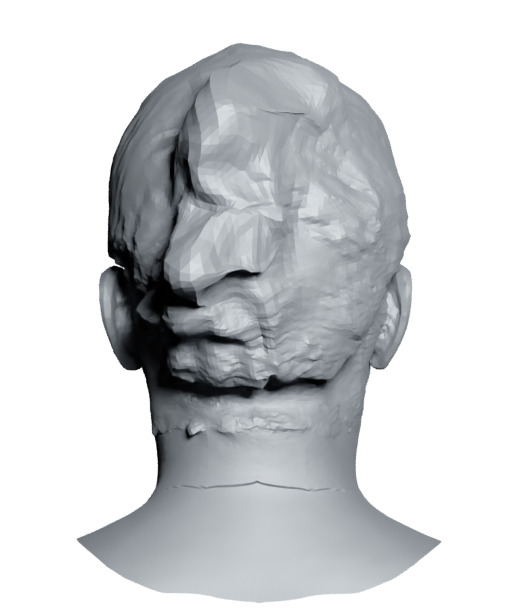} 
        & \includegraphics[width=0.14\linewidth,trim={2cm 0cm 2cm 0cm},clip]{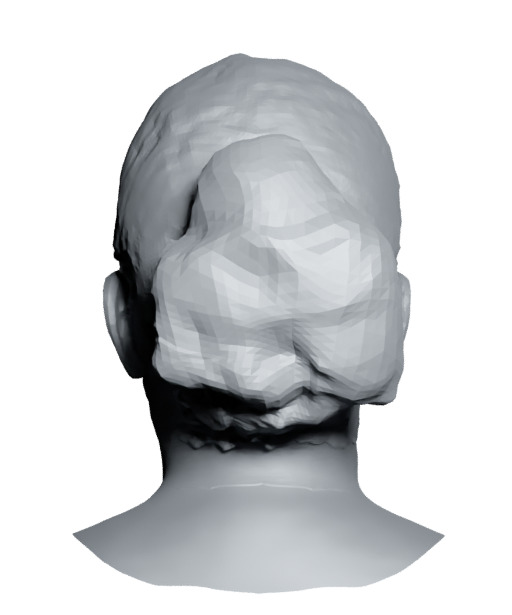}
    \end{tabular}
    \vspace{-0.2cm}
    \captionof{figure}{Ablation over the choice of the UV layout.
    Our method utilizes custom UV layout that allows us to model more consistent geometry by mitigating seam artifacts. 
    %
    }
    \label{fig:uv_layout_ablation}
    \vspace{-0.3cm}
\end{table}

\subsection{Generative model}


For training StyleGAN, we used the \href{https://github.com/nihalsid/stylegan2-ada-lightning}{stylegan2-ada-lightning} implementation with ADA and augmentations turned off and the following hyperparameters:

\begin{table}[h!]
\centering
\resizebox{\linewidth}{!}{
\begin{tabular}{c|c|c|c}
latent dim & \# layers (z $\rightarrow$ w) & G lr  & D lr    \\ \hline
512        & 8                             & \,\,0.002\,\, & \,\,0.001\,\, \\ 
\multicolumn{4}{l}{} \\
$\lambda_{gp}$ & $\lambda_{plp}$ & img size & batch size \\ \hline
\,\,4.0\,\,            & \,\,2.0\,\,            & \,\,$256 \times 256$\,\,           & 8    \\
\end{tabular}
}
\end{table}

The model was trained on four NVIDIA RTX 2080 Ti GPUs.
For training of the StyleGAN, the ground truth UV displacement maps were transformed from $256 \times 256 \times 3$ into $256 \times 256 \times 6$ by first bilinearly upscaling the displacement maps over the width dimension (resulting in the map of shape $256 \times 512 \times 3$) and then splitting it into two three-channel maps by the width dimension.
This effectively means that StyleGAN predicts the face and scalp as two separate three-channel images, which increases the spatial area in the output dedicated to each region.
Additionally, we replaced the facial part of the registered UV maps of subjects in the NPHM dataset with the corresponding facial part from the neutral expression scan of the same person.
This has been done to better support various expressions at the inference time.
We also found it beneficial to disable the StyleGAN noise at the inference time, typically injected into the generator, for the face part of the UV map to smooth out the generated facial displacements relative to the scalp displacements that normally require a higher level of detail.
For the scalp region, the StyleGAN noise is constant, initialized separately for each generator layer before training.
This noise schedule separation is performed via two passes through StyleGAN -- one with zero noise and another with constant noise -- and combining the corresponding results.

\begin{figure*}
    \includegraphics[trim={0cm 7.5cm 0cm 0}, clip, width=\linewidth]{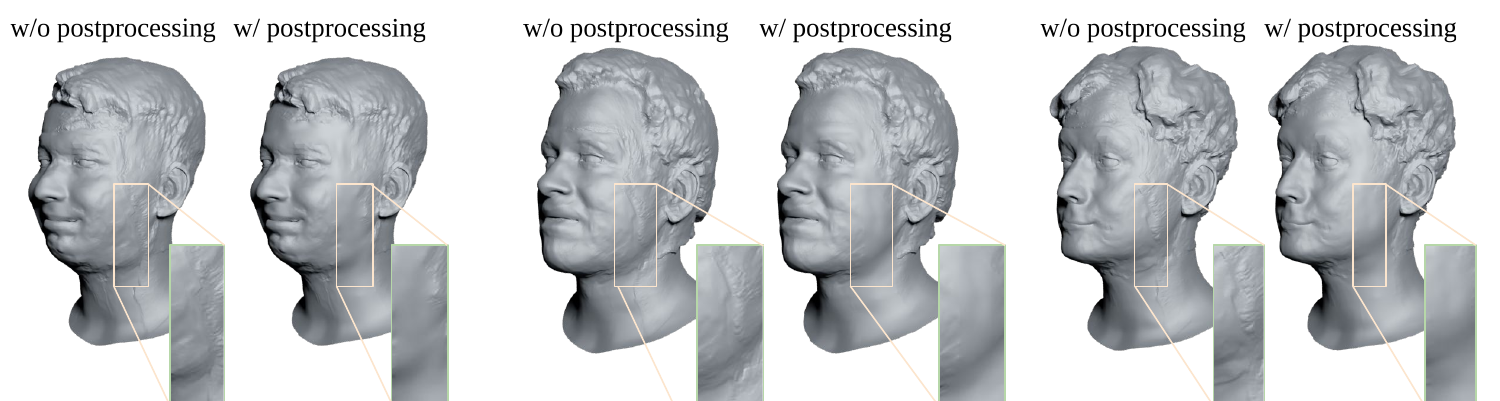}
    \vspace{-0.6cm}
    \caption{
    %
    %
    %
    Comparison w/ and w/o the post-processing procedure that enables smoother transition between the face and scalp parts modeled in unconnected regions of the UV space.
    }
    \label{fig:postprocessing}
    \vspace{-0.3cm}
\end{figure*}

\mypara{Post-processing.} 
To ensure a smooth transition from the face to the scalp part around the seam, we first create a mesh by querying the UV map as is, then identify the seam vertices and apply Laplacian smoothing~\cite{vollmer1999improved} to the mesh in the $K$-vertex vicinity of the seam (vicinity obtained via BFS-style expansion of the seed vertices defining the seam in a graph defined by the mesh edges).
In our experiments, $K = 10$ (corresponding to the subdivided template with roughly 100K vertices), and Laplacian smoothing is repeated 10 times, only affecting the seam vicinity area.
Since the face displacements are modeled with lower smoothness due to the disabled StyleGAN noise in that area, the post-processing technique results in a smooth transition from the relatively smooth face displacements to sharper details in the scalp area while maintaining local geometric consistency.
The visualization is shown in Fig.~\ref{fig:postprocessing}.

\section{Results}
\label{sec:supp_results}

\mypara{Unconditional sampling.} In Fig.~\ref{fig:uncond_more}, we provide more unconditional samples from our model from different viewpoints.
All the samples have been produced by sampling $z \in \mathcal{N}(0, \mathbb{I})$ with a truncation trick~\cite{truncation_trick} with the power $\psi=0.7$.
For the evaluation in the Table 1 in the main text, the implementation of the metrics MMD, JSD, COV from PointFlow~\cite{yang2019pointflow} was used.
Since FaceVerse contains samples grouped by subjects, the nearest neighbor of a ground truth scan is typically a scan of the same subject with a different expression.
Because of that, we only select one ground truth sample per subject (with the same neutral expression for all subjects) to calculate COV.
All FaceVerse scans are used to calculate MMD and JSD.
As a distance measure between individual point clouds, aggregated over multiple observations in MMD and COV, we use Chamfer Distance.

\begin{table}[h!]
    \vspace{0.2cm}
    \centering
    \setlength{\tabcolsep}{0pt}
    \begin{tabular}{cccp{0.1cm}|p{0.1cm}cccp{0.1cm}|p{0.1cm}ccc}
    \multicolumn{3}{c}{Ours} & & & \multicolumn{3}{c}{Ours w/ VAE} & & & \multicolumn{3}{c}{Ours w/ VQ-VAE} \\
    \includegraphics[width=0.105\linewidth,trim={1cm 1cm 0.5cm 1cm},clip]{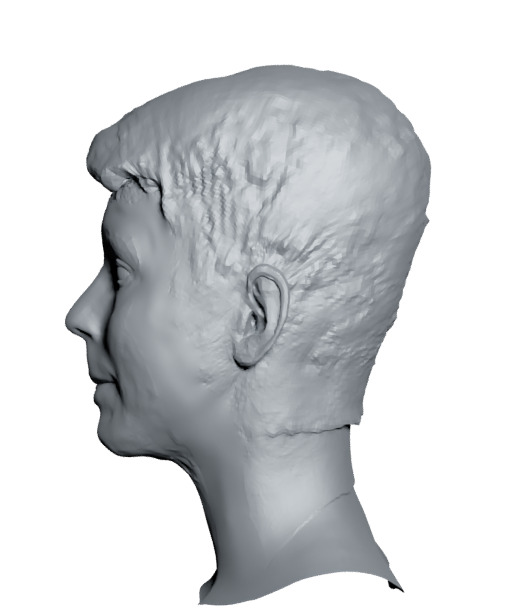} 
    & \includegraphics[width=0.105\linewidth,trim={1cm 1cm 0.5cm 1cm},clip]{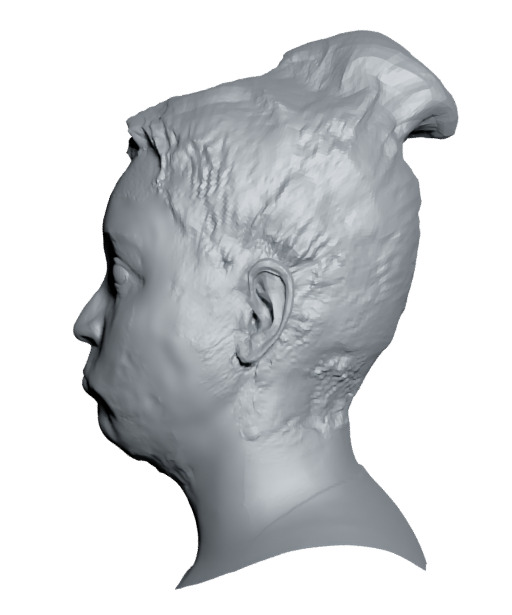} 
    & \includegraphics[width=0.105\linewidth,trim={1cm 1cm 0.5cm 1cm},clip]{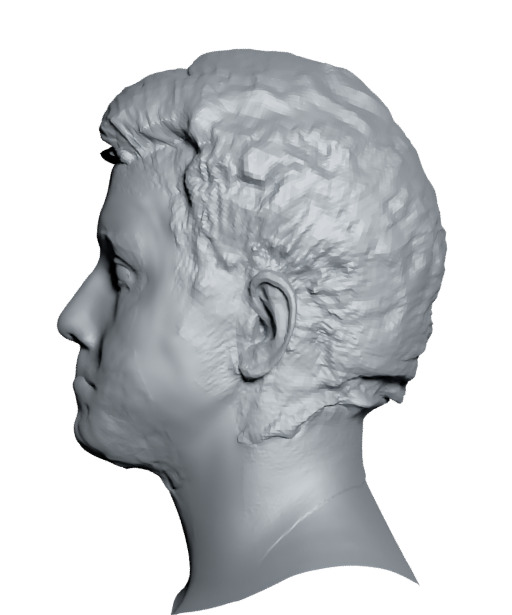} 
    & 
    & 
    & \includegraphics[width=0.105\linewidth,trim={1cm 1cm 0.5cm 1cm},clip]{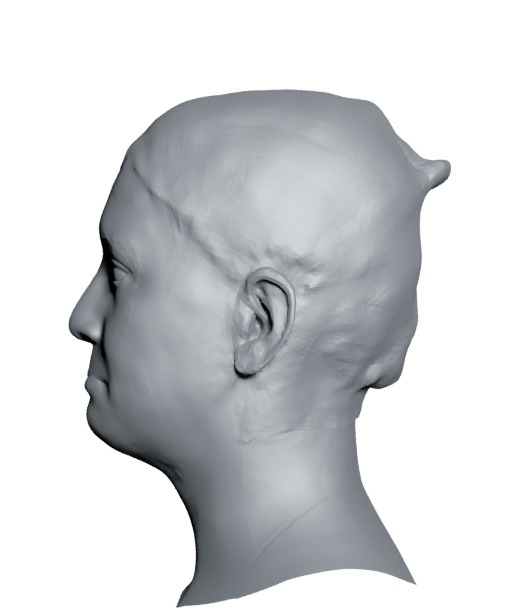} 
    & \includegraphics[width=0.105\linewidth,trim={1cm 1cm 0.5cm 1cm},clip]{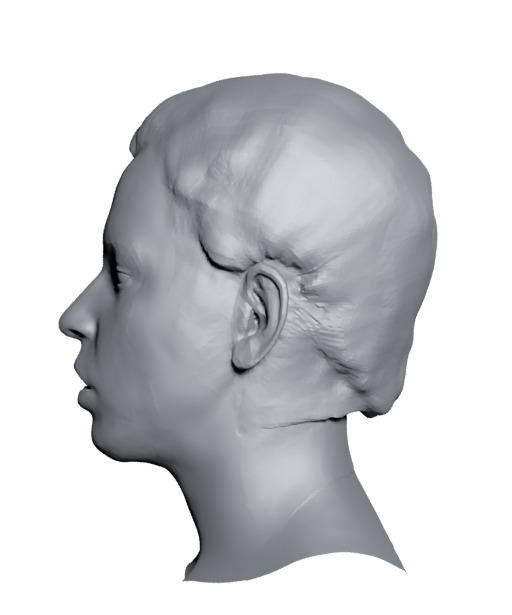} 
    & \includegraphics[width=0.105\linewidth,trim={1cm 1cm 0.5cm 1cm},clip]{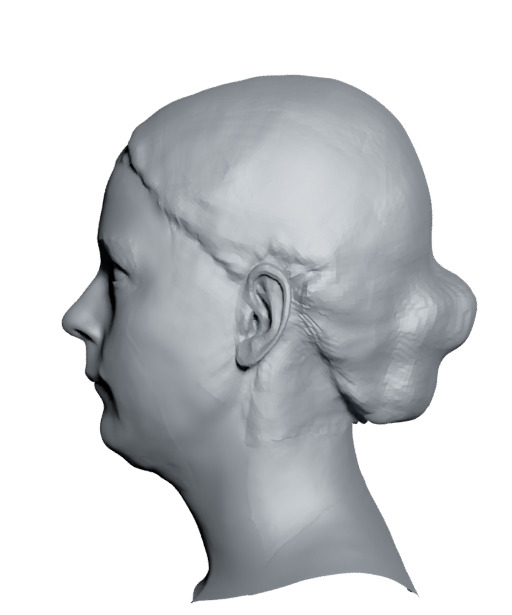} 
    & 
    & 
    & \includegraphics[width=0.105\linewidth,trim={1cm 1cm 0.5cm 1cm},clip]{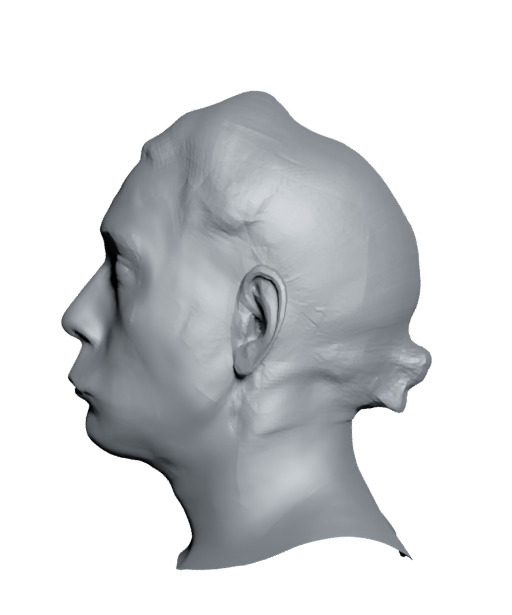}  
    & \includegraphics[width=0.105\linewidth,trim={1cm 1cm 0.5cm 1cm},clip]{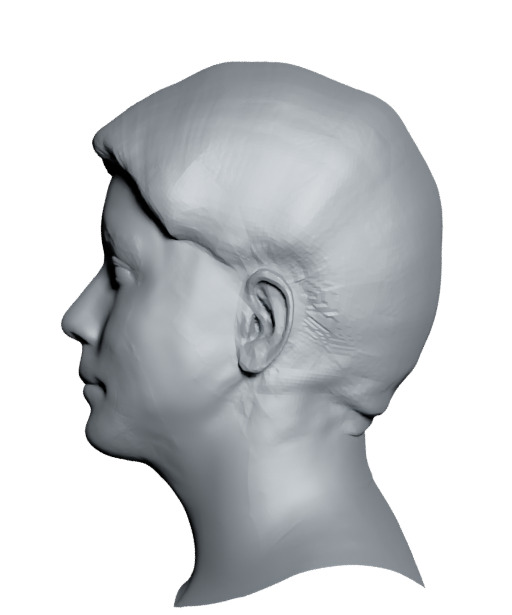} 
    & \includegraphics[width=0.105\linewidth,trim={1cm 1cm 0.5cm 1cm},clip]{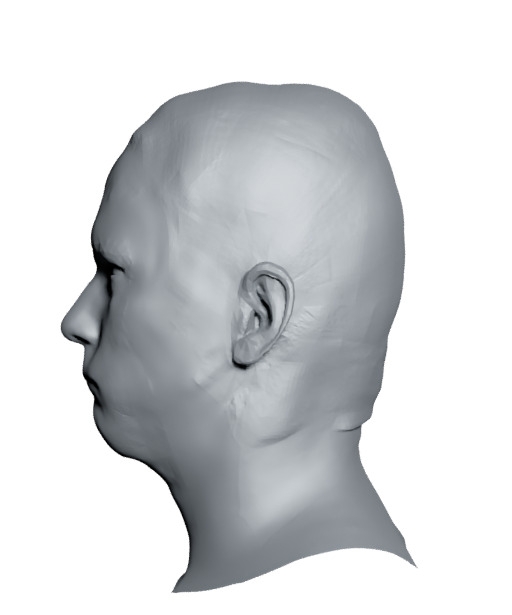} 
    \end{tabular}
    \vspace{-0.2cm}
    \captionof{figure}{
    Ablation over the model design.
    VAE and VQ-VAE both follow the ResNet-18 encoder and decoder architecture, while our method is based on StyleGAN2.
    [VQ-]VAE mostly match the diversity of the training data but not the level of detail.
    }
    \label{fig:arch_ablation}
\end{table}

\mypara{Ablating over the choice of the UV layout.}
%
%
We assess the effect of a manually hand-crafted UV space for FLAME on the quality of generations in Fig.~\ref{fig:uv_layout_ablation}. 
As observed, moving the seam from the vertical middle line, as in the standard UV layout for FLAME, to the face border, allows us to model more consistent and complex geometry without large distinction between a left and a right part.

\mypara{Ablating over the choice of the generative model architecture.} 
The VAE used in our experiments is based on the \href{https://github.com/Lightning-Universe/lightning-bolts}{Lightning Bolts} library.
The encoder follows the ResNet-18 architecture consisting of blocks of 2 convolutions each, with every second convolution with a stride of two (starting from the third) to downsample the activations spatially the increasing number of channels (64 in the first two blocks, then 128, 256, 256, 256, 512, 512 in the next blocks, respectively). 
The Lightning Bolts implementation adds two fully-connected layers on top of the encoder (one for the $\mu$ and one for the $\sigma$ prediction).
The dimension of the latent space equals 512.
The decoder follows the architecture symmetric to the encoder, where the stride two for some convolutions is replaced with a nearest-neighbor 2x upscaling.

For VQ-VAE implementation, we used the implementation of the VQ layer from \href{https://github.com/lucidrains/vector-quantize-pytorch/tree/master}{vector-quantize-pytorch}. 
\href{https://github.com/jzbontar/pixelcnn-pytorch/tree/master}{Pixelcnn-pytorch} served as a basis for the PixelCNN sampler implementation.
Similarly to VAE, ResNet-18 encoder and decoder were used, with the exception that fewer downsampling operations have been used: they were introduced at each second layer (starting from the third), not each first layer.
This is introduced to maintain a trade-off between the autoencoder quality and sampling ability, i.e. not to make PixelCNN operate in a too small latent space.
The spatial resolution of the bottleneck is $32 \times 32$, which we found to be optimal, as the sampling performance of PixelCNN degrades from the top-left corner to the bottom-right corner and it is very noticeable already at the $64 \times 64$ bottleneck spatial resolution.
The number of channels is 64, 128, 128, 32 for each two consecutive blocks, respectively.
VQ-VAE is trained for 10K steps with batch size of 8, which we found to be enough to reach the sufficient visual quality of autoencoding.
To facilitate the sampling, we obtain a dataset of VQ indices and learn PixelCNN to autoregressively sample from those for 200 epochs with a batch size of 32.

Visual comparison of the generative model choices is demonstrated in Fig.~\ref{fig:arch_ablation}.

\begin{table}[b!]
    \resizebox{\linewidth}{!}{
        \begin{tabular}{l|ccc}
        Registration     & \multicolumn{1}{l}{$L_1$-Chamfer $\downarrow$} & \multicolumn{1}{l}{N.C. $\uparrow$} & \multicolumn{1}{l}{F-Score @ 1.5 $\uparrow$} \\ \hline
        FLAME            &    3.33e-1       &   0.763               &      0.266      \\
        Stage 1 only     &    9.20e-2                    &    0.817      &   0.861            \\
        Stage 1+2 (Ours) &    5.68e-2                    &    0.841                 &      0.949        
        \end{tabular}
    }
    \caption{Metrics reflecting the registration quality for our method w.r.t. only using Stage 1 of the registration procedure (vector offsets) and FLAME fits.}
    \label{table:reg_quality}
\end{table}

\mypara{Behavior of the registration procedure.} 
In Fig.~\ref{fig:more_registrations_p1}, \ref{fig:more_registrations_p2}, \ref{fig:more_registrations_p3}, \ref{fig:more_registrations_p4}, we show how the mesh deforms as a result of the vector displacements optimization and normal displacements optimization.

\mypara{Registration quality.}
In Table~\ref{table:reg_quality}, we report exactly the same metrics as in NPHM~\cite{nphm} for both stages of our registration, averaged over identities. 
N.C. refers to Normal Consistency score; see~\cite{nphm} for clarification on all metrics.
We excluded the region below the threshold by vertical axis (-0.30 in standard NPHM coordinate system) in order to not account for non-modeled region (clothes and neck). 
The registration precision is naturally better by all metrics than FLAME that does not provide a hair fit and better face fit.
The necessity of the second stage is also motivated accordingly.
Note that even though the numbers are not directly comparable to the ones in the NPHM paper~\cite{nphm}, since there the evaluation over the face region only was conducted, and for us it is for the whole FLAME surface, they are still in the same scale.
This indicates that the quality attained by non-rigid face refinement procedure in NPHM has been mostly achieved by our registration procedure for both face and hair.

\mypara{Consistency of registrations.} 
In Fig.~\ref{fig:uv_checker}, we demonstrate the analysis as to which template vertices are selected by the registration procedure to cover various regions of different meshes. 
Since we know the UV coordinates of all template vertices, this can be done by rendering the meshes with a \href{https://uvchecker.vinzi.xyz/}{UV checker} texture image.
Note that the long hair parts, such as pony tails, are mostly explained by the same regions of the layout as the vertices they originate from.

\begin{figure}[h!]
    \includegraphics[trim={0.5cm 0.5cm 0cm 0.5cm}, clip, width=\linewidth]{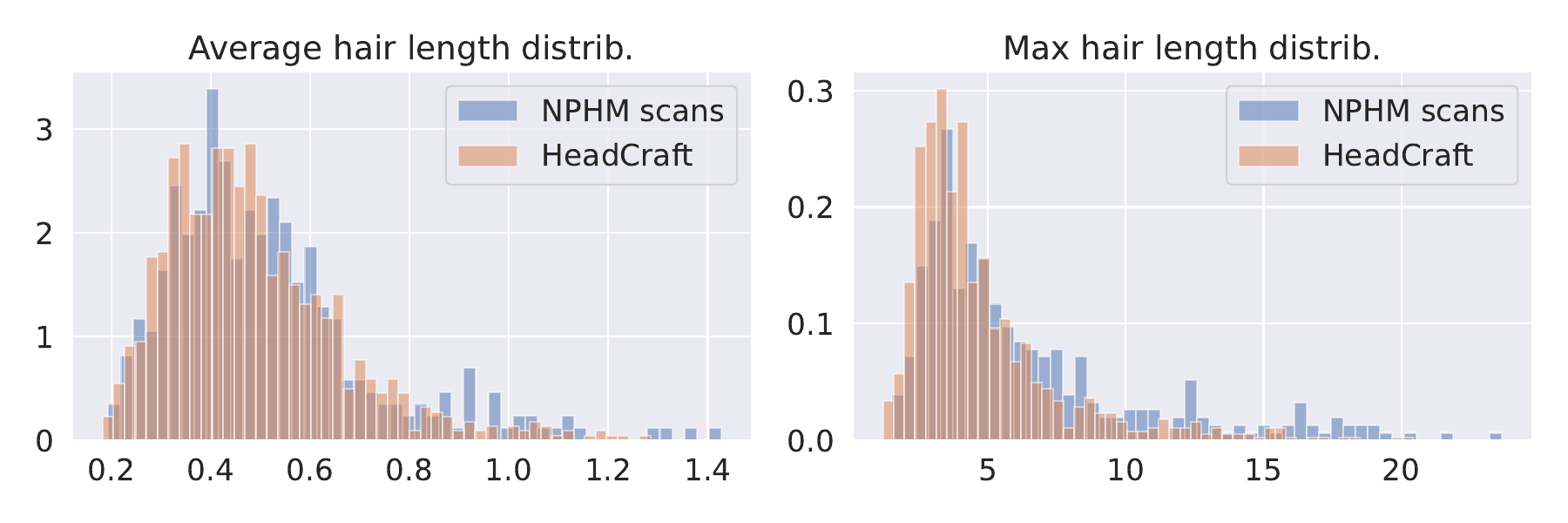}
    \vspace{-0.6cm}
    \caption{
    Comparison of the average offset length distribution in the scalp region for our method's predictions and 3D scans from the NPHM dataset.
    We consider the offset length in the scalp region the approximation of the haircut size (or, simplified, a measure of hair length, without accounting for the hair curvature). 
    %
    %
    The diversity of the average/maximum of this parameter per scan for our generated samples is on par with the training data.
    The horizontal axis quantifies the length in the NPHM coordinate system and the vertical axis stands for the bin height of the histogram.
    }
    \label{fig:hair_length_distrib}
    \vspace{-0.3cm}
\end{figure}

\mypara{Comparison of the hair length.}
In Fig.~\ref{fig:hair_length_distrib}, we show the comparison of the offset length in the scalp region between the ground truth data (\textit{NPHM scans}) and the model predictions (\textit{HeadCraft}).
Since this offset length approximates the haircut size, we use it as a measure of hair length (not accounting for any hair curvature).
This is demonstrated for comparison of the occurrence frequency of long hair samples in the model's predictions w.r.t. the ground truth data distribution.

\subsection{Applications}

\mypara{Fitting the latent code to a full scan.}
To fit the latent to the complete head scan, we have to apply preliminary steps, similar to the ones used to construct the training set.
Firstly fit the FLAME to the scan, then apply our registration procedure to get a UV map $U_{gt}$.
After that, we fit a $w \in \mathcal{W}+ \subset \mathbb{R}^{16 \times 512}$ latent code for the StyleGAN generator $g(w): \mathcal{W}+ \rightarrow \mathbb{R}^{H \times W \times 3} $ to satisfy the following loss terms:

\begin{align*}
    \mathcal{L}_{opt}^{full}&(w | U_{gt}, \boldsymbol{\lambda})  \\
    & = \lambda^\textrm{LPIPS} \cdot \textrm{LPIPS}(g(w), U_{gt})  \\
    & + \lambda^{L_1} \cdot L_1(g(w), U_{gt}),
\end{align*}

\noindent where $L_1(\cdot, \cdot)$ is an average pixel-wise L1 distance between two images and $LPIPS(\cdot, \cdot)$ corresponds to the LPIPS score~\cite{lpips}.
To calculate LPIPS, we cut the $256 \times 256$ UV maps (both predicted $U = g(w)$ and ground truth $U_{gt}$ into sixteen $64 \times 64$ patches, evaluate LPIPS between the respective patches of $U$ and $U_{gt}$, and average the obtained sixteen scores.
The parameters of the loss equal to $\lambda^\textrm{LPIPS}=0.1$ and $\lambda^{L_1}=3$.
The loss is being optimized via Adam algorithm with the learning rate of $10^{-2}$ for 1K steps.
The $w$ is initialized as the average latent predicted by the trained StyleGAN mapping network, evaluated over $10^5$ codes $z \in \mathcal{N}(0, \mathbb{I})$.

Finally, we optimize for the StyleGAN noise (only for the scalp region of the UV space) to better fit the tiny details of the map $U_{gt}$.
This step can be omitted in practice if fitting very high-frequency details is not required.
Exactly the same loss terms are being optimized, this time not with respect to $w$ but with respect to the StyleGAN noise tensors of all generator layers, while $w$ remains fixed.
The optimization is again carried out by Adam with the same learning rate and number of steps.

\mypara{Fitting the latent code to a depth map.} 
Fitting the latent representation to represent a partial observation poses a more challenging problem than trying to represent a full scan, since the resulting displacements must both resemble the original point cloud and complete it in a realistic way.
This requires several changes to the fitting pipeline, described next.

Firstly, prior to applying the registration procedure to register part of the cloud $P$ in the UV space, we identify the mask of points $m \in \{0, 1\}^{|V|}$ that are allowed to be offset by selecting only the points within the convex hull of the point cloud, expanded by 1.5x from its center to account for the possible important regions missing in the point cloud.
The points below a certain horizontal plane are not accounted for when estimating the convex hull to disregard the shoulders and clothing, usually featured in NPHM raw scans.
The level of the horizontal plane is selected as a 30\% quantile of the coordinates of the points along the vertical axis.
Masking out the points too far from the convex hull of the point cloud is especially important when the point cloud covers the minority of the geometry (e.g. if it is coming from a single depth map), since in this case, these points tend to pull in to cover the parts that the points inside the hull cannot explain (e.g. due to the regularizations), and this results in a non-plausible shape.
For the registration procedure itself, stronger regularization parameters for the first stage have been selected, namely $\boldsymbol{\lambda}_\textrm{Stage 1} = (\lambda_\textrm{Stage 1}^\textrm{Chamfer}, \lambda_\textrm{Stage 1}^\textrm{edge}, \lambda_\textrm{Stage 1}^\textrm{lapl}) = (2 \cdot 10^3, 8 \cdot 10^5, 10^5)$.
The correspondence pruning threshold, on the contrary, is raised to 10.0 for the first stage to allow the points to move farther while maintaining higher smoothness of the overall geometry due to stronger regularizations.
For the second stage, the threshold is on the contrary reduced to 0.1 to penalize for large false movement of points along the template normals to explain the individual points of the cloud. 

At the end of the registration, we refine the mask of the points by only selecting those of them that are sufficiently close to the fitted point cloud: $m^\textrm{final}_i = m_i \cdot [d(v_i + D_{\textrm{Stage 2}, i}, nn(v_i + D_{\textrm{Stage 2}, i}, P)) \le t]$, where $t$ defines the proximity threshold, and its optimal value depends on the sparsity of the cloud. 
For the point cloud formed from a dense depth map, we set $t=0.1$, and for a sparse cloud with only $1\%$ points of the original depth map left, we set $t=0.3$.
The regressed displacements and the mask are separately baked in the UV map as two independent images, 3-channel real-valued $U$ and 1-channel binary $M$, respectively.
In Fig.~\ref{fig:partial_registration_vis}, we demonstrate the typical result of the partial registration stages.

\begin{figure*}[t!]
    \centering
    \includegraphics[trim={0cm 7cm 0cm 0}, clip, width=.75\linewidth]{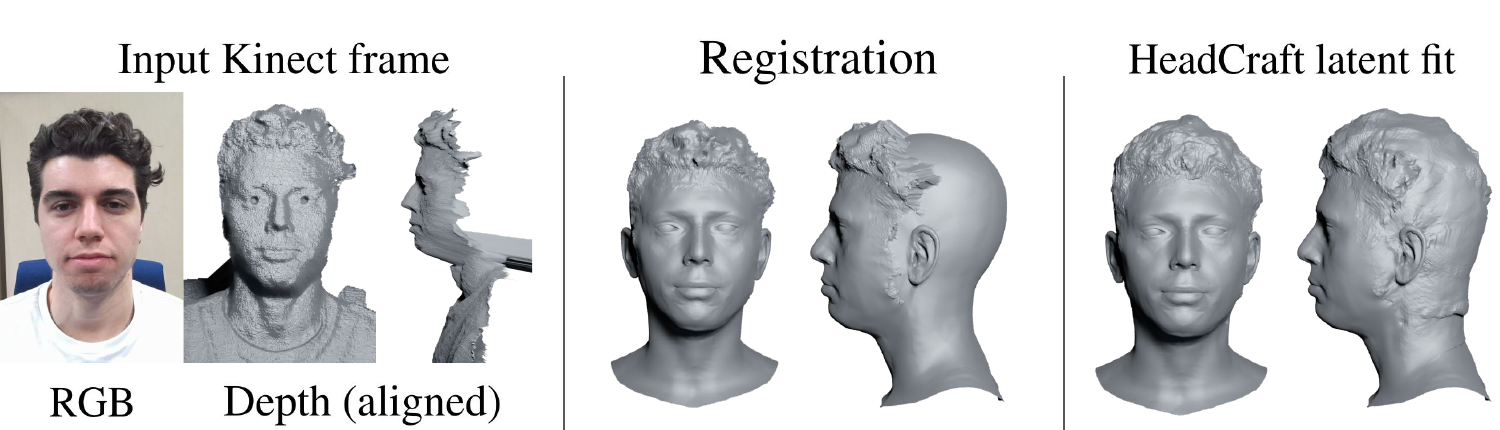}
    \caption{
    An example of the use case for the HeadCraft registration and fitting stages to Kinect depth data.
    An example of a frontal RGB frame is shown on the left (the color information is not used further), along with the observed Kinect depth, converted into a mesh and aligned with the standard FLAME coordinate system.
    We fit FLAME and apply the partial registration procedure, described in Sec.~\ref{sec:supp_results}, to register the displacements in the observed region (middle) and further fit the HeadCraft latent representation to the corresponding part of the UV map.
    }
    \label{fig:kinect}
\end{figure*}

Another important change lies in the latent fitting procedure.
In our observations, the optimization of $w \in \mathcal{W}+$ latent code works great for the visible part but tends to produce displacements closer to the average shape for the non-visible part.
We explain it by not strong enough supervision from the prior during fitting in $\mathcal{W}+$ space.
To mitigate that effect, we first fit the $z \in \mathcal{Z} \subset \mathbb{R}^D$ latent code of the StyleGAN mapping network $map(z):\mathcal{Z} \rightarrow \mathcal{W}+$, obtain the respective $w = map(z) \in \mathcal{W}+$ and regress the delta to the $w$ code: $\Delta w$.
We found that optimizing $z$ code yields much better, yet rougher result of completion, and refining the map by regressing the $\Delta w$ greatly improves fitting of the details.

In more detail, during the first $z$ optimization step, we optimize the following loss:

\begin{align*}
    \mathcal{L}_{opt}^{z}&(z | U_{gt}, \boldsymbol{\lambda})  \\
    & = \lambda^\textrm{LPIPS} \cdot \textrm{LPIPS}(g(map(z)) \cdot M, U_{gt} \cdot M)  \\
    & + \lambda^{L_1} \cdot L_1(g(map(z)) \cdot M, U_{gt} \cdot M),
\end{align*}

Similarly to the $\mathcal{L}_{opt}^{full}$, we use $\lambda^\textrm{LPIPS}=0.1$ and $\lambda^{L_1}=3$. 
The $z$ is initialized from $\mathcal{N}(0, \mathbb{I})$ and further optimized by Adam with the learning rate of $10^{-2}$ for 500 steps.
Here and further, $\textrm{LPIPS}(\cdot, \cdot)$ and $L_1(\cdot, \cdot)$ follow the same expressions as for the full scan fitting.

During the second $\Delta w$ optimization step, we optimize a similar expression with a few additional terms:

\begin{align*}
    \mathcal{L}_{opt}^{\Delta w}&(\Delta w | z, U_{gt}, \boldsymbol{\lambda})  \\
    & = \lambda^\textrm{LPIPS} \cdot \textrm{LPIPS}(g(map(z) + \Delta w) \cdot M, U_{gt} \cdot M)  \\
    & + \lambda^{L_1} \cdot L_1(g(map(z) + \Delta w) \cdot M, U_{gt} \cdot M) \\
    & + \lambda_\textrm{preserve}^\textrm{LPIPS} \cdot \textrm{LPIPS}(g(map(z) + \Delta w) \cdot (1 - M), \\
    & \hspace{2.77cm} g(map(z) \cdot (1 - M)) \\
    & + \lambda_\textrm{preserve}^{L_1} \cdot L_1(g(map(z) + \Delta w) \cdot (1 - M), \\
    & \hspace{2.25cm} g(map(z)) \cdot (1 - M)), \\
\end{align*}

\noindent where $M^\textrm{face}$ is a predefined mask of the face region in the UV space, reduced to the circle including the eyes, nose and mouth.

The third and second ``preserve'' terms are introduced to not let the map guided by the $\Delta w$  optimization deviate much from the output corresponding to the regressed $z$ in non-visible regions, which is essential due to the tendency of convergence to the average shape there when optimizing in the $\mathcal{W}+$ space.
$\lambda^\textrm{LPIPS}=0.1$ and $\lambda^{L_1}=3$ remain the same as before, and $\lambda_\textrm{preserve}^\textrm{LPIPS}=0.01$ and $\lambda_\textrm{preserve}^{L_1}=0.3$ are selected $10\times$ less.
%
%
%
The optimization is carried out by Adam with the same learning rate of $10^{-2}$ for 500 steps.
The $\Delta w$ is initialized with zeros.

Finally, we optimize the StyleGAN noise to improve the details in the visible part.
Despite that we consider this step optional, we found that it helps reconstruct more detail even for a sparse cloud.
We optimize the same expression as $\mathcal{L}_{opt}^z$, with the difference that it is only being optimized w.r.t. the StyleGAN noise tensors (only in the scalp region).
The only modification is the introduced regularization that equals to the sum of the noise tensors L2 norms.
The optimization is carried out by Adam with the same learning rate of $10^{-2}$ for 500 steps.
The coefficient of this regularization is equal to $10^{-5}$.

In the Supplementary Video, we demonstrate more results of fitting the latent to the point clouds with different sparsity.

\mypara{Use case: Kinect data.} 
We demonstrate that the HeadCraft pipeline can be applied to real-world depth scans in Fig.~\ref{fig:kinect}.
For a sample RGB-D image captured by Kinect, we first convert it into a mesh by unprojecting the points with color and connecting the vertices by the triangles constructed from the image pixels.
The resulting mesh with vertex colors is rendered onto three views (frontal, slight left, and slight right) to obtain the facial landmarks from each side via an image-based facial landmark detector~\cite{bulat2017far} and aggregate them (jawline landmarks are obtained from slight left and slight right and the others from the frontal rendering).
FLAME is fitted using these landmarks, and the displacements for the visible part of the mesh are obtained via the partial registration procedure described above and later baked into the UV displacement map.
We also show the result of fitting the latent representation of HeadCraft to the visible part of the scan.
Compared to the aforementioned procedure, we omit LPIPS loss to give L1 more relevance in predicting coarse shape and only supervise the latent in the scalp region.

\mypara{Animation}. 
Here we expand on more details regarding applying displacements to a template, deforming over time.
Compared to the simple unconditional scenario, where the displacements are also applied to a certain FLAME template, we have to introduce two key differences.

First, as mentioned in Subsec.~\ref{subsec:supmat_registration}, to apply the displacements to the template, we apply Butterfly subdivision, the MeshLab implementation of which also smooths the surface.
However, the result of Butterfly is not consistent over various FLAME templates and yields a bit different number of vertices every time.
To solve that, we come up with \emph{consistent subdivision}, i.e. the way to construct the same topology for every FLAME.
To do that, we first apply Butterfly subdivision to an arbitrary scan, and for each vertex after the subdivision, we find which triangle of the original template it belongs to and the barycentric coordinates w.r.t. that triangle.
Later, for every new template, the locations of the subdivided vertices are evaluated based on these triangles and barycentric coordinates.
To handle the seam accurately, we consider each vertex of every triangle after subdivision individually, thus accounting for the duplicate vertices.

An artifact of such procedure is that the smoothness of the surface, introduced in the MeshLab implementation of Butterfly subdivision, cannot be trivially transferred onto a new mesh this way.
Because of this, the surface normals remain the same within the large triangles of the original template even after the subdivision, creating a non-appealing ``tiling'' effect.
To mitigate that, we apply Laplacian smoothing~\cite{vollmer1999improved} in its classical version to smooth the surface.
In order to account for important subtle parts, we apply a different number of Laplacian smoothing iterations to different regions, namely, 3 times to the lips region, 5 times to the face skin (face except mouth, eyeballs and eye surroundings), and 10 times to the scalp and the neck.
Since the realism of mouth, ears, and eyeballs is important for animation, they remain intact.

Second, as mentioned in the main text, we rotate the displacements according to the rotation of the surface normals of the template.
To do that, we first estimate the local basis of the\linebreak TBN space~\cite{learn_opengl_tbn} for each FLAME in a sequence.
This basis defines the normalized tangent $\boldsymbol{t_i^k}$, bitangent $\boldsymbol{b_i^k}$, and normal $\boldsymbol{n_i^k}$, pre-estimated for the $i$-th vertex of the FLAME template $F^k = FLAME(\textrm{shape}, \textrm{exp}^k, \textrm{jaw}^k,$ $\textrm{headpose}^k)$.
In addition, we estimate the TBN basis $(\boldsymbol{t_i^\textrm{neutral}}, \boldsymbol{b_i^\textrm{neutral}},$ $\boldsymbol{n_i^\textrm{neutral}})$ for a FLAME, corresponding to the same person and a neutral expression and pose $F^\textrm{neutral} = FLAME(\textrm{shape}, \boldsymbol{0}, \boldsymbol{0}, \boldsymbol{0})$.
The displacements $D$, queried from the generated UV map $U$, are first transferred from the object space into the neutral TBN space:
$$ D^\textrm{TBN} = \left( (\boldsymbol{t_i^\textrm{neutral}} \cdot \boldsymbol{d_i}), (\boldsymbol{b_i^\textrm{neutral}} \cdot \boldsymbol{d_i}), (\boldsymbol{n_i^\textrm{neutral}} \cdot \boldsymbol{d_i}) \right)_{i=1}^{|D|}$$

Then, for each of the sequence frames, we transfer them into object space, this time w.r.t. the TBN basis of the given frame:

$$ D^\textrm{object}_k = \left(
\begin{bmatrix}
    \boldsymbol{t_i^k}\,\,\,\, \boldsymbol{b_i^k}\,\,\,\, \boldsymbol{n_i^k}
\end{bmatrix} \cdot \boldsymbol{d_i^\textrm{TBN}} 
\right)_{i=1}^{|D|}$$

\noindent (the $\boldsymbol{t_i^k},\, \boldsymbol{b_i^k},\, \boldsymbol{n_i^k},\, \boldsymbol{d_i^\textrm{TBN}}$ vectors above treated as columns).

More examples of animations can be found in the Supplementary Video.




\begin{table*}[h!]
    \vspace{-0.3cm}
    \setlength{\tabcolsep}{0pt}
    \renewcommand{\arraystretch}{0}
    \centering
    \begin{tabular}{lccccccc}
        \raisebox{2\normalbaselineskip}[0pt][0pt]{\rotatebox[origin=c]{90}{Frontal}} & 
        \includegraphics[width=0.12\textwidth,trim={0cm 2cm 1cm 0cm},clip]{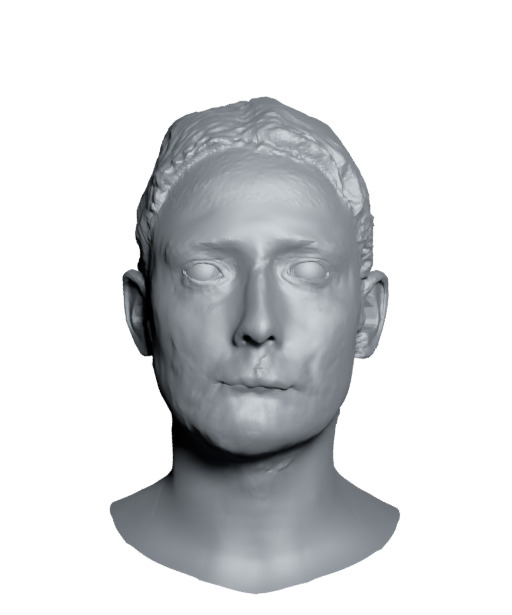} &
        \includegraphics[width=0.12\textwidth,trim={0cm 2cm 1cm 0cm},clip]{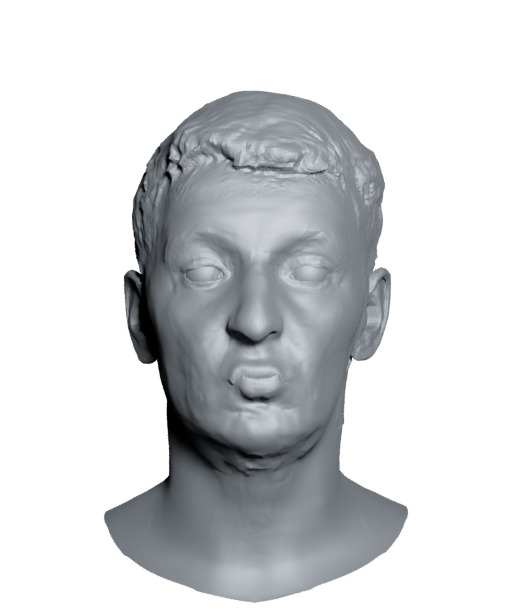} & 
        \includegraphics[width=0.12\textwidth,trim={0cm 2cm 1cm 0cm},clip]{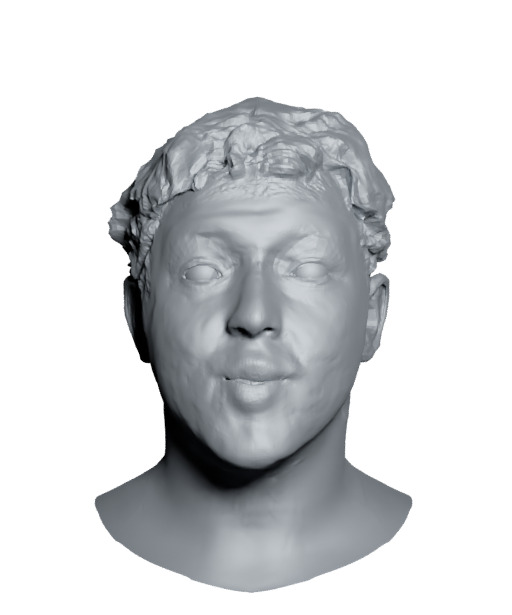} & 
        \includegraphics[width=0.12\textwidth,trim={0cm 2cm 1cm 0cm},clip]{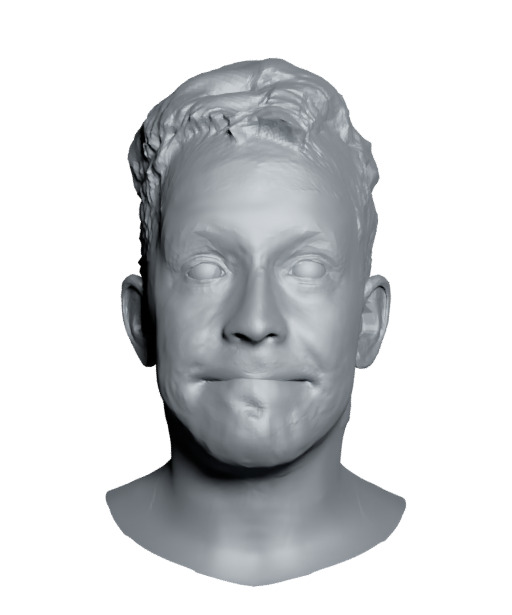} & 
        \includegraphics[width=0.12\textwidth,trim={0cm 2cm 1cm 0cm},clip]{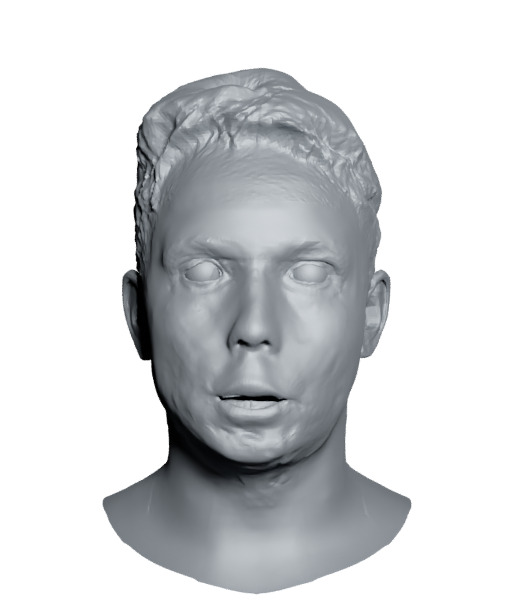} & 
        \includegraphics[width=0.12\textwidth,trim={0cm 2cm 1cm 0cm},clip]{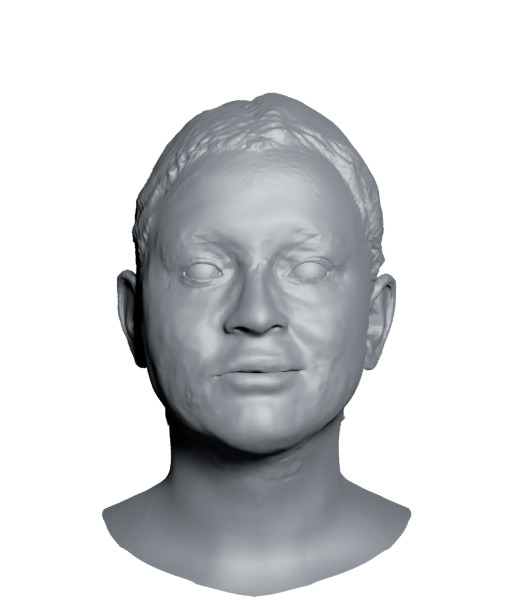} & 
        \includegraphics[width=0.12\textwidth,trim={0cm 2cm 1cm 0cm},clip]{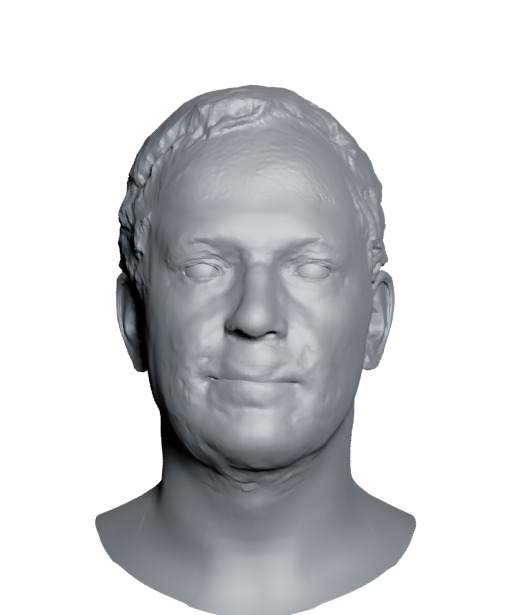} \\
        \raisebox{2.6\normalbaselineskip}[0pt][0pt]{\rotatebox[origin=c]{90}{Left}} & 
        \includegraphics[width=0.12\textwidth,trim={0cm 2cm 1cm 0cm},clip]{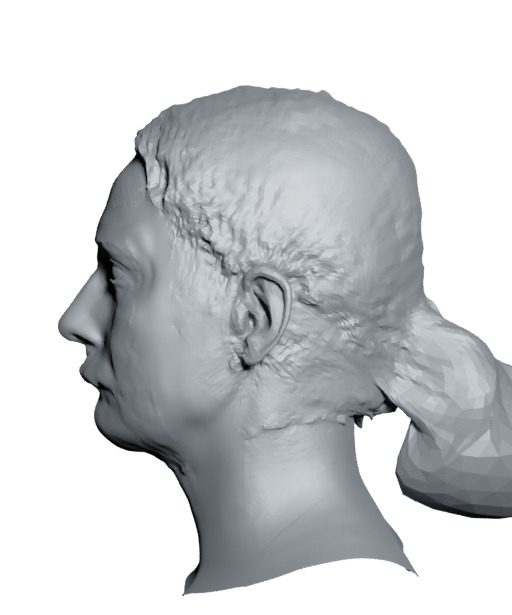} &
        \includegraphics[width=0.12\textwidth,trim={0cm 2cm 1cm 0cm},clip]{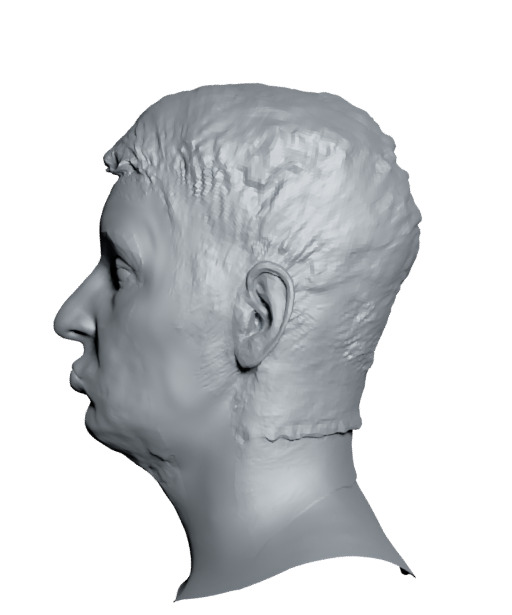} & 
        \includegraphics[width=0.12\textwidth,trim={0cm 2cm 1cm 0cm},clip]{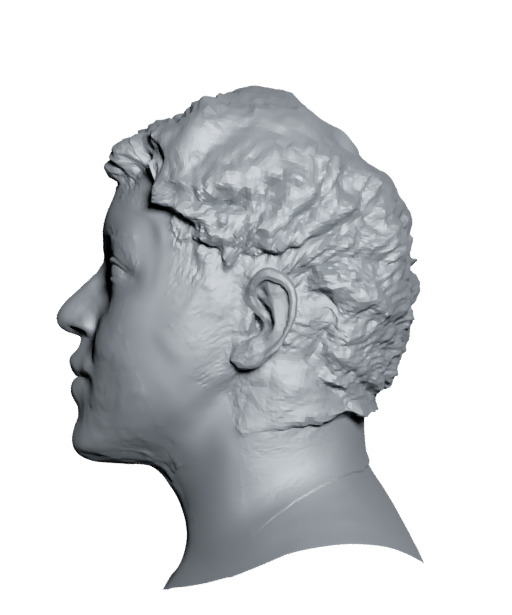} & 
        \includegraphics[width=0.12\textwidth,trim={0cm 2cm 1cm 0cm},clip]{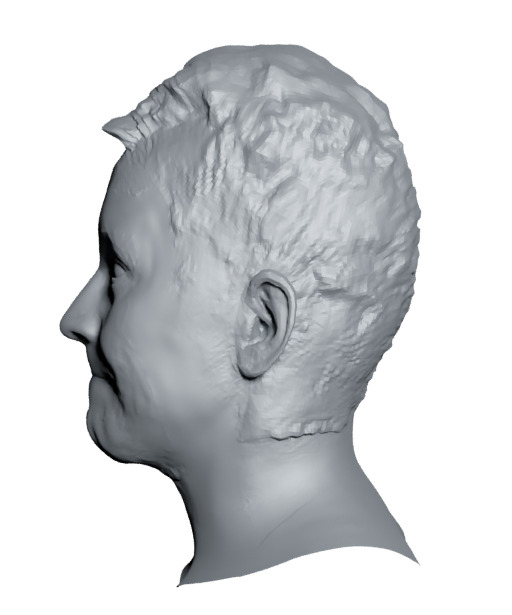} & 
        \includegraphics[width=0.12\textwidth,trim={0cm 2cm 1cm 0cm},clip]{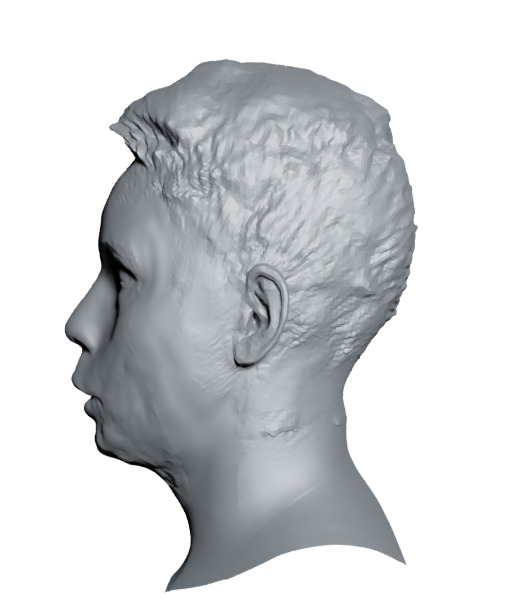} & 
        \includegraphics[width=0.12\textwidth,trim={0cm 2cm 1cm 0cm},clip]{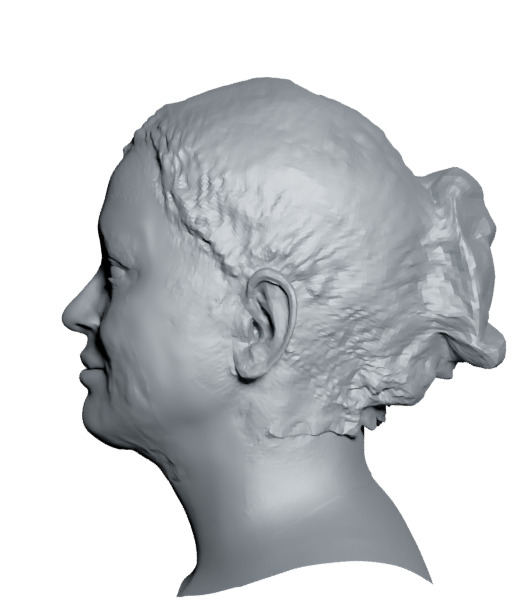} & 
        \includegraphics[width=0.12\textwidth,trim={0cm 2cm 1cm 0cm},clip]{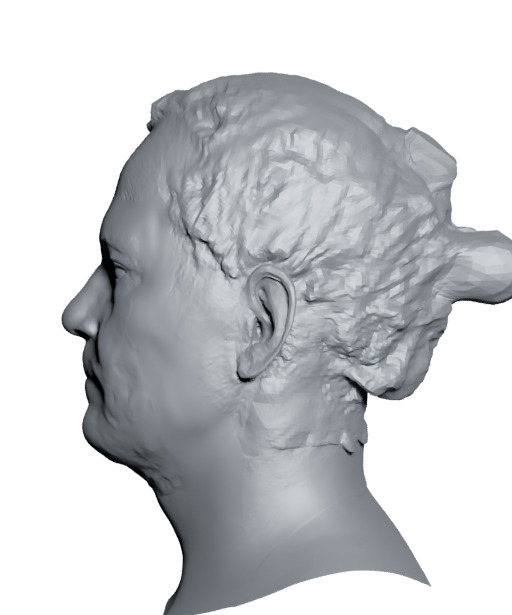} \\
        \raisebox{2.3\normalbaselineskip}[0pt][0pt]{\rotatebox[origin=c]{90}{Right}} & 
        \includegraphics[width=0.12\textwidth,trim={0cm 2cm 1cm 0cm},clip]{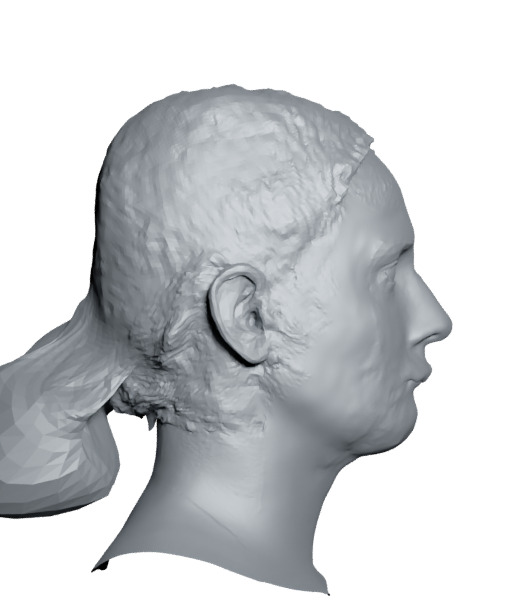} &
        \includegraphics[width=0.12\textwidth,trim={0cm 2cm 1cm 0cm},clip]{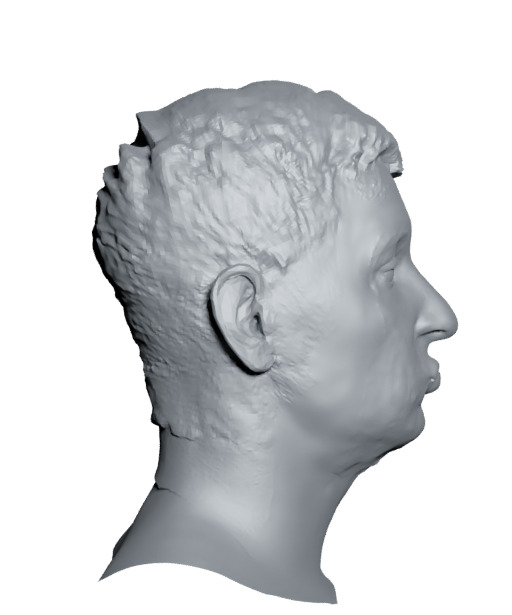} & 
        \includegraphics[width=0.12\textwidth,trim={0cm 2cm 1cm 0cm},clip]{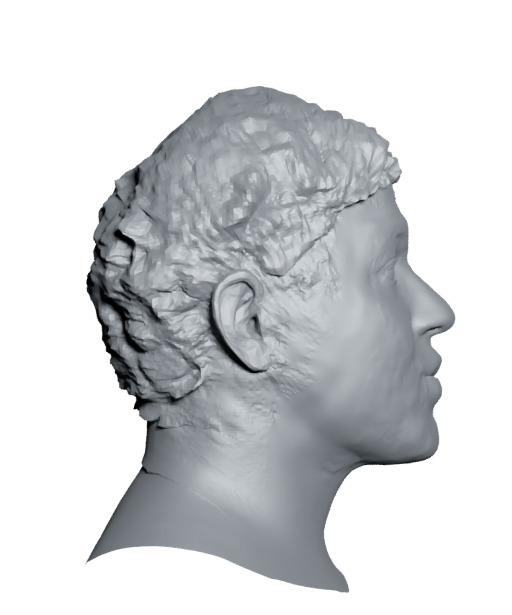} & 
        \includegraphics[width=0.12\textwidth,trim={0cm 2cm 1cm 0cm},clip]{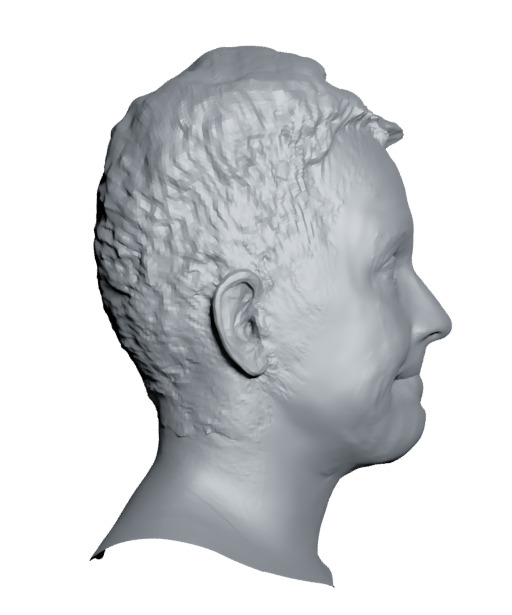} & 
        \includegraphics[width=0.12\textwidth,trim={0cm 2cm 1cm 0cm},clip]{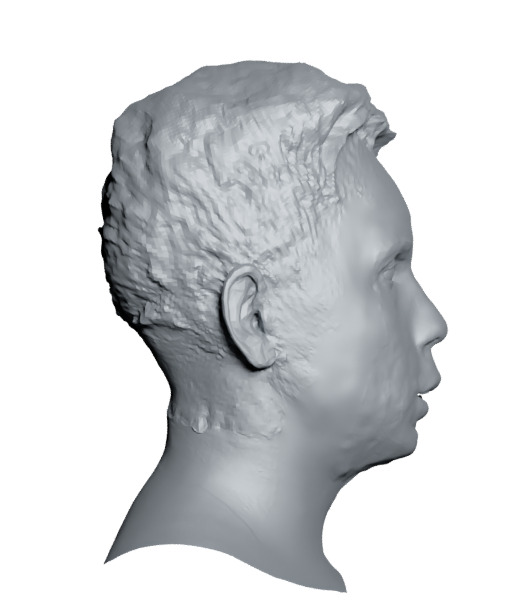} & 
        \includegraphics[width=0.12\textwidth,trim={0cm 2cm 1cm 0cm},clip]{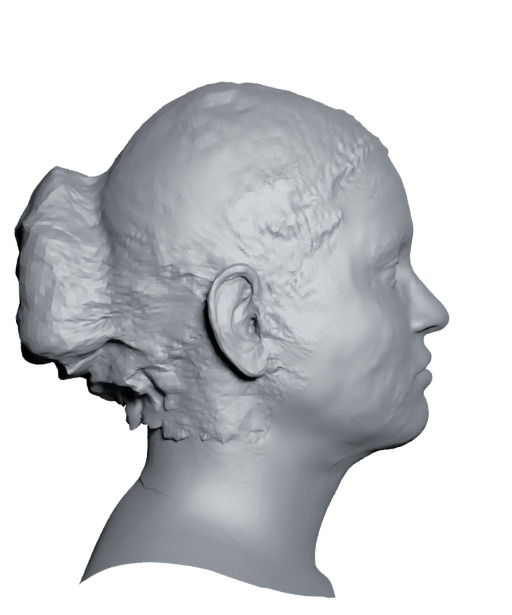} & 
        \includegraphics[width=0.12\textwidth,trim={0cm 2cm 1cm 0cm},clip]{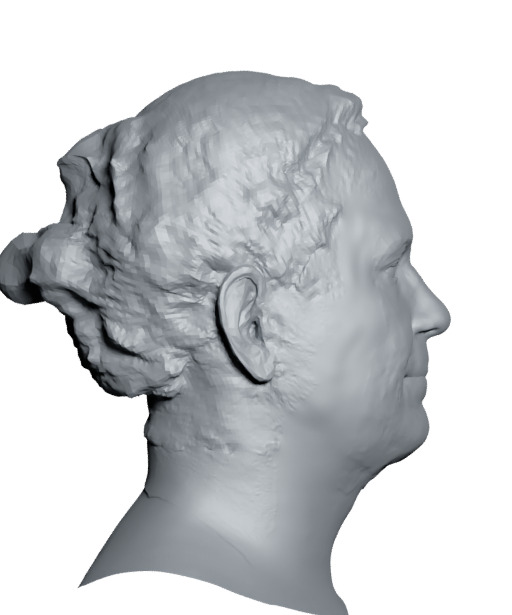} \\
        \raisebox{2.6\normalbaselineskip}[0pt][0pt]{\rotatebox[origin=c]{90}{Back}} &  
        \includegraphics[width=0.12\textwidth,trim={0cm 2cm 1cm 0cm},clip]{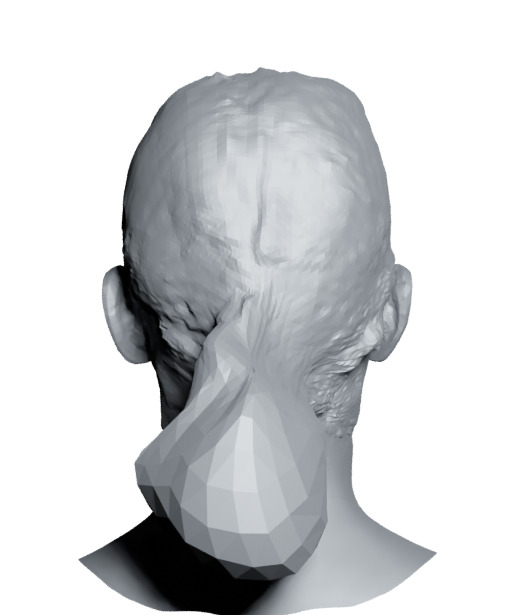} &
        \includegraphics[width=0.12\textwidth,trim={0cm 2cm 1cm 0cm},clip]{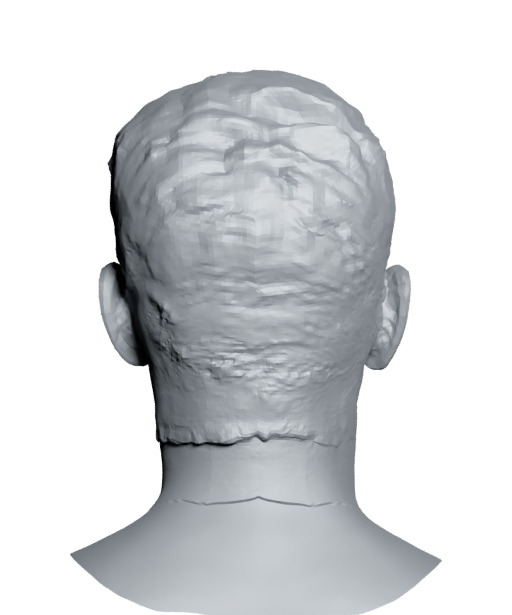} & 
        \includegraphics[width=0.12\textwidth,trim={0cm 2cm 1cm 0cm},clip]{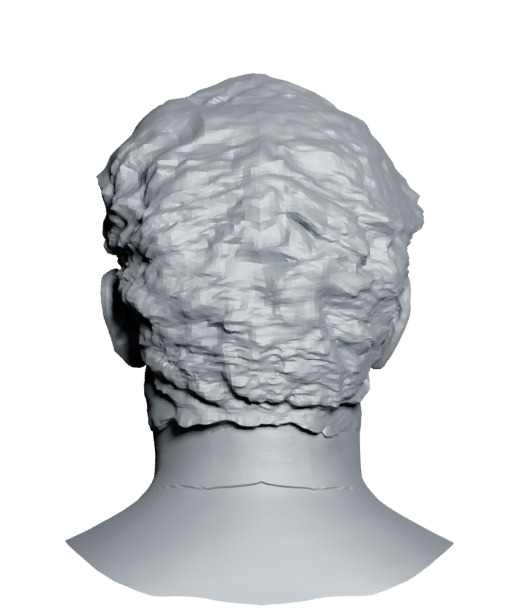} & 
        \includegraphics[width=0.12\textwidth,trim={0cm 2cm 1cm 0cm},clip]{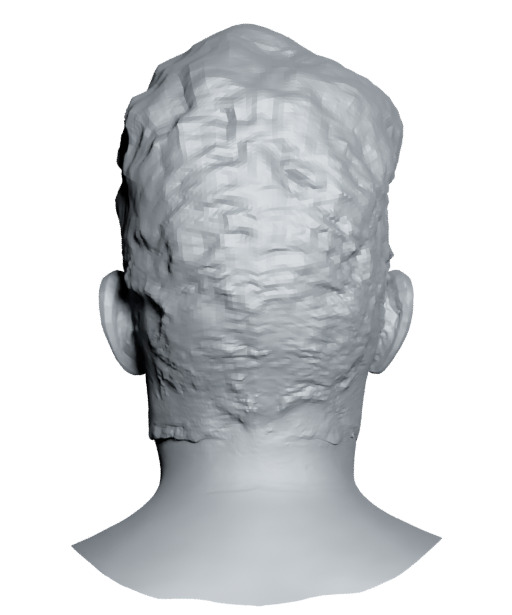} & 
        \includegraphics[width=0.12\textwidth,trim={0cm 2cm 1cm 0cm},clip]{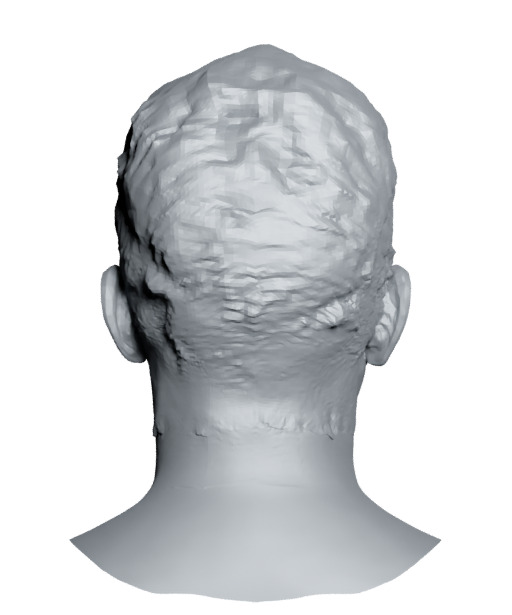} & 
        \includegraphics[width=0.12\textwidth,trim={0cm 2cm 1cm 0cm},clip]{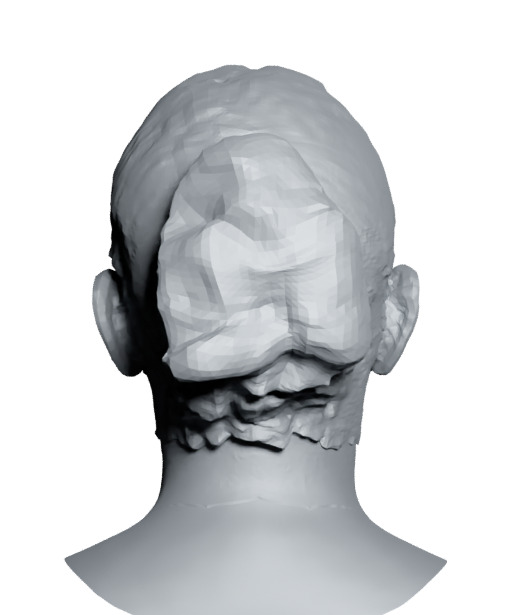} & 
        \includegraphics[width=0.12\textwidth,trim={0cm 2cm 1cm 0cm},clip]{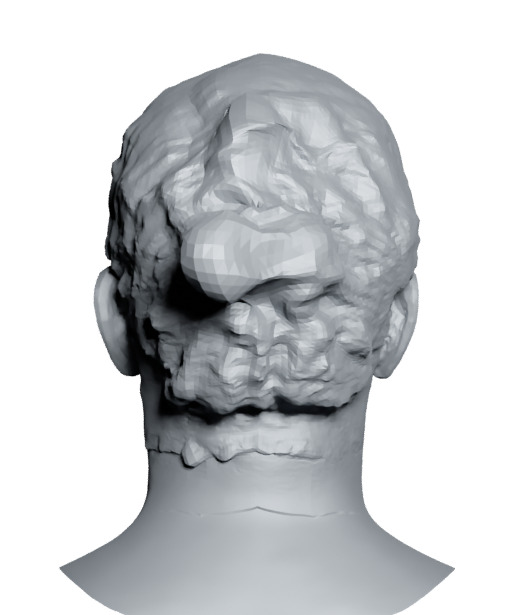} \\[0.5cm]
        \raisebox{2\normalbaselineskip}[0pt][0pt]{\rotatebox[origin=c]{90}{Frontal}} & 
        \includegraphics[width=0.12\textwidth,trim={0cm 2cm 1cm 0cm},clip]{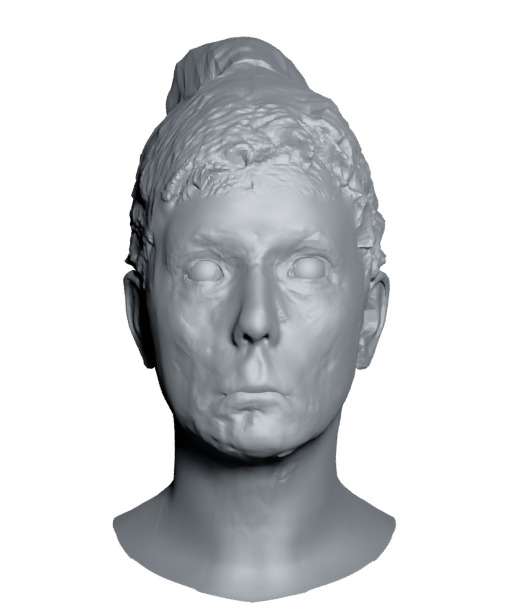} &
        \includegraphics[width=0.12\textwidth,trim={0cm 2cm 1cm 0cm},clip]{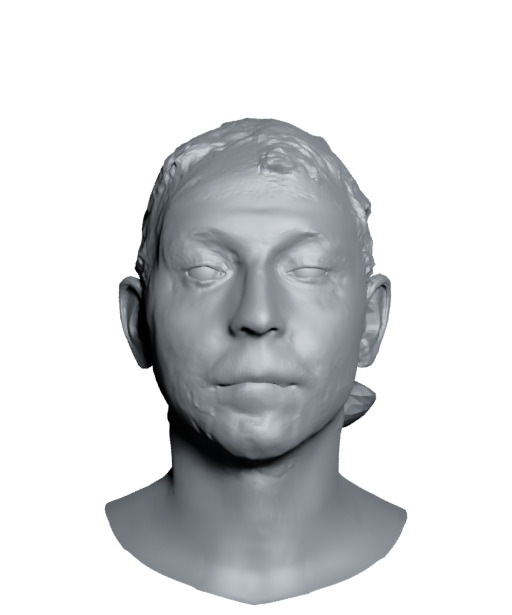} & 
        \includegraphics[width=0.12\textwidth,trim={0cm 2cm 1cm 0cm},clip]{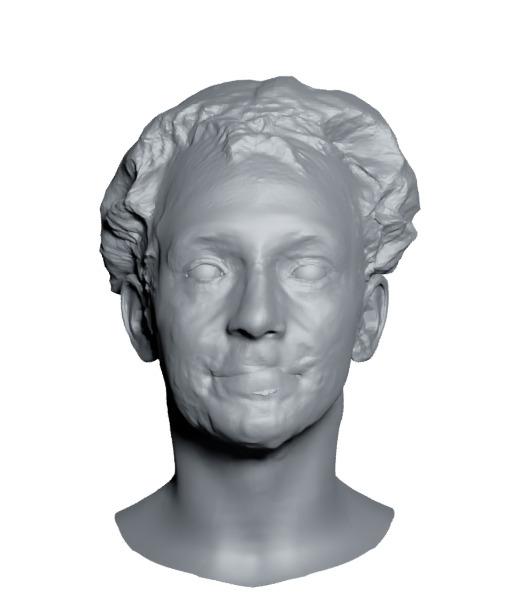} & 
        \includegraphics[width=0.12\textwidth,trim={0cm 2cm 1cm 0cm},clip]{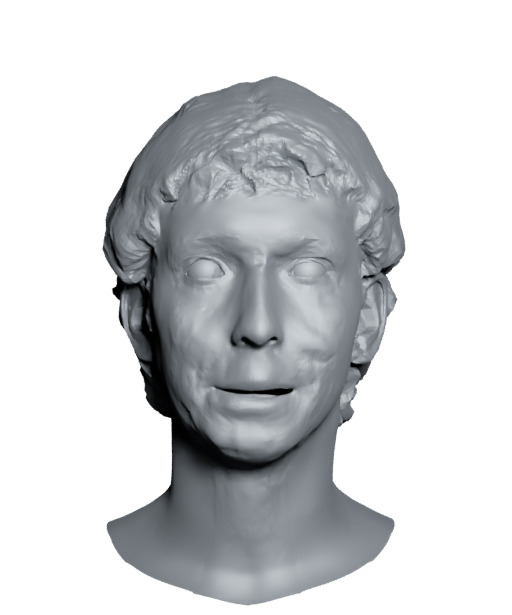} & 
        \includegraphics[width=0.12\textwidth,trim={0cm 2cm 1cm 0cm},clip]{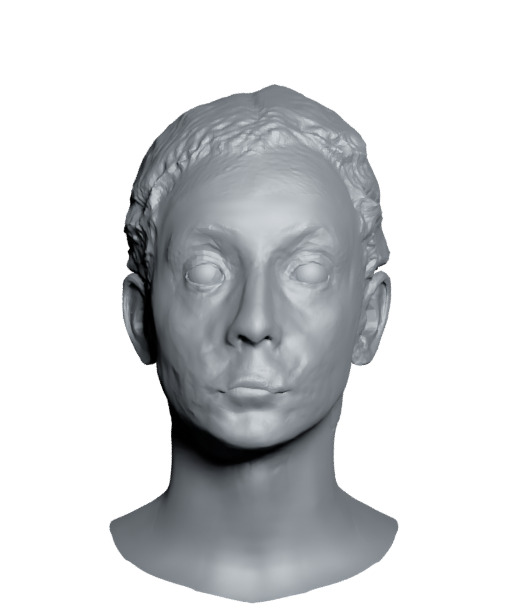} & 
        \includegraphics[width=0.12\textwidth,trim={0cm 2cm 1cm 0cm},clip]{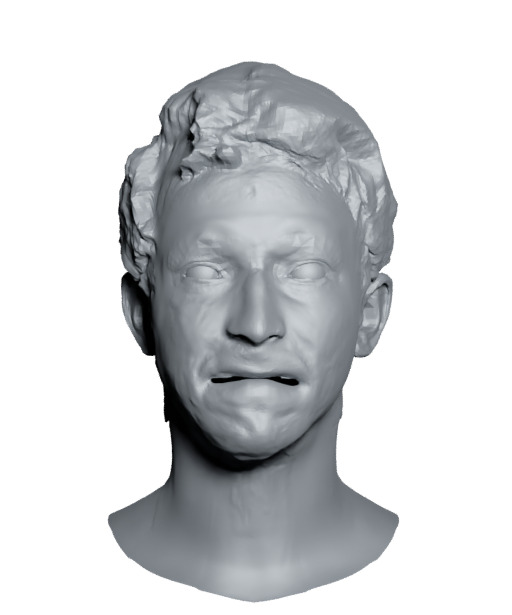} & 
        \includegraphics[width=0.12\textwidth,trim={0cm 2cm 1cm 0cm},clip]{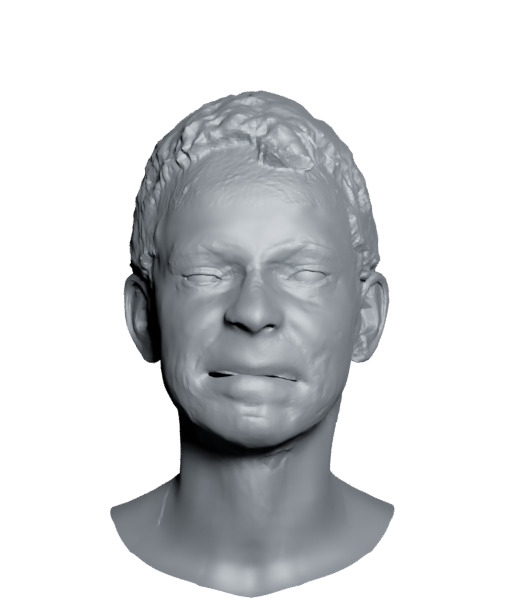} \\
        \raisebox{2.6\normalbaselineskip}[0pt][0pt]{\rotatebox[origin=c]{90}{Left}} & 
        \includegraphics[width=0.12\textwidth,trim={0cm 2cm 1cm 0cm},clip]{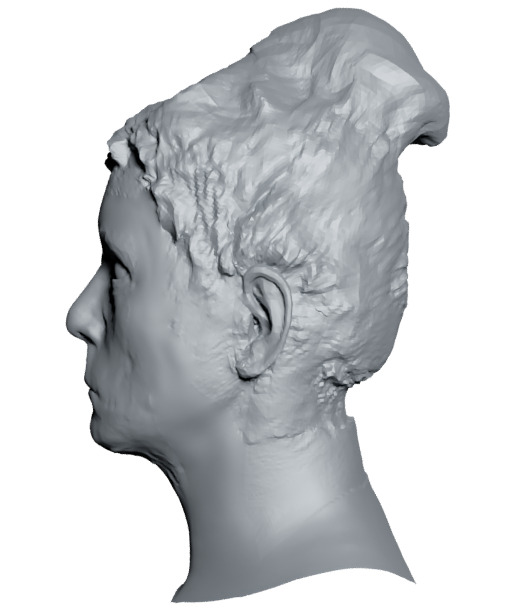} &
        \includegraphics[width=0.12\textwidth,trim={0cm 2cm 1cm 0cm},clip]{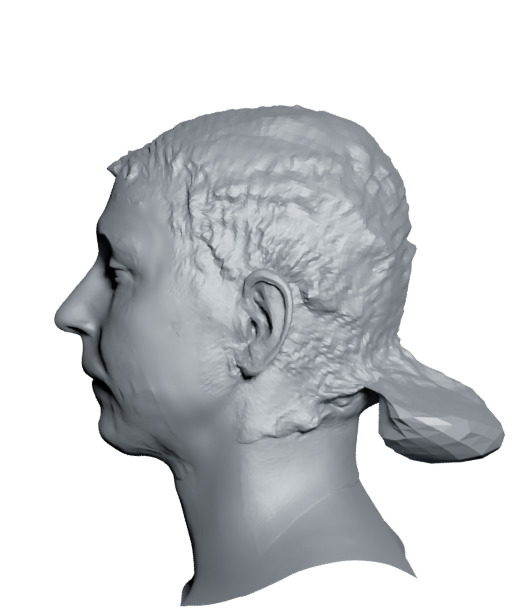} & 
        \includegraphics[width=0.12\textwidth,trim={0cm 2cm 1cm 0cm},clip]{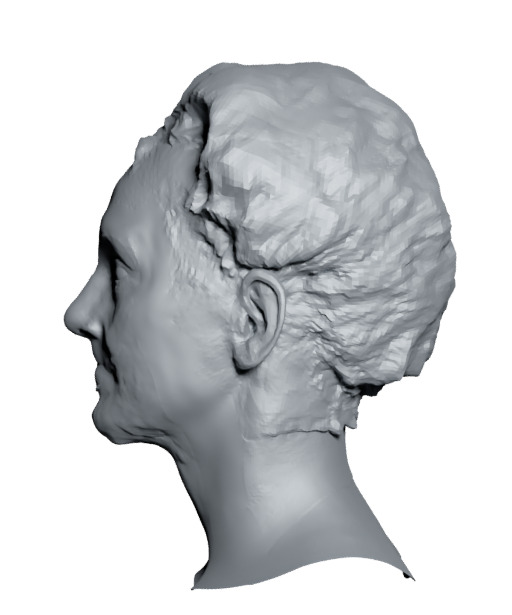} & 
        \includegraphics[width=0.12\textwidth,trim={0cm 2cm 1cm 0cm},clip]{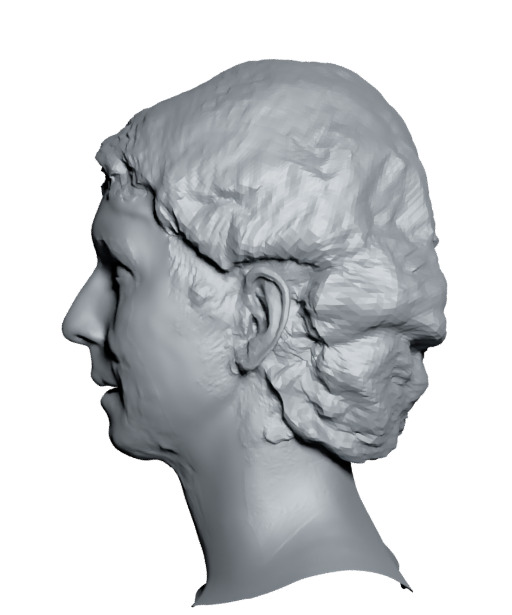} & 
        \includegraphics[width=0.12\textwidth,trim={0cm 2cm 1cm 0cm},clip]{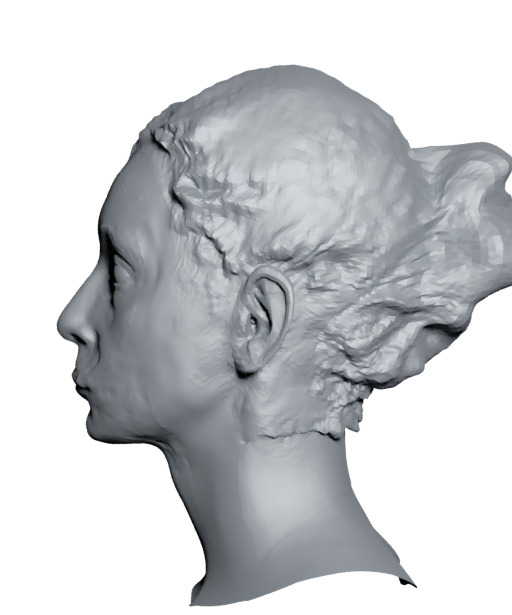} & 
        \includegraphics[width=0.12\textwidth,trim={0cm 2cm 1cm 0cm},clip]{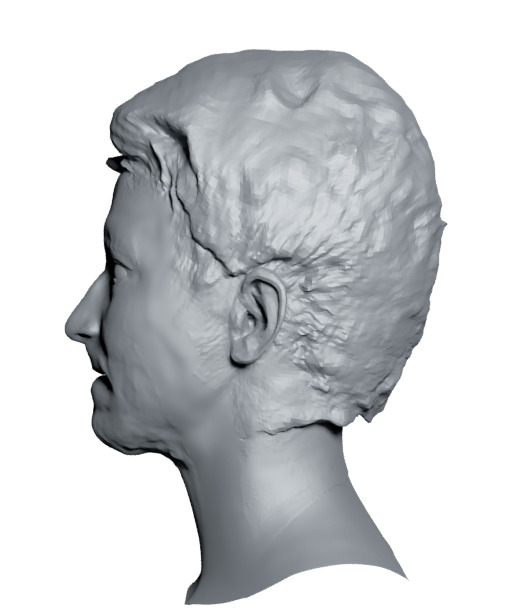} & 
        \includegraphics[width=0.12\textwidth,trim={0cm 2cm 1cm 0cm},clip]{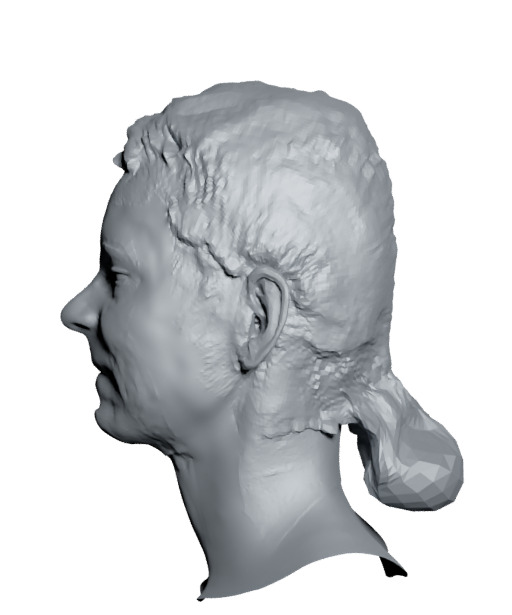} \\
        \raisebox{2.3\normalbaselineskip}[0pt][0pt]{\rotatebox[origin=c]{90}{Right}} & 
        \includegraphics[width=0.12\textwidth,trim={0cm 2cm 1cm 0cm},clip]{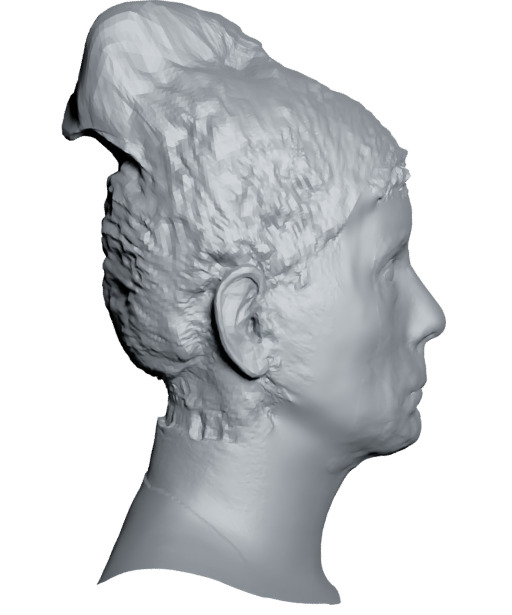} &
        \includegraphics[width=0.12\textwidth,trim={0cm 2cm 1cm 0cm},clip]{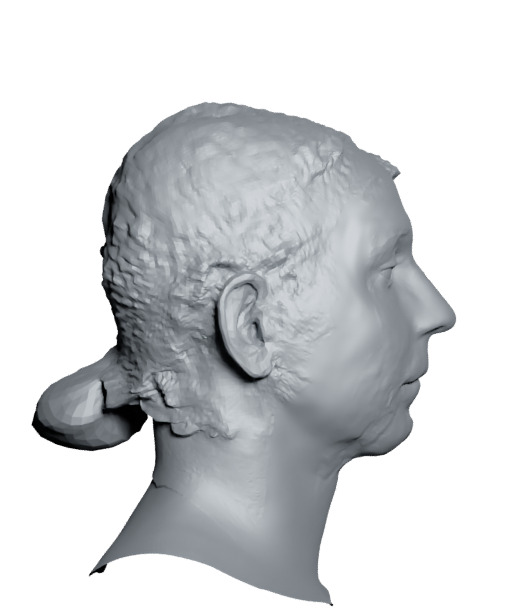} & 
        \includegraphics[width=0.12\textwidth,trim={0cm 2cm 1cm 0cm},clip]{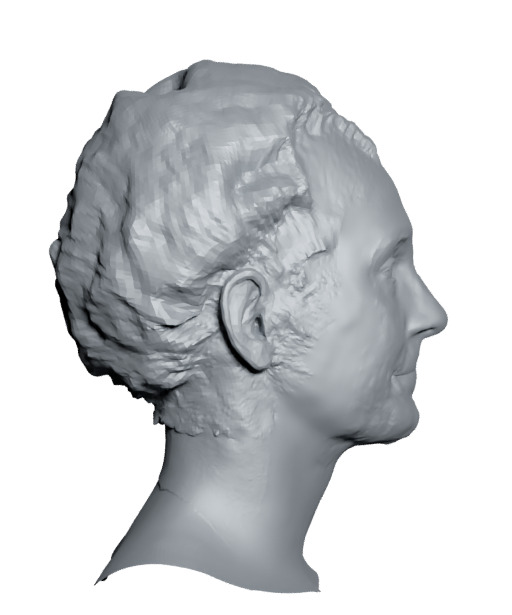} & 
        \includegraphics[width=0.12\textwidth,trim={0cm 2cm 1cm 0cm},clip]{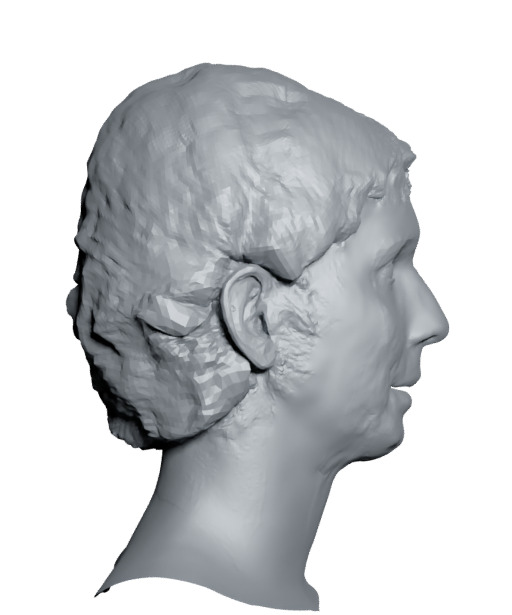} & 
        \includegraphics[width=0.12\textwidth,trim={0cm 2cm 1cm 0cm},clip]{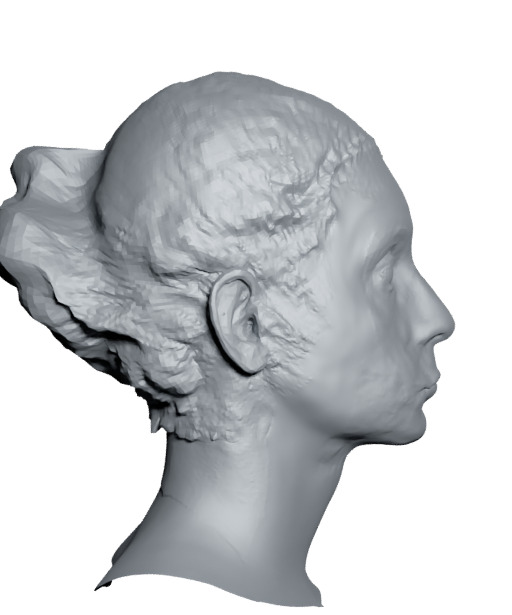} & 
        \includegraphics[width=0.12\textwidth,trim={0cm 2cm 1cm 0cm},clip]{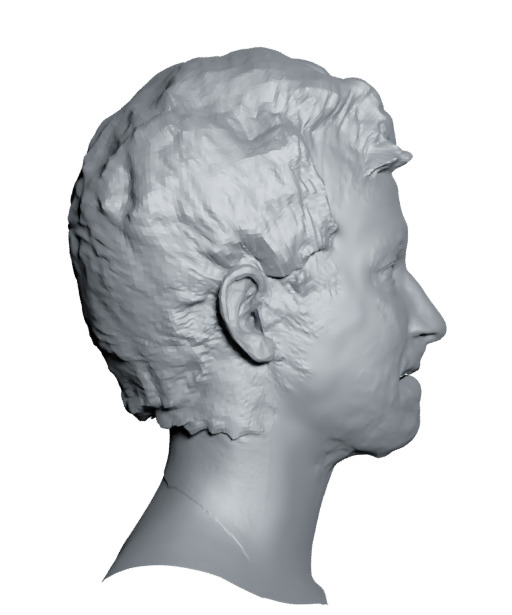} & 
        \includegraphics[width=0.12\textwidth,trim={0cm 2cm 1cm 0cm},clip]{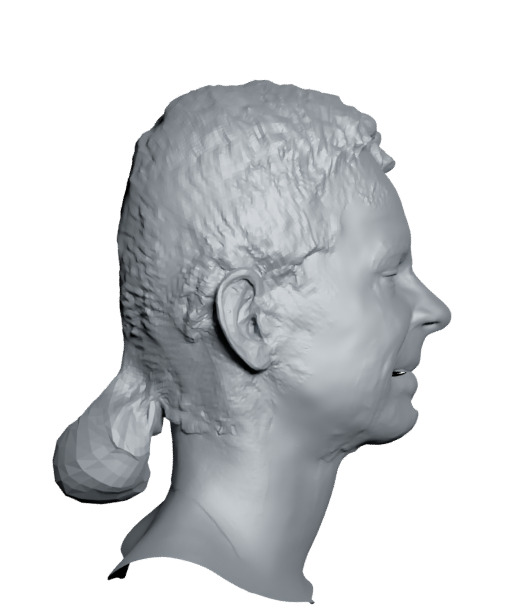} \\
        \raisebox{2.6\normalbaselineskip}[0pt][0pt]{\rotatebox[origin=c]{90}{Back}} &  
        \includegraphics[width=0.12\textwidth,trim={0cm 2cm 1cm 0cm},clip]{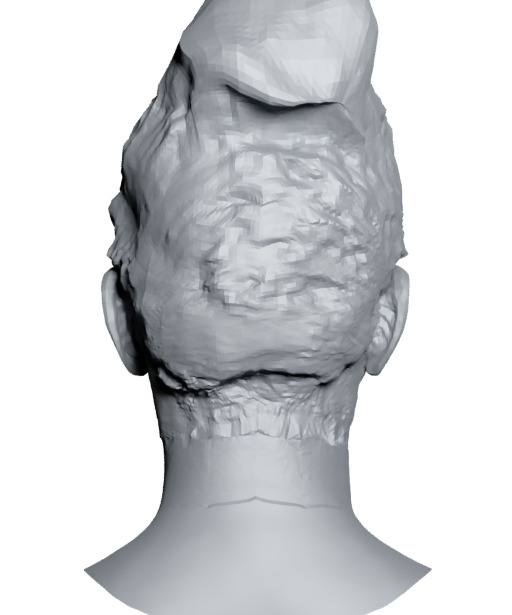} &
        \includegraphics[width=0.12\textwidth,trim={0cm 2cm 1cm 0cm},clip]{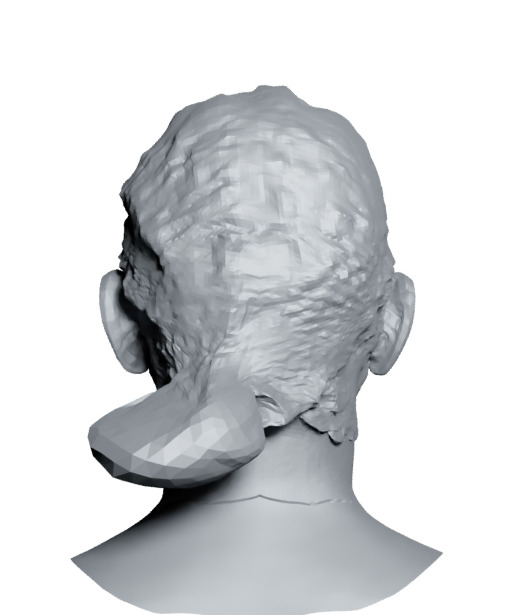} & 
        \includegraphics[width=0.12\textwidth,trim={0cm 2cm 1cm 0cm},clip]{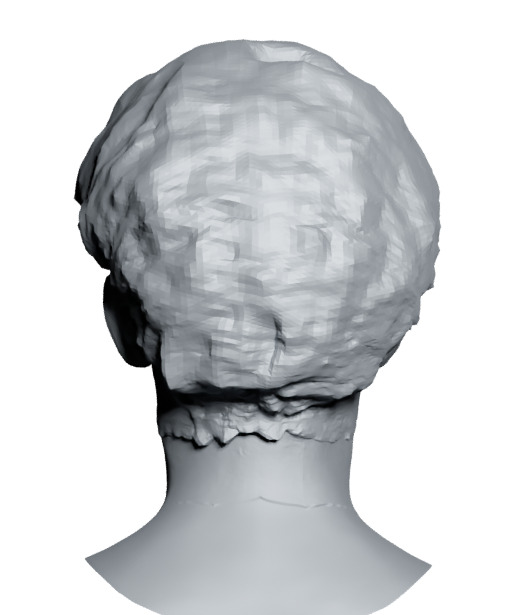} & 
        \includegraphics[width=0.12\textwidth,trim={0cm 2cm 1cm 0cm},clip]{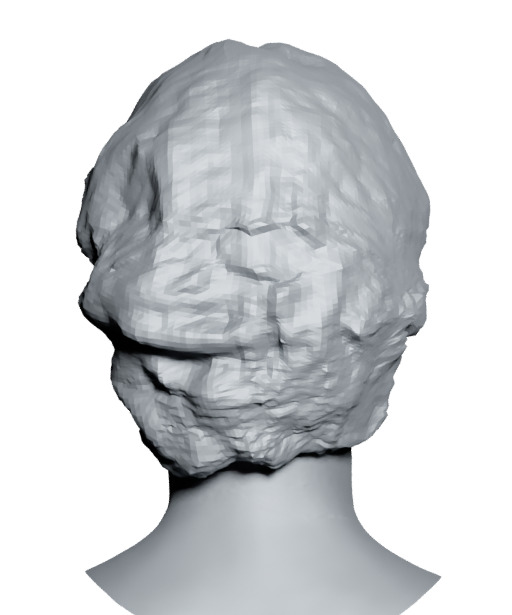} & 
        \includegraphics[width=0.12\textwidth,trim={0cm 2cm 1cm 0cm},clip]{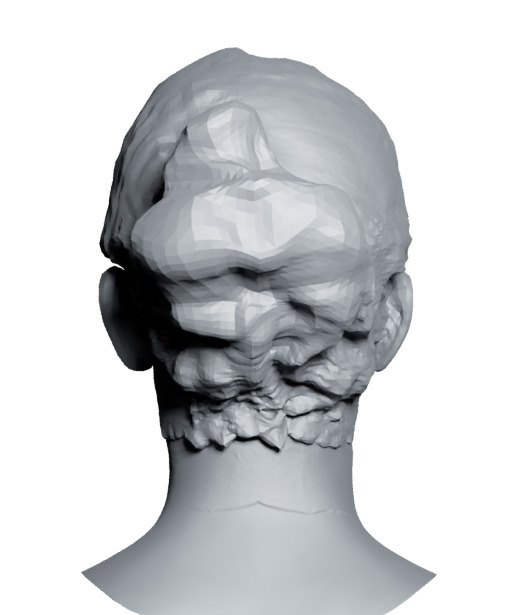} & 
        \includegraphics[width=0.12\textwidth,trim={0cm 2cm 1cm 0cm},clip]{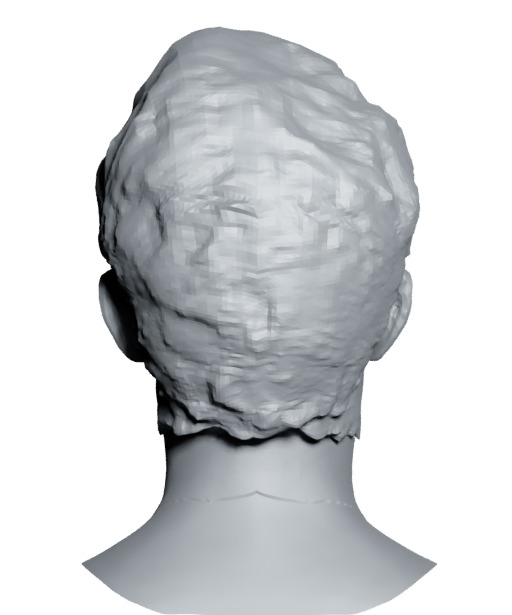} & 
        \includegraphics[width=0.12\textwidth,trim={0cm 2cm 1cm 0cm},clip]{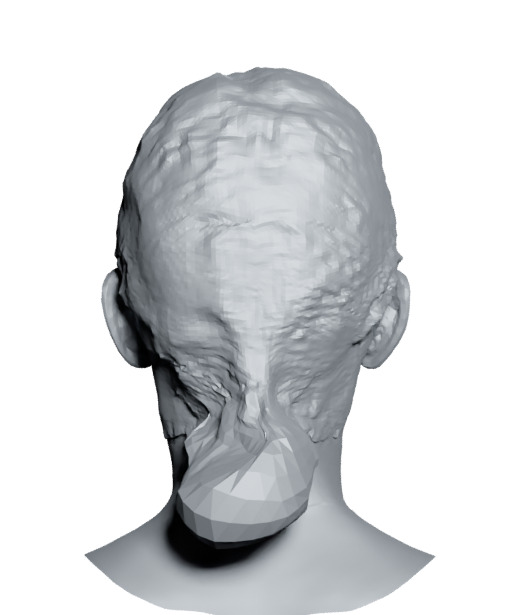} \\
    \end{tabular}
    \captionof{figure}{
    Additional demonstration of the diversity and level of detail of the unconditionally sampled generations from HeadCraft.
    %
    %
    The generations are obtained by randomly sampling $z \sim \mathcal{N}(0, \mathbb{I})$ latent code of the generative model.
    The displacements, returned by the model, are applied to the random FLAMEs sampled from Gaussian distribution with statistics calculated over the NPHM dataset.
    }
    \label{fig:uncond_more}
\end{table*}
\begin{table*}[]
    \setlength{\tabcolsep}{0pt}
    \renewcommand{\arraystretch}{0}
    \centering
    \begin{tabular}{lcccc}
        & FLAME & Stage 1 & Stage 2 & Ground truth \\
        \raisebox{4\normalbaselineskip}[0pt][0pt]{\rotatebox[origin=c]{90}{Frontal}} & 
          \includegraphics[width=0.18\textwidth,trim={0cm 1cm 0cm 1cm},clip]{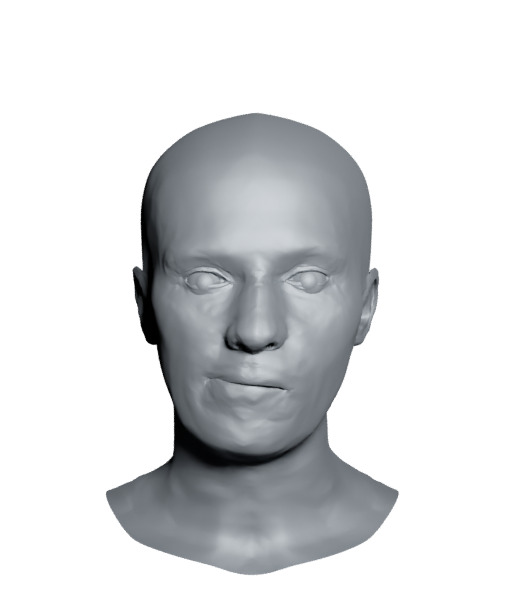} 
        & \includegraphics[width=0.18\textwidth,trim={0cm 1cm 0cm 1cm},clip]{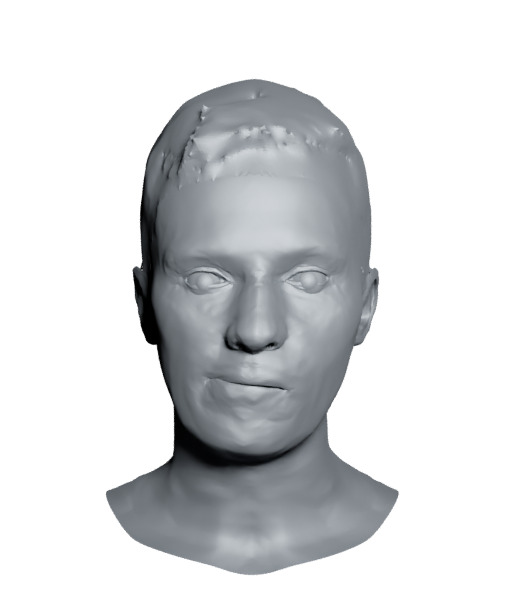}
        & \includegraphics[width=0.18\textwidth,trim={0cm 1cm 0cm 1cm},clip]{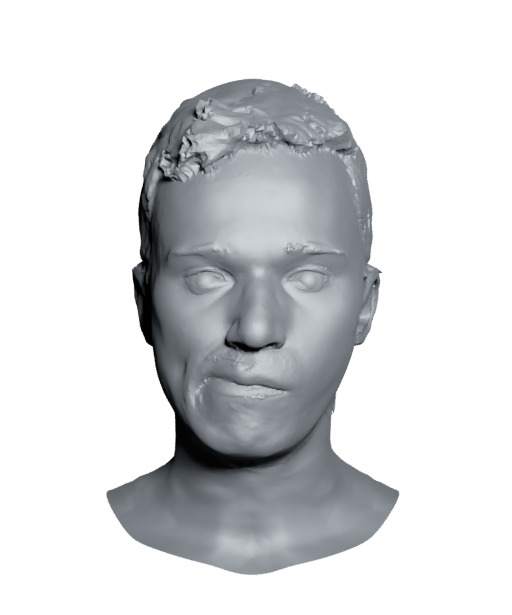}
        & \includegraphics[width=0.18\textwidth,trim={0cm 1cm 0cm 1cm},clip]{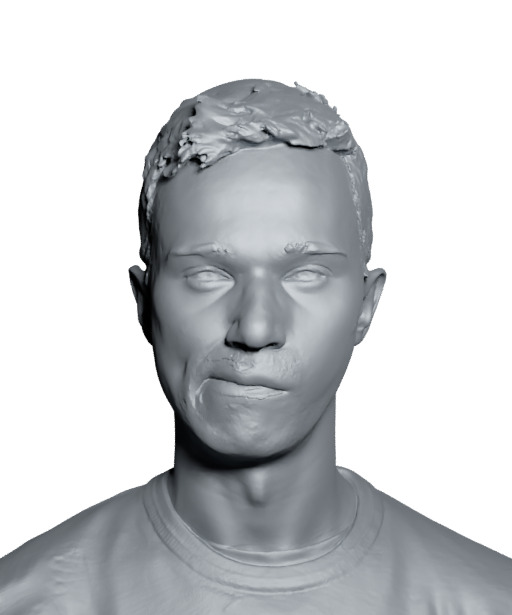}  \\
        \raisebox{4\normalbaselineskip}[0pt][0pt]{\rotatebox[origin=c]{90}{Left}} & 
          \includegraphics[width=0.18\textwidth,trim={0cm 1cm 0cm 1cm},clip]{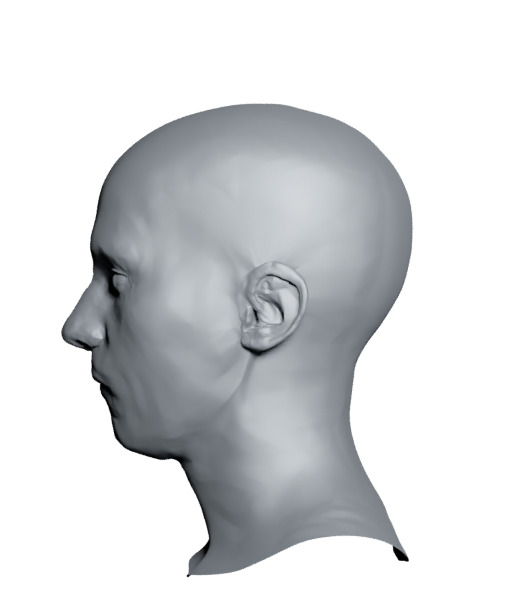} 
        & \includegraphics[width=0.18\textwidth,trim={0cm 1cm 0cm 1cm},clip]{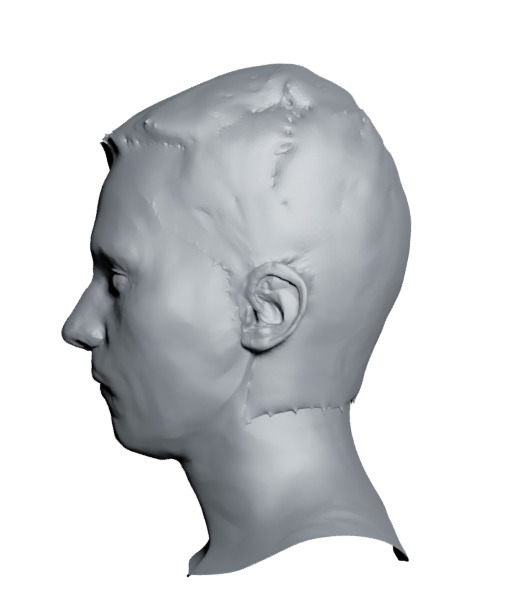}
        & \includegraphics[width=0.18\textwidth,trim={0cm 1cm 0cm 1cm},clip]{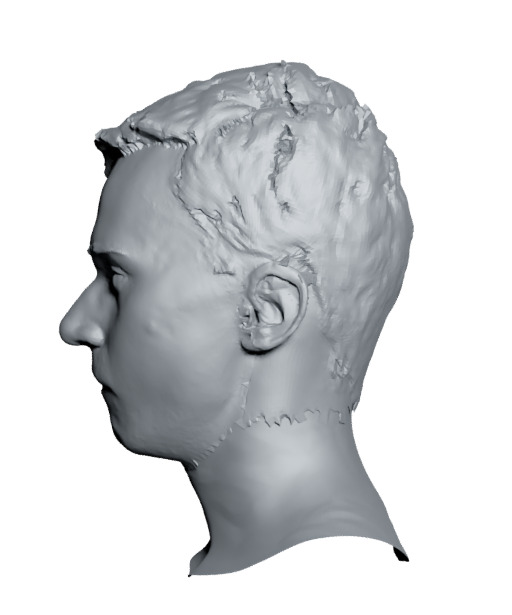}
        & \includegraphics[width=0.18\textwidth,trim={0cm 1cm 0cm 1cm},clip]{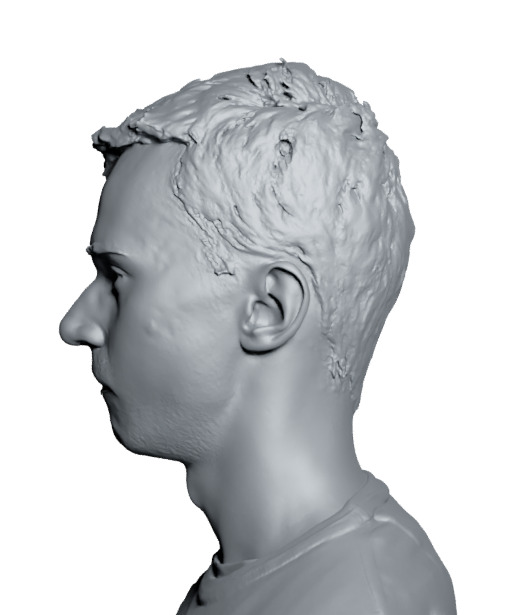}  \\
        \raisebox{4\normalbaselineskip}[0pt][0pt]{\rotatebox[origin=c]{90}{Top}} & 
          \includegraphics[width=0.18\textwidth,trim={0cm 3cm 0cm 3cm},clip]{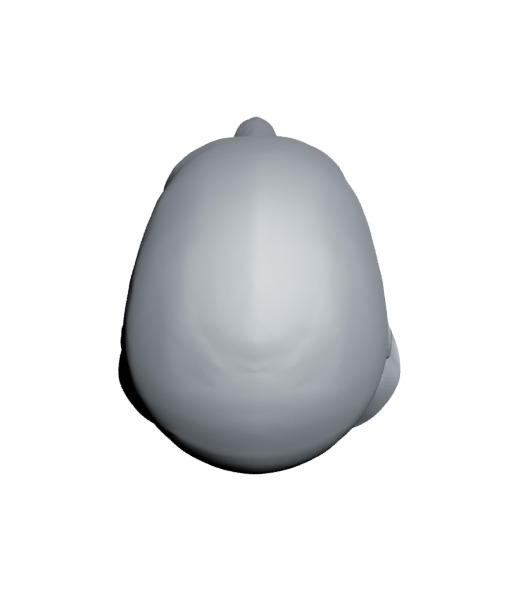} 
        & \includegraphics[width=0.18\textwidth,trim={0cm 3cm 0cm 3cm},clip]{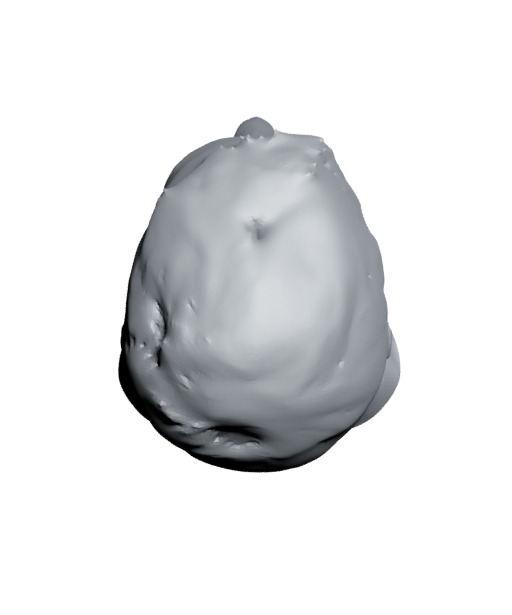}
        & \includegraphics[width=0.18\textwidth,trim={0cm 3cm 0cm 3cm},clip]{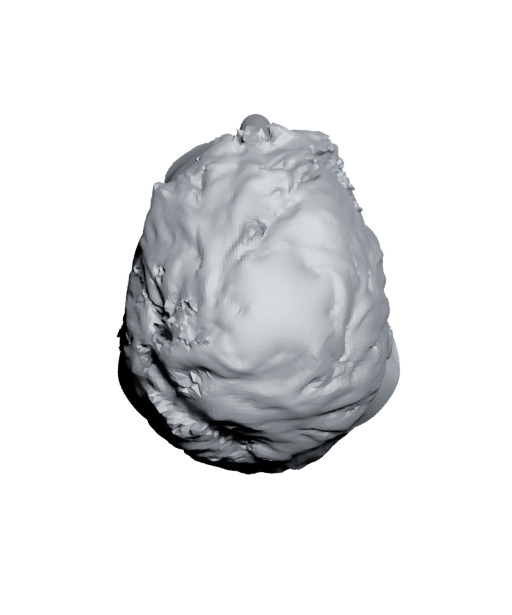}
        & \includegraphics[width=0.18\textwidth,trim={0cm 3cm 0cm 3cm},clip]{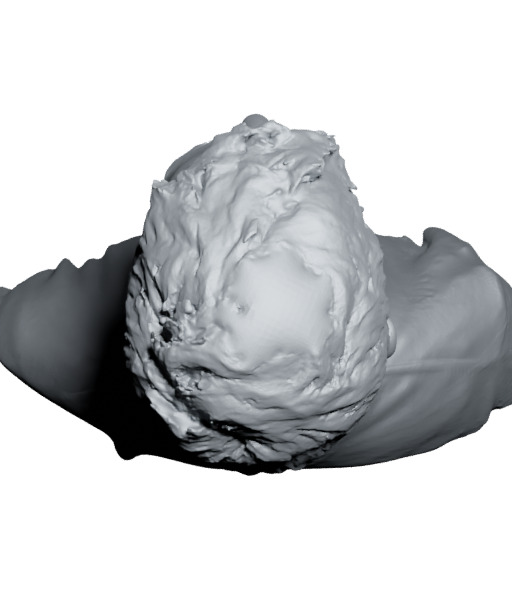}  \\[0.19cm]
        & FLAME & Stage 1 & Stage 2 & Ground truth \\
        \raisebox{4\normalbaselineskip}[0pt][0pt]{\rotatebox[origin=c]{90}{Frontal}} & 
          \includegraphics[width=0.18\textwidth,trim={0cm 1cm 0cm 1cm},clip]{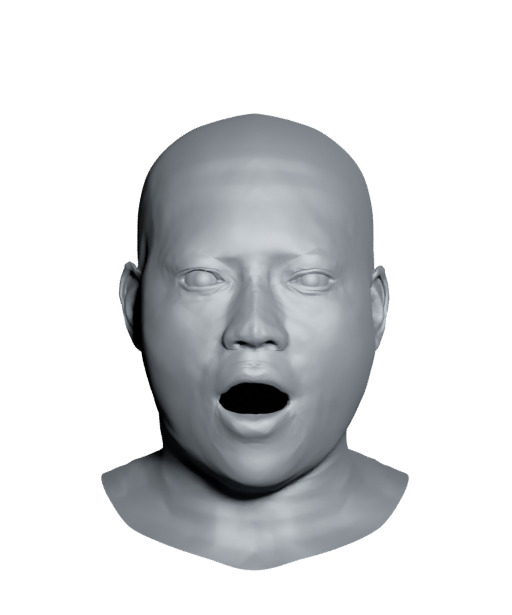} 
        & \includegraphics[width=0.18\textwidth,trim={0cm 1cm 0cm 1cm},clip]{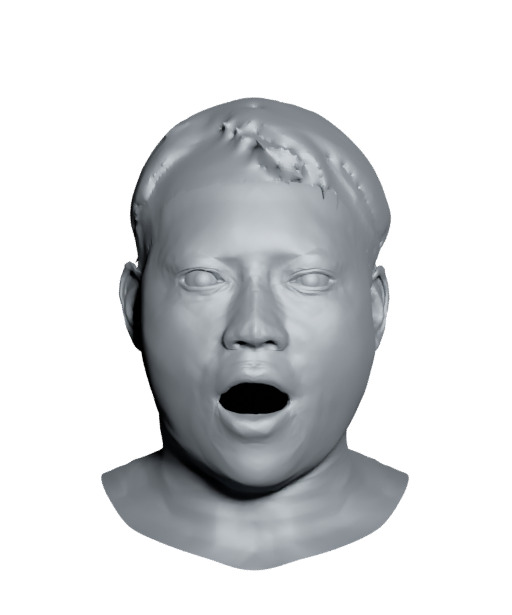}
        & \includegraphics[width=0.18\textwidth,trim={0cm 1cm 0cm 1cm},clip]{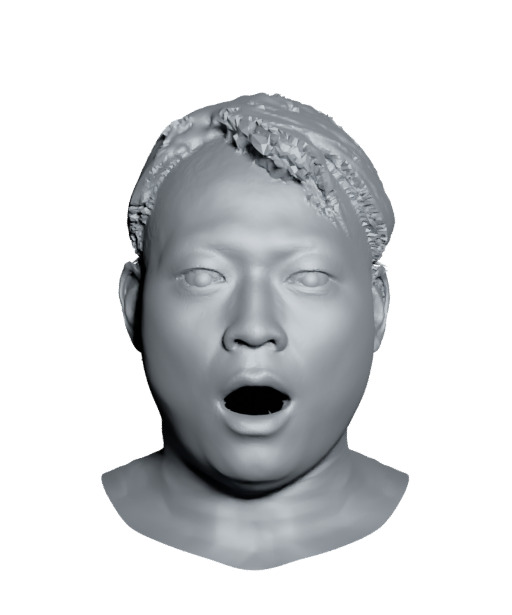}
        & \includegraphics[width=0.18\textwidth,trim={0cm 1cm 0cm 1cm},clip]{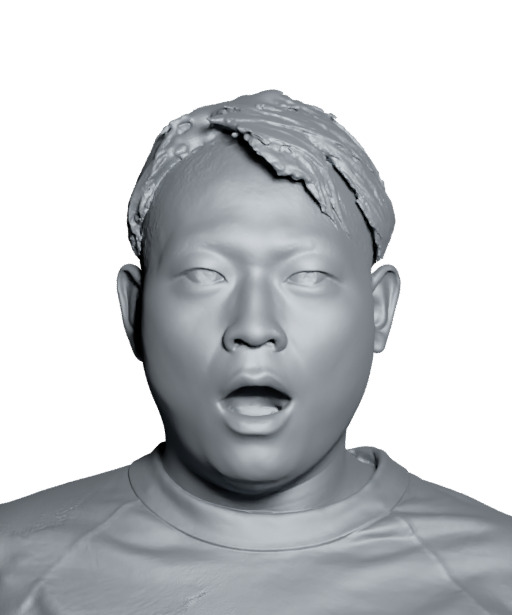}  \\
        \raisebox{4\normalbaselineskip}[0pt][0pt]{\rotatebox[origin=c]{90}{Left}} & 
          \includegraphics[width=0.18\textwidth,trim={0cm 1cm 0cm 1cm},clip]{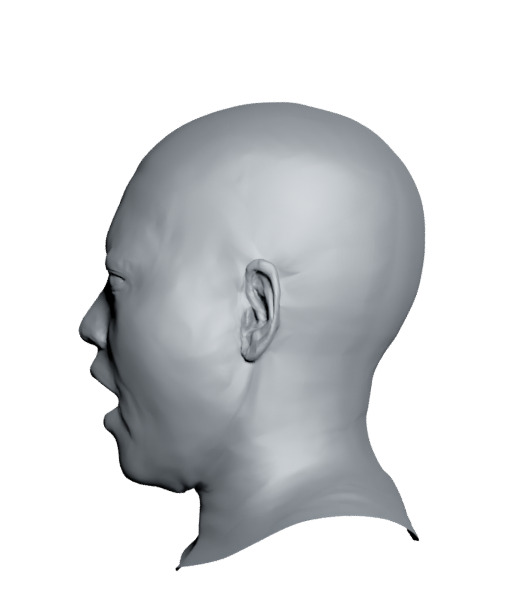} 
        & \includegraphics[width=0.18\textwidth,trim={0cm 1cm 0cm 1cm},clip]{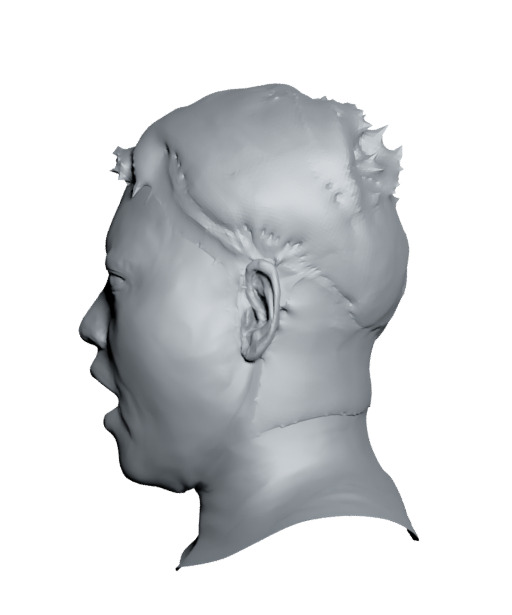}
        & \includegraphics[width=0.18\textwidth,trim={0cm 1cm 0cm 1cm},clip]{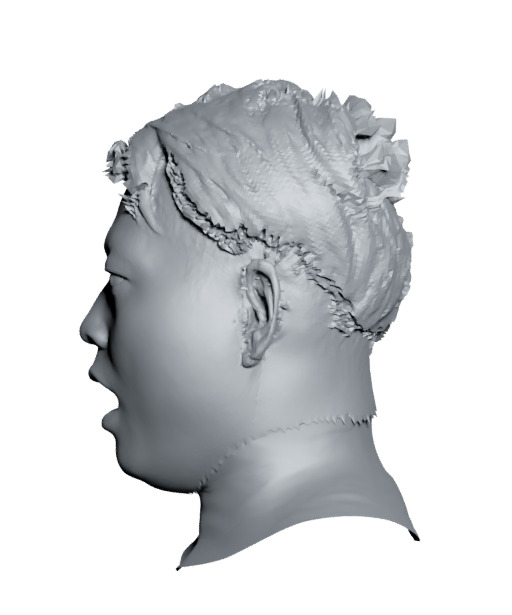}
        & \includegraphics[width=0.18\textwidth,trim={0cm 1cm 0cm 1cm},clip]{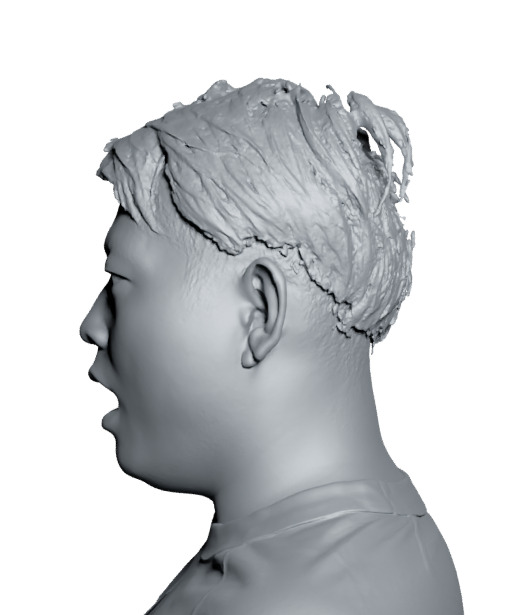}  \\
        \raisebox{3\normalbaselineskip}[0pt][0pt]{\rotatebox[origin=c]{90}{Top}} & 
          \includegraphics[width=0.18\textwidth,trim={0cm 3cm 0cm 3cm},clip]{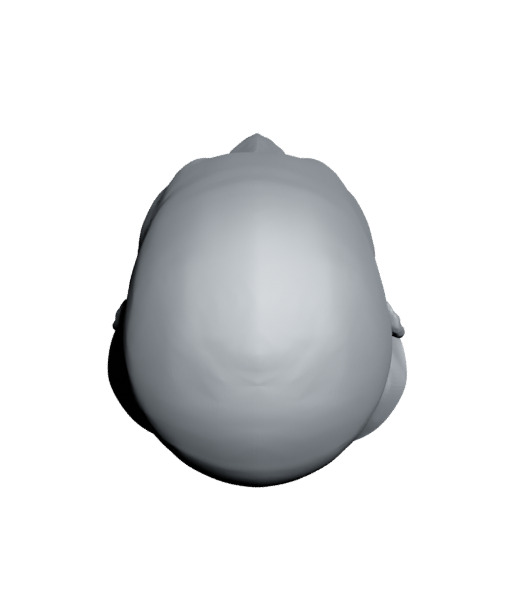} 
        & \includegraphics[width=0.18\textwidth,trim={0cm 3cm 0cm 3cm},clip]{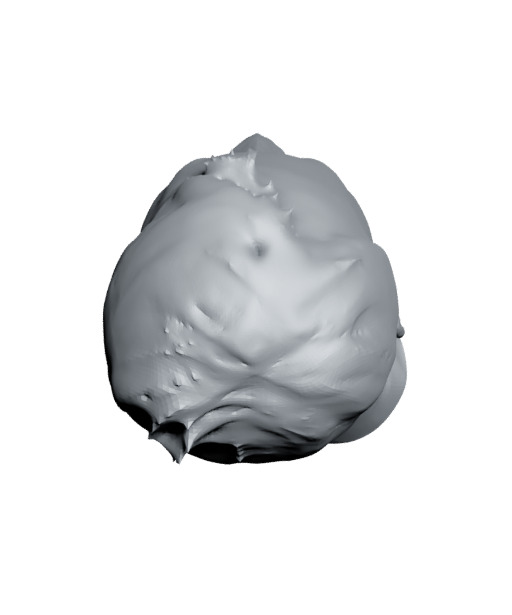}
        & \includegraphics[width=0.18\textwidth,trim={0cm 3cm 0cm 3cm},clip]{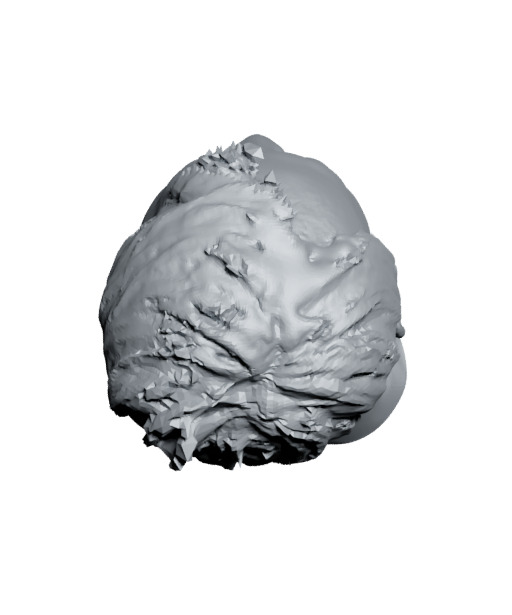}
        & \includegraphics[width=0.18\textwidth,trim={0cm 3cm 0cm 3cm},clip]{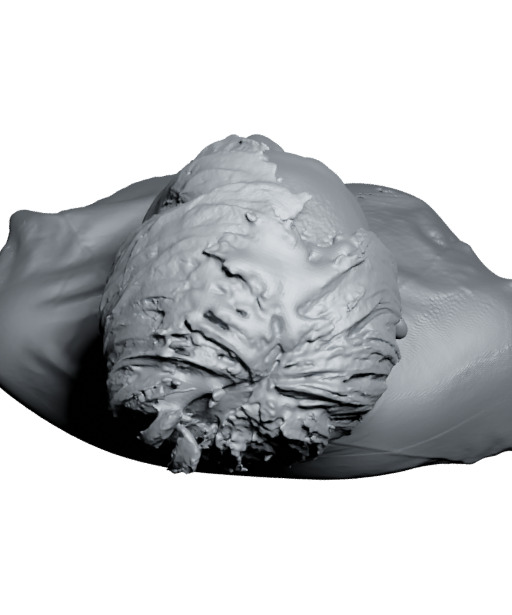}  \\
    \end{tabular}
    \captionof{figure}{
    Additional demonstration of the two-stage registration. 
    %
    %
    \textit{Stage 1} corresponds to the vector displacements regression; \textit{Stage 2} -- to the refinement of the displacements along the normals.
    The second stage significantly improves the level of detail and allows us to match the high-frequency component of the scans, such as strands and subtle face features.
    }
    \label{fig:more_registrations_p1}
\end{table*}
\begin{table*}[]
    \setlength{\tabcolsep}{0pt}
    \renewcommand{\arraystretch}{0}
    \centering
    \begin{tabular}{lcccc}
        & FLAME & Stage 1 & Stage 2 & Ground truth \\
        \raisebox{4\normalbaselineskip}[0pt][0pt]{\rotatebox[origin=c]{90}{Frontal}} & 
          \includegraphics[width=0.18\textwidth,trim={0cm 1cm 0cm 1cm},clip]{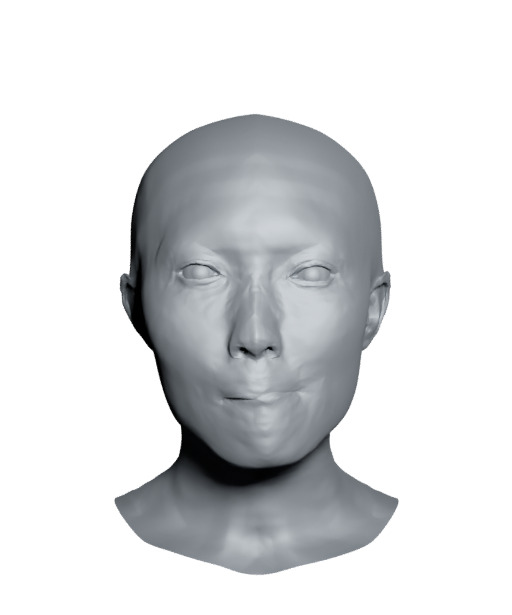} 
        & \includegraphics[width=0.18\textwidth,trim={0cm 1cm 0cm 1cm},clip]{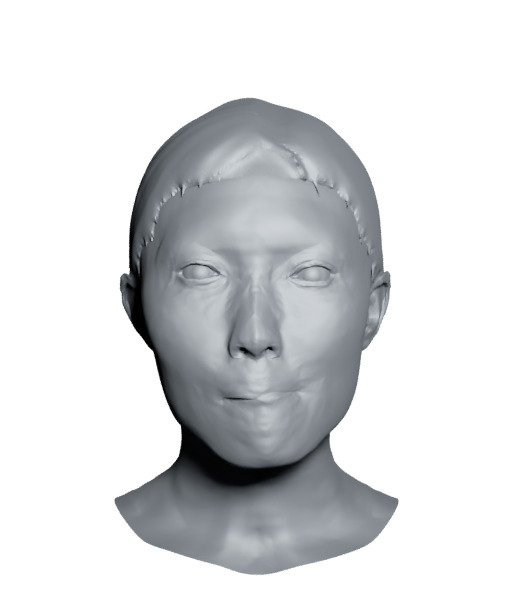}
        & \includegraphics[width=0.18\textwidth,trim={0cm 1cm 0cm 1cm},clip]{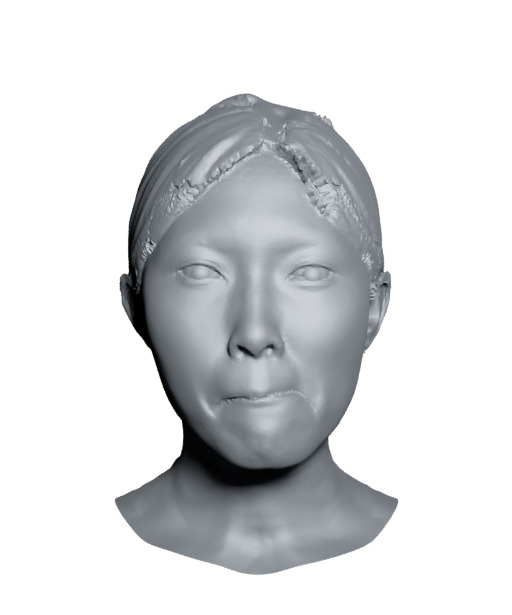}
        & \includegraphics[width=0.18\textwidth,trim={0cm 1cm 0cm 1cm},clip]{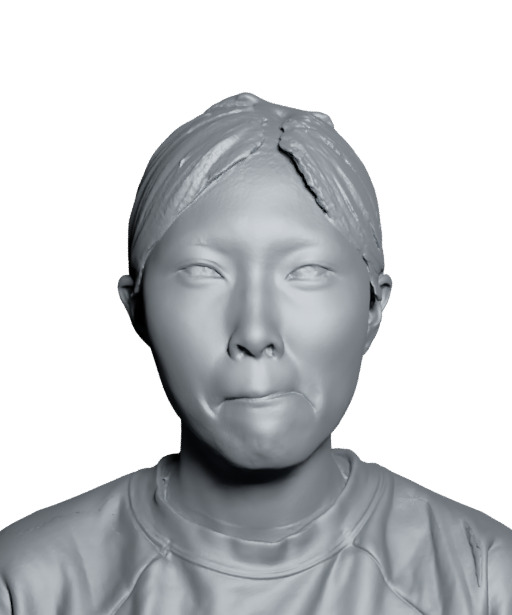}  \\
        \raisebox{4\normalbaselineskip}[0pt][0pt]{\rotatebox[origin=c]{90}{Left}} & 
          \includegraphics[width=0.18\textwidth,trim={0cm 1cm 0cm 1cm},clip]{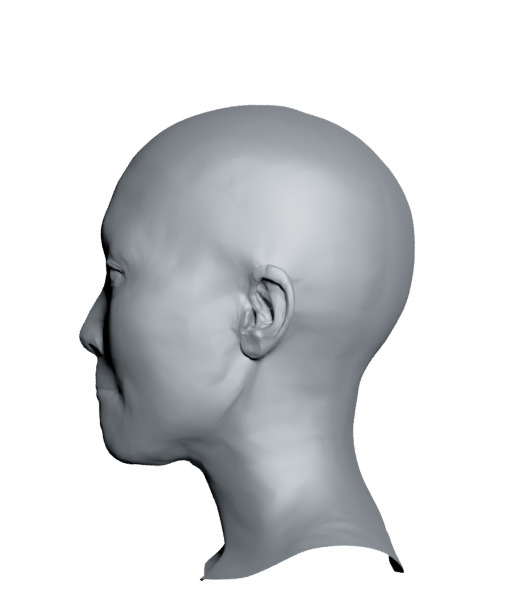} 
        & \includegraphics[width=0.18\textwidth,trim={0cm 1cm 0cm 1cm},clip]{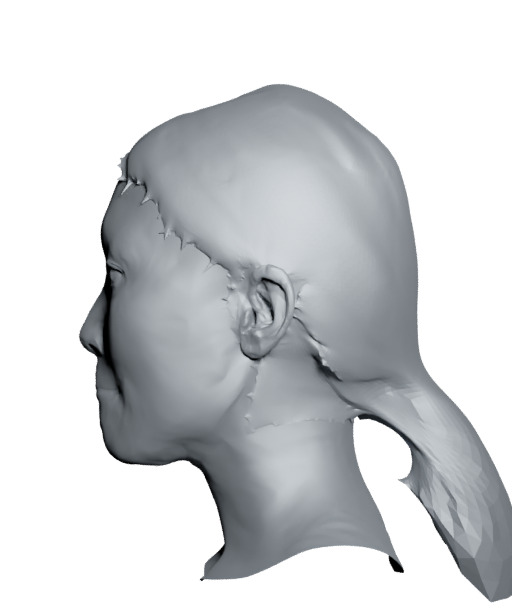}
        & \includegraphics[width=0.18\textwidth,trim={0cm 1cm 0cm 1cm},clip]{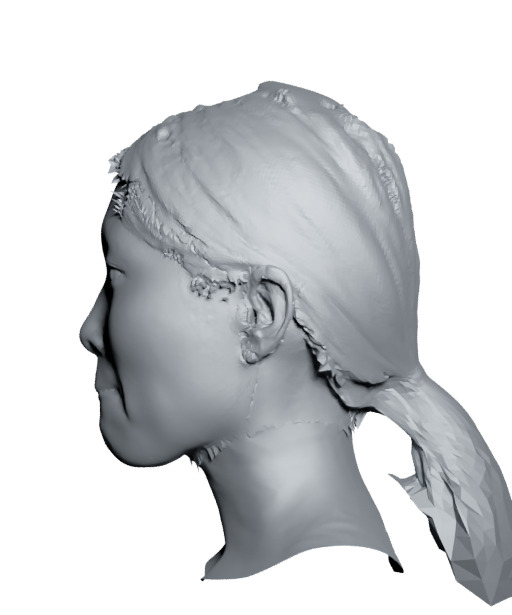}
        & \includegraphics[width=0.18\textwidth,trim={0cm 1cm 0cm 1cm},clip]{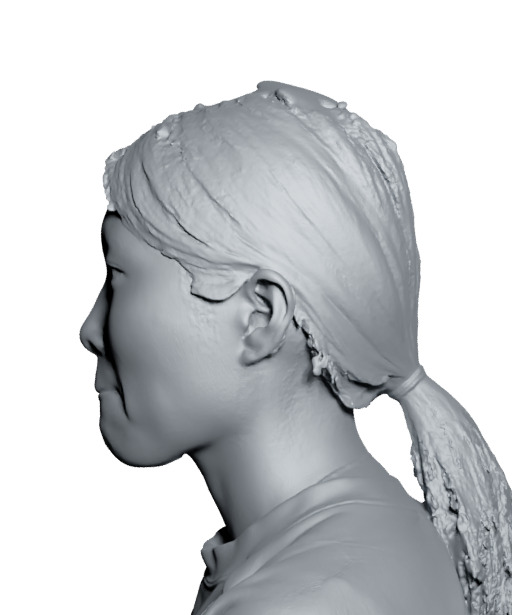}  \\
        \raisebox{4\normalbaselineskip}[0pt][0pt]{\rotatebox[origin=c]{90}{Top}} & 
          \includegraphics[width=0.18\textwidth,trim={0cm 3cm 0cm 3cm},clip]{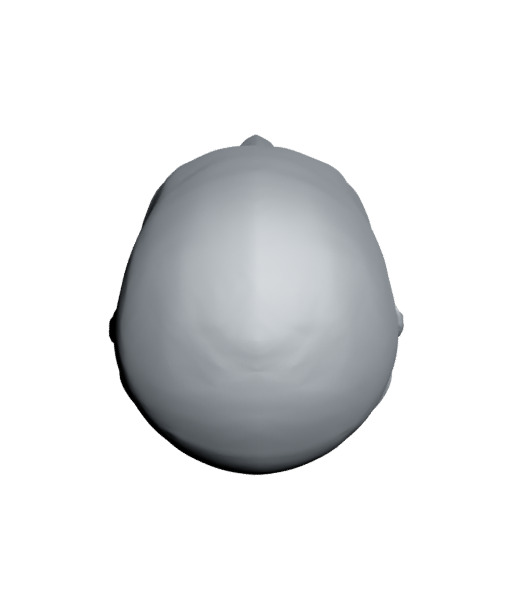} 
        & \includegraphics[width=0.18\textwidth,trim={0cm 3cm 0cm 3cm},clip]{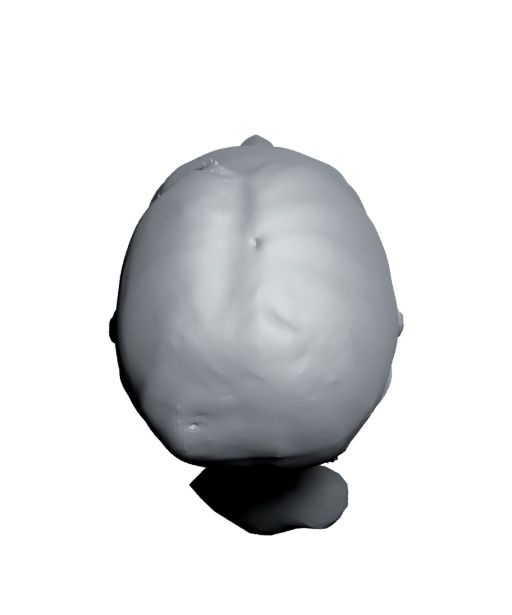}
        & \includegraphics[width=0.18\textwidth,trim={0cm 3cm 0cm 3cm},clip]{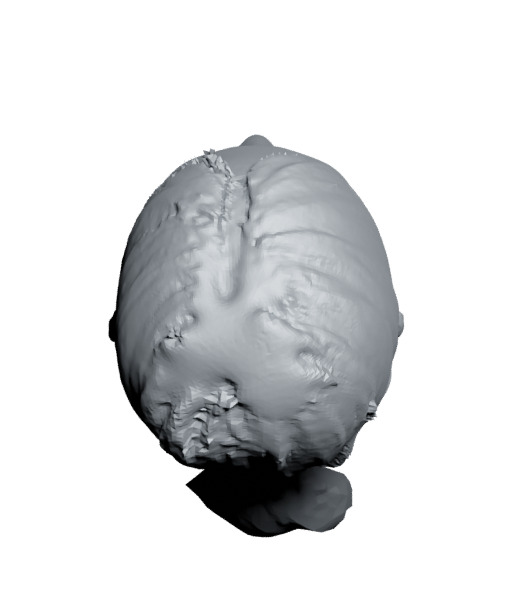}
        & \includegraphics[width=0.18\textwidth,trim={0cm 3cm 0cm 3cm},clip]{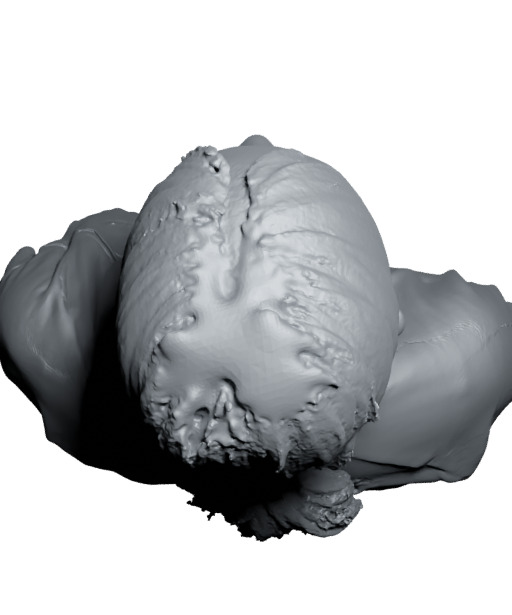}  \\[0.19cm]
        & FLAME & Stage 1 & Stage 2 & Ground truth \\
        \raisebox{4\normalbaselineskip}[0pt][0pt]{\rotatebox[origin=c]{90}{Frontal}} & 
          \includegraphics[width=0.18\textwidth,trim={0cm 1cm 0cm 1cm},clip]{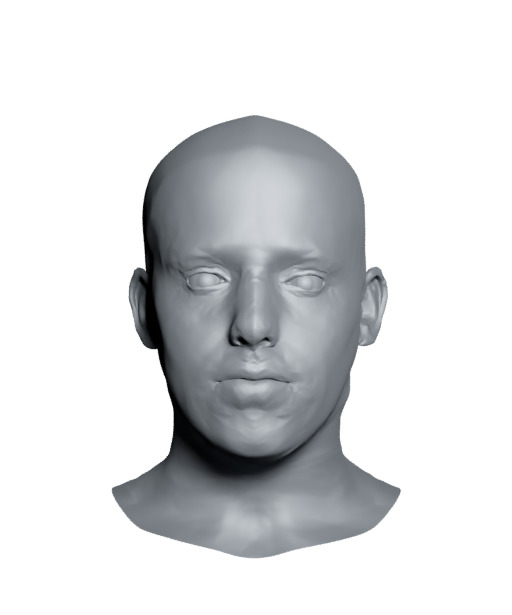} 
        & \includegraphics[width=0.18\textwidth,trim={0cm 1cm 0cm 1cm},clip]{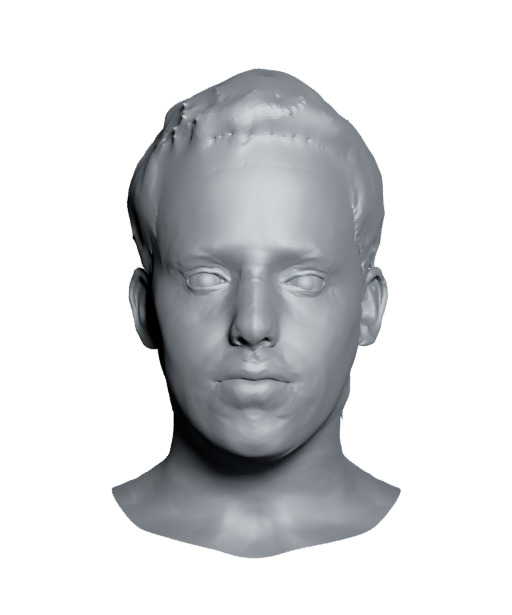}
        & \includegraphics[width=0.18\textwidth,trim={0cm 1cm 0cm 1cm},clip]{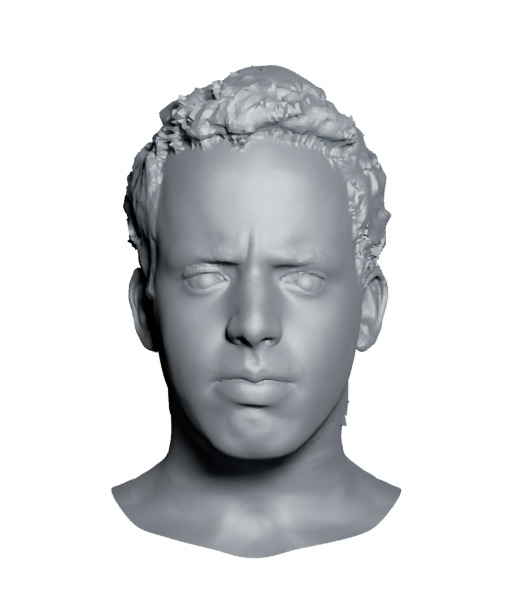}
        & \includegraphics[width=0.18\textwidth,trim={0cm 1cm 0cm 1cm},clip]{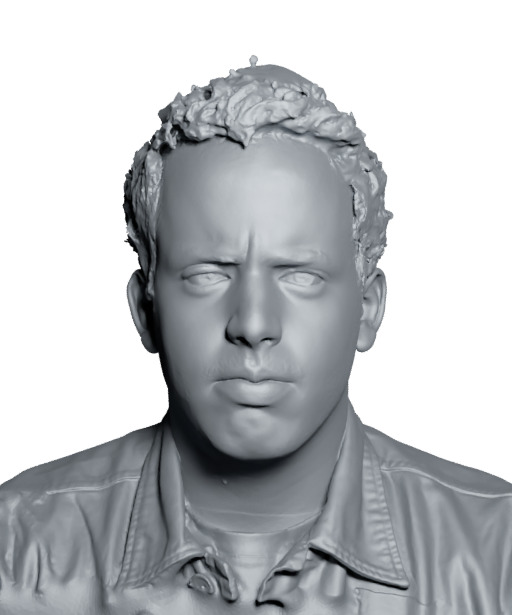}  \\
        \raisebox{4\normalbaselineskip}[0pt][0pt]{\rotatebox[origin=c]{90}{Left}} & 
          \includegraphics[width=0.18\textwidth,trim={0cm 1cm 0cm 1cm},clip]{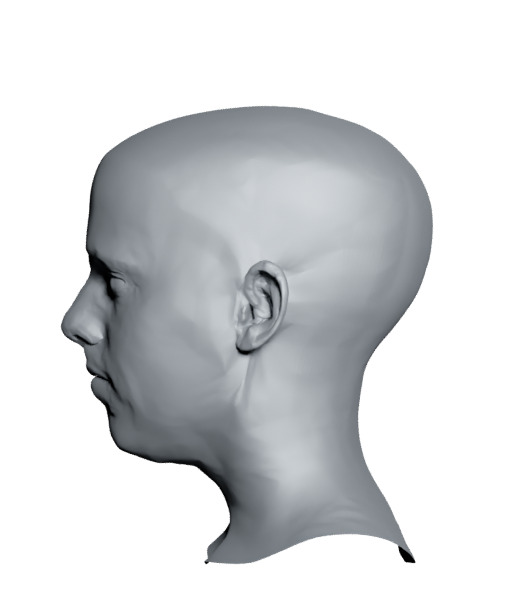} 
        & \includegraphics[width=0.18\textwidth,trim={0cm 1cm 0cm 1cm},clip]{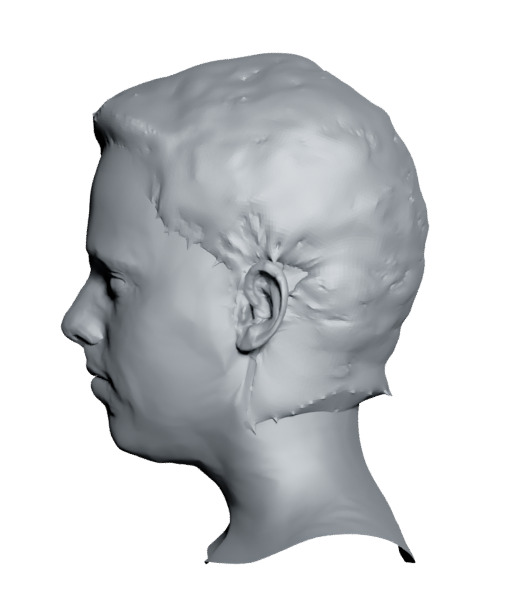}
        & \includegraphics[width=0.18\textwidth,trim={0cm 1cm 0cm 1cm},clip]{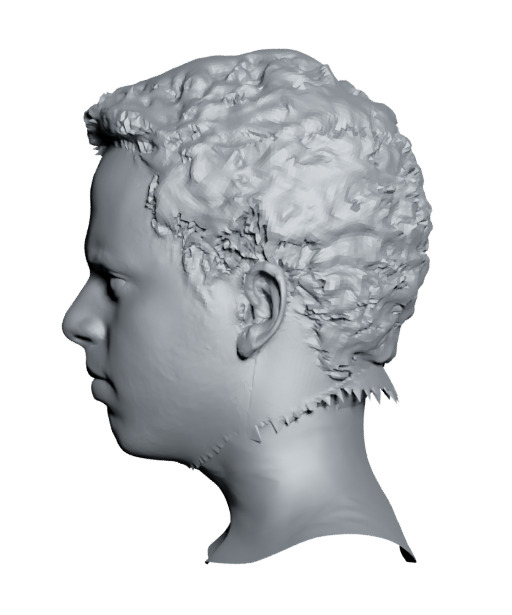}
        & \includegraphics[width=0.18\textwidth,trim={0cm 1cm 0cm 1cm},clip]{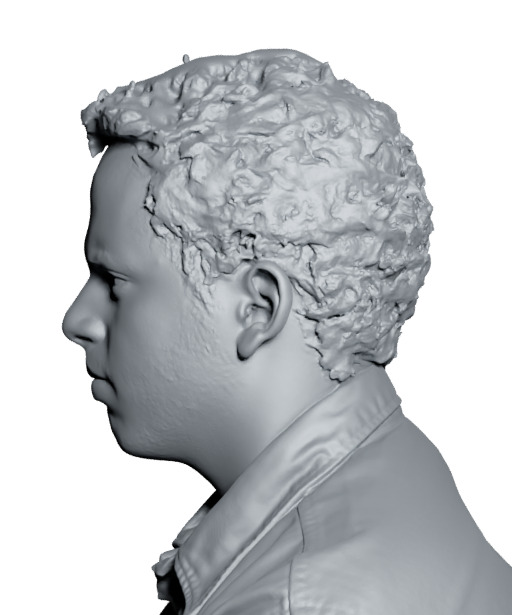}  \\
        \raisebox{3\normalbaselineskip}[0pt][0pt]{\rotatebox[origin=c]{90}{Top}} & 
          \includegraphics[width=0.18\textwidth,trim={0cm 3cm 0cm 3cm},clip]{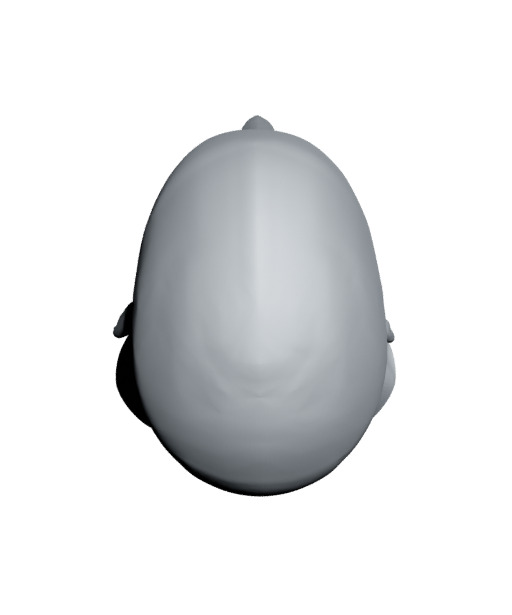} 
        & \includegraphics[width=0.18\textwidth,trim={0cm 3cm 0cm 3cm},clip]{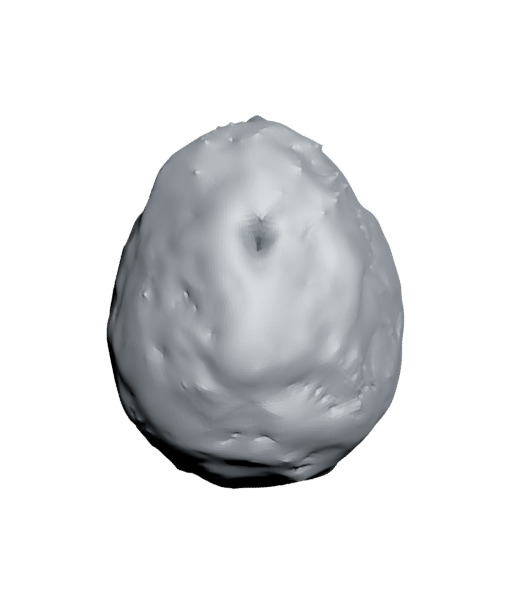}
        & \includegraphics[width=0.18\textwidth,trim={0cm 3cm 0cm 3cm},clip]{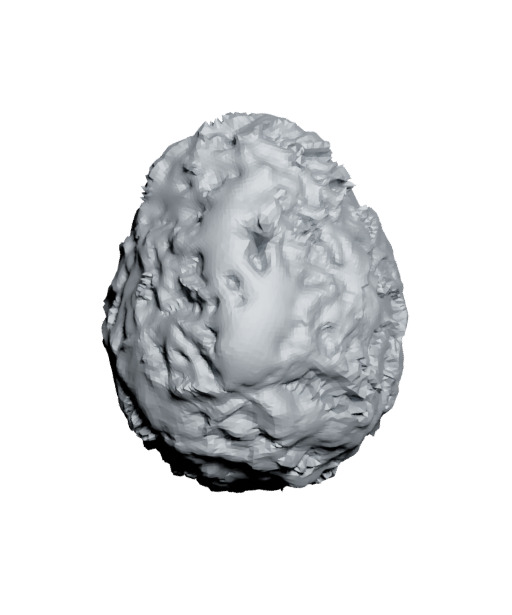}
        & \includegraphics[width=0.18\textwidth,trim={0cm 3cm 0cm 3cm},clip]{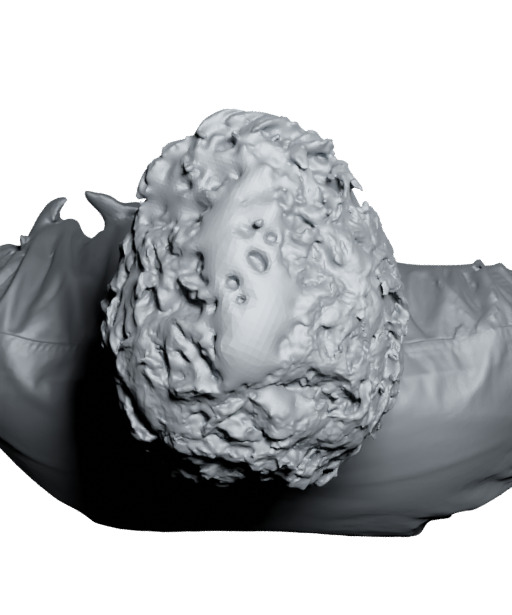}  \\
    \end{tabular}
    \captionof{figure}{
    Additional demonstration of the two-stage registration. 
    %
    %
    \textit{Stage 1} corresponds to the vector displacements regression; \textit{Stage 2} -- to the refinement of the displacements along the normals.
    The second stage significantly improves the level of detail and allows us to match the high-frequency component of the scans, such as strands and subtle face features.
    }
    \label{fig:more_registrations_p2}
\end{table*}
\begin{table*}[]
    \setlength{\tabcolsep}{0pt}
    \renewcommand{\arraystretch}{0}
    \centering
    \begin{tabular}{lcccc}
        & FLAME & Stage 1 & Stage 2 & Ground truth \\
        \raisebox{4\normalbaselineskip}[0pt][0pt]{\rotatebox[origin=c]{90}{Frontal}} & 
          \includegraphics[width=0.18\textwidth,trim={0cm 1cm 0cm 1cm},clip]{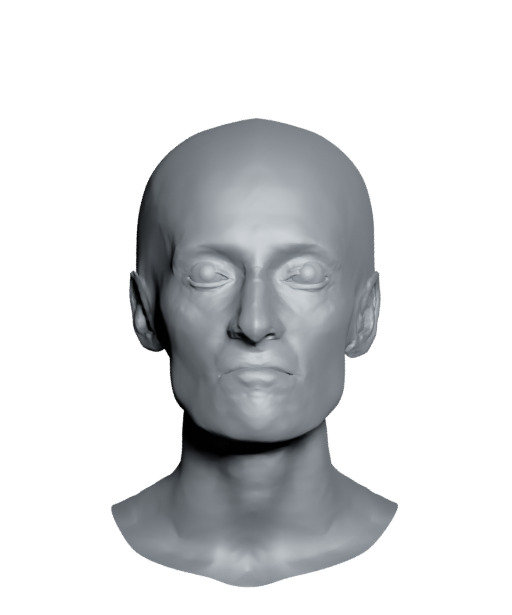} 
        & \includegraphics[width=0.18\textwidth,trim={0cm 1cm 0cm 1cm},clip]{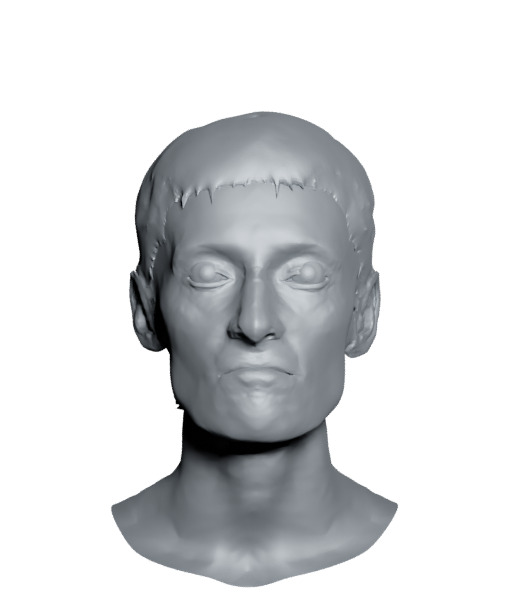}
        & \includegraphics[width=0.18\textwidth,trim={0cm 1cm 0cm 1cm},clip]{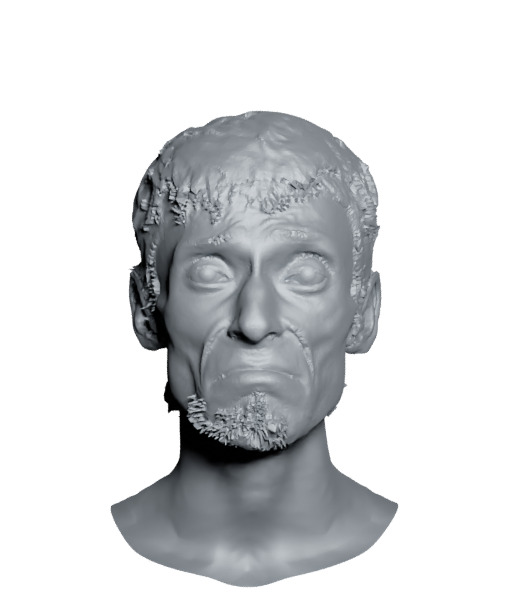}
        & \includegraphics[width=0.18\textwidth,trim={0cm 1cm 0cm 1cm},clip]{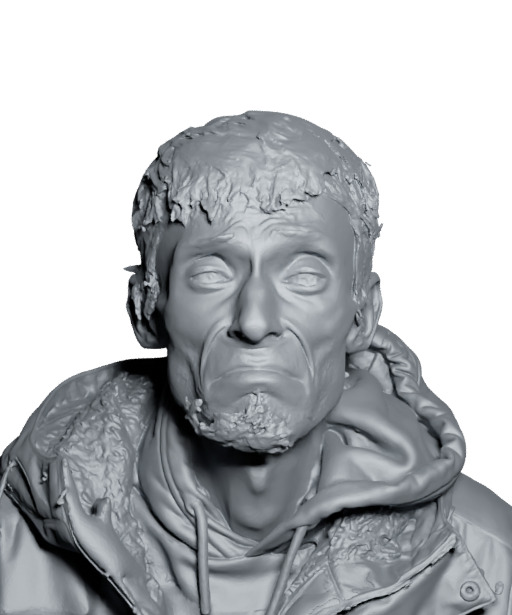}  \\
        \raisebox{4\normalbaselineskip}[0pt][0pt]{\rotatebox[origin=c]{90}{Left}} & 
          \includegraphics[width=0.18\textwidth,trim={0cm 1cm 0cm 1cm},clip]{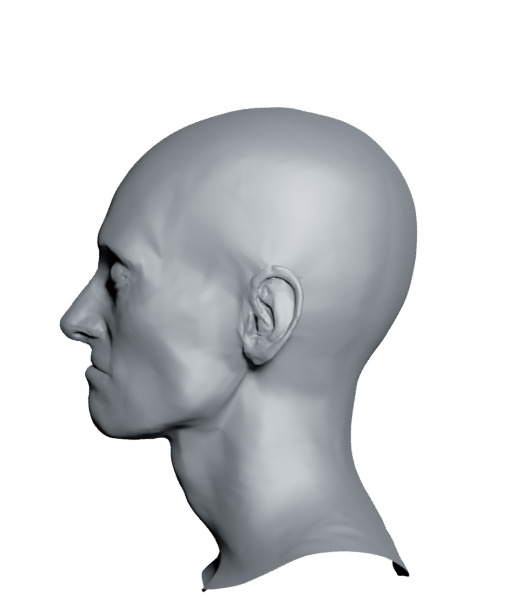} 
        & \includegraphics[width=0.18\textwidth,trim={0cm 1cm 0cm 1cm},clip]{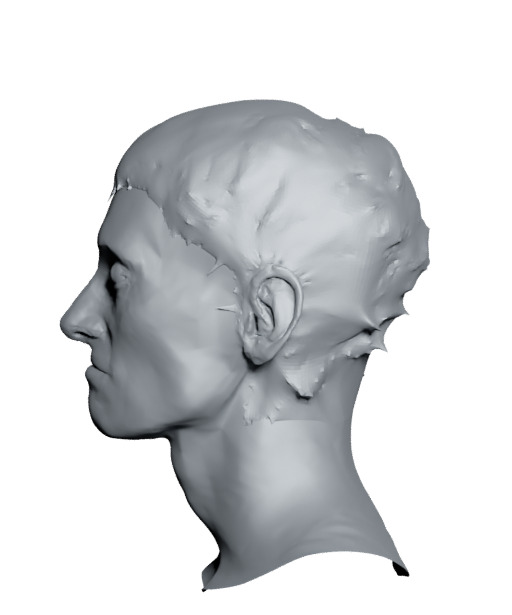}
        & \includegraphics[width=0.18\textwidth,trim={0cm 1cm 0cm 1cm},clip]{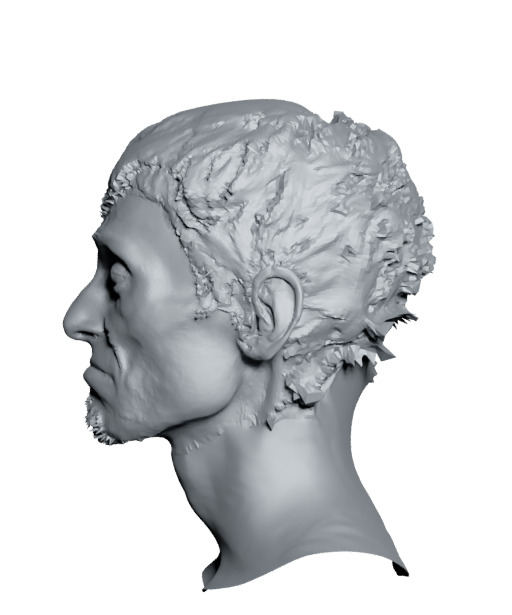}
        & \includegraphics[width=0.18\textwidth,trim={0cm 1cm 0cm 1cm},clip]{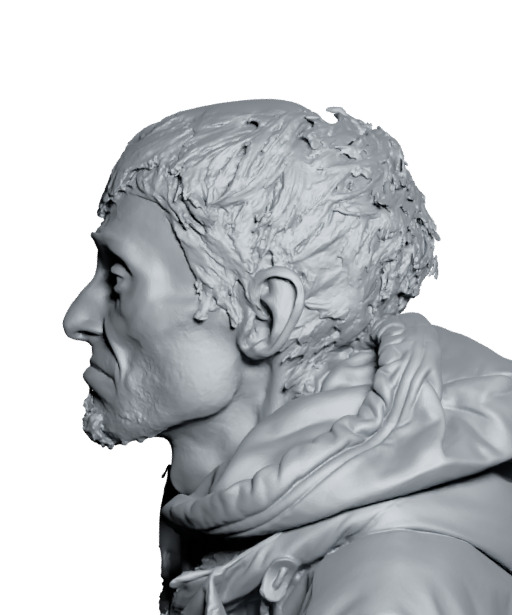}  \\
        \raisebox{4\normalbaselineskip}[0pt][0pt]{\rotatebox[origin=c]{90}{Top}} & 
          \includegraphics[width=0.18\textwidth,trim={0cm 3cm 0cm 3cm},clip]{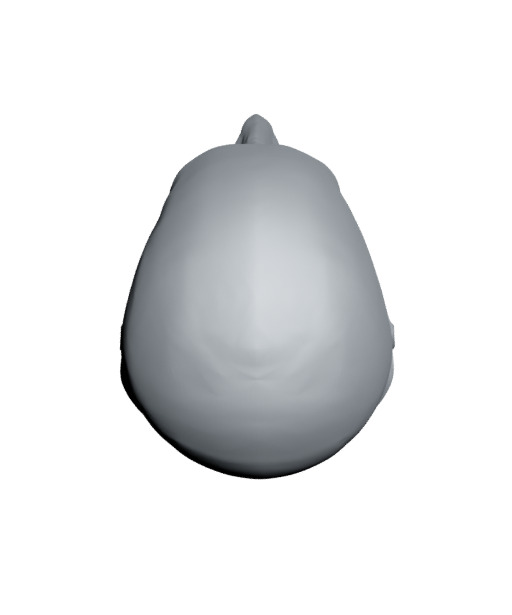} 
        & \includegraphics[width=0.18\textwidth,trim={0cm 3cm 0cm 3cm},clip]{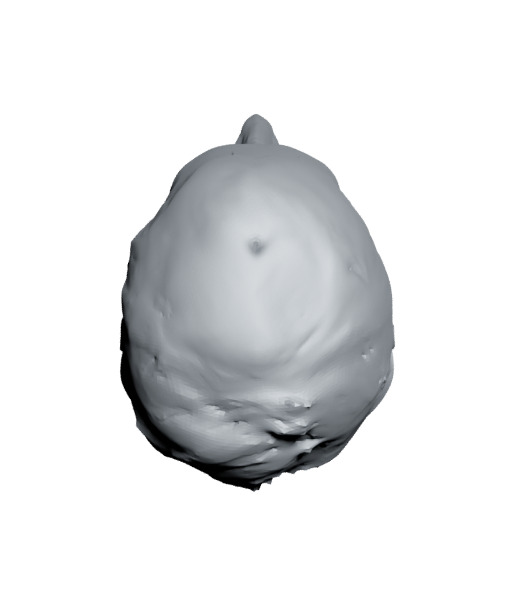}
        & \includegraphics[width=0.18\textwidth,trim={0cm 3cm 0cm 3cm},clip]{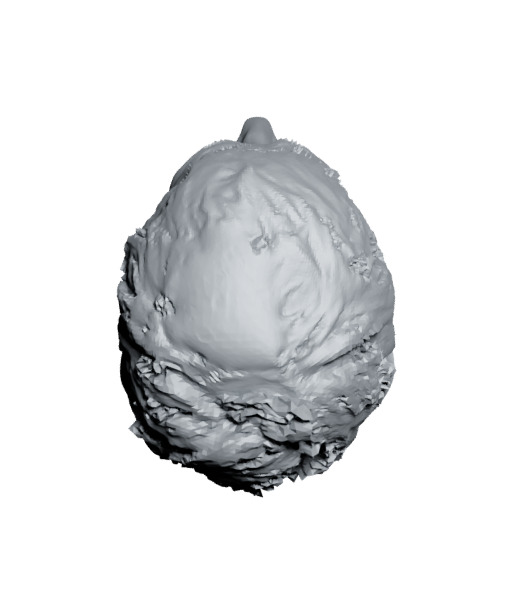}
        & \includegraphics[width=0.18\textwidth,trim={0cm 3cm 0cm 3cm},clip]{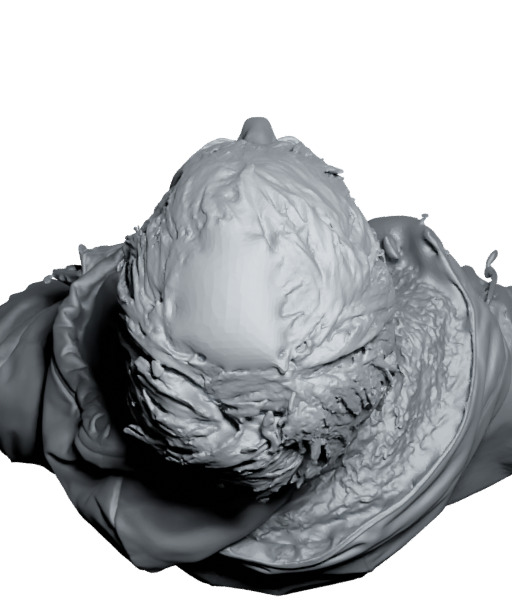}  \\[0.19cm]
        & FLAME & Stage 1 & Stage 2 & Ground truth \\
        \raisebox{4\normalbaselineskip}[0pt][0pt]{\rotatebox[origin=c]{90}{Frontal}} & 
          \includegraphics[width=0.18\textwidth,trim={0cm 1cm 0cm 1cm},clip]{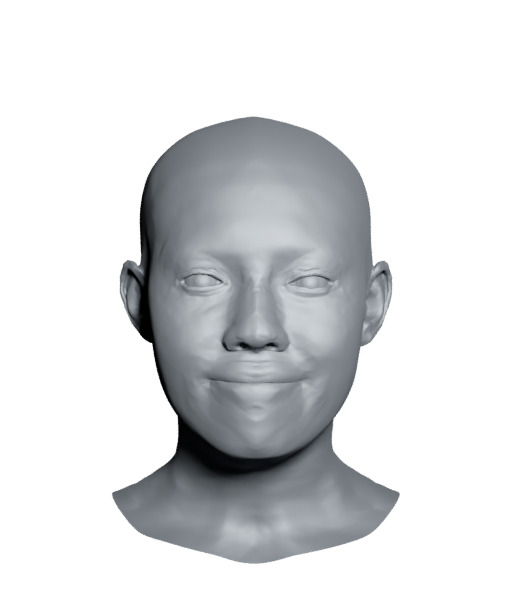} 
        & \includegraphics[width=0.18\textwidth,trim={0cm 1cm 0cm 1cm},clip]{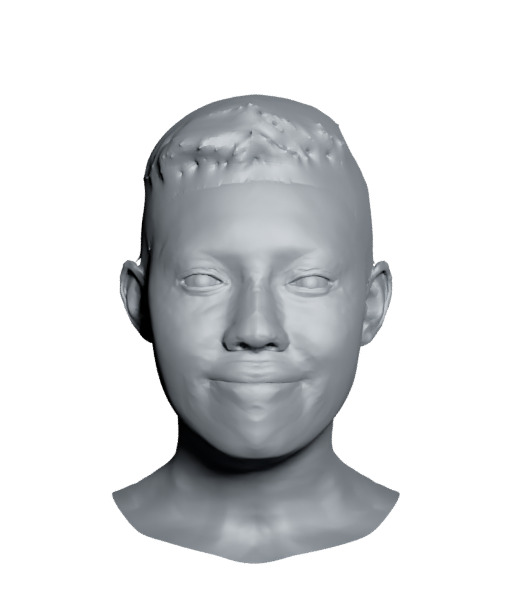}
        & \includegraphics[width=0.18\textwidth,trim={0cm 1cm 0cm 1cm},clip]{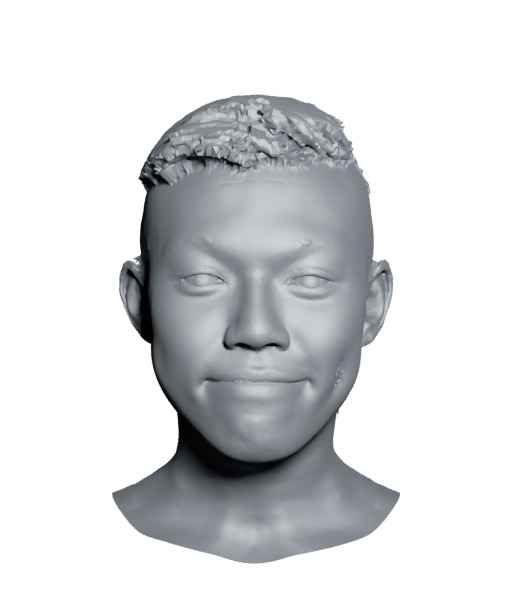}
        & \includegraphics[width=0.18\textwidth,trim={0cm 1cm 0cm 1cm},clip]{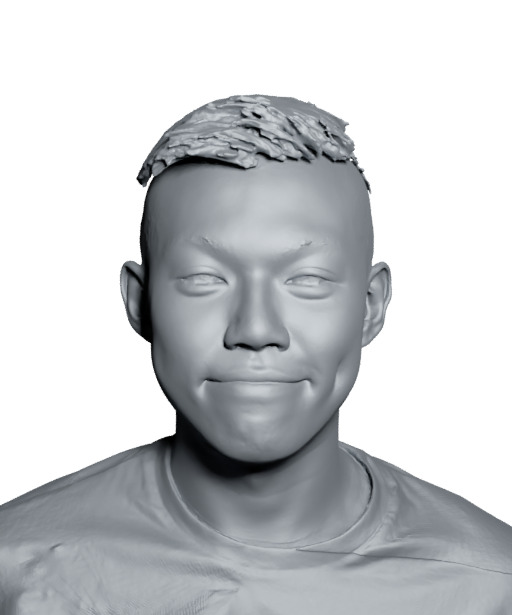}  \\
        \raisebox{4\normalbaselineskip}[0pt][0pt]{\rotatebox[origin=c]{90}{Left}} & 
          \includegraphics[width=0.18\textwidth,trim={0cm 1cm 0cm 1cm},clip]{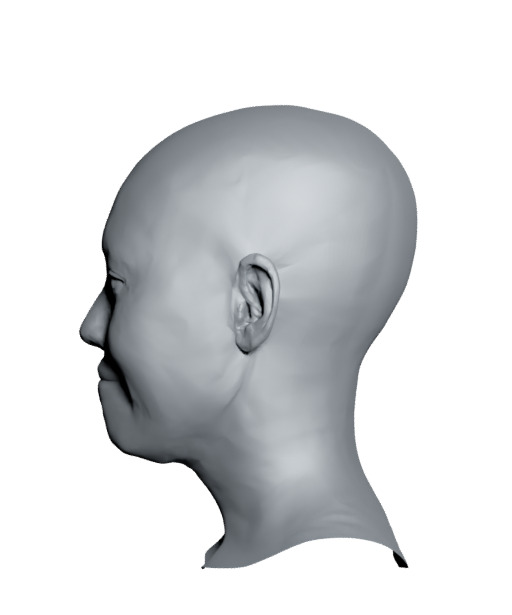} 
        & \includegraphics[width=0.18\textwidth,trim={0cm 1cm 0cm 1cm},clip]{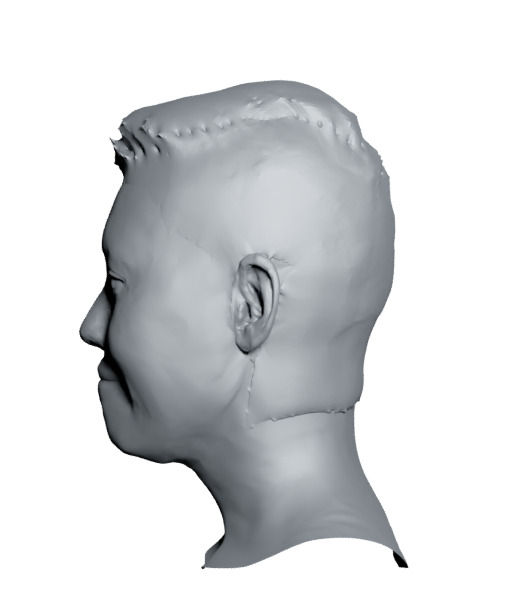}
        & \includegraphics[width=0.18\textwidth,trim={0cm 1cm 0cm 1cm},clip]{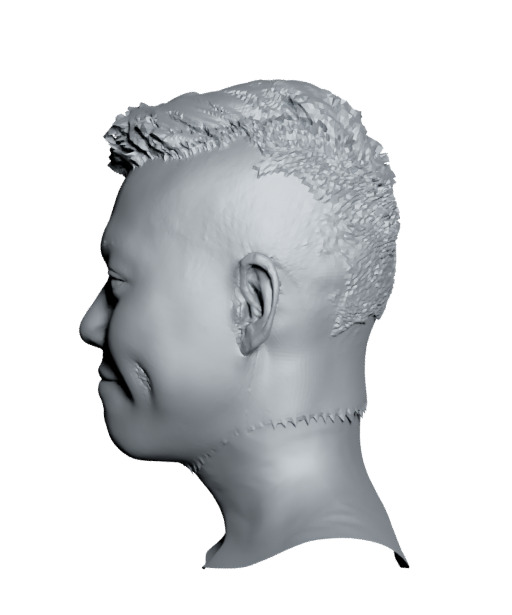}
        & \includegraphics[width=0.18\textwidth,trim={0cm 1cm 0cm 1cm},clip]{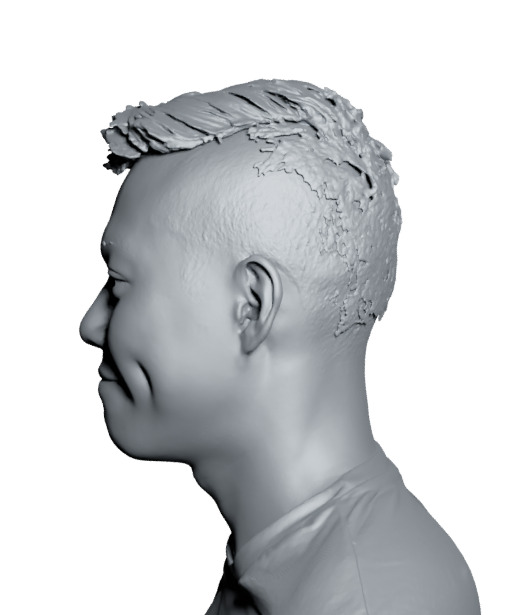}  \\
        \raisebox{3\normalbaselineskip}[0pt][0pt]{\rotatebox[origin=c]{90}{Top}} & 
          \includegraphics[width=0.18\textwidth,trim={0cm 3cm 0cm 3cm},clip]{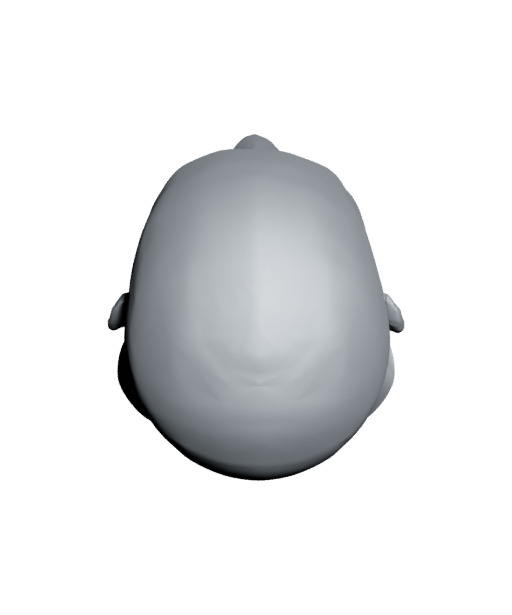} 
        & \includegraphics[width=0.18\textwidth,trim={0cm 3cm 0cm 3cm},clip]{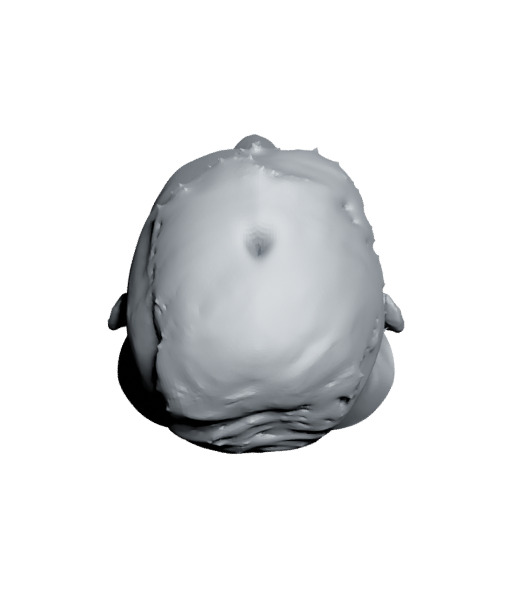}
        & \includegraphics[width=0.18\textwidth,trim={0cm 3cm 0cm 3cm},clip]{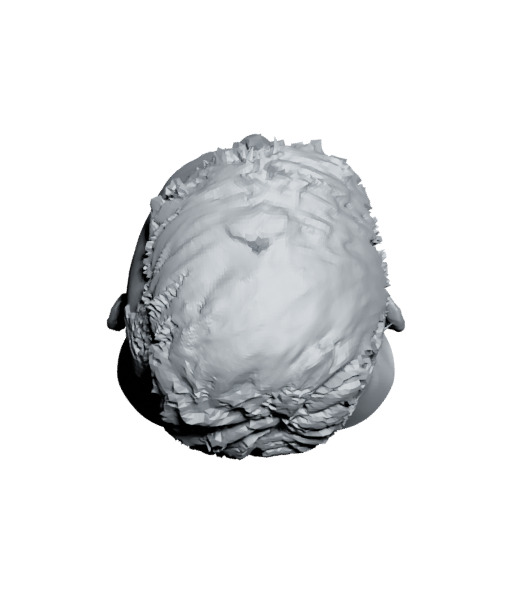}
        & \includegraphics[width=0.18\textwidth,trim={0cm 3cm 0cm 3cm},clip]{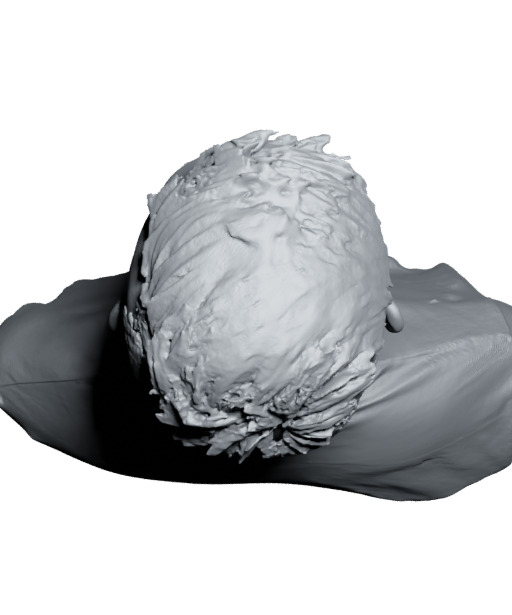}  \\
    \end{tabular}
    \captionof{figure}{
    Additional demonstration of the two-stage registration. 
    %
    %
    \textit{Stage 1} corresponds to the vector displacements regression; \textit{Stage 2} -- to the refinement of the displacements along the normals.
    The second stage significantly improves the level of detail and allows us to match the high-frequency component of the scans, such as strands and subtle face features.
    }
    \label{fig:more_registrations_p3}
    \vspace{-0.19cm}
\end{table*}
\begin{table*}[]
    \setlength{\tabcolsep}{0pt}
    \renewcommand{\arraystretch}{0}
    \centering
    \begin{tabular}{lcccc}
        & FLAME & Stage 1 & Stage 2 & Ground truth \\
        \raisebox{4\normalbaselineskip}[0pt][0pt]{\rotatebox[origin=c]{90}{Frontal}} & 
          \includegraphics[width=0.18\textwidth,trim={0cm 1cm 0cm 1cm},clip]{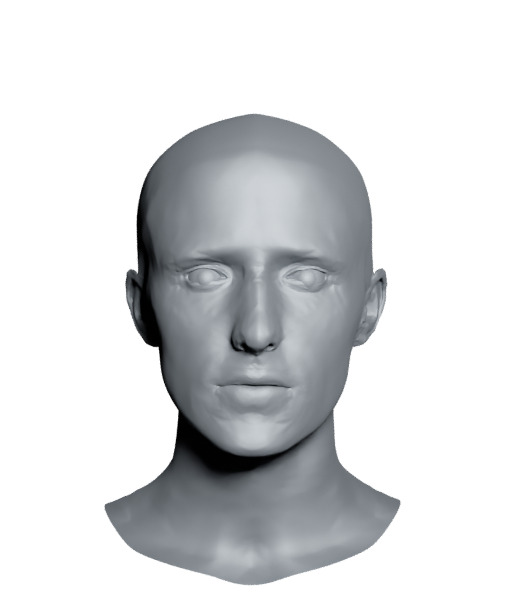} 
        & \includegraphics[width=0.18\textwidth,trim={0cm 1cm 0cm 1cm},clip]{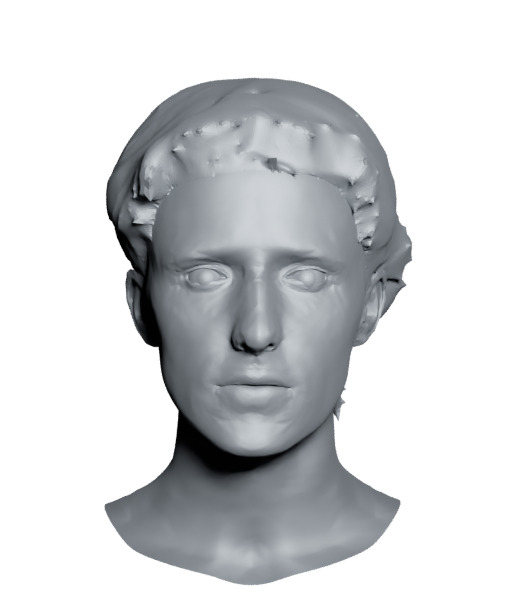}
        & \includegraphics[width=0.18\textwidth,trim={0cm 1cm 0cm 1cm},clip]{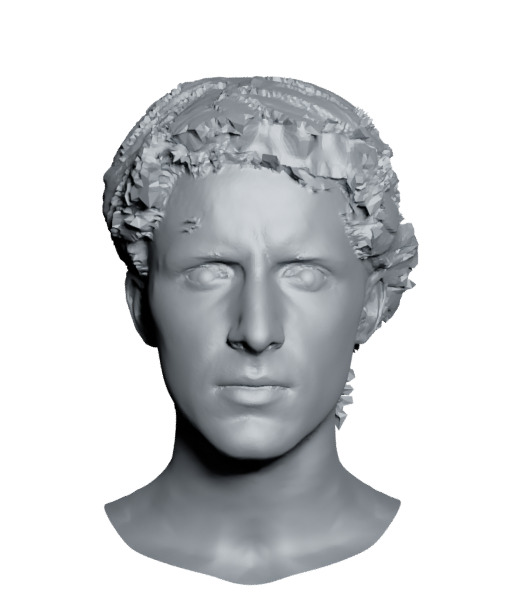}
        & \includegraphics[width=0.18\textwidth,trim={0cm 1cm 0cm 1cm},clip]{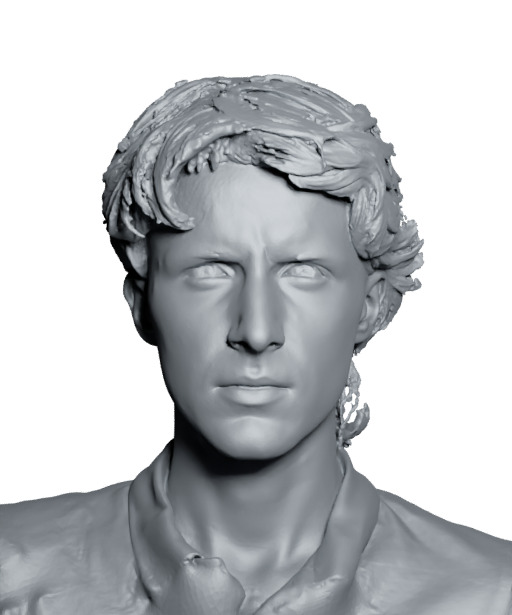}  \\
        \raisebox{4\normalbaselineskip}[0pt][0pt]{\rotatebox[origin=c]{90}{Left}} & 
          \includegraphics[width=0.18\textwidth,trim={0cm 1cm 0cm 1cm},clip]{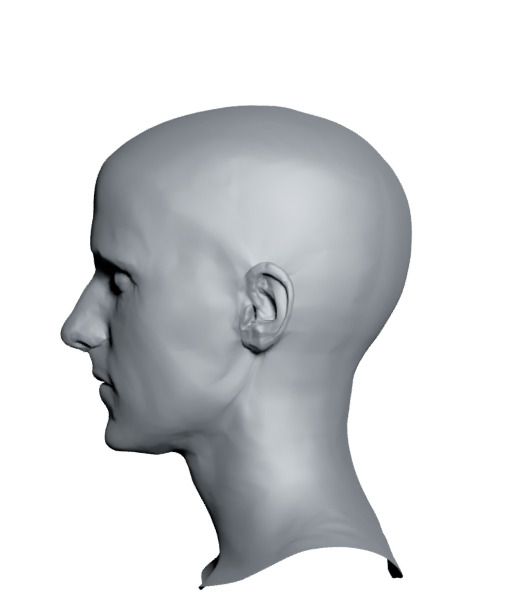} 
        & \includegraphics[width=0.18\textwidth,trim={0cm 1cm 0cm 1cm},clip]{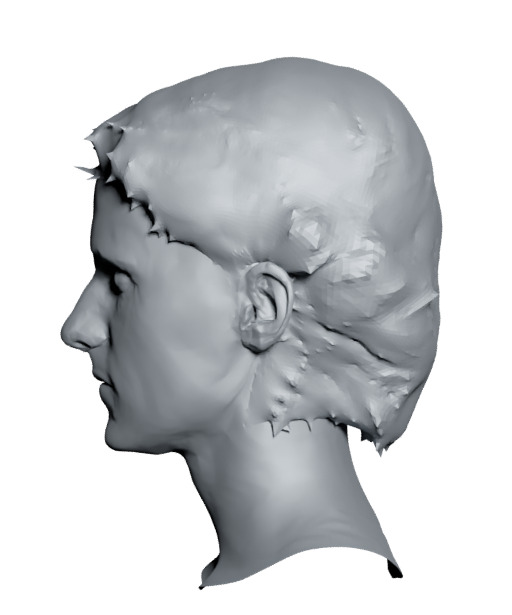}
        & \includegraphics[width=0.18\textwidth,trim={0cm 1cm 0cm 1cm},clip]{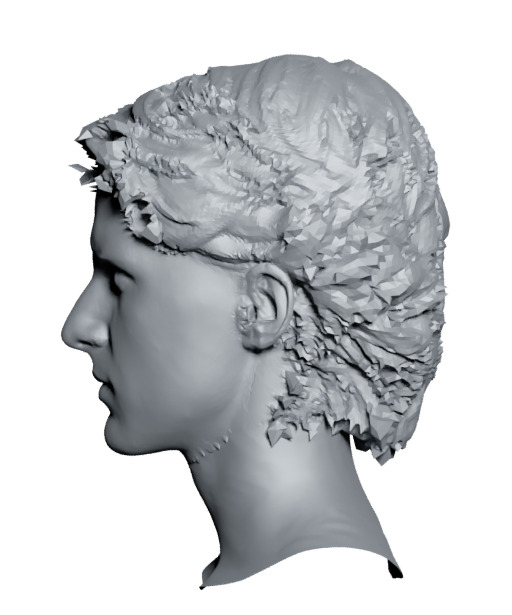}
        & \includegraphics[width=0.18\textwidth,trim={0cm 1cm 0cm 1cm},clip]{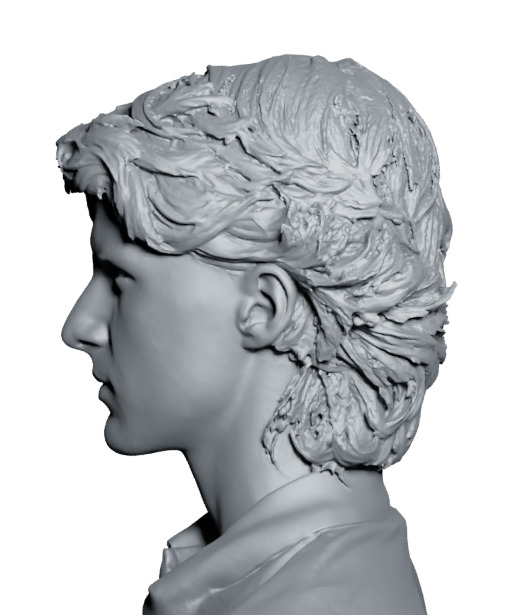}  \\
        \raisebox{4\normalbaselineskip}[0pt][0pt]{\rotatebox[origin=c]{90}{Top}} & 
          \includegraphics[width=0.18\textwidth,trim={0cm 3cm 0cm 3cm},clip]{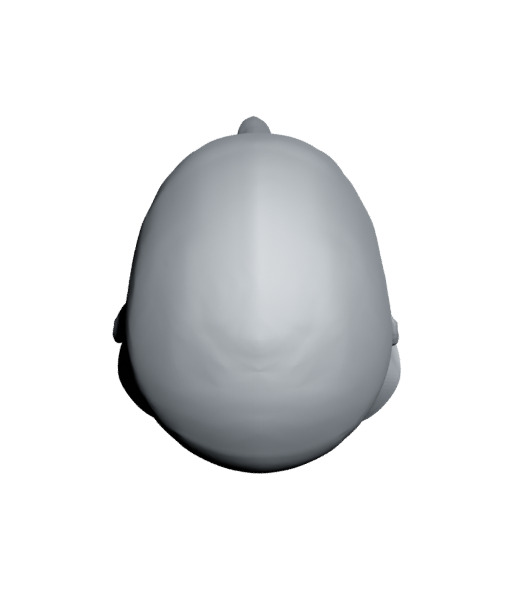} 
        & \includegraphics[width=0.18\textwidth,trim={0cm 3cm 0cm 3cm},clip]{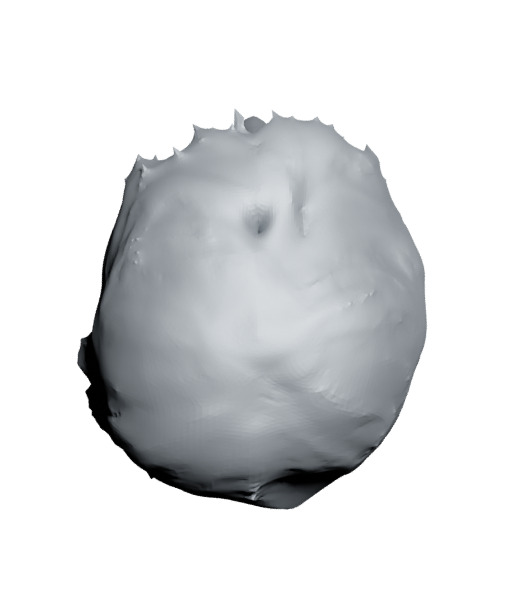}
        & \includegraphics[width=0.18\textwidth,trim={0cm 3cm 0cm 3cm},clip]{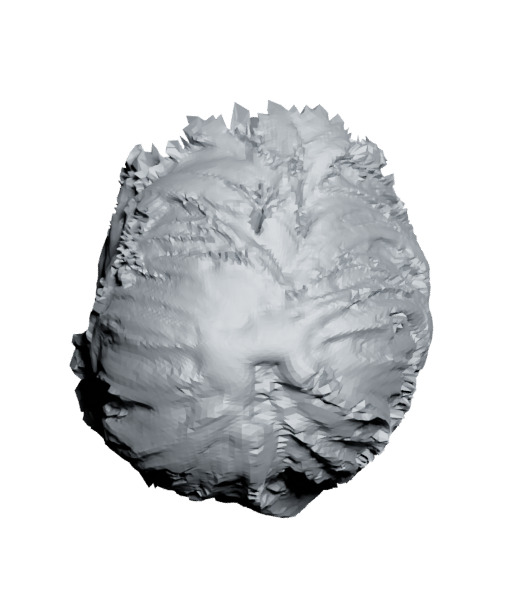}
        & \includegraphics[width=0.18\textwidth,trim={0cm 3cm 0cm 3cm},clip]{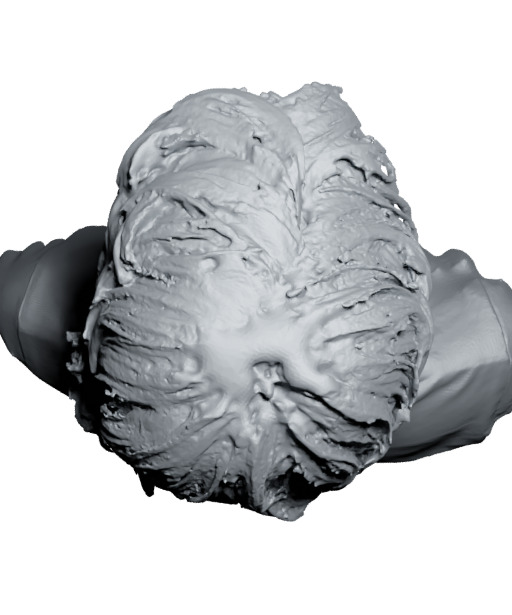} \\[0.19cm]
        & FLAME & Stage 1 & Stage 2 & Ground truth \\
        \raisebox{4\normalbaselineskip}[0pt][0pt]{\rotatebox[origin=c]{90}{Frontal}} & 
          \includegraphics[width=0.18\textwidth,trim={0cm 1cm 0cm 1cm},clip]{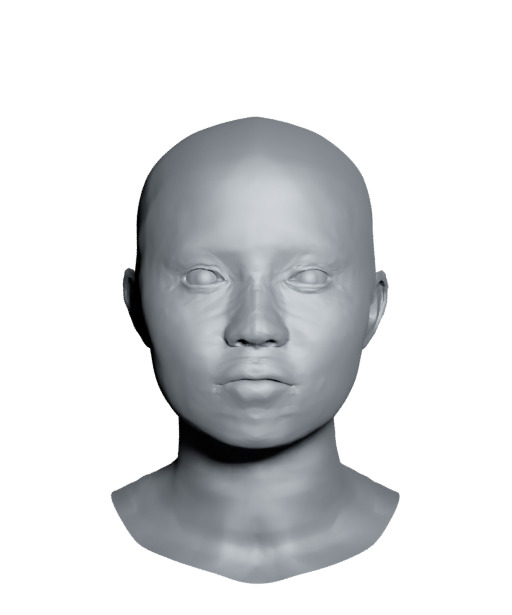} 
        & \includegraphics[width=0.18\textwidth,trim={0cm 1cm 0cm 1cm},clip]{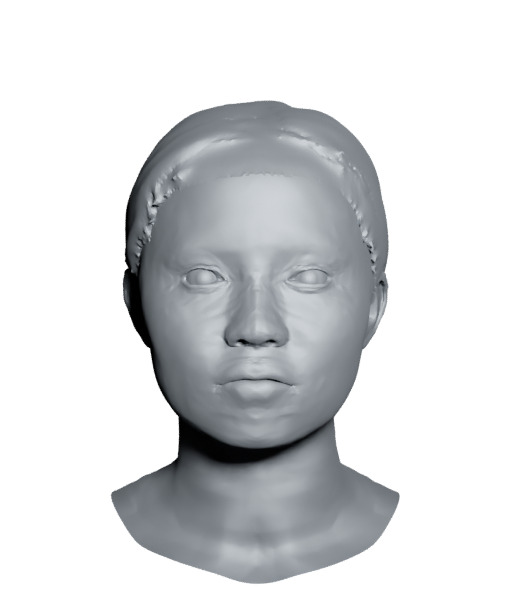}
        & \includegraphics[width=0.18\textwidth,trim={0cm 1cm 0cm 1cm},clip]{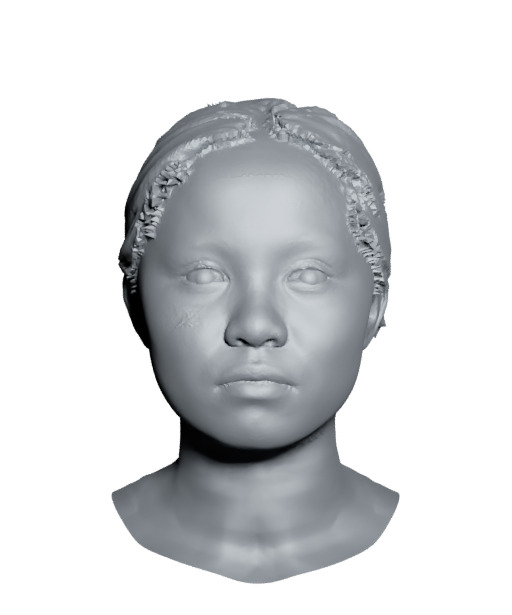}
        & \includegraphics[width=0.18\textwidth,trim={0cm 1cm 0cm 1cm},clip]{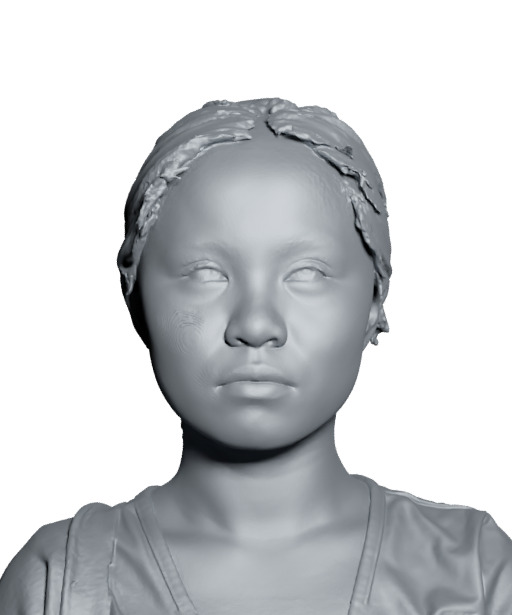}  \\
        \raisebox{4\normalbaselineskip}[0pt][0pt]{\rotatebox[origin=c]{90}{Left}} & 
          \includegraphics[width=0.18\textwidth,trim={0cm 1cm 0cm 1cm},clip]{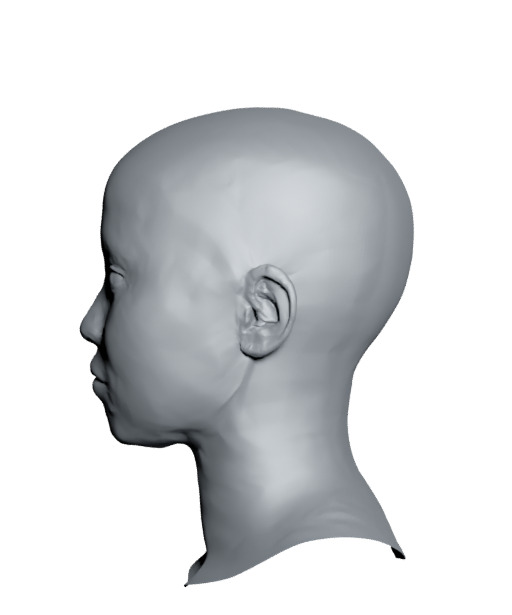} 
        & \includegraphics[width=0.18\textwidth,trim={0cm 1cm 0cm 1cm},clip]{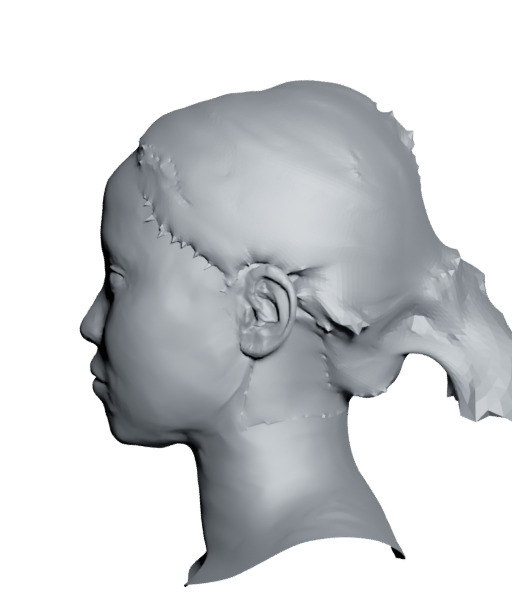}
        & \includegraphics[width=0.18\textwidth,trim={0cm 1cm 0cm 1cm},clip]{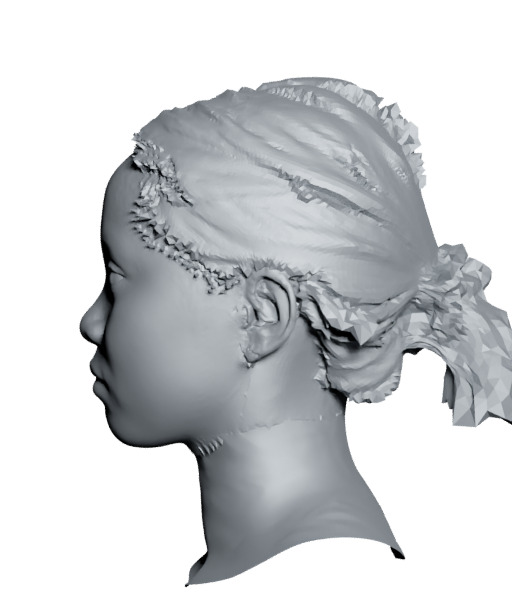}
        & \includegraphics[width=0.18\textwidth,trim={0cm 1cm 0cm 1cm},clip]{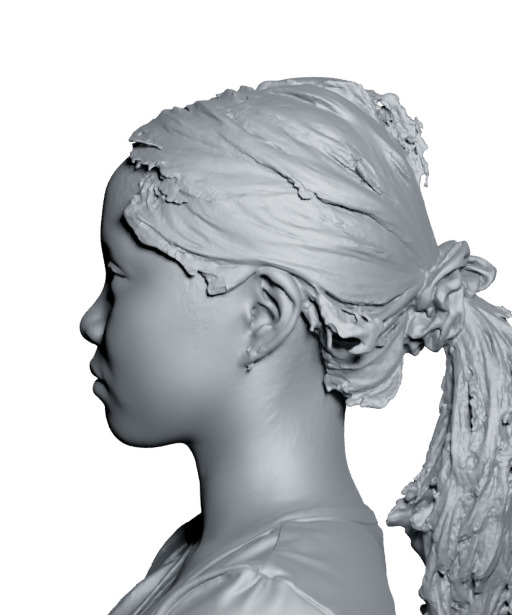}  \\
        \raisebox{3\normalbaselineskip}[0pt][0pt]{\rotatebox[origin=c]{90}{Top}} & 
          \includegraphics[width=0.18\textwidth,trim={0cm 3cm 0cm 3cm},clip]{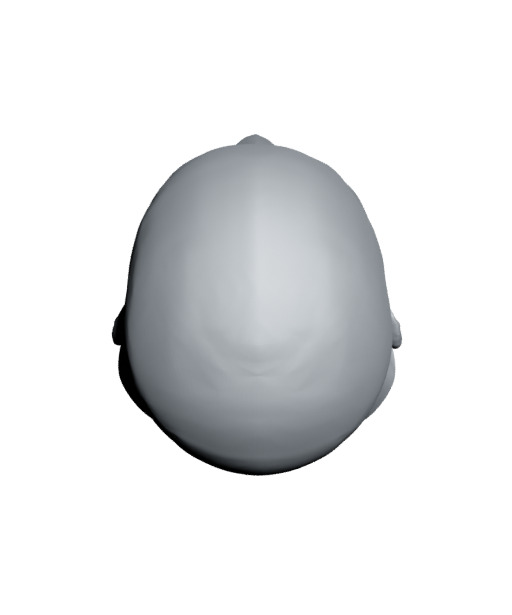} 
        & \includegraphics[width=0.18\textwidth,trim={0cm 3cm 0cm 3cm},clip]{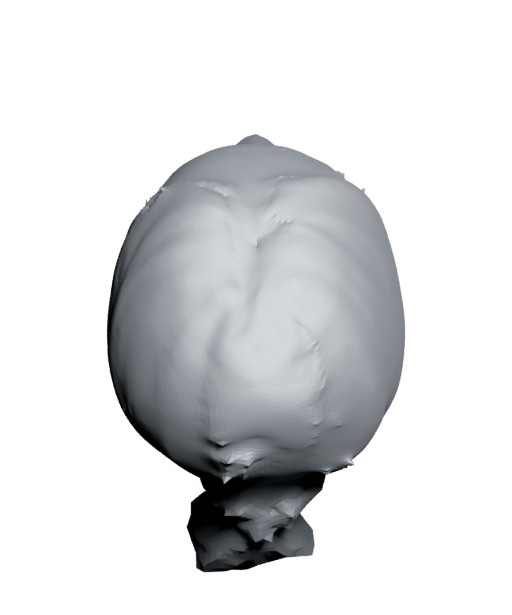}
        & \includegraphics[width=0.18\textwidth,trim={0cm 3cm 0cm 3cm},clip]{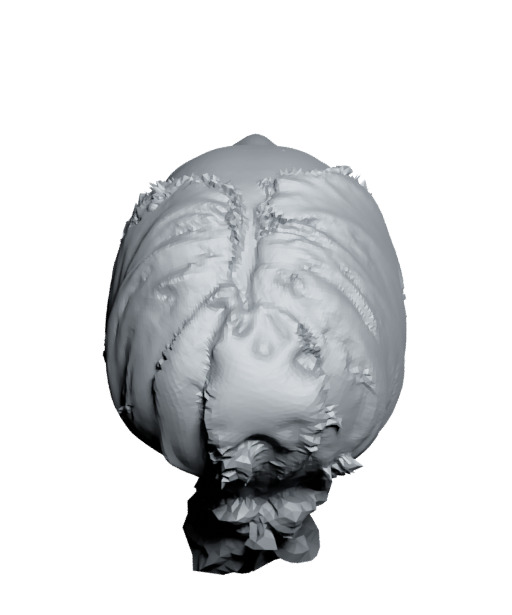}
        & \includegraphics[width=0.18\textwidth,trim={0cm 3cm 0cm 3cm},clip]{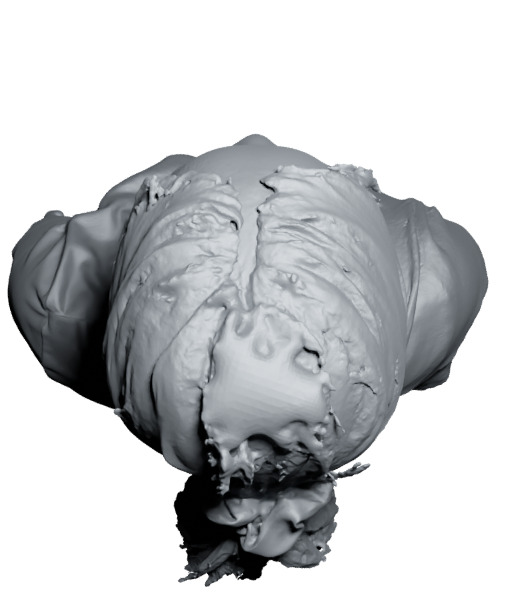}  \\
    \end{tabular}
    \captionof{figure}{
    Additional demonstration of the two-stage registration. 
    %
    %
    \textit{Stage 1} corresponds to the vector displacements regression; \textit{Stage 2} -- to the refinement of the displacements along the normals.
    The second stage significantly improves the level of detail and allows us to match the high-frequency component of the scans, such as strands and subtle face features.
    }
    \label{fig:more_registrations_p4}
    \vspace{-0.19cm}
\end{table*}
\begin{table*}[]
    \vspace{-0.3cm}
    \setlength{\tabcolsep}{0pt}
    \renewcommand{\arraystretch}{0}
    \centering
    \begin{tabular}{lcccccc}
        & Input p.c. & FLAME & Stage 1 & Stage 2 & Masking $m^\textrm{final}$ & Ground truth \\
        \raisebox{3\normalbaselineskip}[0pt][0pt]{\rotatebox[origin=c]{90}{Frontal}} & 
        \includegraphics[width=0.16\textwidth,trim={7cm 1cm 7cm 2.5cm},clip]{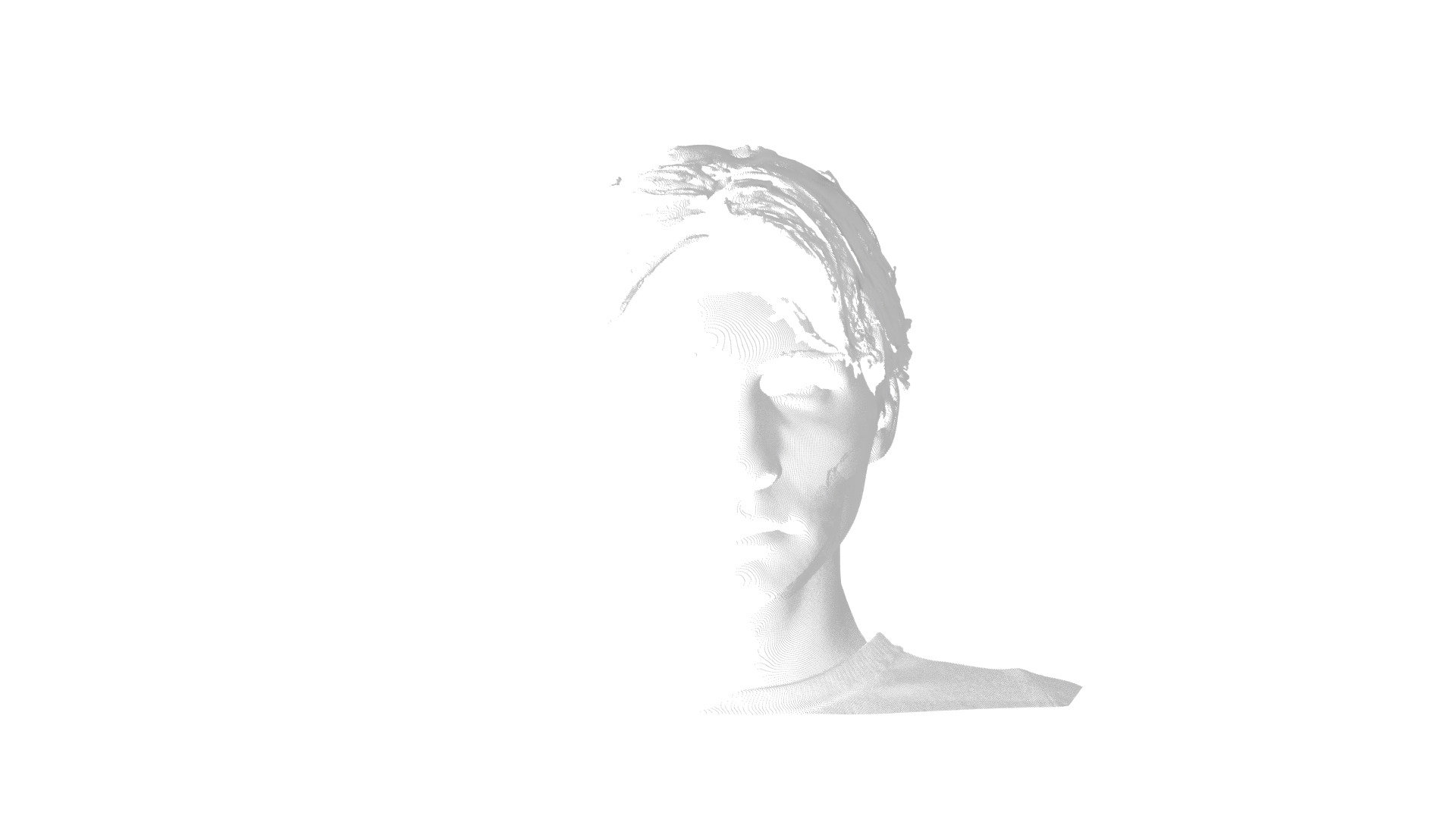} &
        \includegraphics[width=0.16\textwidth,trim={0cm 2cm 1cm 0cm},clip]{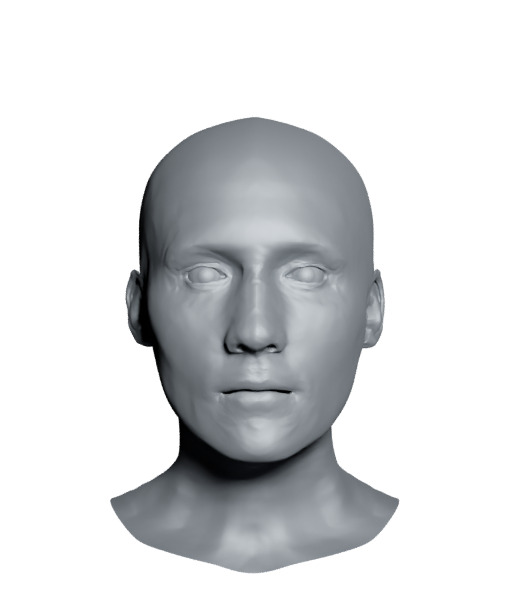} & 
        \includegraphics[width=0.16\textwidth,trim={0cm 2cm 1cm 0cm},clip]{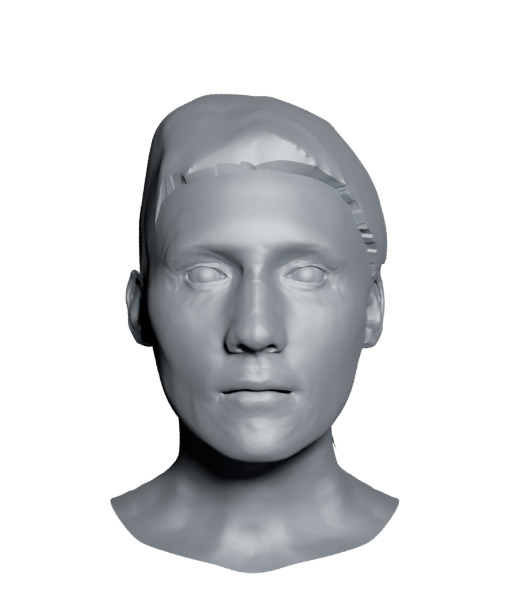} & 
        \includegraphics[width=0.16\textwidth,trim={0cm 2cm 1cm 0cm},clip]{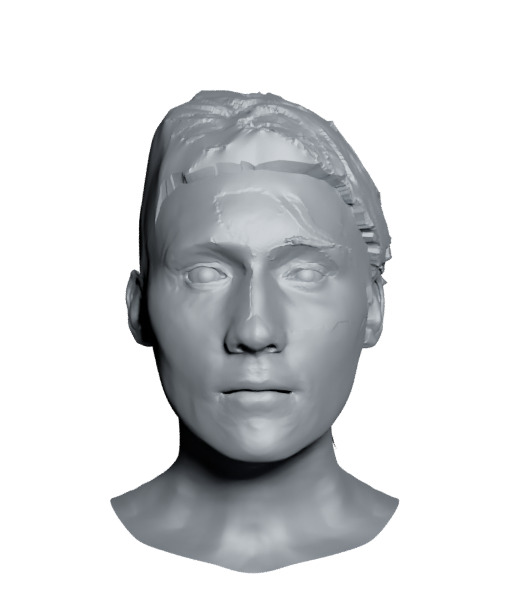} & 
        \includegraphics[width=0.16\textwidth,trim={0cm 2cm 1cm 0cm},clip]{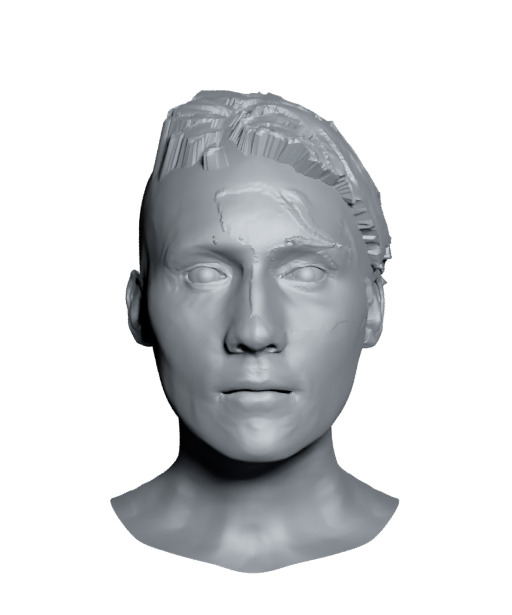} & 
        \includegraphics[width=0.16\textwidth,trim={0cm 2cm 1cm 0cm},clip]{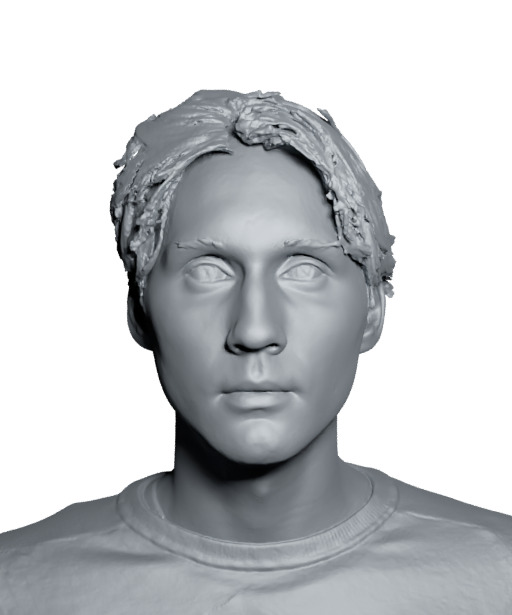}  \\
        \raisebox{3\normalbaselineskip}[0pt][0pt]{\rotatebox[origin=c]{90}{Left}} & 
        \includegraphics[width=0.16\textwidth,trim={7cm 1cm 7cm 2.5cm},clip]{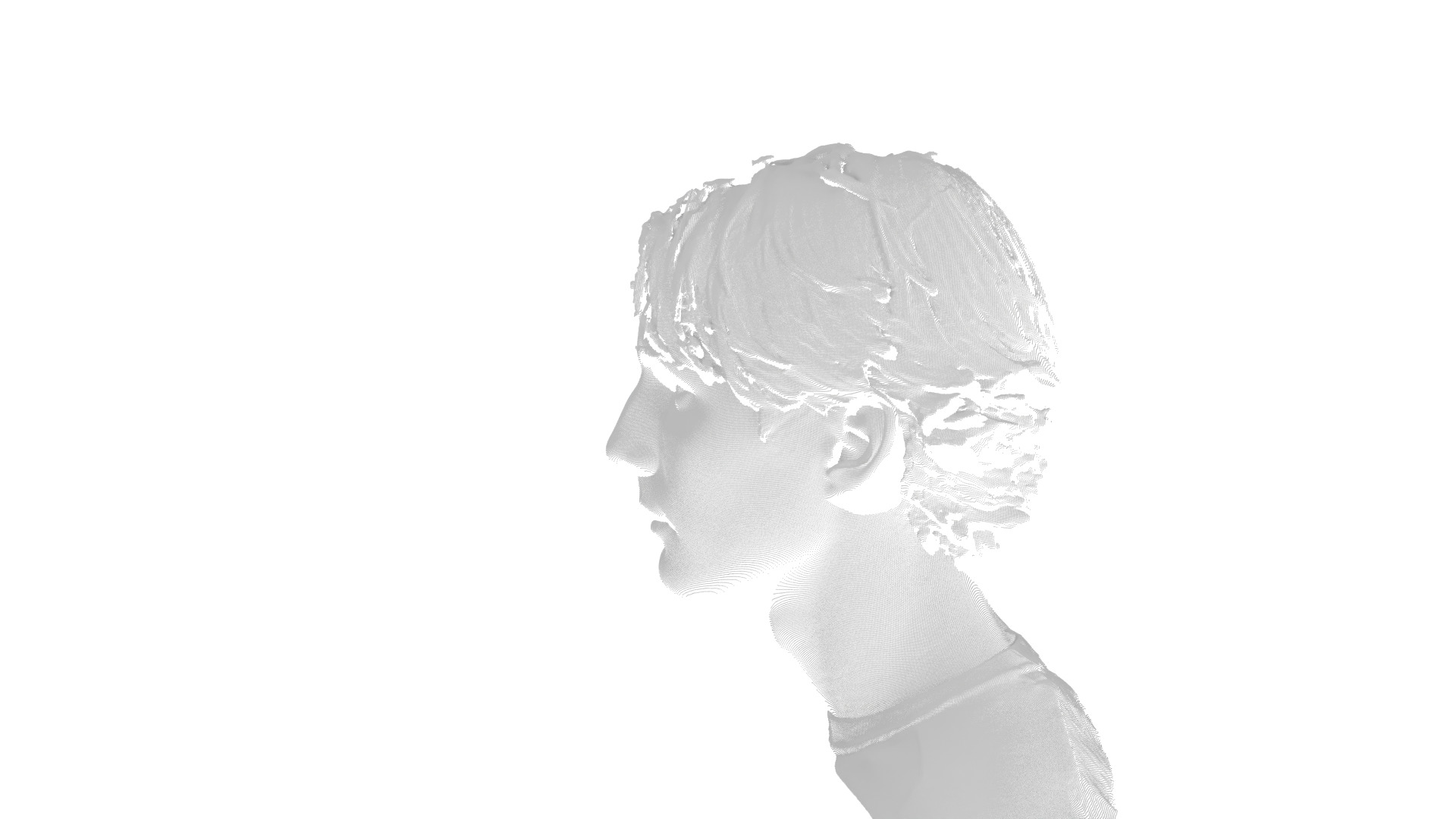} &
        \includegraphics[width=0.16\textwidth,trim={0cm 2cm 1cm 0cm},clip]{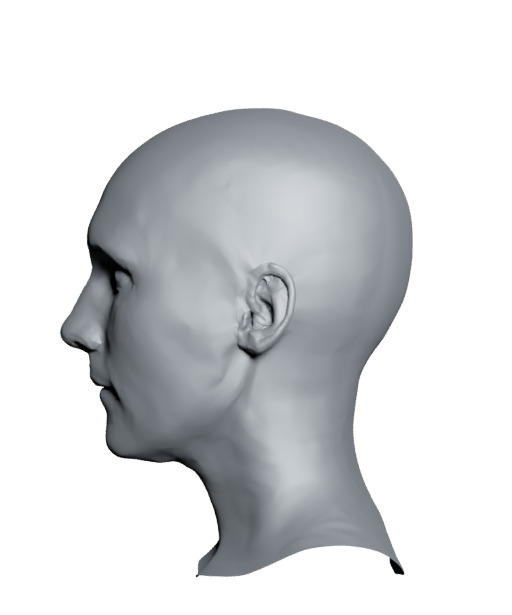} & 
        \includegraphics[width=0.16\textwidth,trim={0cm 2cm 1cm 0cm},clip]{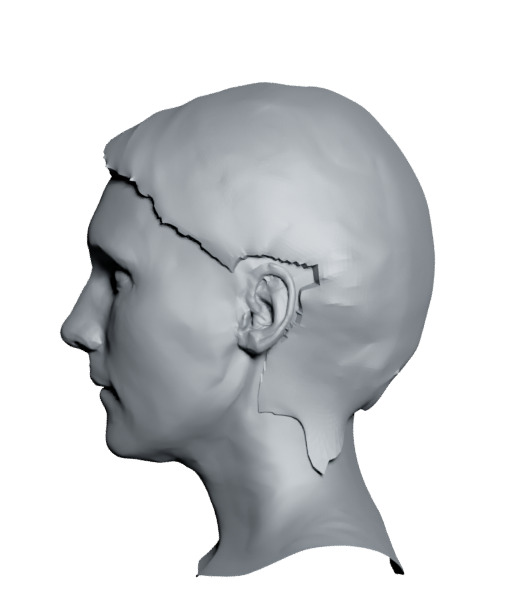} & 
        \includegraphics[width=0.16\textwidth,trim={0cm 2cm 1cm 0cm},clip]{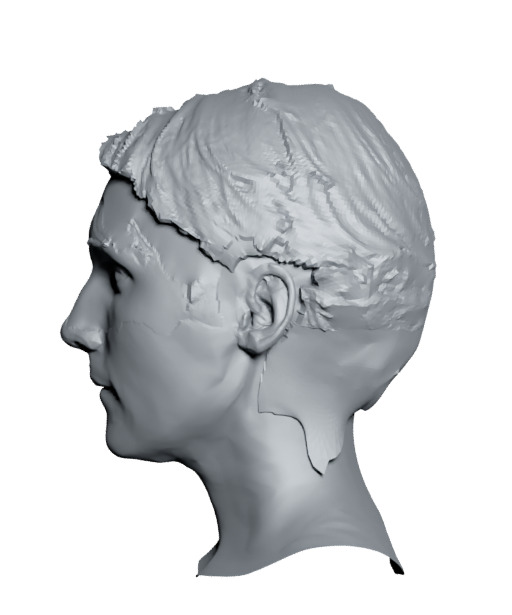} & 
        \includegraphics[width=0.16\textwidth,trim={0cm 2cm 1cm 0cm},clip]{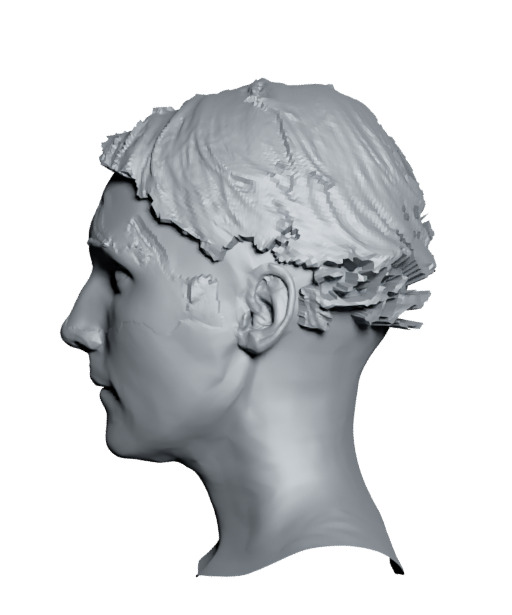} & 
        \includegraphics[width=0.143\textwidth,trim={1cm 2cm 2cm 0cm},clip]{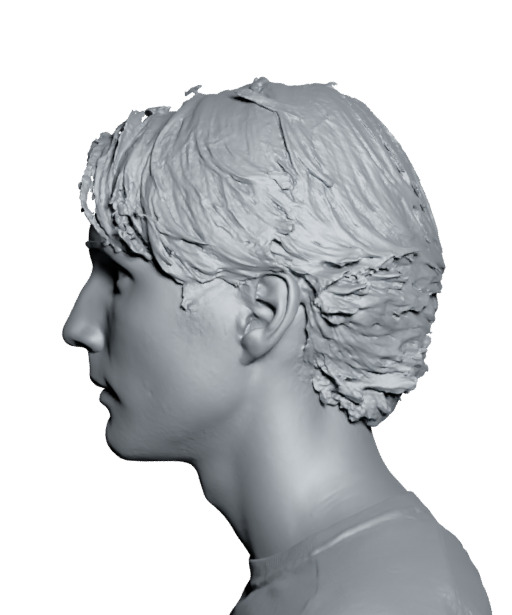}  \\
        \raisebox{3\normalbaselineskip}[0pt][0pt]{\rotatebox[origin=c]{90}{Top}} & 
        \includegraphics[width=0.16\textwidth,trim={7cm 1cm 7cm 2.5cm},clip]{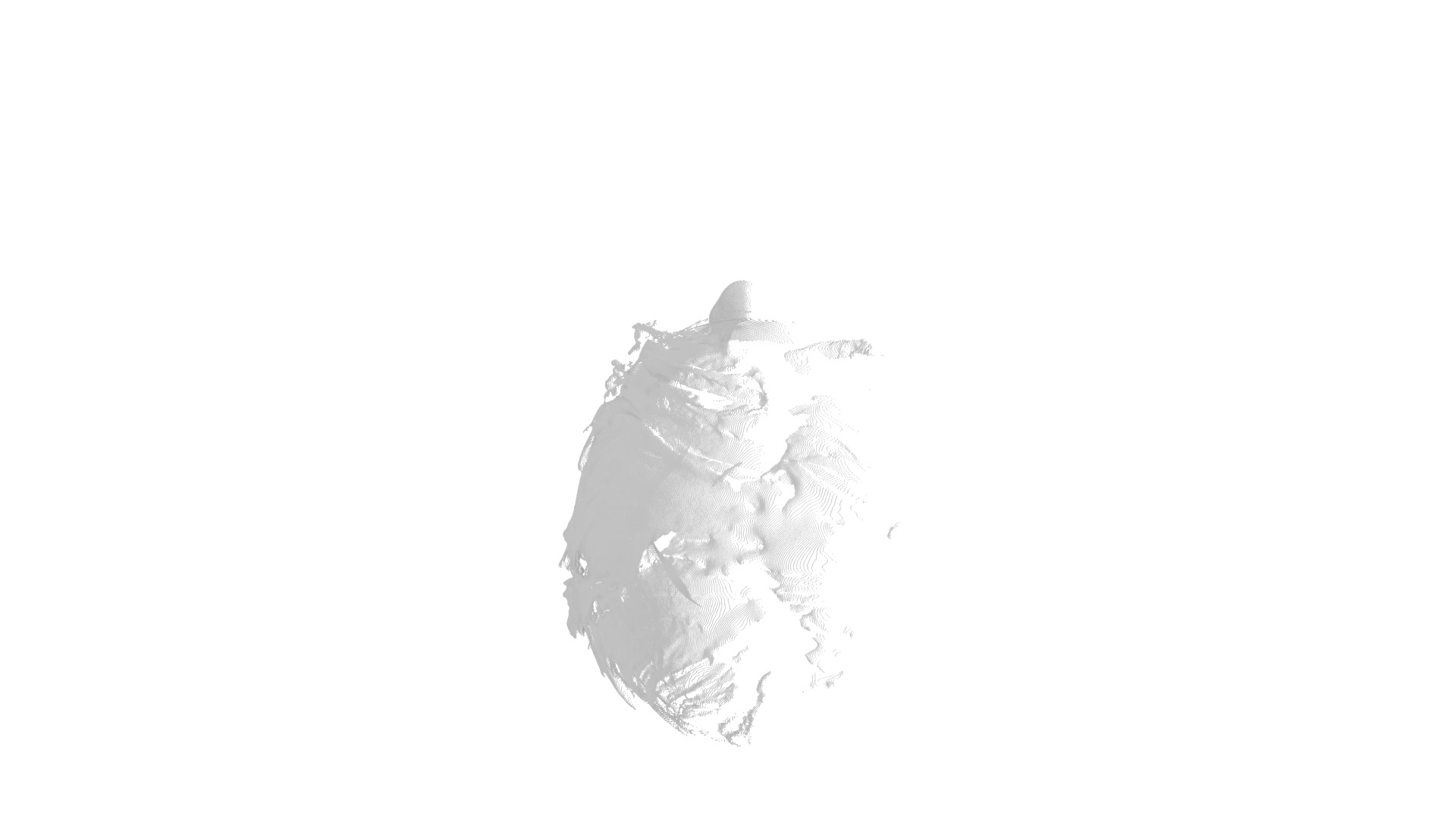} &
        \includegraphics[width=0.16\textwidth,trim={0cm 2cm 1cm 0cm},clip]{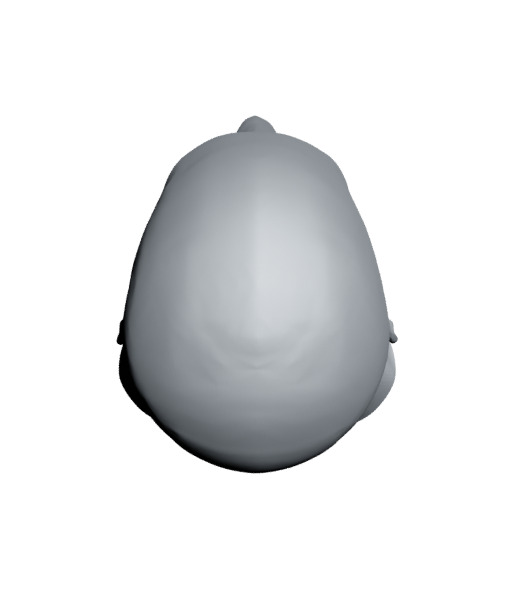} & 
        \includegraphics[width=0.16\textwidth,trim={0cm 2cm 1cm 0cm},clip]{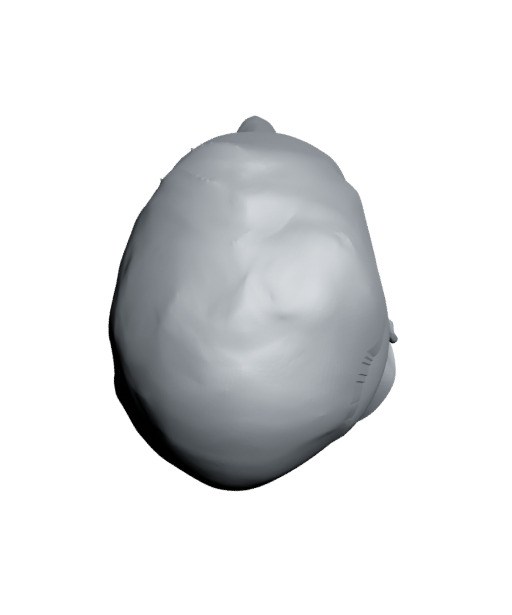} & 
        \includegraphics[width=0.16\textwidth,trim={0cm 2cm 1cm 0cm},clip]{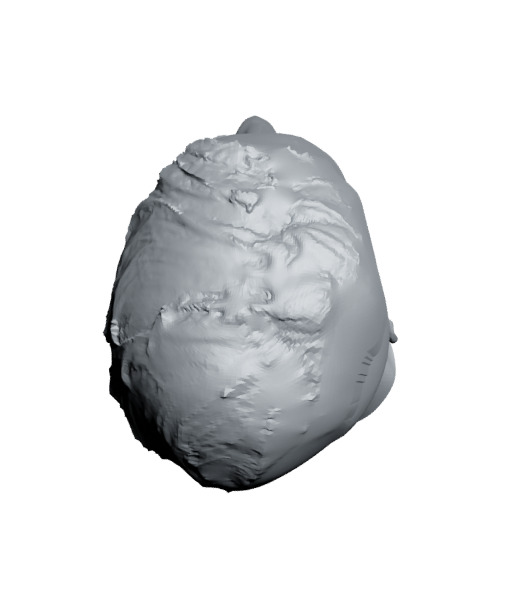} & 
        \includegraphics[width=0.16\textwidth,trim={0cm 2cm 1cm 0cm},clip]{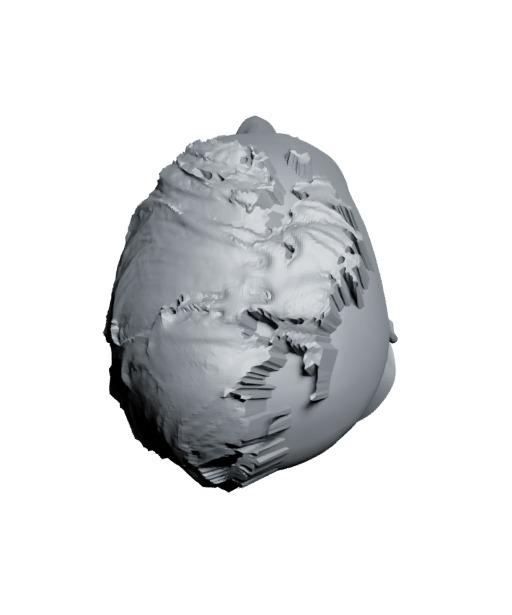} & 
        \includegraphics[width=0.16\textwidth,trim={0cm 2cm 1cm 0cm},clip]{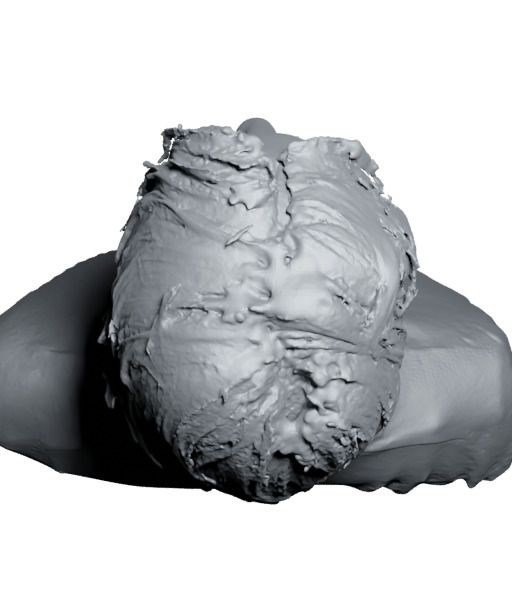}  \\
        %
        %
        & Input p.c. & FLAME & Stage 1 & Stage 2 & Masking $m^\textrm{final}$ & Ground truth \\
        \raisebox{3\normalbaselineskip}[0pt][0pt]{\rotatebox[origin=c]{90}{Frontal}} & 
        \includegraphics[width=0.16\textwidth,trim={7cm 1cm 7cm 2.5cm},clip]{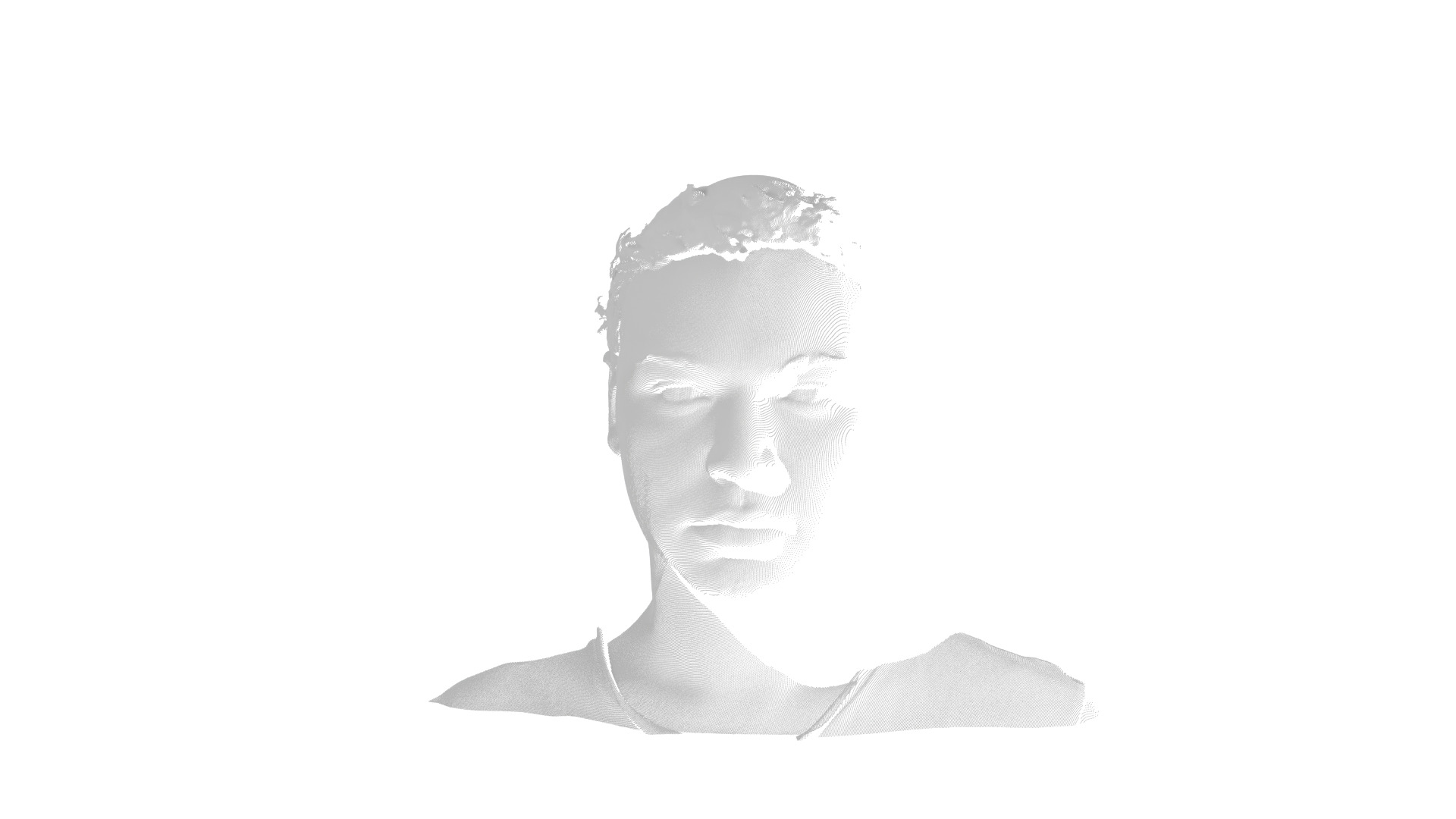} &
        \includegraphics[width=0.16\textwidth,trim={0cm 2cm 1cm 0cm},clip]{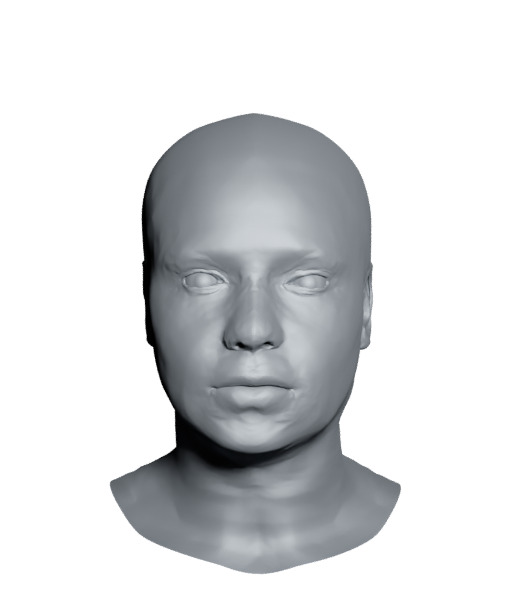} & 
        \includegraphics[width=0.16\textwidth,trim={0cm 2cm 1cm 0cm},clip]{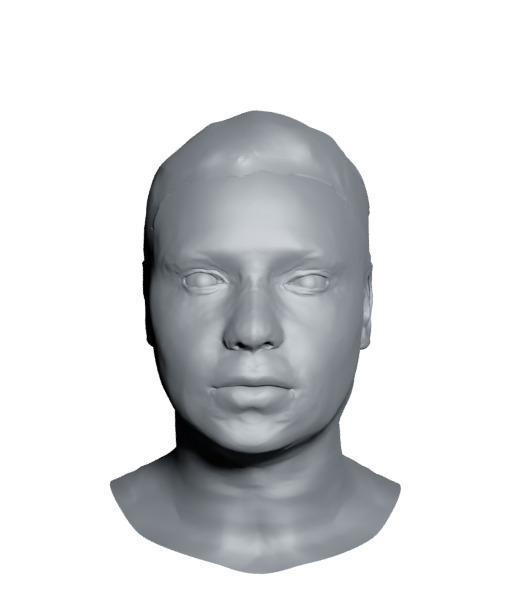} & 
        \includegraphics[width=0.16\textwidth,trim={0cm 2cm 1cm 0cm},clip]{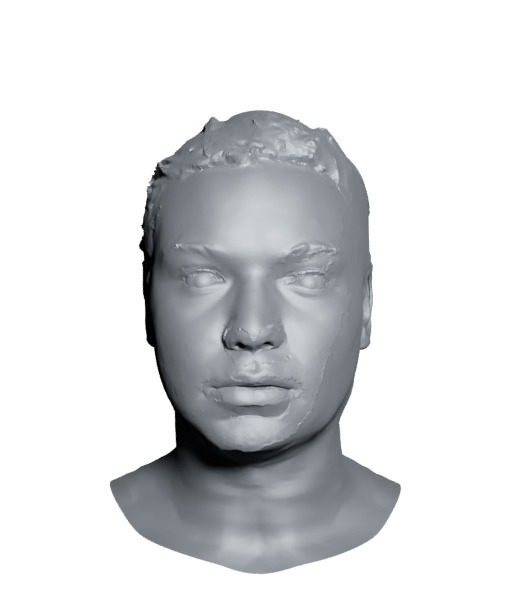} & 
        \includegraphics[width=0.16\textwidth,trim={0cm 2cm 1cm 0cm},clip]{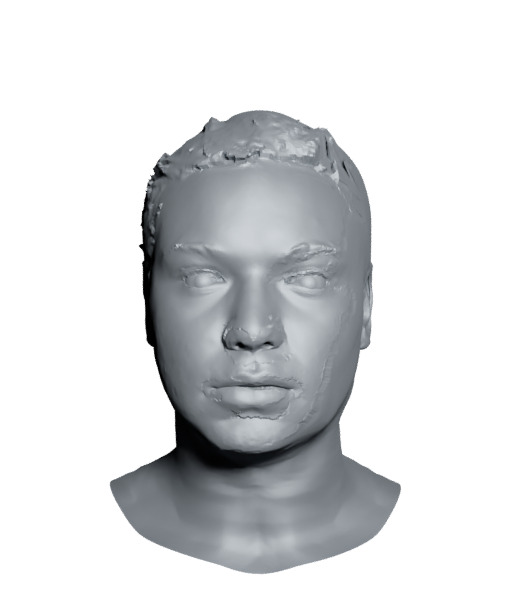} & 
        \includegraphics[width=0.16\textwidth,trim={0cm 2cm 1cm 0cm},clip]{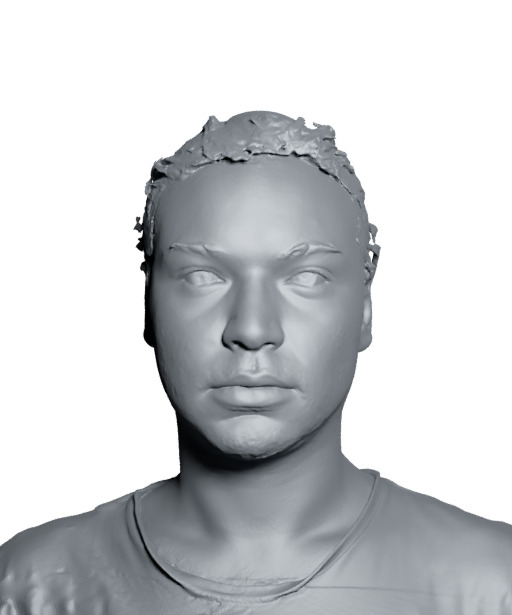}  \\
        \raisebox{2.5\normalbaselineskip}[0pt][0pt]{\rotatebox[origin=c]{90}{Left}} & 
        \includegraphics[width=0.16\textwidth,trim={7cm 1cm 7cm 2.5cm},clip]{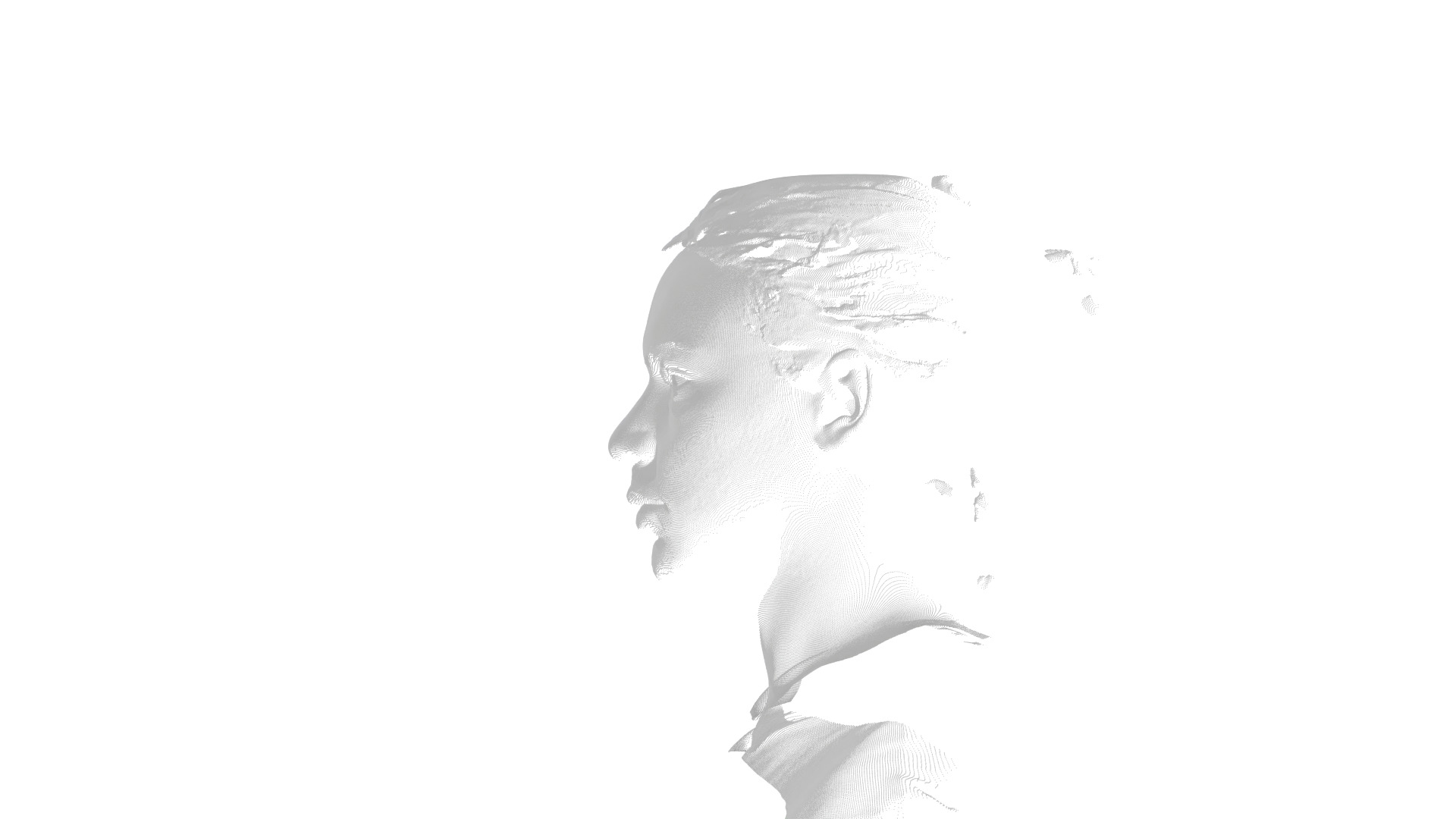} &
        \includegraphics[width=0.16\textwidth,trim={0cm 2cm 1cm 0cm},clip]{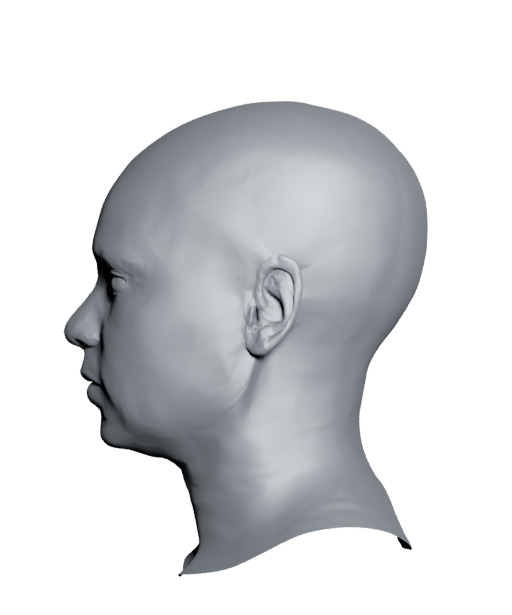} & 
        \includegraphics[width=0.16\textwidth,trim={0cm 2cm 1cm 0cm},clip]{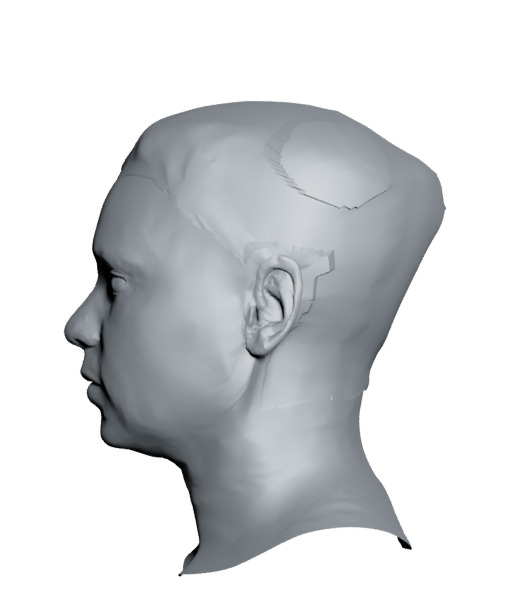} & 
        \includegraphics[width=0.16\textwidth,trim={0cm 2cm 1cm 0cm},clip]{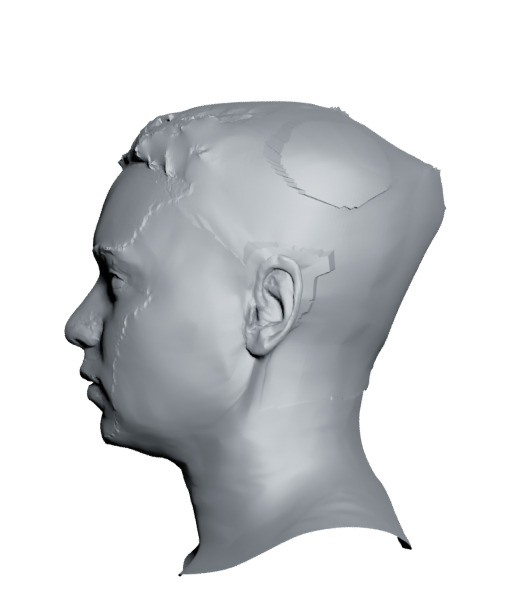} & 
        \includegraphics[width=0.16\textwidth,trim={0cm 2cm 1cm 0cm},clip]{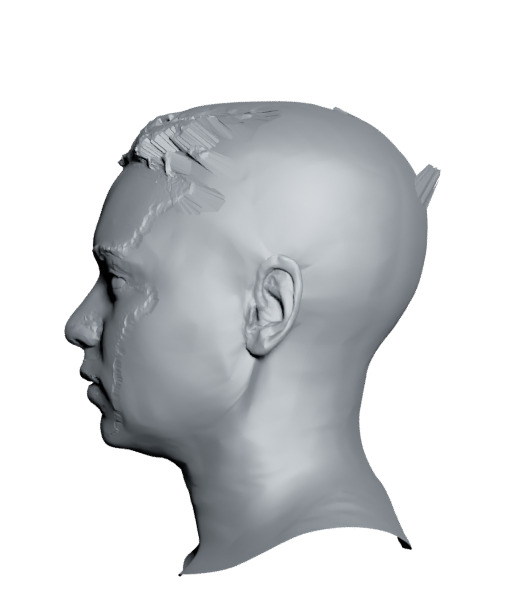} & 
        \includegraphics[width=0.143\textwidth,trim={1cm 2cm 2cm 0cm},clip]{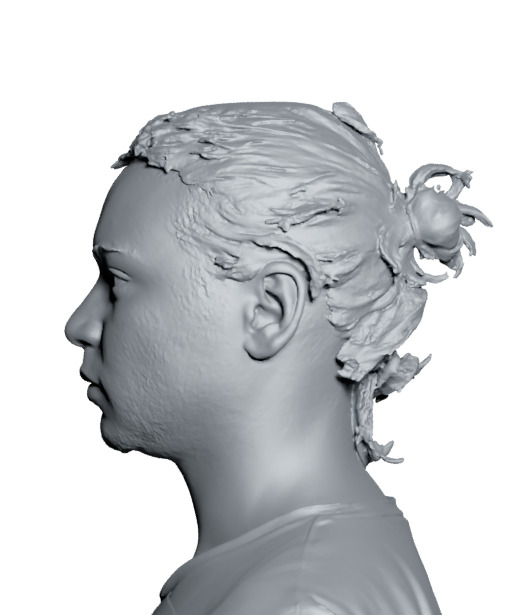}  \\
        \raisebox{3\normalbaselineskip}[0pt][0pt]{\rotatebox[origin=c]{90}{Top}} & 
        \includegraphics[width=0.16\textwidth,trim={7cm 1cm 7cm 2.5cm},clip]{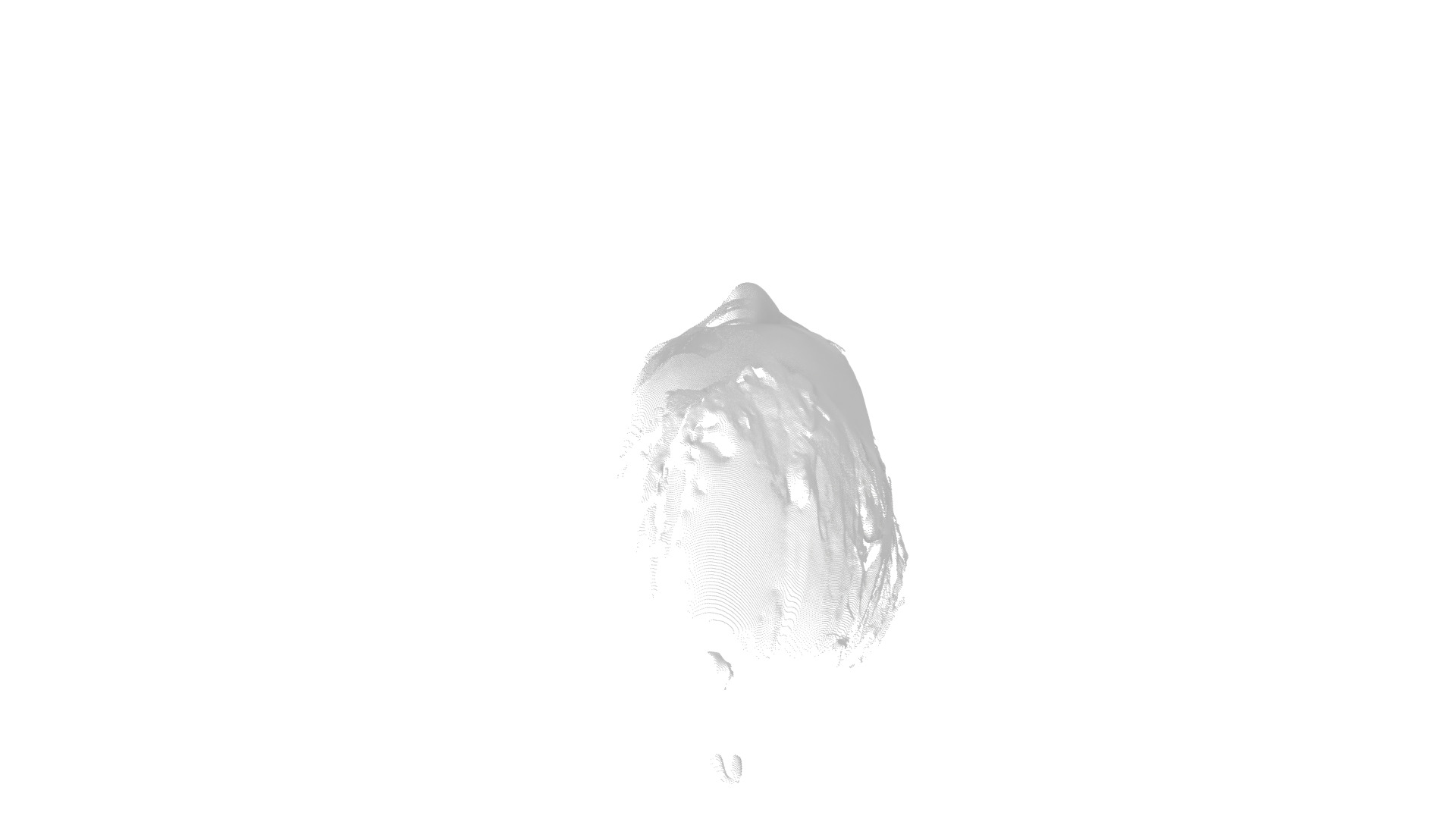} &
        \includegraphics[width=0.16\textwidth,trim={0cm 2cm 1cm 0cm},clip]{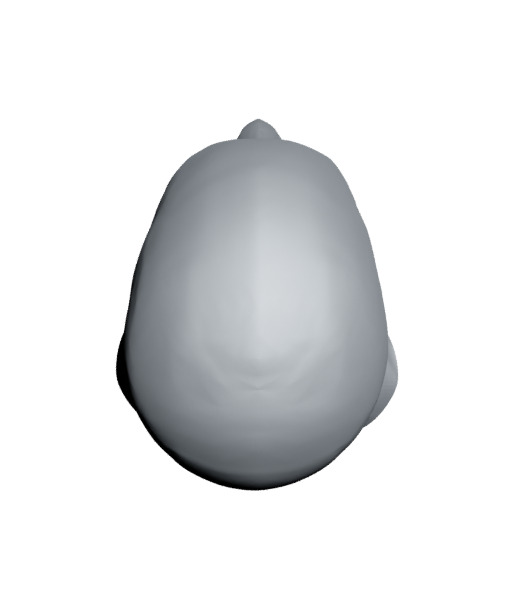} & 
        \includegraphics[width=0.16\textwidth,trim={0cm 2cm 1cm 0cm},clip]{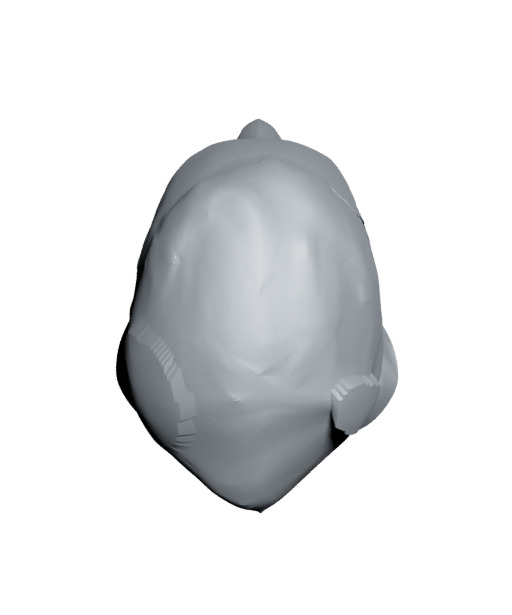} & 
        \includegraphics[width=0.16\textwidth,trim={0cm 2cm 1cm 0cm},clip]{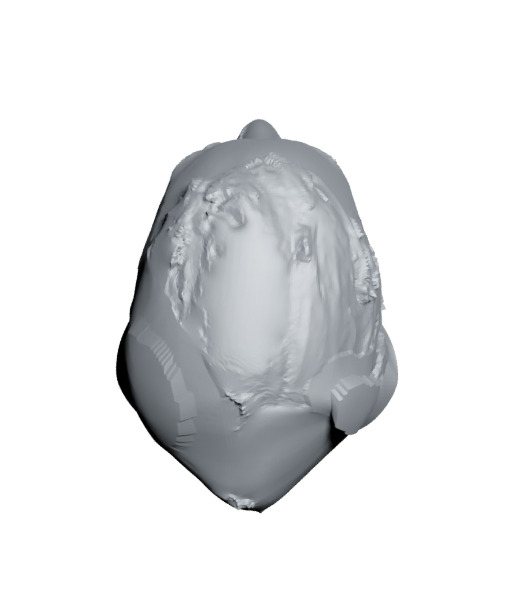} & 
        \includegraphics[width=0.16\textwidth,trim={0cm 2cm 1cm 0cm},clip]{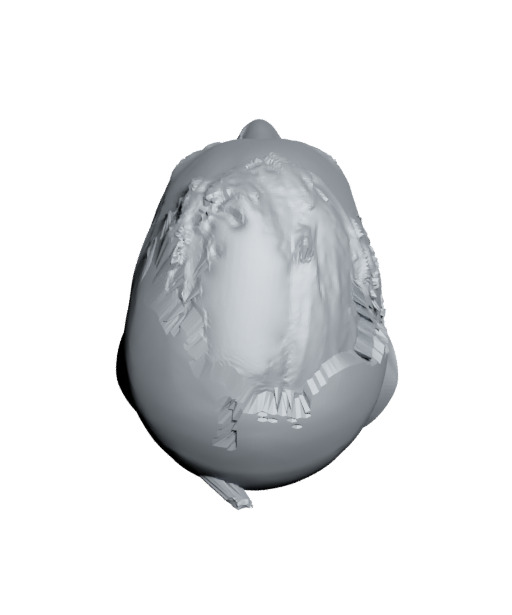} & 
        \includegraphics[width=0.16\textwidth,trim={0cm 2cm 1cm 0cm},clip]{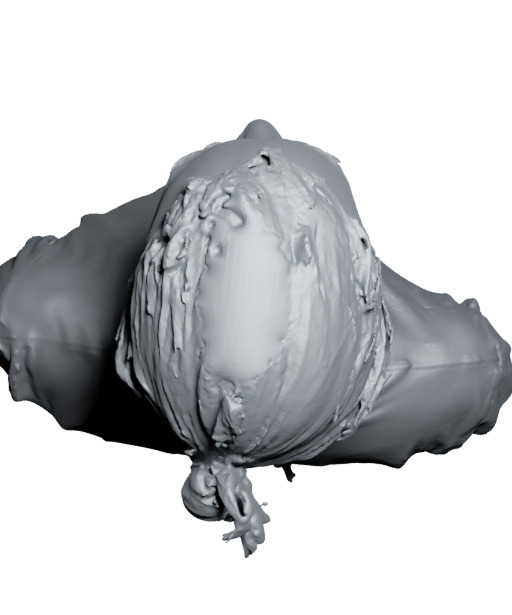}  \\
    \end{tabular}
    \vspace{-0.25cm}
    \captionof{figure}{
    Demonstration of the stages of the partial registration procedure required to fit a part of the scan.
    %
    %
    The key difference between this procedure and the standard registration used to generate training data for HeadCraft is in the presence of only a part of the scan, e.g. a point cloud coming from the depth map.
    To overcome that obstacle, the displacements are being estimated only in the convex hull of the point cloud, and are subsequently filtered out by a separate mask $m^\textrm{final}$, leaving only the displacements close enough the ground truth scan (others are nullified in this visualization).
    }
    \label{fig:partial_registration_vis}
\end{table*}
\begin{table*}[]
    \begin{center}
        \begin{tabular}{c}
            \includegraphics[width=.9\textwidth,trim={0cm 3cm 0cm 3cm},clip]{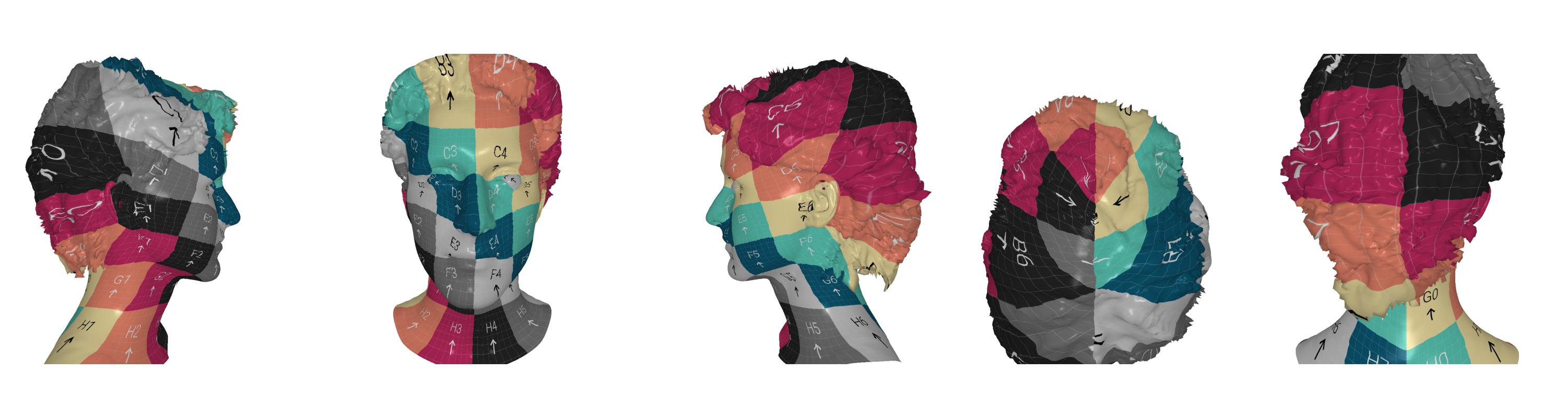} \\
            \includegraphics[width=.9\textwidth,trim={0cm 3cm 0cm 3cm},clip]{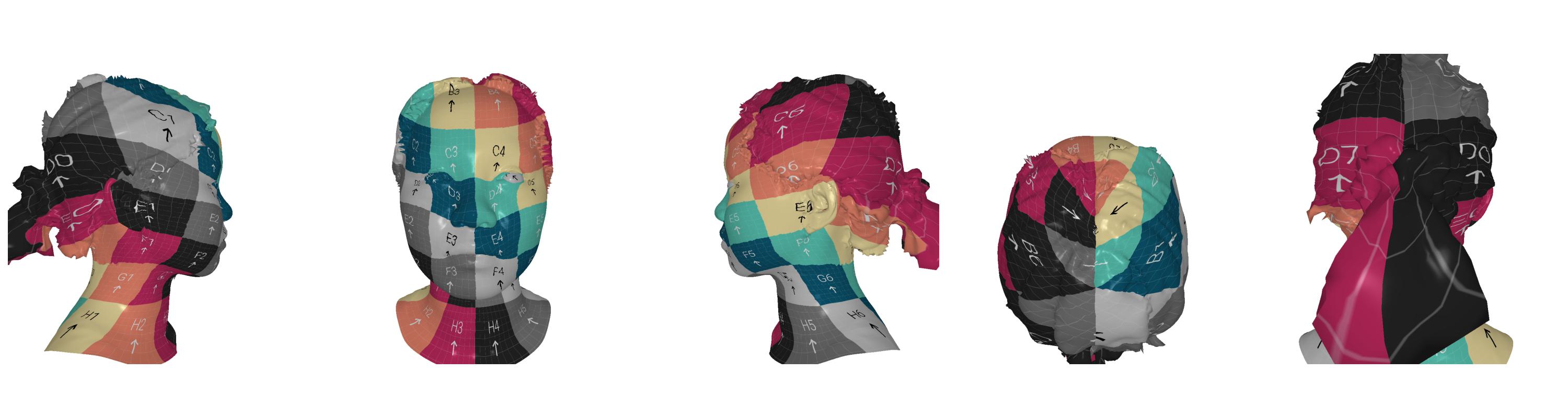} \\
            \includegraphics[width=.9\textwidth,trim={0cm 3cm 0cm 3cm},clip]{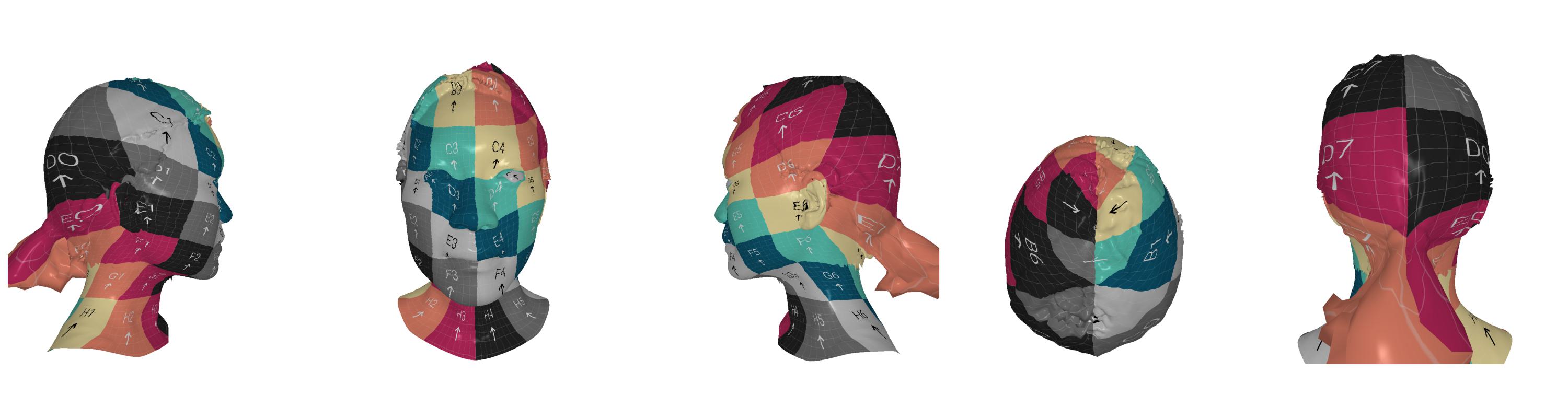} \\
            \includegraphics[width=.9\textwidth,trim={0cm 3cm 0cm 3cm},clip]{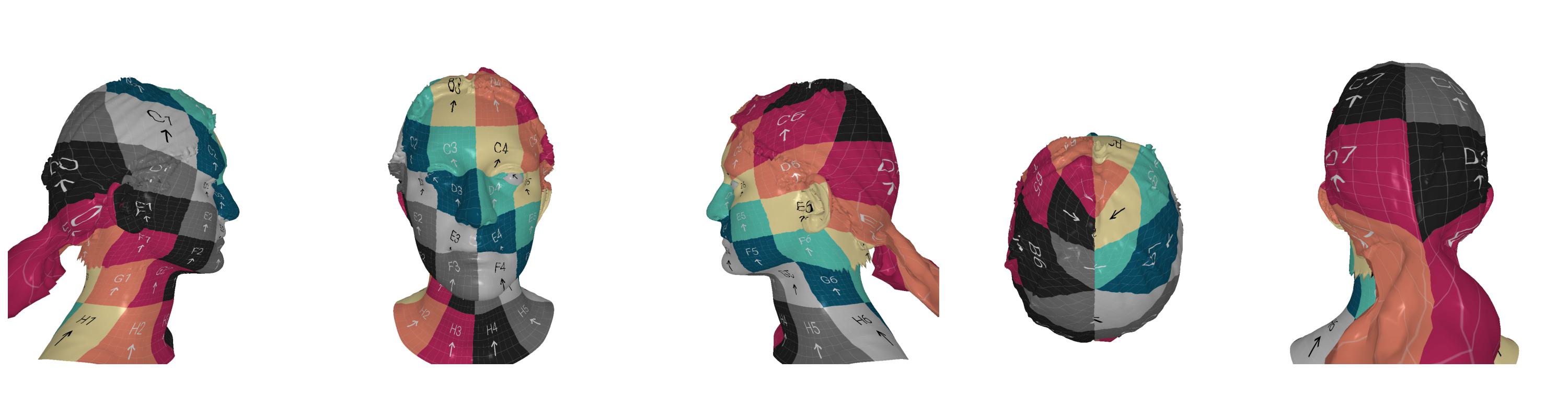} \\
            \includegraphics[width=.9\textwidth,trim={0cm 3cm 0cm 3cm},clip]{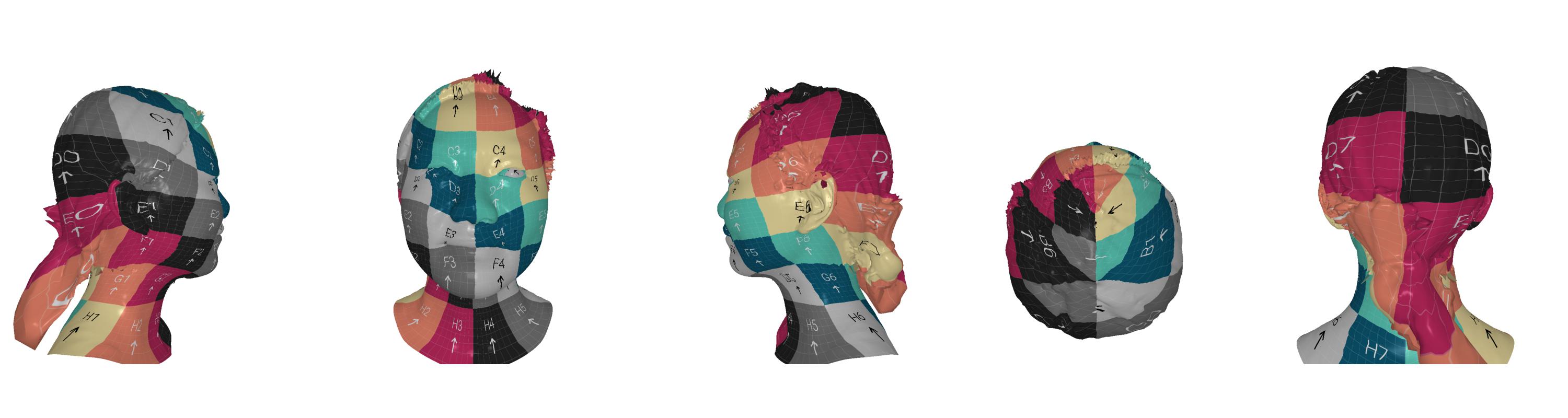} \\
        \end{tabular}
    \captionof{figure}{
    Consistency analysis of the registration. 
    We demonstrate which template vertices are offset with the registration procedure 
    to cover various regions of different meshes. 
    Since we know the UV coordinates of all template vertices, this can be done by rendering the meshes with a \textit{UV checker} texture image.
    For clarity of the visualization, the texture is applied to the standard FLAME layout.
    Note that the long hair parts, such as pony tails, are mostly explained by the same regions of the layout as the vertices they originate from.
    }
    \label{fig:uv_checker}
    \end{center}
\end{table*}
\end{appendix}

\end{document}